\documentclass[10pt,twocolumn,letterpaper]{article}

\usepackage[pagenumbers]{cvpr} 

\usepackage[accsupp]{axessibility}
\definecolor{colCD}{RGB}{0, 47, 167}%
\definecolor{colFONE}{RGB}{230, 110, 40}%

\newcommand{\yes}{\textcolor{ForestGreen!70}{\faCheck}}
\newcommand{\no}{\textcolor{red!75!black!60}{\faTimes}}

\newcommand{\greyrule}{\arrayrulecolor{black!30}\midrule\arrayrulecolor{black}}

\newcommand{\posenc}[3]{
    \node[draw, circle, minimum size=#3 cm, thick] (posenc) at (#1,#2) {};
    \draw[domain=0:360,samples=100,smooth,variable=\x, thick] 
    plot({#1+#3/2+\x/360*#3 - #3},{#2+sin(\x)*(0.3*#3)});
    \draw[domain=0:360,samples=100,smooth,variable=\x, thick] 
    plot({#1+0+#3/2+\x/360*#3 - #3},{#2+sin(2*\x)*(0.3*#3)});
    }

\newcommand{\SEP}{\texttt{SEP}}
\newcommand{\EOS}{\texttt{EoS}}
\newcommand{\MASK}{\texttt{MASK}}
\newcommand{\cS}{\mathcal{S}}
\newcommand{\bR}{\mathbb{R}}

\DeclareMathOperator{\Lin}{Lin}
\DeclareMathOperator{\MLP}{MLP}
\DeclareMathOperator{\type}{type}

\newcommand{\negphantom}[1]{\settowidth{\dimen0}{#1}\hspace*{-\dimen0}}

\newcommand{\jetbar}[1]{\def\colormapheight{#1}\rotatebox{0}{
\pgfplotsset{
  /pgfplots/colormap={viridis_mpl}{
    rgb255(0cm)=(185,175,50), 
    rgb255(1cm)=(110,175,120),
    rgb255(2cm)=(150,160,170),
    rgb255(3cm)=(160,120,130) ,
    rgb255(4cm)=(160,25,140),
  }
}
\begin{tikzpicture}
  \begin{axis}[
    hide axis,
    width=0pt, height=0pt,
    scale only axis,
    colormap name=viridis_mpl, 
    colorbar,
    colorbar style={
      at={(0,0)}, anchor=south west,
      width=.2cm,
      height=\colormapheight,
      ytick=\empty,
      yticklabels=\empty,
      yticklabel style={font=\tiny},
      major tick length=1.5pt,
      line width=.05mm,
      grid style={draw=none},
    },
    point meta min=0,
    point meta max=4,
  ]
    \addplot [draw=none] coordinates {(0,0)};
  \end{axis}
\end{tikzpicture}}}

\usepackage{fontawesome5}
\usepackage{makecell}
\usepackage{multicol,multirow}
\usepackage[table,RGB]{xcolor}
\usepackage{siunitx}
\usepackage{calc}
\usepackage{paralist}
\usepackage{adjustbox}
\usepackage{microtype}
\usepackage{balance}

\usepackage{tikz}
\usepackage{tikz-3dplot}
\usetikzlibrary{positioning,shapes,calc,3d,fit}
\usepackage{pgfplots}
\pgfplotsset{compat=1.14}

\usepackage[skip=3pt]{caption}
\renewcommand{\arraystretch}{0.99}

\tdplotsetmaincoords{60}{30}
\definecolor{cvprblue}{rgb}{0.21,0.49,0.74}
\usepackage[pagebackref,breaklinks,colorlinks,allcolors=cvprblue]{hyperref}

\def\paperID{1218} 
\def\confName{CVPR}
\def\confYear{2025}

\title{Order Matters: 3D Shape Generation from Sequential VR Sketches
}

\author{Yizi Chen \textsuperscript{1}\thanks{Equal contribution.}
\qquad
Sidi Wu \textsuperscript{1}\footnotemark[1]
\qquad
Tianyi Xiao \textsuperscript{1}
\qquad
Nina Wiedemann \textsuperscript{1}
\qquad
Loic Landrieu\textsuperscript{2}
\vspace{.5em}
\\
{\textsuperscript{1} \small  ETH Zurich
}
\quad
{
\textsuperscript{2} \small  LIGM, ENPC, IP Paris, Univ Gustave Eiffel, CNRS
}
}

\begin{document}
\maketitle

\begin{abstract}
VR sketching lets users explore and iterate on ideas directly in 3D, offering a faster and more intuitive alternative to conventional CAD tools.
However, existing sketch-to-shape models ignore the temporal ordering of strokes, discarding crucial cues about structure and design intent.
We introduce \textsc{VRSketch2Shape}, the first framework and multi-category dataset for generating 3D shapes from \emph{sequential} VR sketches.
Our contributions are threefold: (i) an automated pipeline that generates sequential VR sketches from arbitrary shapes, (ii) a dataset of over $20$k synthetic and $900$ hand-drawn sketch–shape pairs across four categories, and (iii) an order-aware sketch encoder coupled with a diffusion-based 3D generator.
Our approach yields higher geometric fidelity than prior work, generalizes effectively from synthetic to real sketches with minimal supervision, and performs well even on partial sketches. All data and models are released open-source on \url{https://chenyizi086.github.io/VRSketch2Shape_website/}.
\end{abstract}    
\section{Introduction}
\label{sec:intro}

Creating high-quality 3D content is central to architecture and industrial design and is well supported by powerful CAD tools such as Blender~\cite{willis2020fusion,liu2024point2cad}. 
However, these tools have a steep learning curve and are optimized for precision, making them ill-suited for rapid ideation and early-stage exploration---key steps in the creative process. 
Recent research has therefore explored text-conditioned generative models for 3D shape synthesis~\cite{sanghi2022clip,fu2022shapecrafter,lin2023magic3d}, but natural language remains too ambiguous to specify complex geometries~\cite {sangkloy2022sketch,yu2016sketch}.

\begin{figure}[t]
    \centering
    \input{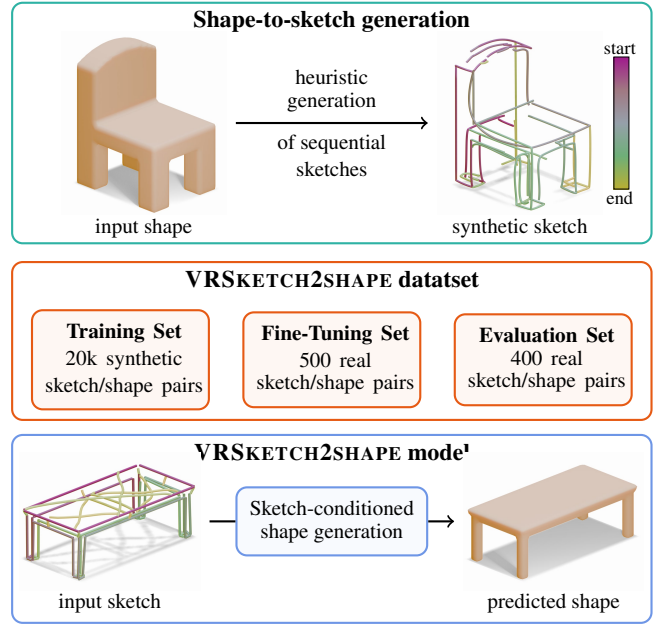}
\caption{{\bf {Overview of Contributions.}} 
We propose: (i)~a learning-free pipeline to generate realistic sequential 3D sketches from arbitrary shapes; 
(ii)~the open-access \textsc{VRSketch2Shape} dataset, with $20$k synthetic and $900$ hand-drawn sketch–shape pairs; and 
(iii)~an order-aware, diffusion-based model to generate high-fidelity 3D shapes from sequential VR sketches.}
\label{fig:teaser}
\end{figure}

\paragraph{Sketch-Based 3D Design.}
Sketching provides a fast and intuitive way to express spatial concepts. 
Early work relied on single- or multi-view \textit{2D} sketches for shape generation~\cite{bandyopadhyay2024doodle,guillard2021sketch2mesh,zhang2021sketch2model,zheng2023locally}. 
With the advent of commodity VR/AR systems, \textit{3D} sketching has emerged as a natural and immersive alternative~\cite{chen2024rapid,luo20233d}: drawing directly in 3D space eliminates perspective ambiguities and occlusions inherent to 2D sketches, while enjoying a more natural and immersive design experience.

\paragraph{Open Challenges.}
Despite its promise, VR sketch–conditioned shape generation faces three main challenges:  
(i)~\textbf{Data scarcity.} Collecting paired VR sketches and 3D meshes is costly; the only public benchmark~\cite{luo2021data} includes just $1{,}005$ sketch–chair pairs from a single category.  
(ii)~\textbf{Geometric misalignment.} Human-annotated sketches naturally include spatial inaccuracies from perspective and depth-perception errors, resulting in imperfect alignment with target shapes and complicating both training and evaluation.
(iii)~\textbf{Temporal information loss.} Existing pipelines to create shapes from VR sketches~\cite{luo2021data,chen2024rapid} but treat them as unordered point clouds, thus discarding stroke order and length; yet these signals encode important information about connectivity, structure, and design intent. 

\paragraph{\textsc{VRSketch2Shape}.}
We propose a new framework to generate 3D shapes from \emph{sequential VR sketches}.  
We model a sketch as a sequence of \emph{strokes}, each itself an ordered sequence of 3D \emph{points}.  
Building on this formulation, we make three primary contributions:
\begin{compactitem}
    \item \textbf{Synthetic Sketch Generation.} An automatic pipeline that produces sequential sketches from arbitrary 3D shapes, yielding over $20$k paired samples for large-scale training.
 \item \textbf{Real Sketch Collection.}
A custom VR sketching interface with surface snapping to reduce drawing errors. Using this tool, we produced $900$ VR sketches across four categories, each annotated with complete stroke and point ordering.
    \item \textbf{Order-Aware Shape Generation.} A sketch encoder that models stroke sequences using a modified BERT architecture~\cite{devlin2019bert}, coupled with \textsc{SDFusion}~\cite{cheng2023sdfusion} for diffusion-based shape generation.
\end{compactitem}

\paragraph{Results.}
The \textsc{VRSktech2Shape} model outperforms prior work by a large margin on both existing and newly collected benchmarks.  
Trained solely on synthetic sketches, it generalizes effectively to real sketches with little or no fine-tuning, highlighting both the robustness of our model and the utility of our synthetic data.  
Moreover, the model remains stable with partial sketches, enabling cross-modal shape completion---an ability that could greatly accelerate interactive 3D design workflows.

\section{Related work}

We first review prior work on 3D shape generation from conventional modalities such as text and images (\cref{sec:related:gene}), followed by sketch-based methods (\cref{sec:related:sketch}).
We then discuss related approaches for sketch generation and encoding (\cref{sec:related:synthetic}).

\subsection{Classical 3D Shape Generation}
\label{sec:related:gene}
\paragraph{Generative Approaches.}
Early work on 3D shape generation explored a range of paradigms, including Generative Adversarial Networks (GANs)~\cite{achlioptas2018gan, chen2019gan, wu2016gan, wu2020gan, zheng2022gan}, 
Variational Autoencoders (VAEs)~\cite{park2019vaegen, cheng2022vaegen}, 
and auto-regressive models~\cite{yan2022autoreg, mittal2022autoreg}.  
Recent advances have shifted toward diffusion-based approaches, which produce high-fidelity 3D content in the form of point clouds~\cite{kong2022pcdiffusion, luo2021pcdiffusion}, 
voxel occupancy grids~\cite{zhou2021voxeldiffusion}, or meshes~\cite{Liu2023MeshDiffusion}.

\paragraph{Implicit Representations.}
Unlike explicit 3D formats, implicit neural fields offer continuous surfaces, compact storage, and theoretically infinite resolution.  
Recent efforts have used diffusion methods to generate signed distance functions (SDFs)~\cite{cheng2023sdfusion, nam2022neuraldiffusion} or neural radiance fields~\cite{poole2022dreamfusion, metzer2023latent}.  
To improve scalability, several diffusion models operate in latent space~\cite{cheng2023sdfusion, nam2022neuraldiffusion, rombach2022high}.  
Generation can be conditioned by images~\cite{cheng2023sdfusion, liu2023one, liu2023zero, shi2023mvdream, tang2023make, tian2023shapescaffolder} or text prompts~\cite{cheng2023sdfusion, fu2022shapecrafter, chen2024it3d, lin2023magic3d}, enabling controllable 3D shape synthesis.


\subsection{Sketch-Based 3D Shapes Generation }
\label{sec:related:sketch}
\paragraph{2D sketches.}
Sketching has recently emerged as a powerful modality for 3D shape generation and editing, enabling users to specify geometry through intuitive freehand input.  
Early work learned deterministic mappings from sketches to 3D shapes~\cite{igarashi2006teddy, karpenko2006smoothsketch,
wang20223d, chen2023deep3dsketch+, zang2023deep3dsketch+}, whereas recent methods adopt diffusion-based generative models~\cite{bandyopadhyay2024doodle, zheng2023locally}.  
To mitigate the inherent ambiguity of single-view sketches, some approaches incorporated camera parameters or viewpoint conditioning~\cite{zhang2021sketch2model, chen2023deep3dsketch, guillard2021sketch2mesh, zheng2023locally}, while others leveraged multi-view sketches for higher geometric accuracy~\cite{lun20173d, delanoy20183d}.

\paragraph{VR sketches.}
More recently, \textit{3D} sketches drawn in virtual reality (VR) have been explored for shape generation, providing a more immersive and spatially intuitive design interface~\cite{luo2021data, luo20233d, chen2024rapid, gu2025vrsketch2gaussian}.
Existing methods represent VR sketches as 3D point clouds and align their latent representations with those of point clouds sampled from target shapes~\cite{luo20233d} or with 2D renderings of 3D shapes~\cite{chen2024rapid}.
However, these approaches ignore the sequential nature of sketches, whereas the \textsc{VRSketch2Shape} explicitly models the temporal order of strokes and points.



\begin{figure*}[t]
  \centering
  \input{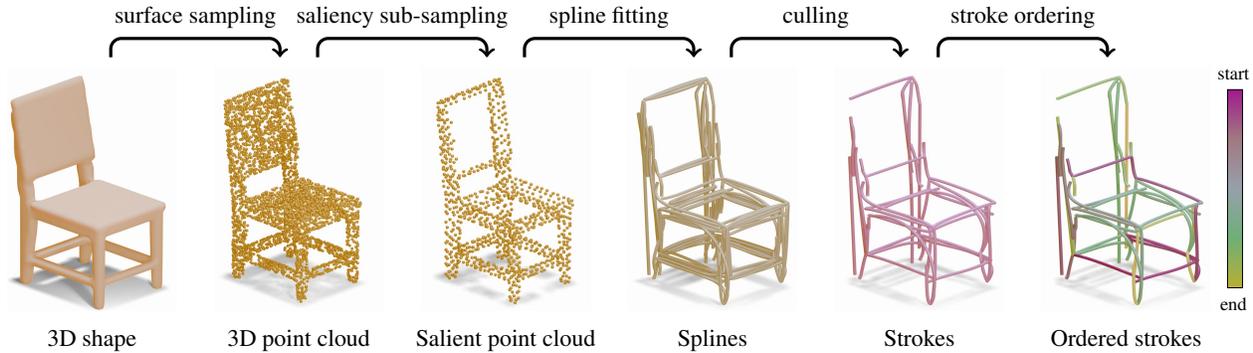}
  \caption{{\bf Synthetic Sketch Generation.} 
  We propose a heuristic, learning-free pipeline for generating 3D sequential sketches from 3D shapes. 
  We first uniformly sample points on the surface and retain only \emph{salient} points. 
  Bézier splines are then fitted through these points to form candidate strokes, which are subsequently merged and simplified. 
  Finally, we order both points and stroke to obtain temporally sequential  3D sketches.}
  \label{fig:synthetic}
\end{figure*}

\begin{table}[]
        \caption{{\bf 3D Sketch Datasets.} \textsc{VRsketch2shape} is the first open-access collection that spans multiple object categories and includes both synthetic and real VR sketches. $\text{CD}$ measures the asymmetric Chamfer distance between the real sketch and shapes.}
    \label{tab:dataset}
    \small 
\begin{tabular}{l@{}c@{\;}c@{\;}c@{\;}c@{\;}c} \toprule
     & open- & cate- &  number of & CD $\times 1000$ \\
     & access& gories & sketches & Sk $\mapsto$ Sh \\ \midrule
3DVRChair \cite{luo2021data} & \yes & 1 & 1005 real & 55.6\\ 
KO3D+ \cite{chen2024rapid} & \no & 6 & 4,200 real & -\\
VRSS \cite{gu2025vrsketch2gaussian} & \no & 55 & 2097 real & -
\\ \greyrule
\makecell[c]{\bf \textsc{VRSketch2shape}\\ \bf  (ours)} & \yes & 4 &  \makecell{20,838 synthetic \\ + 900 real} & 5.5\\\bottomrule
\end{tabular}

    \centering
\end{table}

\subsection{VR Sketch Synthesis and Representation}
\label{sec:related:synthetic}
%
\paragraph{Classical 3D Sketch Synthesis.}
Rendering sketches of 3D shapes is a long-standing problem in computer graphics and vision.
Classical methods primarily aim at perceptual realism~\cite{lee2007line} or stylistic expressiveness~\cite{winkenbach1994pen,hahnlein2022cad2sketch}.
In contrast, our goal is not to produce visually realistic sketches, but to extract structurally faithful VR lines that can serve as supervision for the inverse sketch-to-shape problem.
Consequently, design choices in our pipeline are guided by the performance of downstream models trained on the resulting synthetic sketches rather than by their visual quality.

Related lines of work study procedural CAD~\cite{khan2024text2cad} or boundary-representation models~\cite{xu2024brepgen}, where shapes are defined through parametric modeling operations (\eg, extrusion, revolution) applied to geometric primitives and curves (\eg, cylinders, splines).
While such representations also encode sequential geometric structure, they assume structured CAD inputs, whereas our setting targets informal VR sketches produced during early-stage ideation.

\paragraph{Deep Sketch Synthesis.}
Early work on sketch generation focused on 2D sketches, using image-to-image translation networks~\cite{li2019photo,song2018learning}, VAEs~\cite{ha2017neural}, or auto-regressive models~\cite{bhunia2022doodleformer}.
These approaches typically required large datasets of human sketches for supervision.
To mitigate this limitation, subsequent methods directly optimized vectorized representations under the guidance of pretrained vision–language~\cite{vinker2022clipasso,vinker2023clipascene,radford2021learning} or diffusion models~\cite{xing2023diffsketcher,arar2025swiftsketch}.
Other works addressed sketch completion using GANs~\cite{liu2019sketchgan} or transformers~\cite{lin2020sketchbert}.
Building on these advances, recent approaches extend sketch generation to 3D, synthesizing parametric curves from text, single-view, or multi-view images and optimizing their parameters via pretrained image models~\cite{choi20243doodle,wang2025viewcraft3d,zhang2024diff3ds}. In contrast, our synthetic sketch generation pipeline relies purely on geometric heuristics and is entirely training-free.

\paragraph{Sketch Encoding.}
2D sketches are typically processed either as images using convolutional networks~\cite{yu2015sketchanet, choi20243doodle, guillard2021sketch2mesh, zheng2023locally}, or as sequences using recurrent or transformer-based models~\cite{vaswani2017attention} to capture their temporal and structural logic~\cite{ha2017sketchrnn, lin2020sketchbert}.
In contrast, most existing approaches for 3D VR sketches still represent them as unordered point clouds~\cite{luo2021data, chen2024rapid, gu2025vrsketch2gaussian} and apply point-based encoders such as PointNet++~\cite{qi2017pointnet++}, thereby discarding the intrinsic stroke order.
However, VR sketches are inherently sequential, as the drawing order encodes meaningful cues about connectivity, structure, and design intent. In this work, we encode VR sketches directly as ordered sequences of 3D points, allowing our model to exploit both their spatial geometry and the temporal dependencies of the sketching process.

\section{\textsc{VRsketch2shape} Dataset}

We introduce \textsc{VRSketch2Shape}, a dataset of real and synthetic sequential VR sketches.
We first describe the collection of $900$ real VR sketches aligned with ShapeNet models (\cref{sc:real}), and then detail our automatic synthetic sketch generation pipeline (\cref{sc:synthetic}).

\paragraph{Setup.}
We define a VR sketch as a collection of \emph{3D polylines} (or \emph{strokes}), each represented by a temporally ordered sequence of 3D points drawn in a single continuous motion.  
Our dataset preserves both stroke order and point order, providing temporal information often discarded in prior work that treats sketches as unordered point sets.

\begin{figure*}
    \centering
    \resizebox{\linewidth}{!}{
    \input{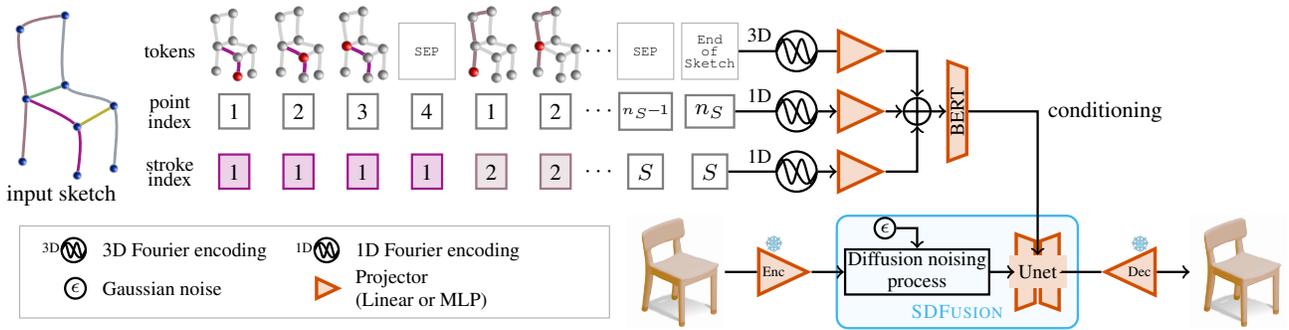}
    }
\caption{{\bf \textsc{VRSketch2Shape} Model.} 
An input VR sketch is tokenized into a sequence of points organized along ordered strokes.
Each 3D point is encoded using 3D Fourier features and an MLP, while stroke and point indices are encoded with 1D Fourier features followed by a linear projection. 
The resulting embeddings are summed and passed through a lightweight BERT encoder. 
The encoded token sequence is then used to condition \textsc{SDFusion}, a diffusion-based 3D shape generation model.}
    \label{fig:pipeline}
\end{figure*}

\subsection{Real Data Collection}
\label{sc:real}

We built a Unity-based VR interface that allows participants to visualize and sketch directly over a reference 3D model.  
A key challenge in VR sketching is depth ambiguity: without guidance, users often draw strokes that float in front of or behind the surface, resulting in imprecise and hard-to-use annotations.  
To mitigate this issue, we implemented a \emph{surface-snapping} mechanism that projects each drawn point onto the underlying 3D model along the shortest path, ensuring geometric alignment between the sketch and the object.  
As shown in \cref{tab:dataset}, this snapping step produces sketches that are substantially more faithful to the input shapes—as measured by asymmetric sketch-to-shape Chamfer distance, leading to a more reliable and less noisy benchmark for sketch-conditioned generation.

We recruited 15 participants, who completed a short tutorial before sketching multiple objects from ShapeNet.  
The resulting dataset contains $900$ sketches across four categories: $300$ chairs, $200$ tables, $200$ cabinets, and $200$ airplanes.  
With an average of 15 minutes per sketch, data collection required approximately $225$ person-hours.


\subsection{Synthetic Sketch Generation.}
\label{sc:synthetic}
Because collecting real sketches is expensive and tedious, we propose a fully automatic pipeline to generate synthetic VR sketches from 3D meshes (\cref{fig:synthetic}), producing $20{,}838$ samples in roughly {10} hours on a standard work station.

\paragraph{Extracting Salient Points.}
We begin by uniformly sampling 2048 points on the surface of the input mesh.
Sketches typically emphasize visually prominent geometric features such as edges, corners, and holes. We emphasize regions of high curvature and structural significance by extracting the \emph{salient point cloud} using Sharp Edge Sampling (SES)~\cite{chen2025dora} and a curvature threshold of 15. 

\paragraph{Recovering Strokes.}
We then fit Bézier splines to the salient point cloud using EMAP~\cite{Li2024CVPRneuraledge} with a maximum degree of $2$ and minimum segment length of $12$.  
The points along each spline form the individual strokes.  
Next, we apply a culling stage inspired by the approach proposed by Liu \etal \cite{liu2018strokeaggregator}. We first remove redundant points in near-linear segments with a cosine distance threshold of $0.04$. Finally, we merge strokes whose endpoints lie within a threshold of 2\% of the normalized shape size. 

\paragraph{Ordering Strokes.}
To approximate human drawing order, we connect stroke endpoints based on spatial proximity and perform a depth-first traversal of the resulting connectivity graph.  
We introduce stochasticity by skipping  nearest connections with a probability of $10\%$, yielding coherent yet varied stroke sequences.

\paragraph{Dataset Structure.}
The proposed \textsc{VRSketch2Shape} dataset is organized into four parts:
\begin{compactitem}
    \item \textbf{Synthetic training set:} $20{,}838$ sketch–shape pairs generated with our automatic pipeline.
    \item \textbf{Real fine-tuning set:} 500 sketch–shape pairs (200 chairs and 100 per other category) for domain adaptation from synthetic to real sketches.
    \item \textbf{Real evaluation set:} 400 pairs (100 per category) reserved for final quantitative and qualitative evaluation.
\end{compactitem}

\section{\textsc{VRsketch2shape} Model}
%

%
%
In this section, we present our model for generating 3D shapes from sequential VR sketches. The overview is shown in \cref{fig:pipeline}.
We first describe our sketch encoder based on \textsc{BERT} (\cref{sec:bert}), then explain how it interfaces with the diffusion-based shape generator \textsc{SDFusion} (\cref{sec:sdiffusion}).  
\begin{figure*}[ht]
    \centering
        \resizebox{\linewidth}{!}{
\begin{tabular}{l@{\;}c@{\;}c@{\;}c@{\;}c@{\;}c@{\;}c@{\;}c@{\!\!}r}
\rotatebox{90}{\small \bf  \quad Real sketch }&
\includegraphics[height=0.115\textheight]{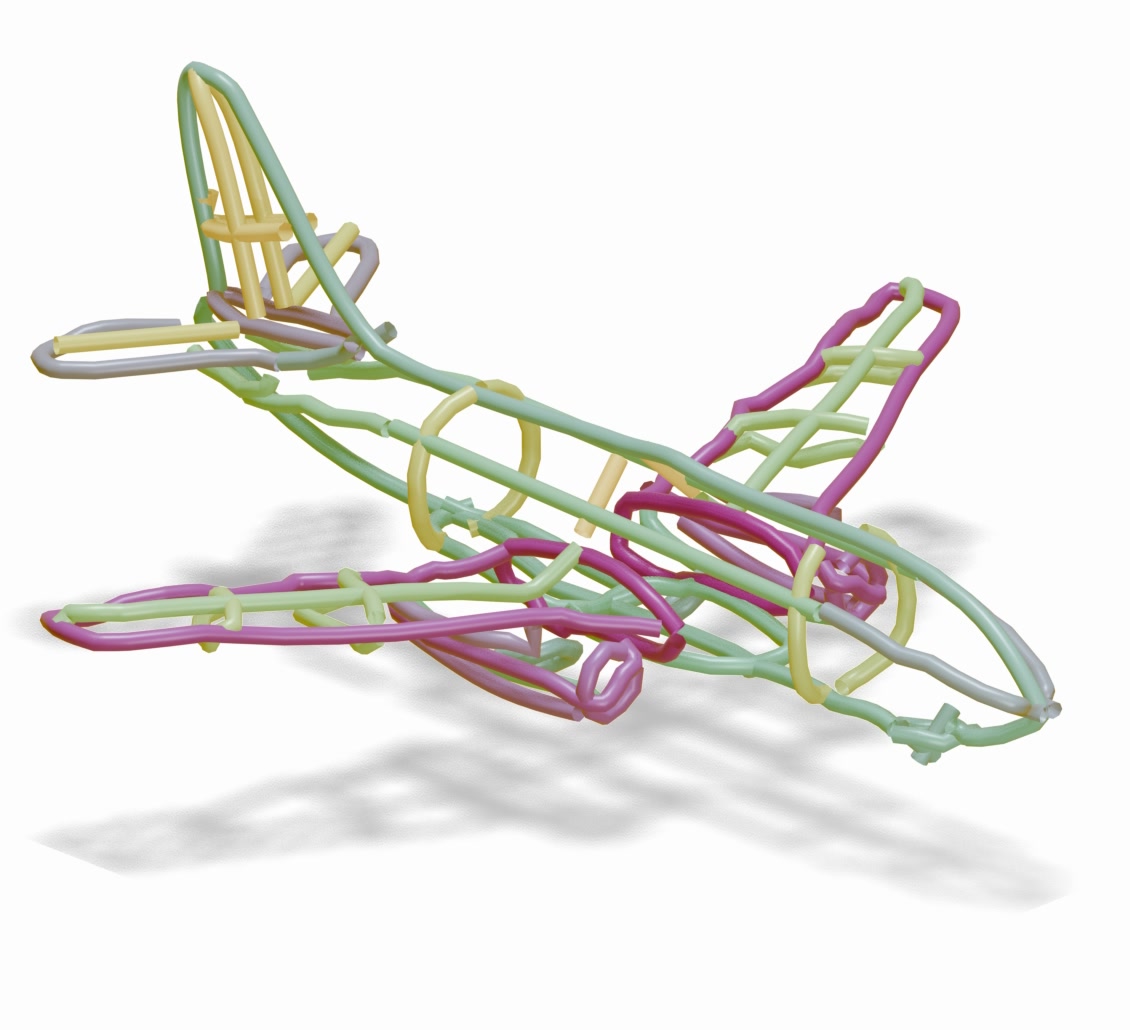} &
\includegraphics[height=0.115\textheight]{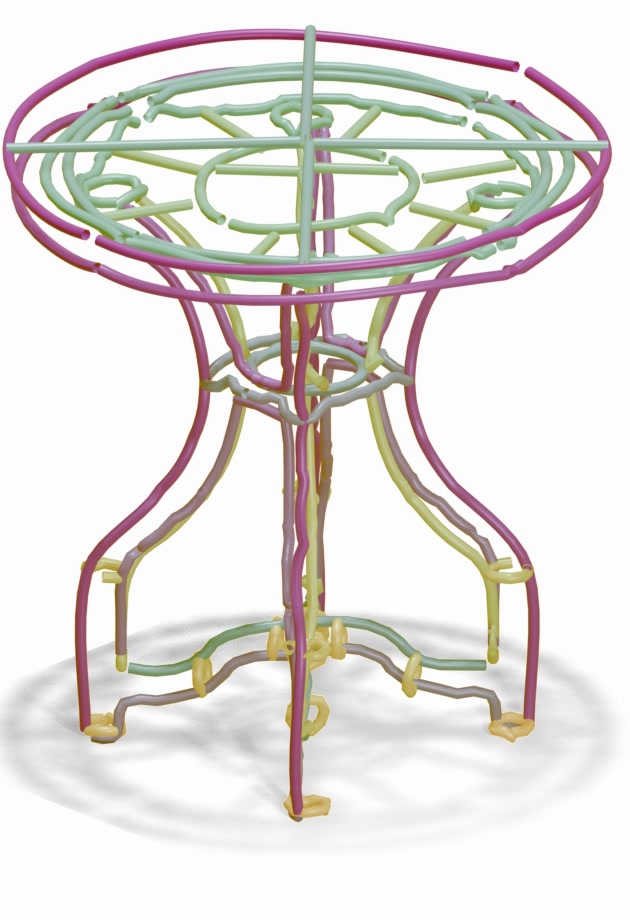} &
\includegraphics[height=0.115\textheight]{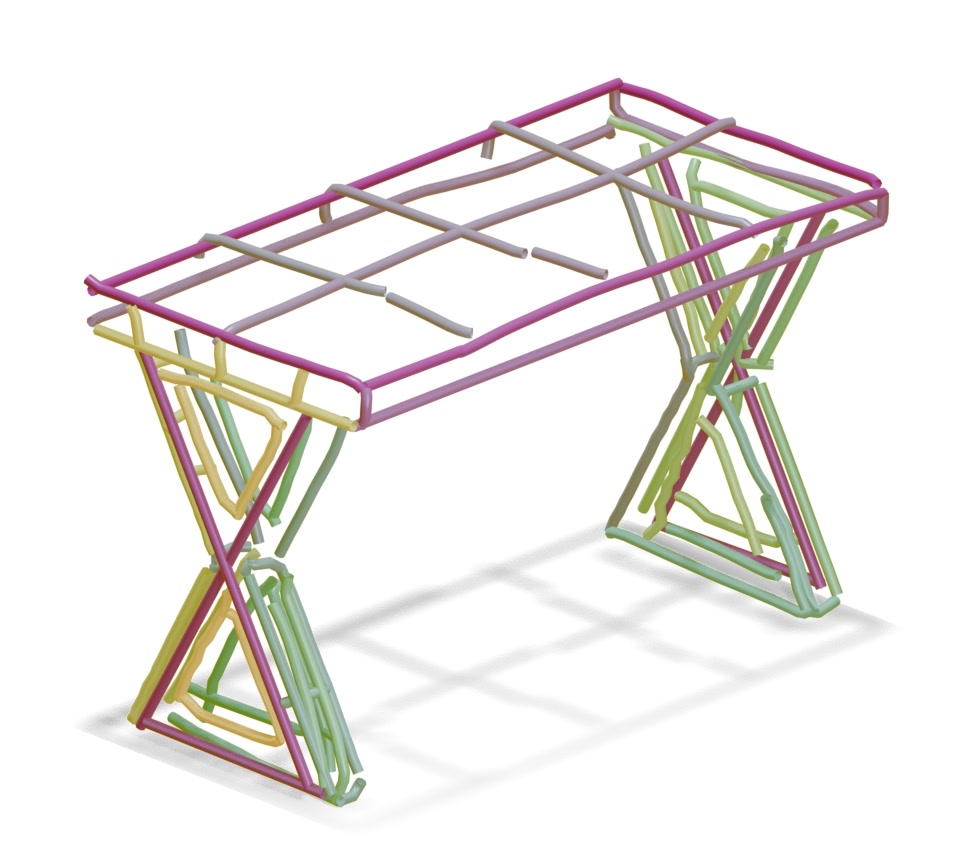} 
&
\includegraphics[height=0.115\textheight]{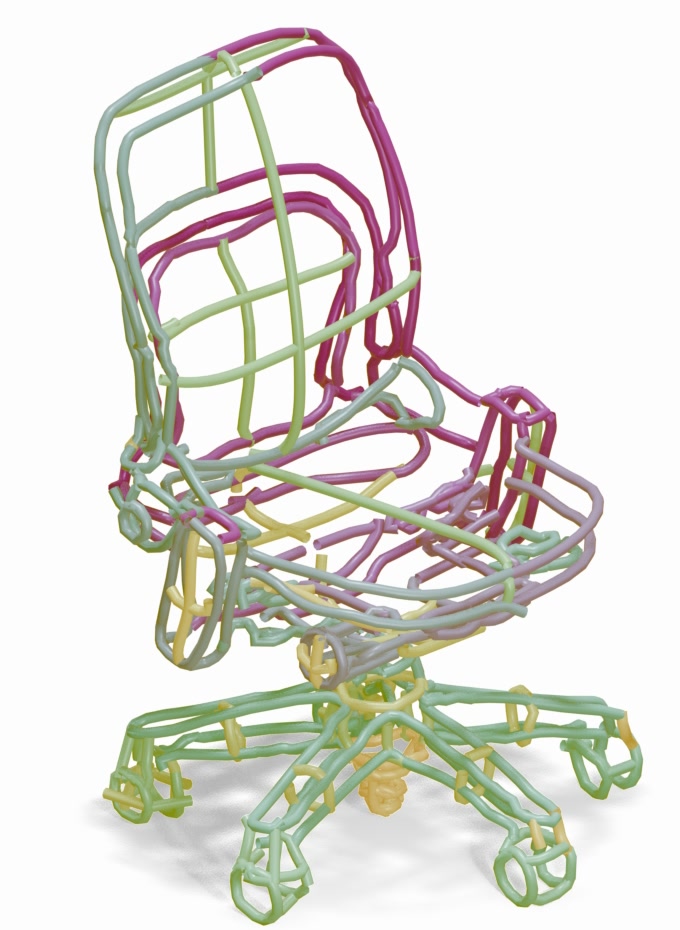} 
&
\includegraphics[height=0.115\textheight]{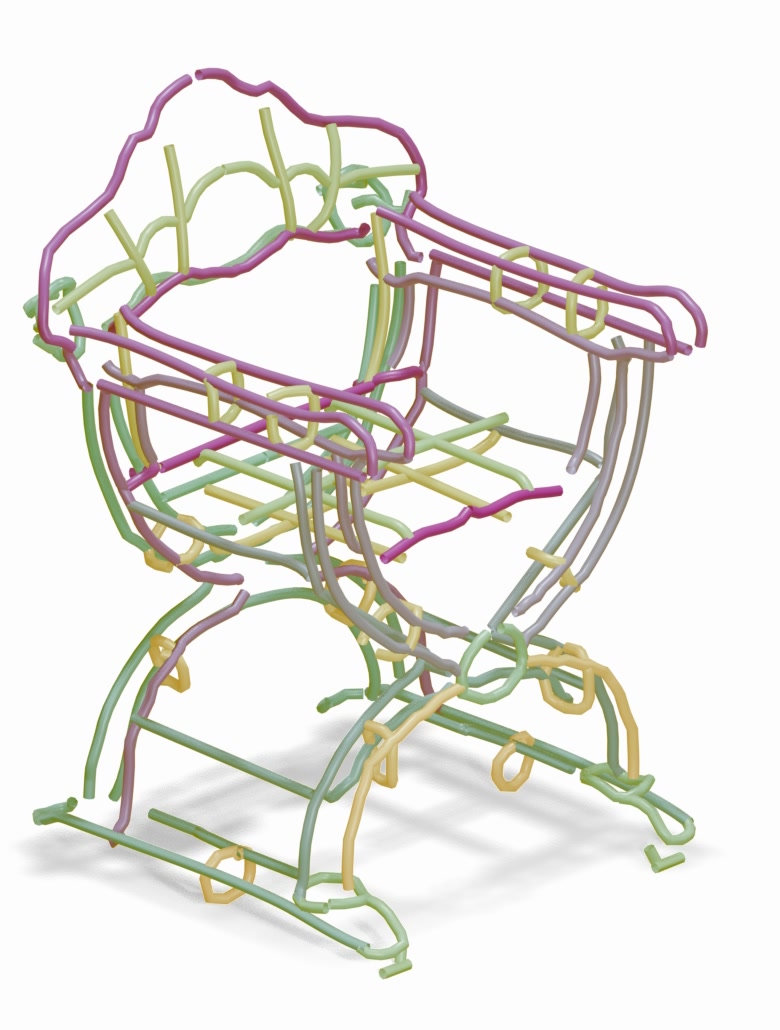}
&
\includegraphics[height=0.115\textheight]{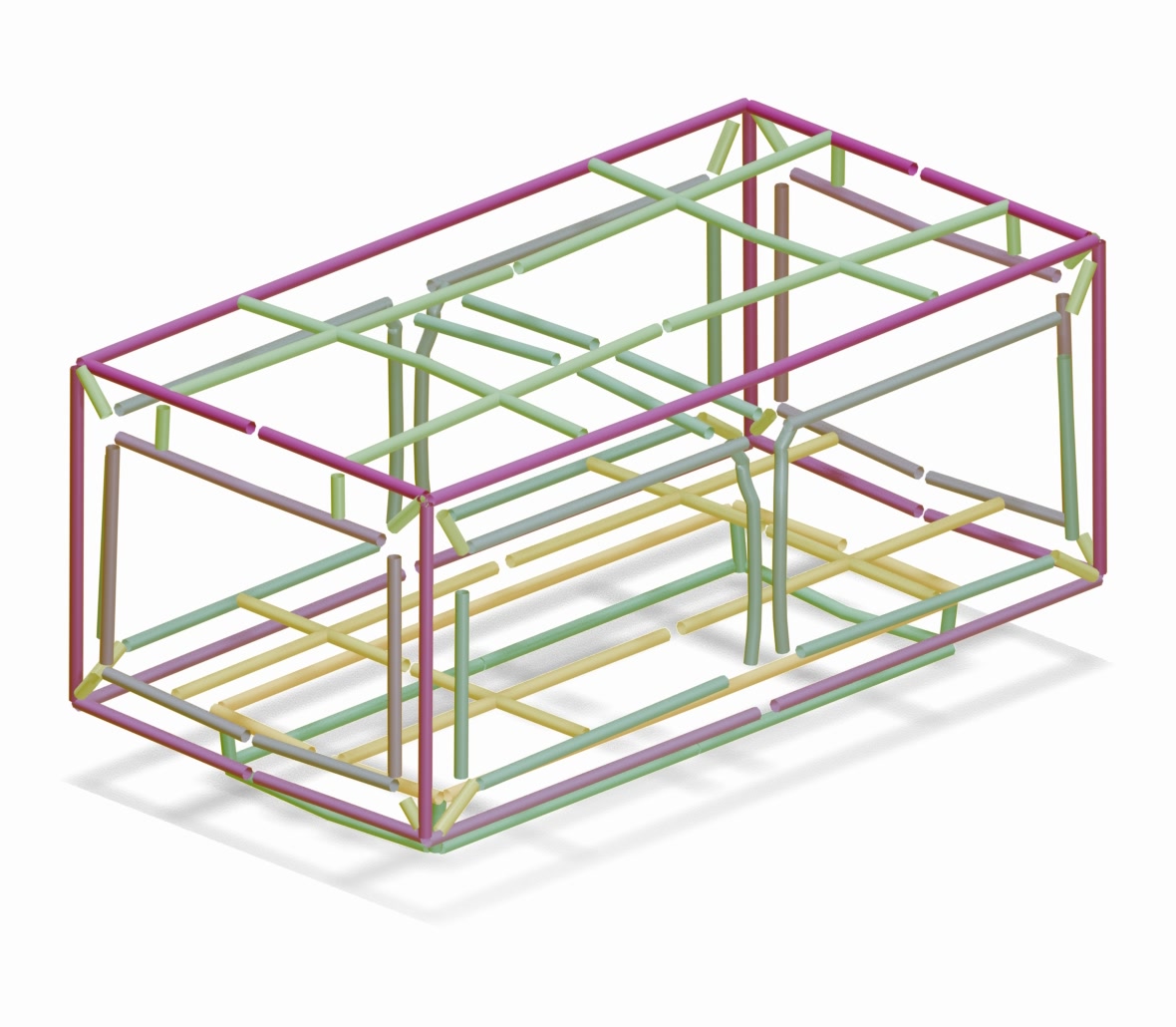}
&
\includegraphics[height=0.115\textheight]{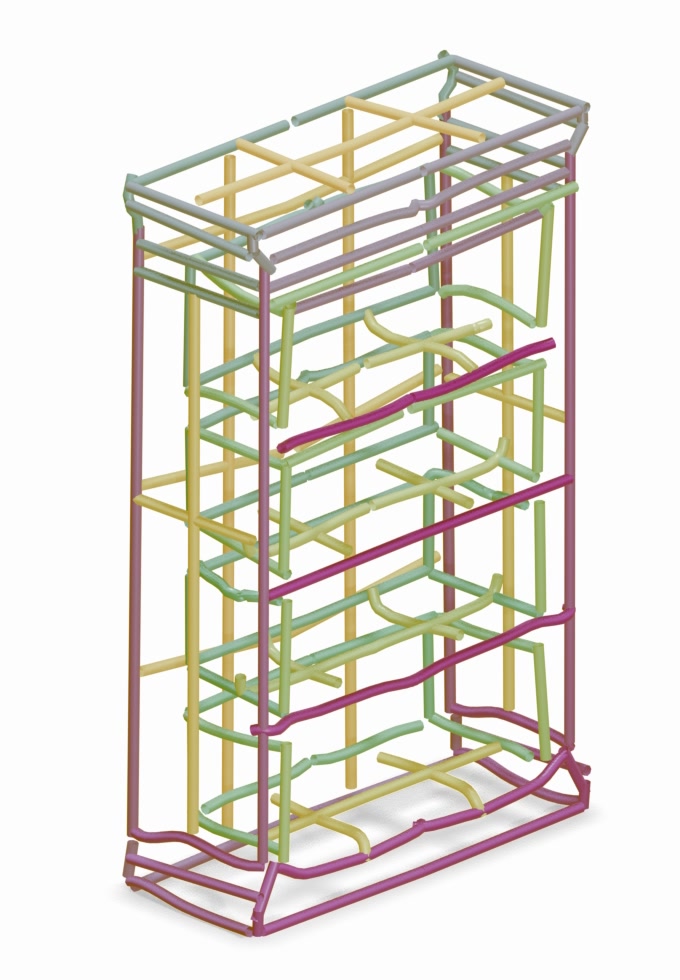}
&
\begin{tikzpicture}[baseline=-0.040\textheight]
      \node[draw=none, inner sep=0pt] (jet) at (0,0) {\rotatebox{0}{\def\colormapheight{50}\rotatebox{0}{\input{figures/jetcolorbar}}}};
    \node[above=-0mm of jet]{\scriptsize \!start};
    \node[below=-9.0mm of jet.south]{\scriptsize \!end};
\end{tikzpicture}
\\[-4mm]
\rotatebox{90}{\small \bf \qquad  GT shape} &
\includegraphics[height=0.115\textheight]{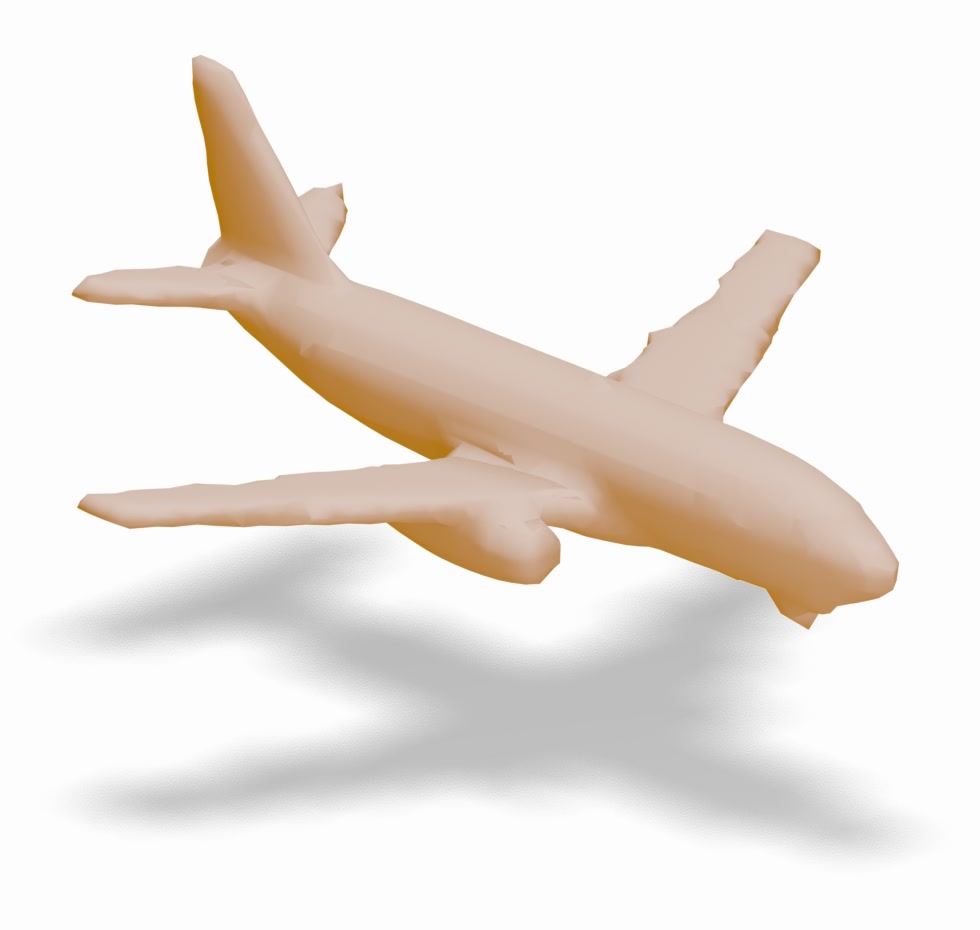} &
\includegraphics[height=0.115\textheight]{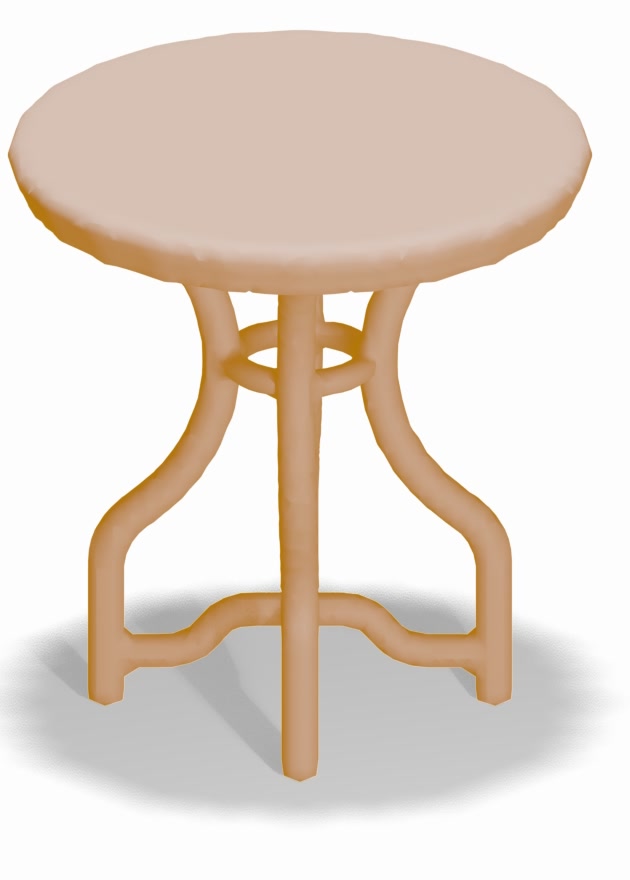} &
\includegraphics[height=0.115\textheight]{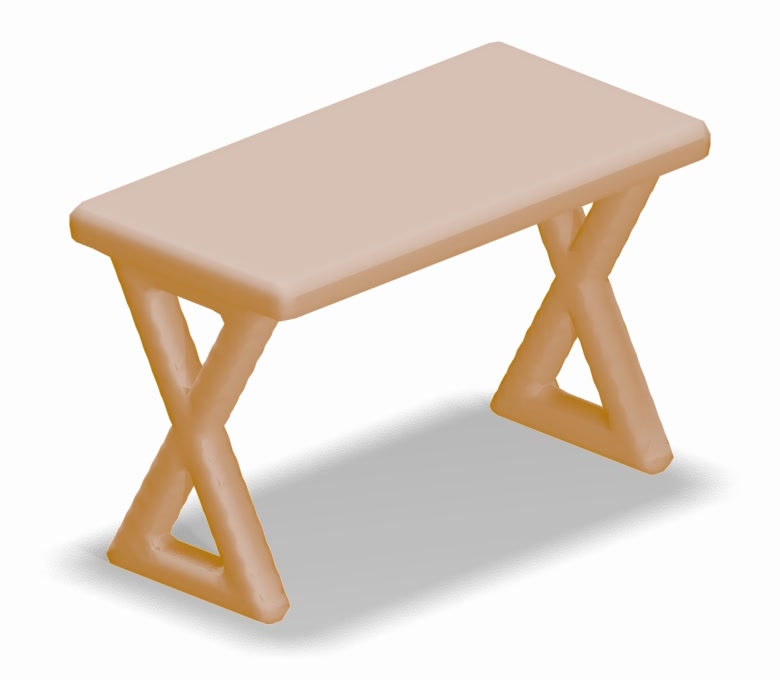} 
&
\includegraphics[height=0.115\textheight]{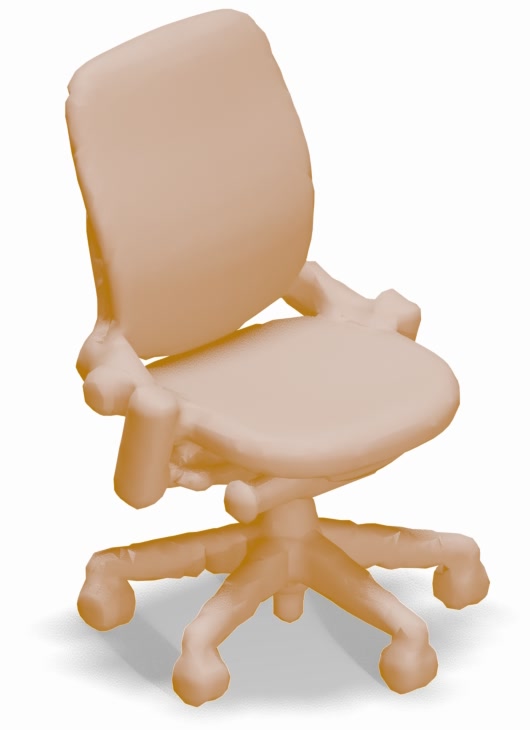}
&
\includegraphics[height=0.115\textheight]{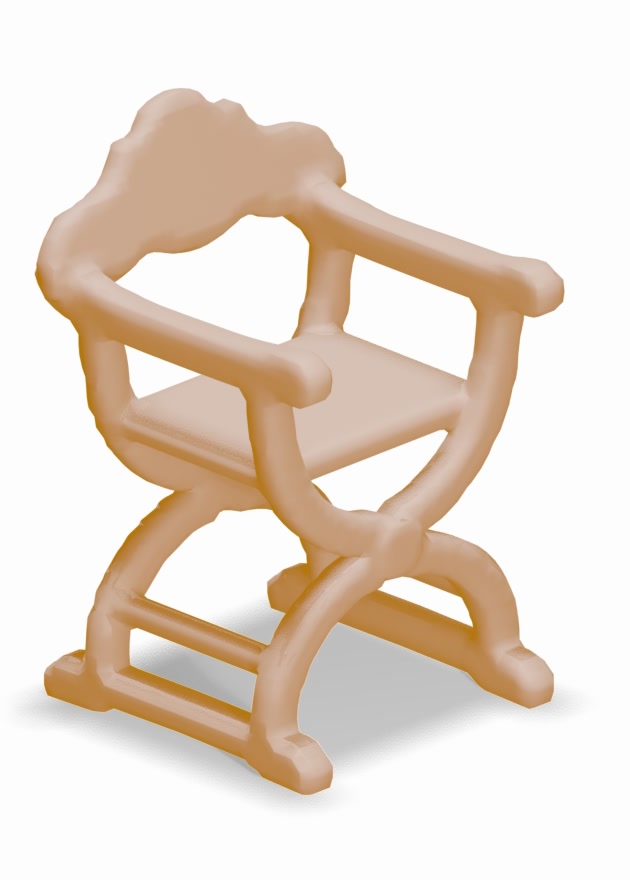}
&
\includegraphics[height=0.115\textheight]{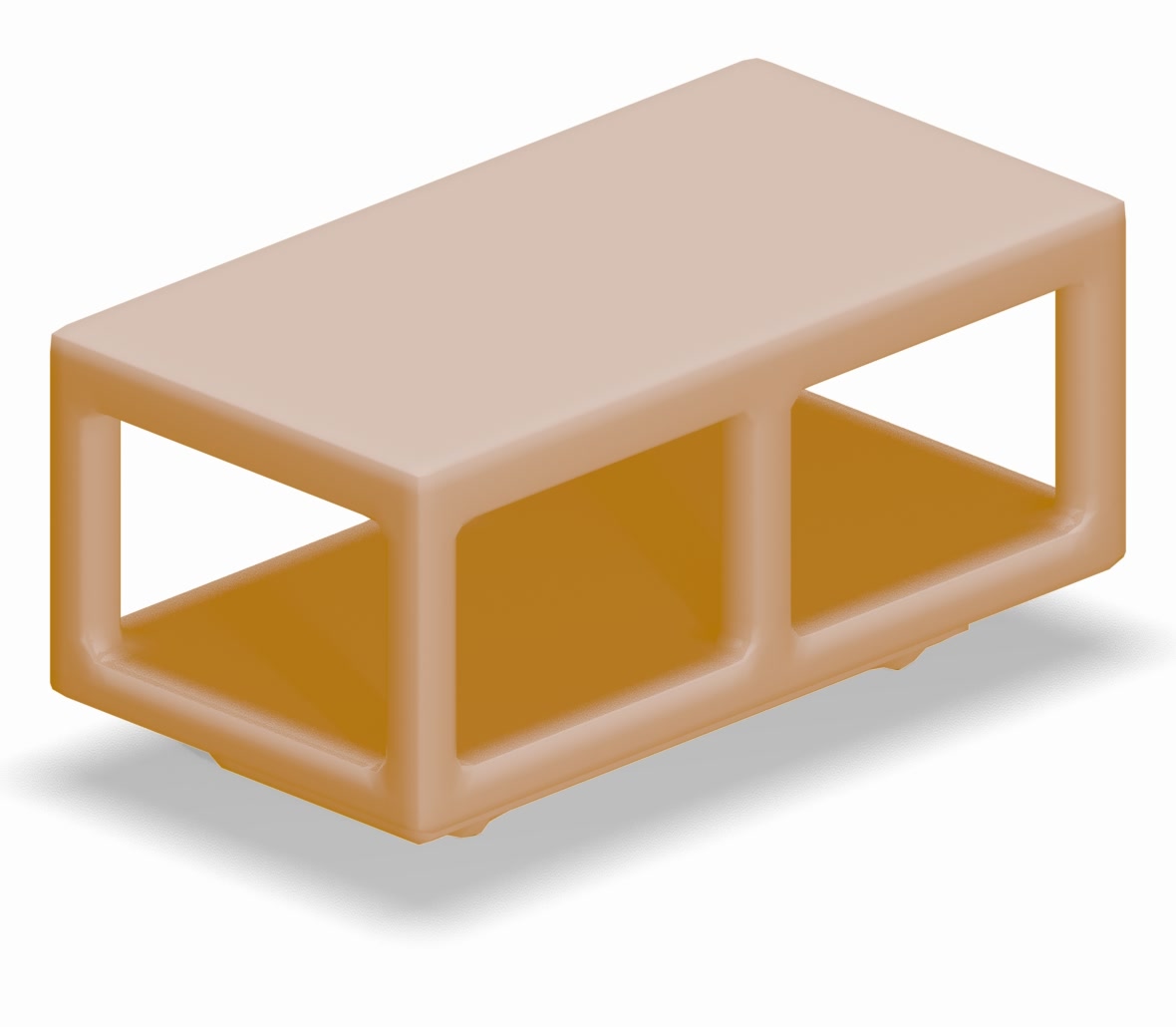}
&
\includegraphics[height=0.115\textheight]{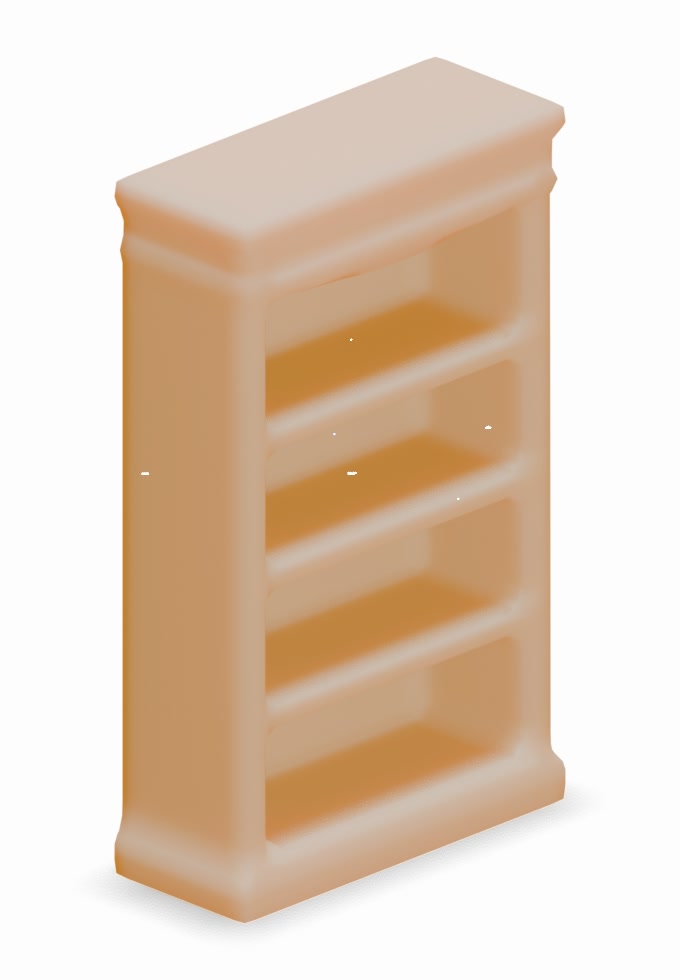}
\\
[-1mm]
\rotatebox{90}{\small \bf \;\; Our prediction} &
\includegraphics[height=0.115\textheight]{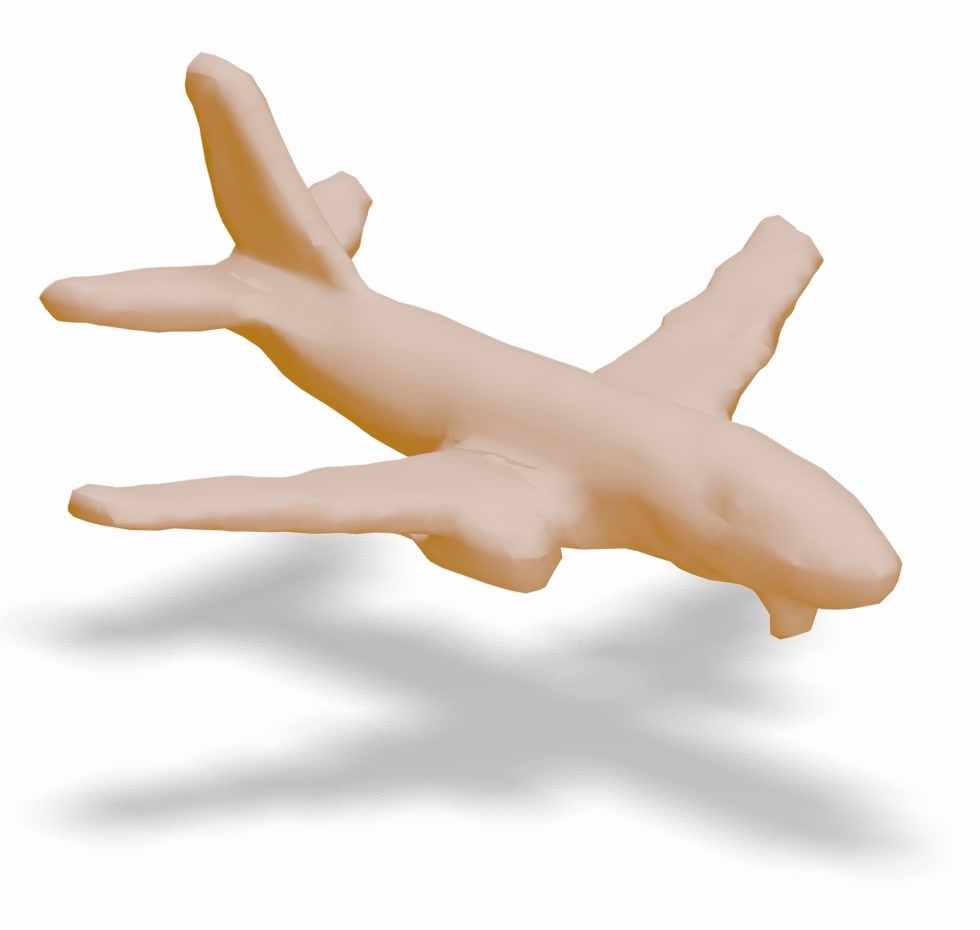} &
\includegraphics[height=0.115\textheight]{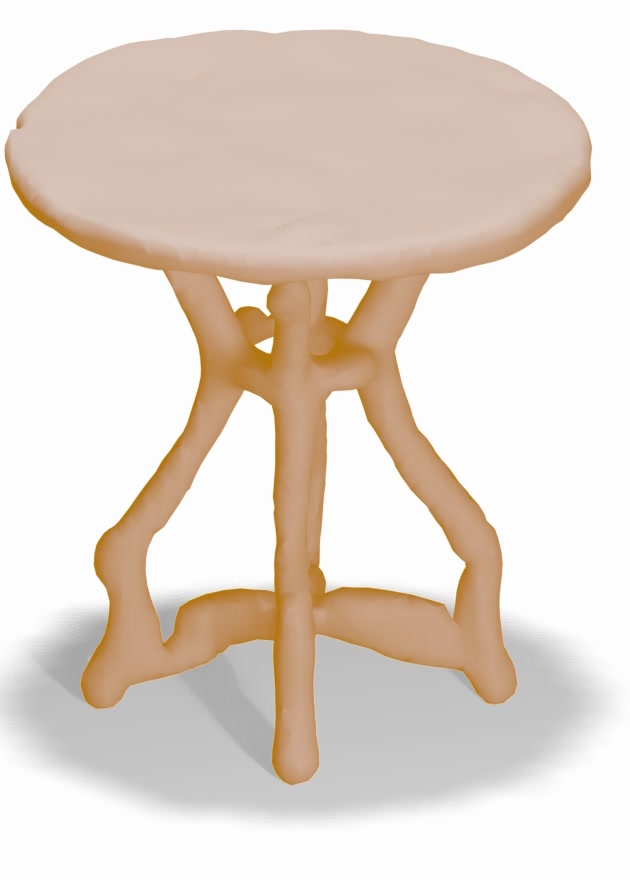} &
\includegraphics[height=0.115\textheight]{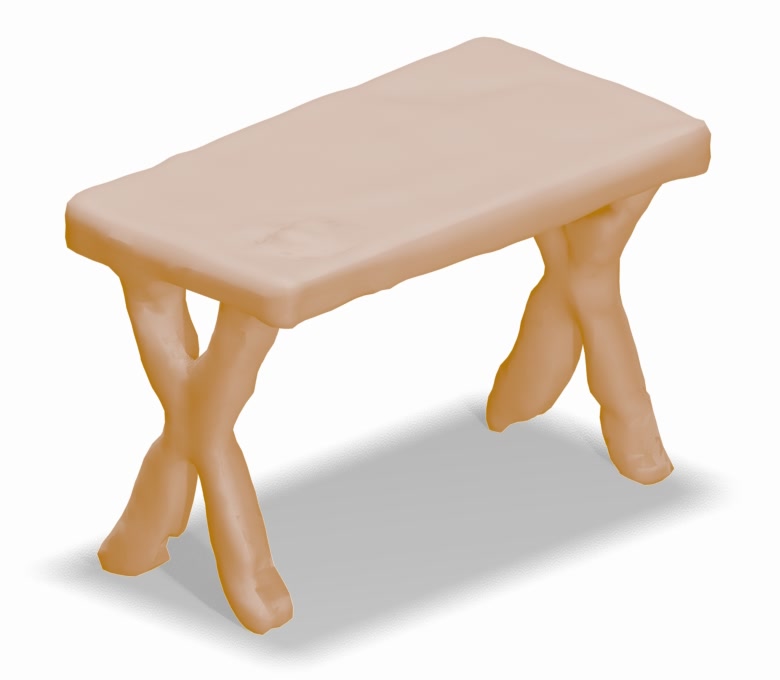}
&
\includegraphics[height=0.115\textheight]{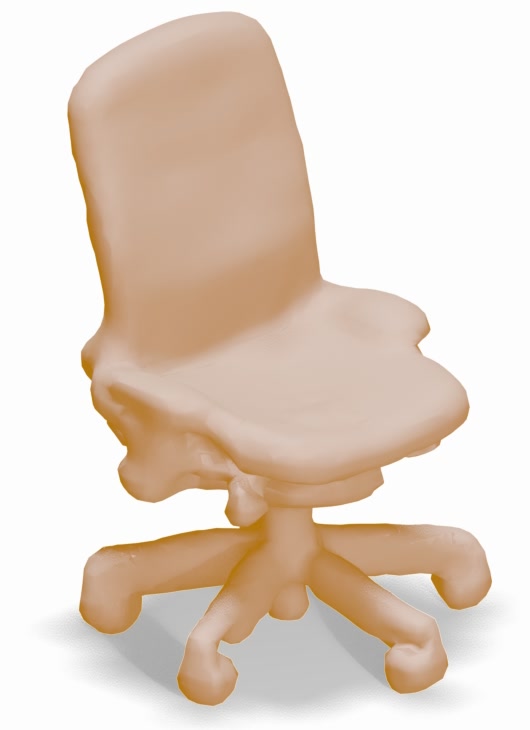} 
&
\includegraphics[height=0.115\textheight]{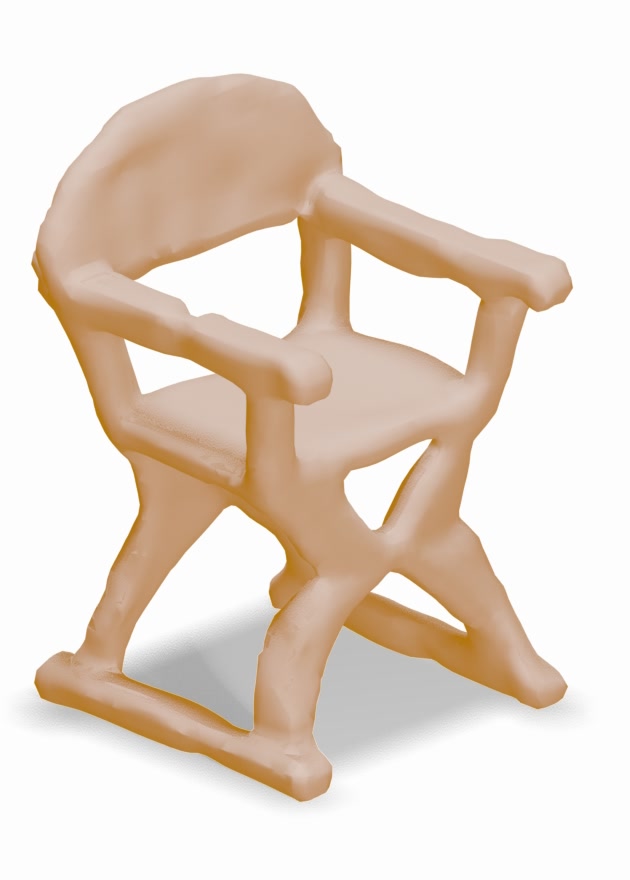}
&
\includegraphics[height=0.115\textheight]{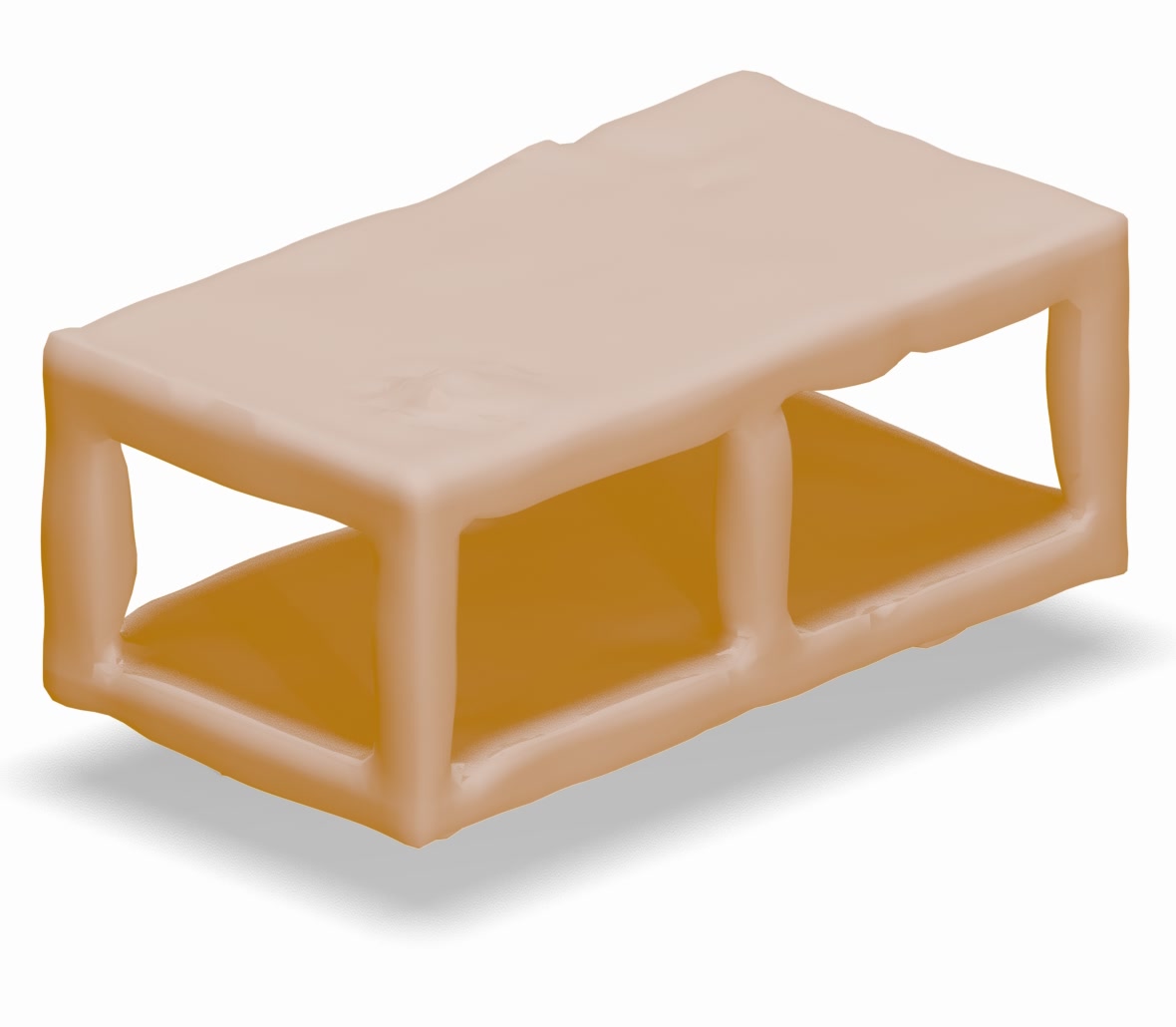}
&
\includegraphics[height=0.115\textheight]{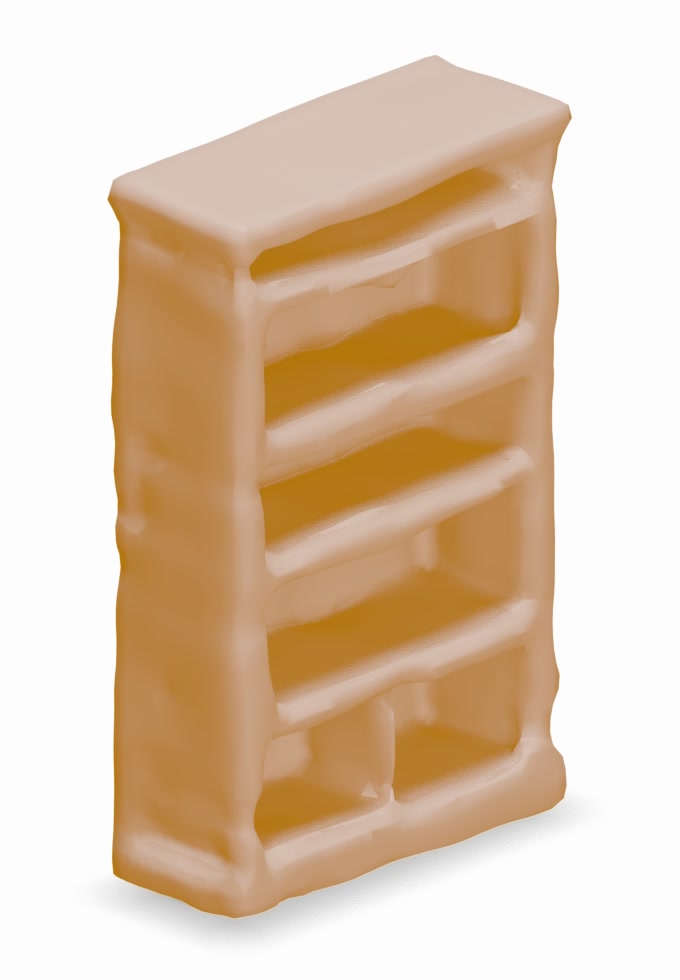}
\\
[-3mm]
\rotatebox{90}{\small \bf Luo's prediction} &
\includegraphics[height=0.115\textheight]{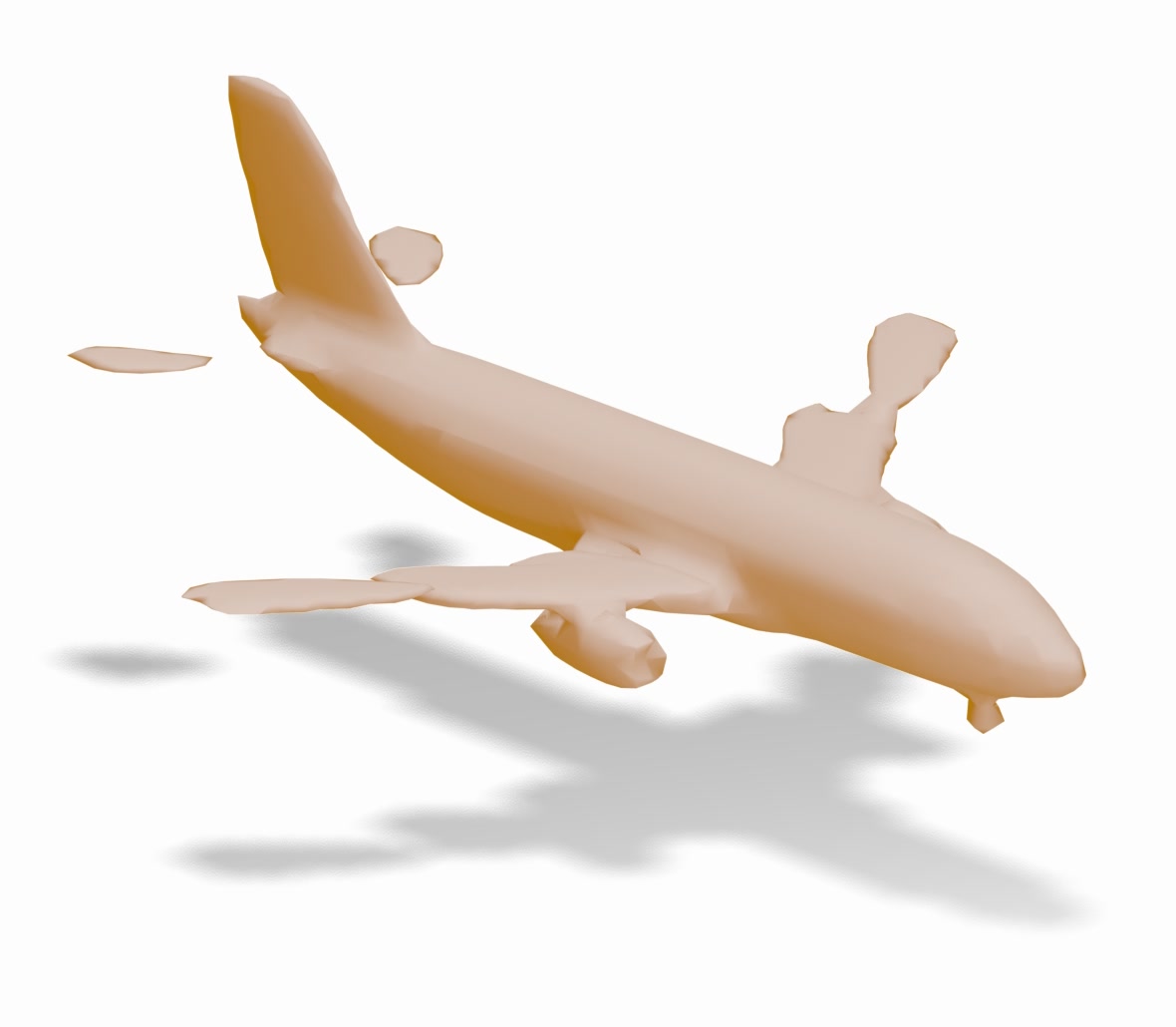} &
\includegraphics[height=0.115\textheight]{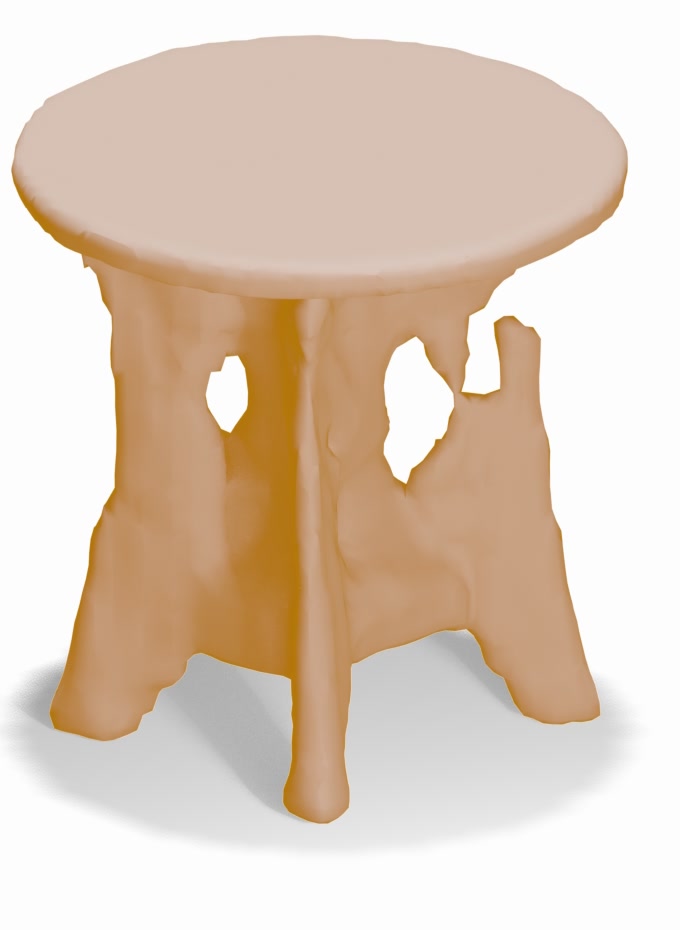} &
\includegraphics[height=0.115\textheight]{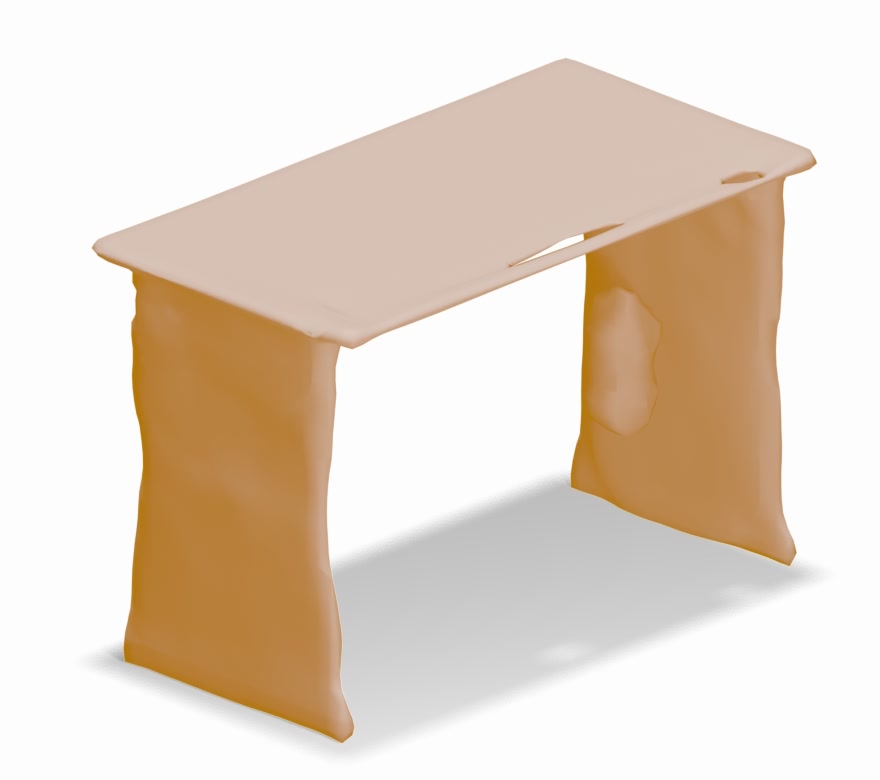}
&
\includegraphics[height=0.115\textheight]{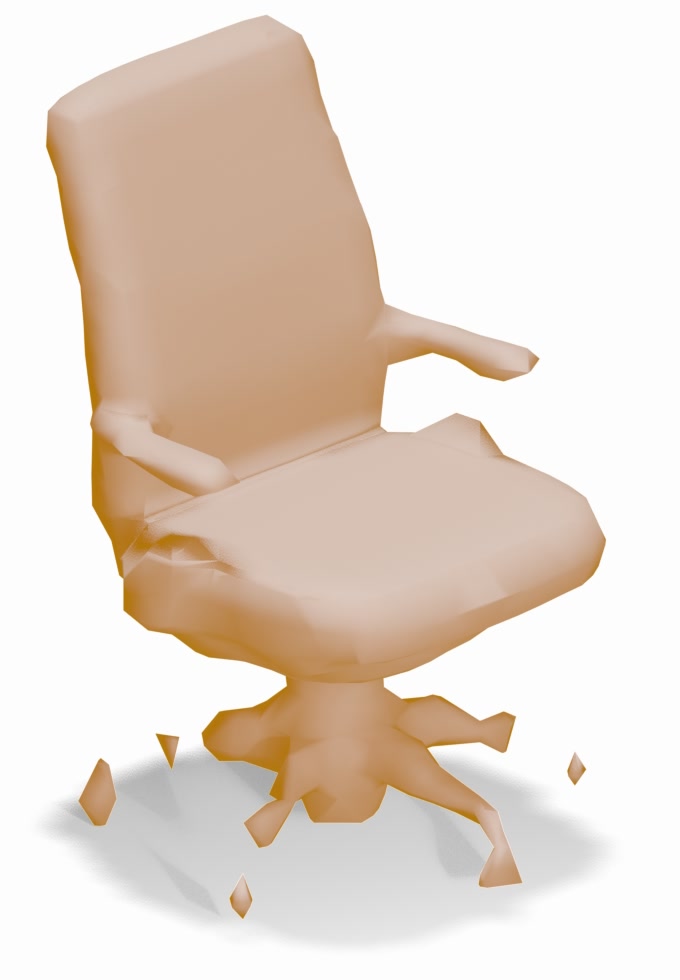} 
&
\includegraphics[height=0.115\textheight]{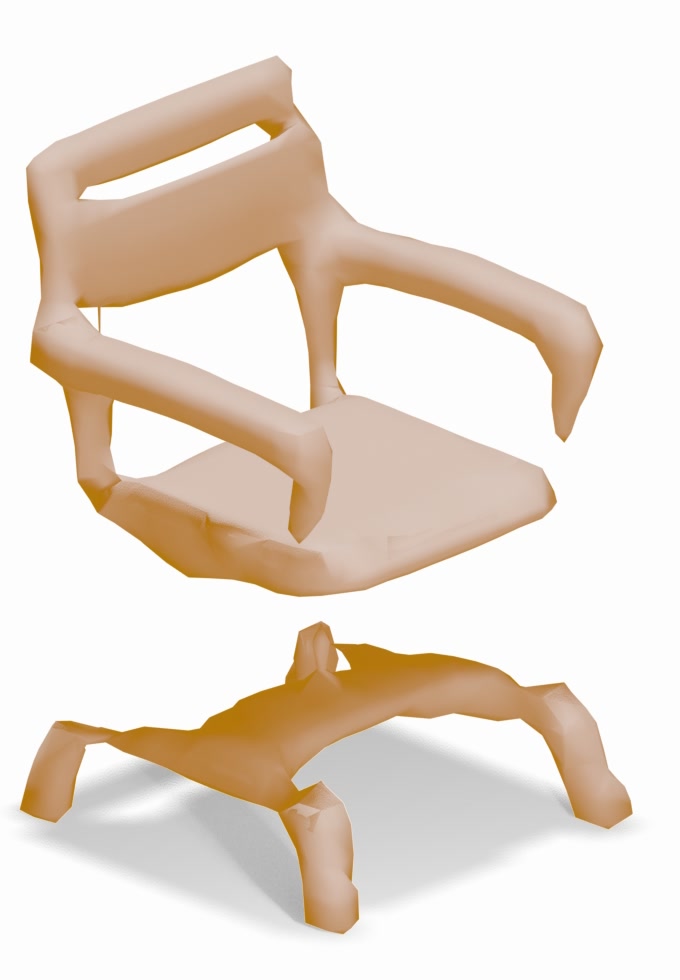}
&
\includegraphics[height=0.115\textheight]{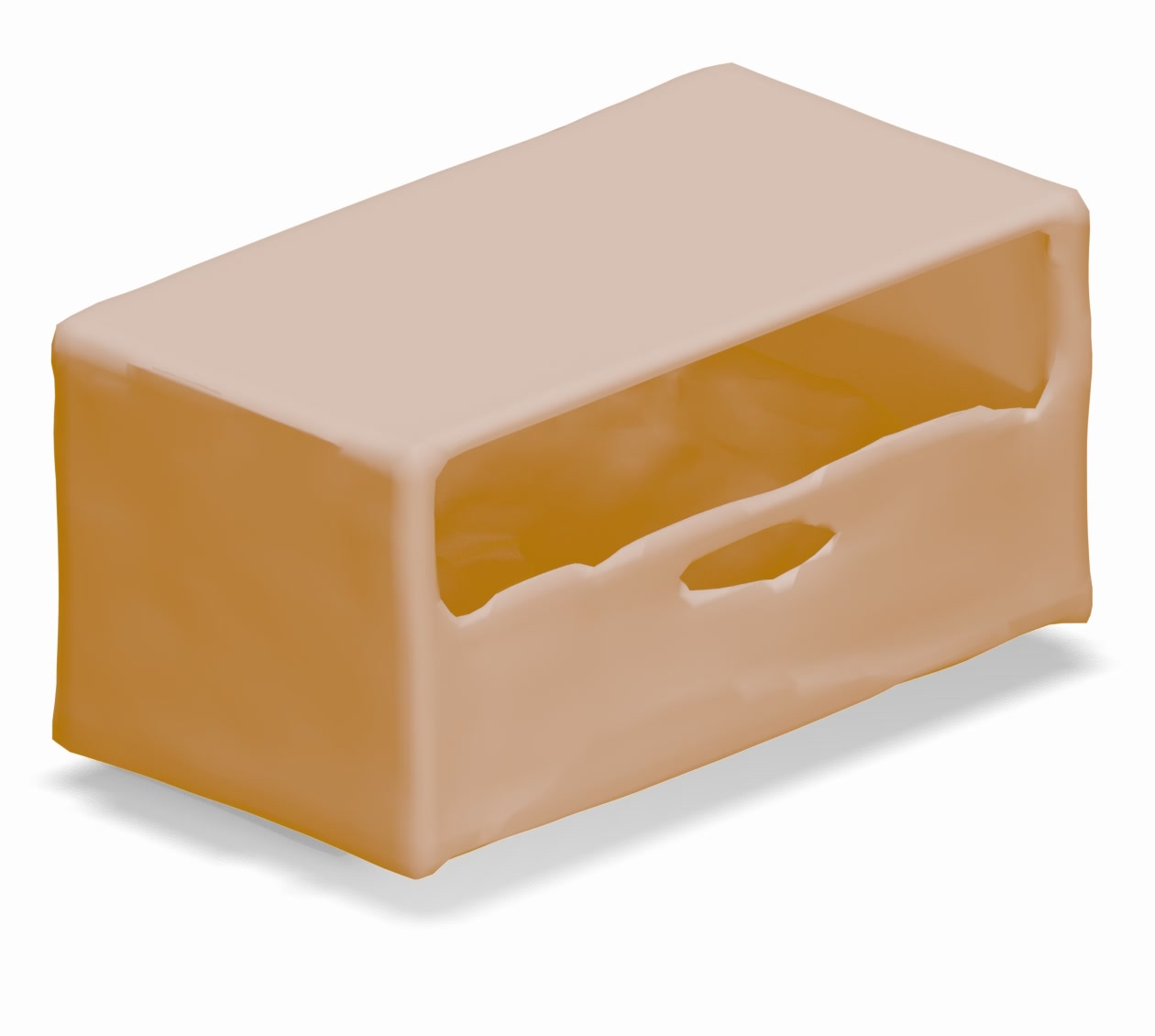}
&
\includegraphics[height=0.115\textheight]{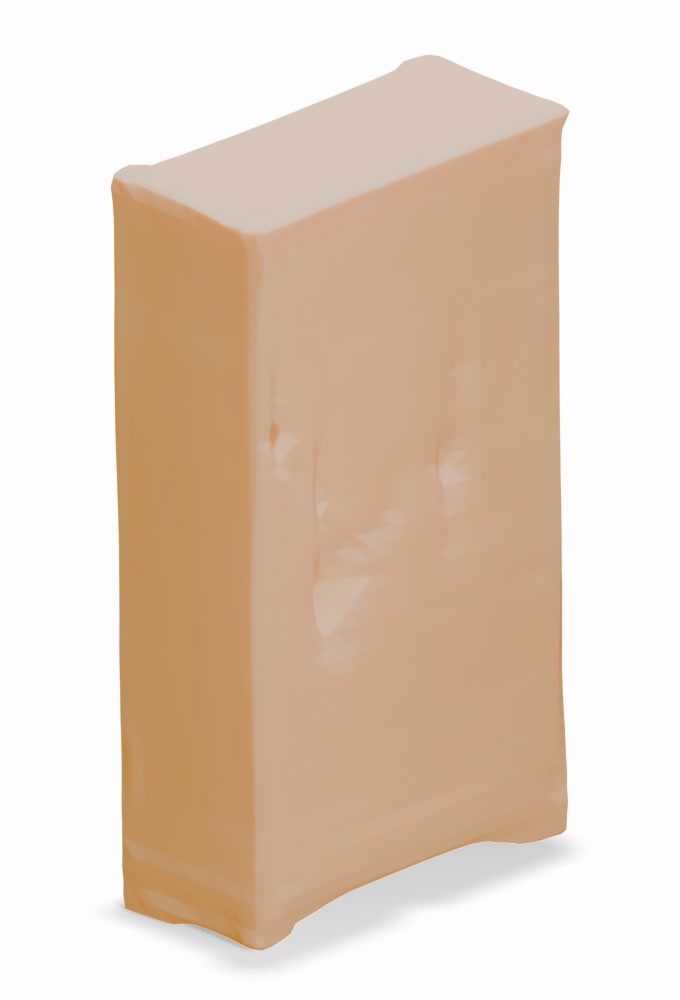}
\end{tabular}
} 
 \if 1 0
\begin{tabular}{c@{\;}c@{\;}c@{\;}c}
Real sketch  & GT shape &  Our prediction & Luo's \cite{luo2021data} predicted shape\\
\includegraphics[width=.23\linewidth]{images/qualitative/qualitative_airplane_sketch_clipped.jpg}
&
\includegraphics[width=.23\linewidth]{images/qualitative/qualitative_airplane_gt_clipped.jpg}
&
\includegraphics[width=.23\linewidth]{images/qualitative/qualitative_airplane_ours_clipped.jpg}
&
\includegraphics[width=.23\linewidth]{images/qualitative/qualitative_airplane_luo_clipped.jpg}
\\
\includegraphics[width=.23\linewidth]{images/qualitative/qualitative_table_sketch_clipped.jpg}
&
\includegraphics[width=.23\linewidth]{images/qualitative/qualitative_table_gt_clipped.jpg}
&
\includegraphics[width=.23\linewidth]{images/qualitative/qualitative_table_ours_clipped.jpg}
&
\includegraphics[width=.23\linewidth]{images/qualitative/qualitative_table_luo_clipped.jpg}
\\
\includegraphics[width=.23\linewidth]{images/qualitative/qualitative_cabinet_sketch_clipped.jpg}
&
\includegraphics[width=.23\linewidth]{images/qualitative/qualitative_cabinet_gt_clipped.jpg}
&
\includegraphics[width=.23\linewidth]{images/qualitative/qualitative_cabinet_ours_clipped.jpg}
&
\includegraphics[width=.23\linewidth]{images/qualitative/qualitative_cabinet_luo_clipped.jpg}
\\
\includegraphics[width=.23\linewidth]{example-image}
&
\includegraphics[width=.23\linewidth]{example-image}
&
\includegraphics[width=.23\linewidth]{example-image}
&
\includegraphics[width=.23\linewidth]{example-image}
\end{tabular}
\fi

%

%
\caption{{\bf Qualitative Illustrations.}
Comparison between our method and Luo~\etal~\cite{luo20233d} on the real test set of \textsc{VRSketch2Shape}.
Both models are pretrained on the same synthetic sketches and fine-tuned on real data.
Our approach generates shapes that are more detailed, structurally accurate, and topologically faithful to the target geometry.}
\label{fig:quali}
\end{figure*}
\subsection{Encoding 3D Sketches}
\label{sec:bert}
We treat each sketch as a sequence and encode it with a transformer-based architecture inspired by BERT~\cite{devlin2019bert}: the sketch is tokenized, embedded, enriched with positional encodings, and processed by several transformer blocks.

\paragraph{Sketch tokenization.}
A VR sketch consists of an ordered set of strokes, each formed by a sequence of 3D points.
We introduce two special tokens: \SEP marks the end of a stroke, and \EOS~(End of Sketch) marks the end of the entire sketch.
A sketch $\cS$ with $S$ strokes is thus tokenized as:
\begin{align}
\!\cS\!=\!
\Big[
p^1_1,\cdots\!,p^1_{n_1},\SEP,\cdots\!,p^S_{1}, \cdots\!,p^S_{n_S},\SEP,\EOS
\Big]~,
\end{align}
where $n_s$ is the number of points in stroke $s$, and each point $p=(x,y,z)\in[0,1]^3$ stores normalized 3D coordinates.
We denote by $p_i^s$ the $i$-th point of the $s$-th stroke.

\paragraph{Spatial embedding.}
Following Mildenhall \etal~\cite{mildenhall2021nerf}, we map each coordinate of $p=(x,y,z)$ through a Fourier feature encoding, known to  better capture high-frequency geometric details:
\begin{align}
\Phi_{\text{spa}}(t)
=
\big[
\sin(2^\ell \pi t),
\cos(2^\ell \pi t)
\big]_{\ell=0}^{L-1}
\in \bR^{2L},
\end{align}
where $L$ is the number of frequency bands and  $[\cdot,\cdot]$ denotes the feature-wise concatenation operator.
We concatenate the encoded coordinates and map them to the model dimension $D$ with $\MLP_{\text{spa}}\!:\bR^{6L}\!\mapsto\!\bR^D$:
\begin{align}
E_{\text{spa}}(p)
=
\MLP_{\text{spa}}\!
\left(
\big[
\Phi_{\text{spa}}(x),
\Phi_{\text{spa}}(y),
\Phi_{\text{spa}}(z)
\big]
\right)~.
\end{align}
The embeddings of the separator tokens
$E_{\text{spa}}(\SEP)$ and $E_{\text{spa}}(\EOS)$ are learned as free parameters in $\bR^D$.

\begin{table*}[t]
    \centering
    \begin{tabular}{c@{}c}
         \begin{minipage}{0.25\linewidth}
         \vspace{2mm}
             \caption{{\bf Quantitative results.} 
    Comparison of sketch-to-shape generation methods on the public \textsc{3DVRChair} dataset and our proposed \textsc{VRSketch2Shape} dataset. 
    $\star$~use 2D renders of sketches.}
        \label{tab:quantitative}
         \end{minipage}
&
\vspace{-2mm}
 \begin{minipage}{0.75\linewidth}\centering
\resizebox{.85\linewidth}{!}
{
    \begin{tabular}{l cc cc cc}\toprule
    & \multicolumn{2}{c}{\multirow{2}{*}{\makecell{3DVRChair \\ \cite{luo2021data} }}} & \multicolumn{4}{c}{\bf VRSketch2Shape (ours)} \\ 
    \cmidrule(lr){4-7}
    & & & \multicolumn{2}{c}{chair only} & \multicolumn{2}{c}{all categories} \\\cmidrule(lr){2-3} \cmidrule(lr){4-5} \cmidrule(lr){6-7}
    & F-score $\uparrow$ & CD\tiny{$\times 1000$}\normalsize $\downarrow$  & F-score $\uparrow$  & CD\tiny{$\times 1000$} \normalsize $\downarrow$  & F-score $\uparrow$  & CD\tiny{$\times 1000$}\normalsize  $\downarrow$  \\ \midrule
     LAS-diffusion$^\star$ \cite{zheng2023locally}& 26.1  & 66.0  & 37.0  & 51.1   &  40.2  & 27.1  \\\greyrule
     Luo \etal \cite{luo20233d} & 26.6 & 35.5 & 42.2 & 13.4 & 48.8 & 13.0 \\
    \bf VRSketch2Shape (ours) & \bf 31.1 & \bf 25.8 & \bf 64.3 & \bf \hphantom{5} 4.0  & \bf 69.8 & \bf \hphantom{1}4.8 \\
    \bottomrule
    \end{tabular}
    }

 \end{minipage}
     \end{tabular}
\end{table*}

\paragraph{Sequence Embeddings.}
Order matters at two levels: stroke index $s$ and within-stroke point index $i$.
We encode both positions using the sinusoidal encoding from the original Transformers~\cite{vaswani2017attention}:
\begin{align}
\Phi_{\text{seq}}(t)
\!=\!
\left[
\sin
\!\left(
\tfrac{t}{10,000^{2d/D}}
\right),
\cos
\!\left(
\tfrac{t}{10,000^{2d/D}}
\right)
\right]_{d=0}^{D/2-1}
\!\!\!.
\end{align}
Stroke and point embeddings are obtained with the linear projections
$\Lin_{\text{stroke}}$ and $ \Lin_{\text{point}}:\bR^D\!\mapsto\!\bR^D$:
\begin{align}
E_{\text{stroke}}(s)
&= \Lin_{\text{stroke}}\!\big(\Phi_{\text{seq}}(s)\big)\\
E_{\text{point}}(i)
&= \Lin_{\text{point}}\!\big(\Phi_{\text{seq}}(i)\big)~.
\end{align}
\paragraph{Final Token Embedding.}
For a point token $p_i^s$, we sum the spatial, stroke, and point embeddings:
\begin{align}
E(p_i^s)
~=~
E_{\text{spa}}(p_i^s)
+ E_{\text{stroke}}(s)
+ E_{\text{point}}(i)~.
\end{align}
\paragraph{Augmentation strategies.} 
We apply the following sketch-specific stochastic data augmentations during training:
\begin{compactitem}
    \item {\bf Stroke dropping.} Randomly mask $15\%$ of the strokes.
    \item {\bf Point dropping.} Randomly mask $30\%$ of the points within the remaining strokes.
    \item {\bf Stroke swapping.} Randomly swap $20\%$ of the strokes with another stroke in the sketch.
\end{compactitem}
The masked tokens are replaced with a learnable token \MASK~
such that $E_{\text{spa}}(\MASK) \in \bR^D$
%
\paragraph{Differences From SketchBERT.}
While our encoder shares structural similarities with the two-dimensional \textsc{SketchBERT}~\cite{lin2020sketchbert}, it differs in three key aspects:
(i) Point coordinates are represented via spatial Fourier features rather than raw positions,
(ii) Stroke delimiters are treated as learned tokens instead of concatenated one-hot flags, and
(iii) Continuous Fourier-based encodings replace fixed lookup tables, allowing flexible handling of variable-length and user-dependent sketch styles.

\subsection{Diffusion-Based Shape Generation}\label{sec:sdiffusion}
We condition 3D shape generation on the sketch embeddings using \textsc{SDFusion}~\cite{cheng2023sdfusion}, a latent diffusion model originally designed for text- and image-guided shape synthesis.
Our sketch encoder interfaces directly with the diffusion model: the sequence of tokens produced by the \textsc{BERT}-style encoder serves as the conditioning input to \textsc{SDFusion}.

During training, each ground-truth 3D shape is first voxelized and encoded into a compact latent representation using a pretrained 3D VQ-VAE~\cite{van2017neural}.
Gaussian noise is then added to this latent through the forward diffusion process, and a U-Net~\cite{ronneberger2015u} is trained to predict the denoised latent.
The VQ-VAE remains frozen, while the U-Net and our sketch encoder are optimized jointly using an $\ell_2$ reconstruction loss between the predicted and target latents, following the approach of~\cite{rombach2022high}.
At inference time, we encode the sketch and apply the reverse denoising process with $100$ DDIM steps \cite{songdenoising} starting from random noise to synthesize the corresponding 3D shape.
Unlike previous approaches that require multi-stage training or modality alignment steps, our framework is trained end-to-end in a single stage.

\paragraph{Implementation Details.}
We encode spatial coordinates using Fourier features with $L\!=\!10$ frequencies per axis, concatenated and projected to a $D\!=\!256$-dimensional embedding through a 2-layer MLP with 256 hidden units. The BERT-style transformer has 6 layers, 8 attention heads, and a feed-forward with an inner width ratio of 1.

We train all models with AdamW~\cite{loshchilovdecoupled} (default parameters) using a base learning rate of $10^{-4}$ and ReduceLROnPlateau decay with a patience of $10$ epochs and decay of $0.5$. 
We use a batch size of 16 for synthetic pretraining and 12 for real-data fine-tuning. 
The model is pretrained for 200 epochs on synthetic sketches and optionally fine-tuned for 300 epochs on real sketches. 
A dropout rate of 0.1 is applied in the transformer encoder.

\if 1 0
\subsection{Cross-Modal Shape Completion}\label{sec:sdiffusion}
We consider another setting where the sketch is incomplete: only the first strokes are visible, and task the model with predicting the entire shape. To do so, we add the option of training the BERT model with a missing token prediction task.

We consider the following sketch: 
$
\cS=\{s_1, \cdots, s_N
\}
$
characterized by their 
where $s_i$ can be either a seperator token or a 3D point $p_i \in [0,1]^3$. We associate a label to each token $\type(s_i)$ defined as $0$ if $s_i$ is a 3D point, $1$ if $s_i=\texttt{SEP}$, and $2$ if $s_i=\texttt{EoS}$.

We mask a portion \texttt{mask} of input token by replacing them with a learned \texttt{mask} token in $\mathbb{R}^D$. The masked inputs is given to our encoder, yielding the embeddings $\{E_1, \cdots E_N\}$.
We add a linear layer 
$\Lin_\text{type}: \bR^D \mapsto \mathbb{R}^3$ tasked with predicting the label of token $s_i$, and $\Lin_\text{pos}: \bR^D \mapsto [0,1]^3$ tasked with predicting the position of the point $p_i$ \emph{if $s_i$ is a 3D point}.

We then supervise the completion task with the following loss:
\begin{align}
    \mathcal{L}_\text{complete}
    &=&
    \sum_{i \in \texttt{masked}}
    H(\type(, \Lin_\text{type}(E_i)
    +
    \lambda \mathbb{1}[t_i=0] \Vert p_i - \Lin_\text{pos} \Vert^2~,
\end{align}
where $\lambda>0$ is an hyperparameter taken as $0.1$. If the model is trained for sketch completion, we add this loss to the reconstruction loss of SDFusion.

In practice, to complete a partial sketch, we remove the \texttt{EoS} token and add $100$ mask tokens at the end. We then 
\fi

\section{Numerical Experiments}
%
%
%
We present our evaluation setting (\cref{sec:dataset}), our results (\cref{sec:results}), and finally an ablation study (\cref{sec:ablation}).
\subsection{Datasets and Evaluation Metrics}
\label{sec:dataset}
We evaluate our approach on two datasets: 3DVRCHAIR~\cite{luo2021data} and our proposed VRSKETCH2SHAPE.

\paragraph{\textsc{3DVRChair}~\cite{luo2021data} .}
This dataset is the only other publicly available benchmark for VR sketch–based 3D generation. 
It contains $1{,}005$ real sketch–shape pairs from the \textit{chair} category only. 
We use the official split of 803 samples for training and 202 for evaluation. 

\paragraph{\textsc{VRSketch2Shape} dataset.}
We also evaluate on our proposed datasetby first training on the synthetic subset and evaluating under two adaptation protocols:
\begin{compactitem}
    \item \textbf{Zero-shot adaptation.} The model is evaluated directly on the real sketch test set to assess the synthetic-to-real generalization gap.
    \item \textbf{Few-shot adaptation.} The model is fine-tuned on a subset or the entirety of the fine-tuning set.
\end{compactitem}

\paragraph{Metrics.}
Following standard practice~\cite{luo2021data}, we evaluate the generated 3D shapes using two metrics:
\begin{compactitem}
    \item \textbf{Chamfer Distance (CD).} We uniformly sample $N=4096$ points from the ground-truth surface  and $N$ points from the generated surface, and compute the mean bidirectional distance.
    \item \textbf{F-score.}
We evaluate the geometric accuracy of generated shapes using the F-score, which combines precision and recall~\cite{tatarchenko2019single}. We use a threshold $\delta = 0.02$ to reflect the inherent imprecison of manual sketching. 
\end{compactitem}

For all metrics, we follow the evaluation protocol of~\cite{luo2021data}: the predicted shapes are first aligned to the ground truth by translating and scaling them such that the diagonal of their bounding box coincides.

\subsection{Results and Analysis}
\label{sec:results}
We report quantitative results in \cref{tab:quantitative}, comparing our approach with all publicly available baselines. 
We conduct four main experiments.

\paragraph{Experiment on \textsc{3DVRChair}.}
We train our model on the training split of \textsc{3DVRChair} and evaluate on its test set.  
We compare against the official pretrained weights of Luo~\etal~\cite{luo2021data} as well as a retrained 2D-based LAS-Diffusion baseline (see \cref{sec:ablation} for implementation details).  
As shown in \cref{tab:quantitative}, our method outperforms competing approaches by a large margin, reducing CD by more than $60\%$ and improving the F-score by over $40\%$ on our dataset.
We were unable to evaluate Chen~\etal~\cite{chen2023deep3dsketch+} due to the absence of publicly released code or checkpoints, and their reported results rely on an unspecified evaluation protocol, making direct comparison impossible.

\paragraph{Experiment on \textsc{VRSketch2Shape}.}
We train all models on our synthetic subset, fine-tune on the real fine-tuning subset, and evaluate on the held-out real test set. 
We report results both on the \textit{chair} category (for comparability with \textsc{3DVRChair}) and across all four categories.  
Our method achieves the best performance in both settings, confirming the benefit of our proposed model.  
Interestingly, Luo~\etal\ also perform better on our dataset than on theirs, likely due to the reduced ambiguity of our automatically aligned synthetic sketches.  
These findings jointly validate the effectiveness of our model and the utility 
 of our dataset for robust VR sketch–based shape generation.

 We visualize in \cref{fig:quali} representative synthetic and real sketches from our test set, along with 3D reconstructions predicted by our model and by Luo~\etal.
The generated shapes closely match the ground-truth geometry and preserve object topology and details more faithfully.
We note that our reconstructions sometimes appear slightly oversmooth, which we attribute to optimizing only in the latent space of a pretrained, frozen 3D VQ-VAE.

\begin{figure}[t]
    \centering
    \resizebox{\linewidth}{!}{
\begin{tikzpicture}
\begin{axis}[
    width=1\linewidth,
    height=4.5cm,
    xlabel={\# of real training samples per category},
    ylabel={\textcolor{colCD}{$\leftarrow$ CD $\times1000$ }},
    ymin=0, ymax=6,
    xtick=data,
    xticklabel style={align=center},
    symbolic x coords={0,50,100,200},
    ytick={0, 2.5, 5},
    axis y line*=left,
    axis x line=bottom,
    yticklabel style={colCD},
    ylabel near ticks,
    enlargelimits=0.00,
    xlabel style={yshift=0ex},
    ylabel style={xshift=0em},
    legend cell align=left,
]
\addplot[colCD, thick, mark=diamond*, mark size=2.5pt, mark options={solid}, smooth, tension=0.5] coordinates {
    (0,5.1) (50,4.4) (100,4.4) (200,4.1)
};

\end{axis}

\begin{axis}[
    width=1\linewidth,
    height=4.5cm,
    axis y line*=right,
    axis x line=none,
    ylabel={\textcolor{colFONE}{F1 $\rightarrow$}},
    ymin=50, ymax=70,
    xtick=data,
    symbolic x coords={0,50,100,200},
    ytick={50,60,70},
    ylabel near ticks,
    yticklabel style={colFONE},
    enlargelimits=0.00,
    legend cell align=left,
]
\addplot[colFONE, thick, mark=diamond*, mark size=2.5pt, mark options={solid}, smooth, tension=0.5] coordinates {
    (0,56.8) (50,61.8) (100,63.9) (200,64)
};

\end{axis}
\end{tikzpicture}
}
    \vspace{-2mm}
    \caption{{\bf Few-Shot Adaptation.} 
 \textcolor{colCD}{Chamfer Distance} and  \textcolor{colFONE}{F1-score} on our real test set as a function of the number of real sketches used to fine-tune a model pretrained on synthetic data.
    }
    \label{fig:few-shot}
\end{figure}
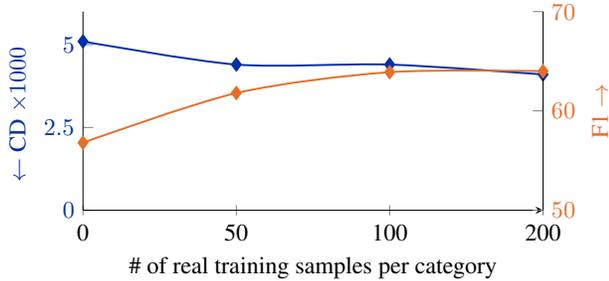

\paragraph{Few-shot Synthetic-to-Real Adaptation.}
We evaluate in \cref{fig:few-shot} the effect of fine-tuning our model trained on synthetic sketches with real sketches.
We take our model pretrained on synthetic data only, and fine-tune it (or not). 
Performance improves steadily as more real sketches are introduced, but as few as 50 sketches per category suffice to reach near-optimal results. 
Remarkably, even in the zero-shot setting (no fine-tuning), our model performs strongly, demonstrating that the synthetic sketches generated by our heuristic pipeline provide effective supervision for real-world generalization.

\paragraph{Free-hand Sketches.}
Real creative workflows involve \emph{free-hand} sketching, \ie users draw without any reference model.  
This raises an important question:  
\emph{Can a model trained primarily on synthetic and reference-guided sketches generalize to free-hand inputs?}
We conducted a user study evaluating sketches drawn with and without reference shapes, comprising 40 sketches: 20 free-hand and 20 reference-guided. From these, we constructed 100 sketch-shape pairs: 40 reconstructions generated by our model, 40 by Luo~\etal, and the 20 ground-truth shapes used for the reference-guided sketches. 
A total of 38 participants were each shown 50 randomly ordered sketch-shape pairs, yielding 1{,}900 ratings, in which sketch-shape correspondence was evaluated on a 1-5 Likert scale.
As reported in \cref{tab:study}, our method significantly outperforms Luo~\etal\ and maintains strong performance on free-hand sketches.
\cref{fig:suppl_freehand} shows that our model's robust performance on free-hand sketches.

\begin{table}[t]
    \centering
        \caption{{\bf User Study.} Likert scale: \textit{5-Excellent}: faithful geometry;
\textit{4-Good}: minor artifacts;
\textit{3-Acceptable:} recognizable, missing details;
\textit{2-Poor}: weak correspondence;
\textit{1-Failed}.} 
\small{
    \begin{tabular}{lcc}\toprule
    & w. reference & free-hand \\ \midrule
    Luo's prediction & 2.76 $\pm$ 0.84 & 2.02 $\pm$ 0.89 \\
    Ours prediction & \bf 3.92 $\pm$ 0.74 & \bf 3.60
    $\pm$ 1.01\\\greyrule
    Reference shape& 4.76 $\pm$ 0.50 & - \\\bottomrule
    \end{tabular}}
    \label{tab:study}
\end{table}


\begin{figure}[h]
    \centering
    \resizebox{\linewidth}{!}{
\begin{tabular}{l@{\,}c@{\,}c@{\,}c@{\,}c@{\,}}
\rotatebox{90}{\small \bf \qquad  Free-hand Sketch}&
\includegraphics[height=0.15\textheight]{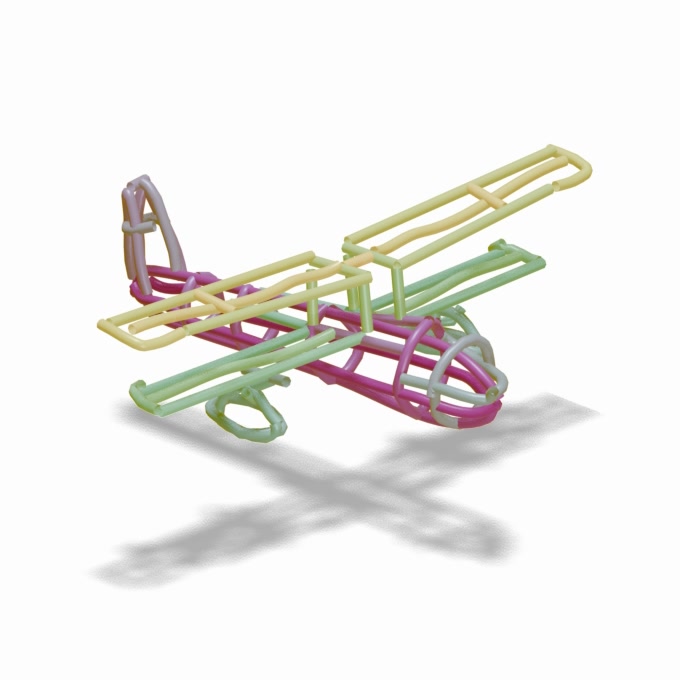} &
\includegraphics[height=0.15\textheight]{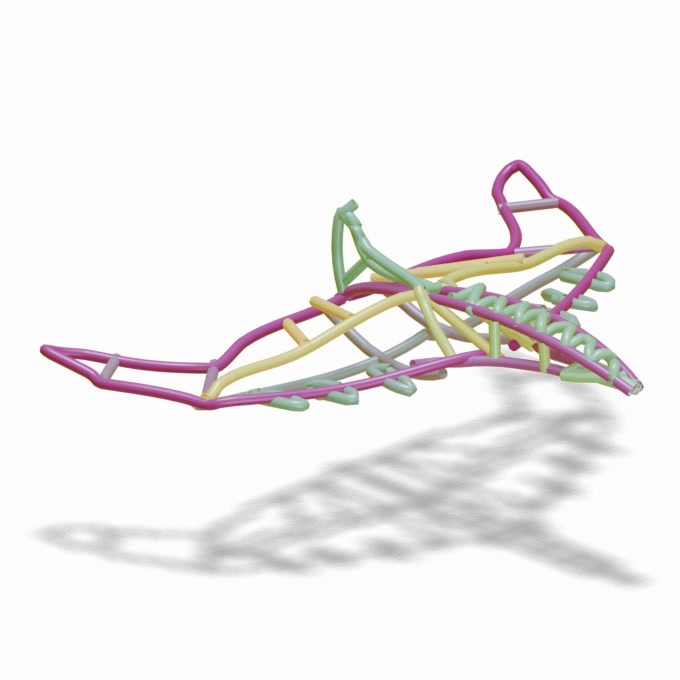} &
\includegraphics[height=0.15\textheight]{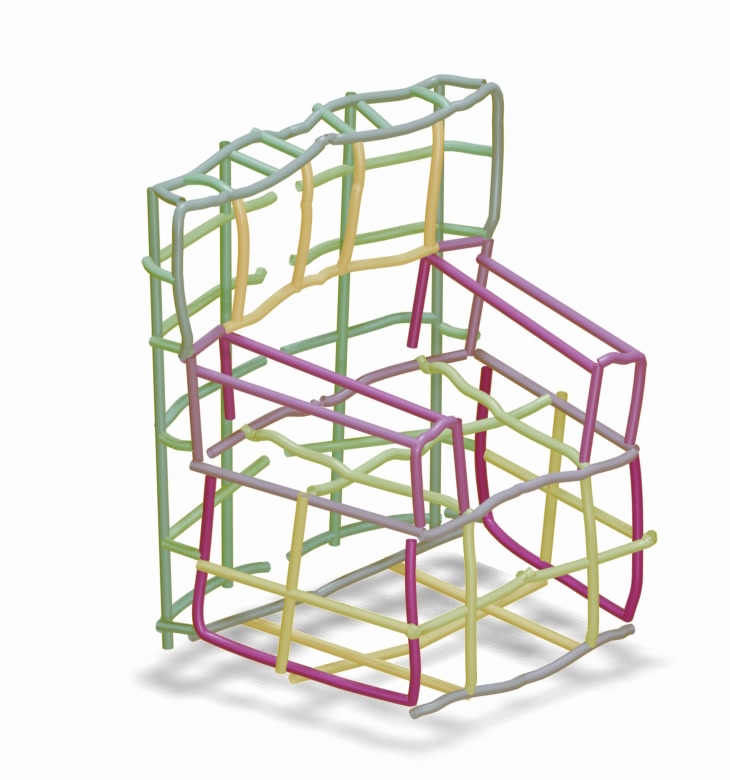} 
&
\includegraphics[height=0.15\textheight]{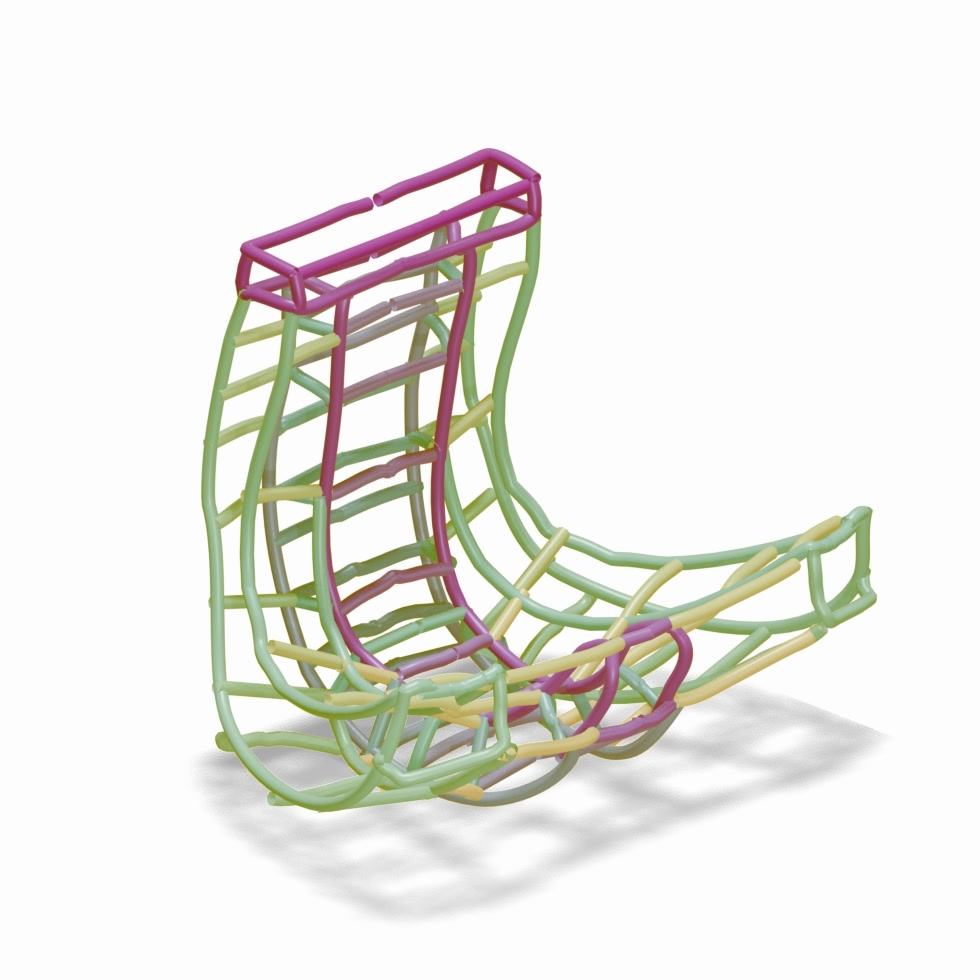} 
\\[-4mm]
\rotatebox{90}{\small \bf \qquad Our prediction} &
\includegraphics[height=0.15\textheight]{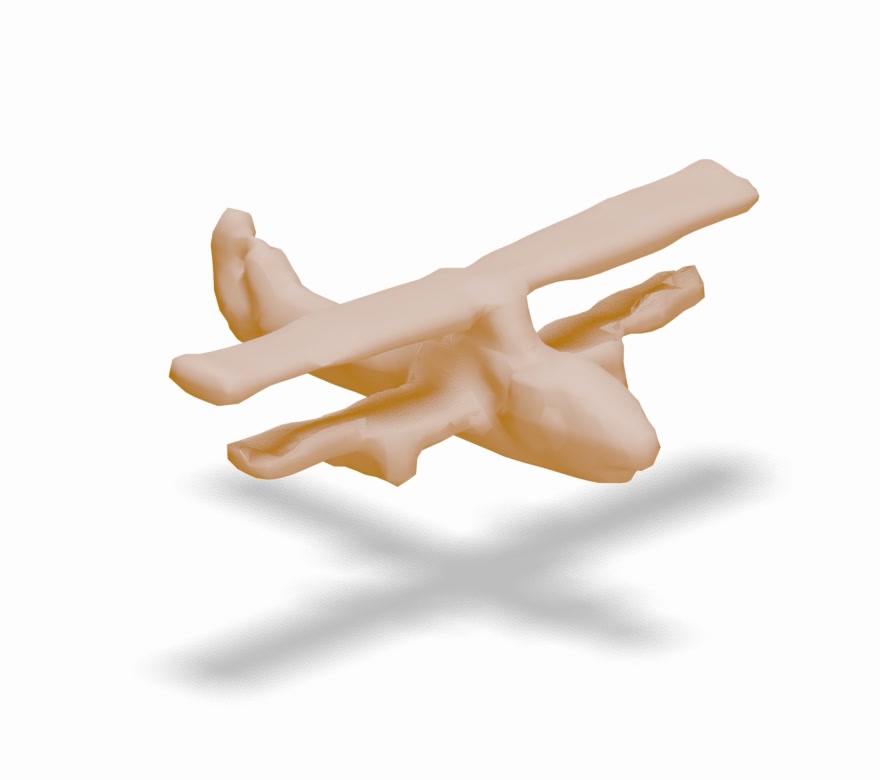} &
\includegraphics[height=0.15\textheight]{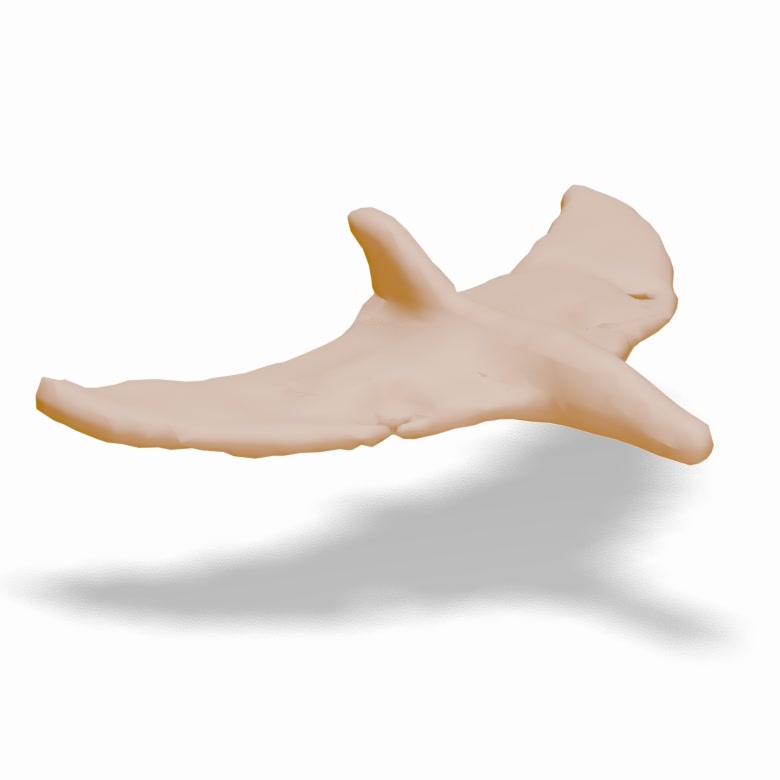} &
\includegraphics[height=0.15\textheight]{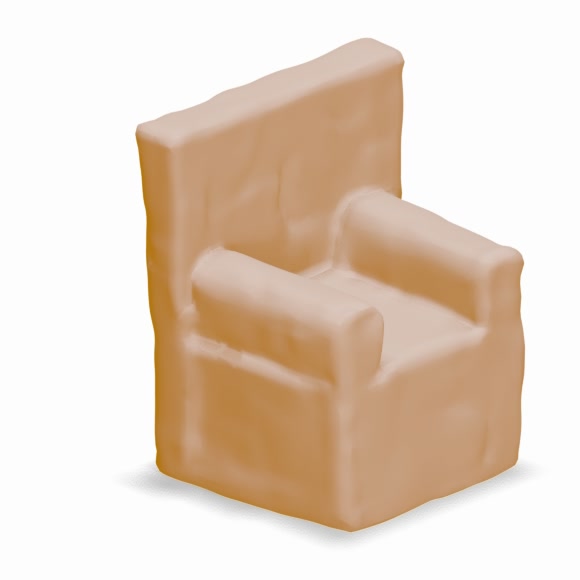} 
&
\includegraphics[height=0.15\textheight]{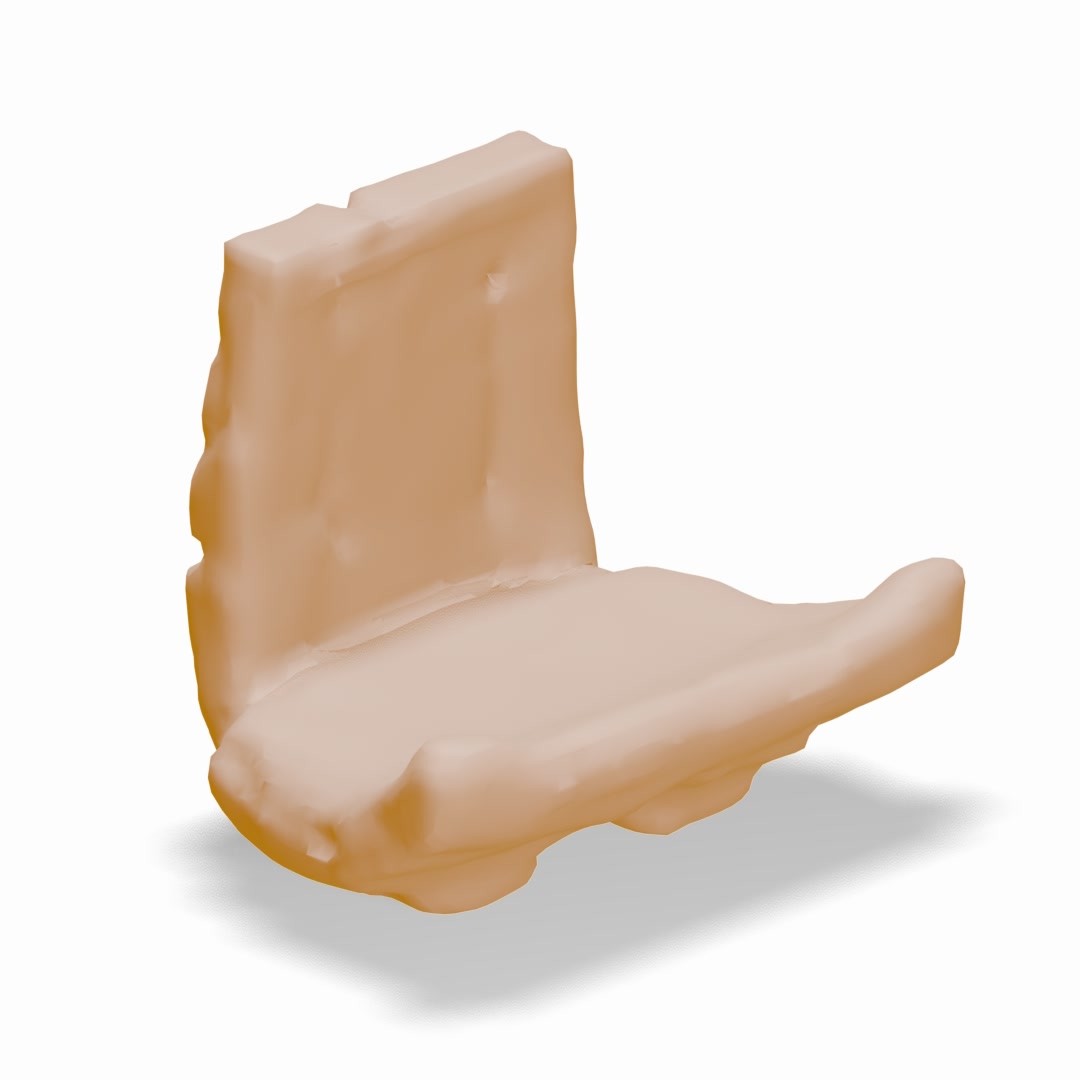}
\\[-4mm]
\rotatebox{90}{\small \bf \qquad Free-hand Sketch} &
\includegraphics[height=0.15\textheight]{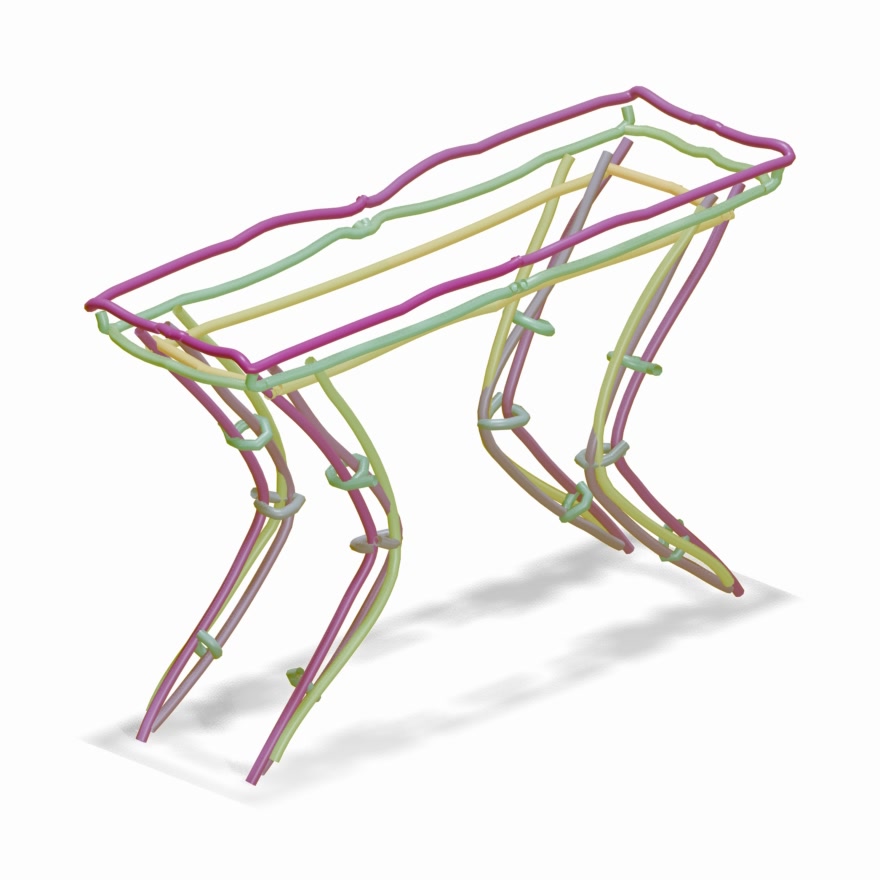} &
\includegraphics[height=0.15\textheight]{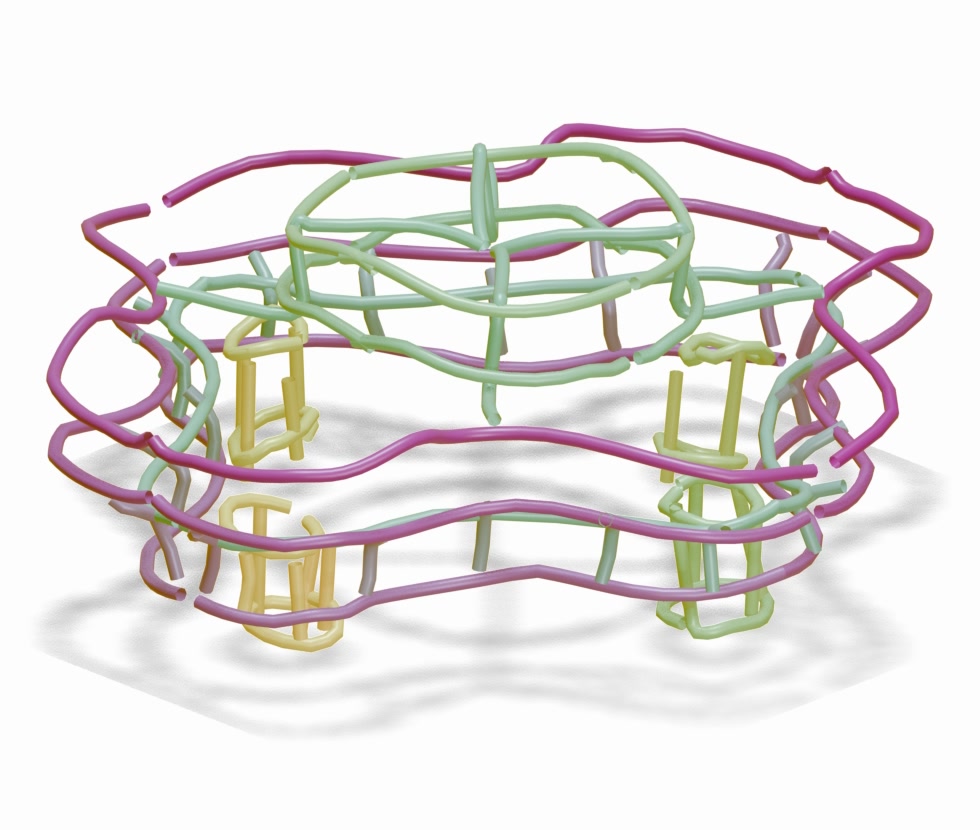} &
\includegraphics[height=0.15\textheight]{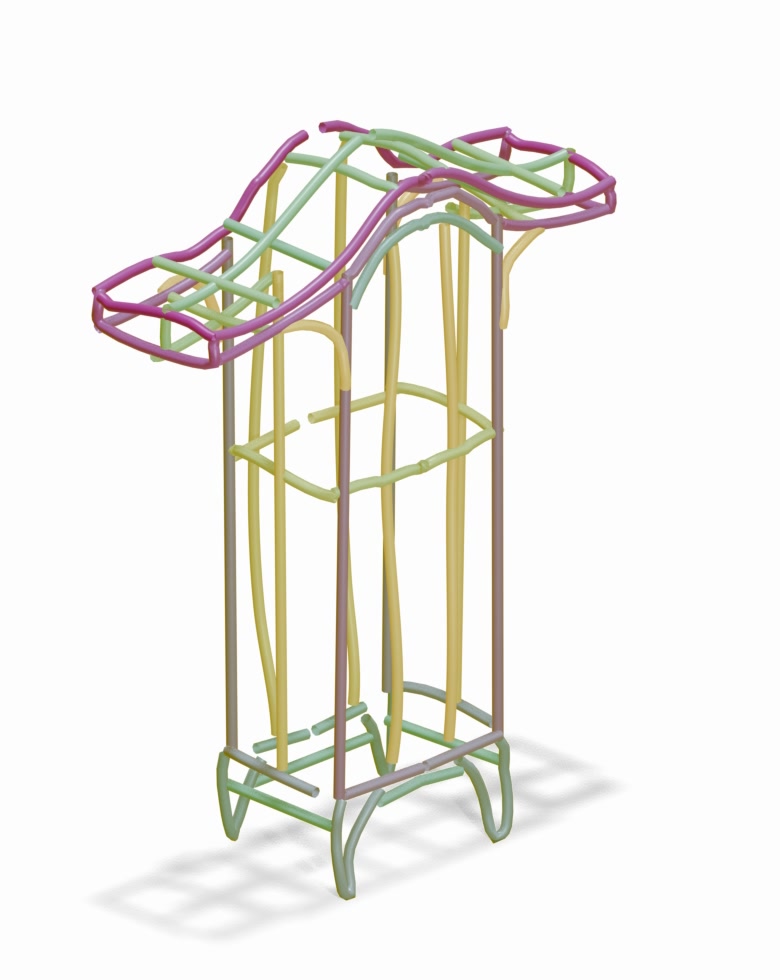} 
&
\includegraphics[height=0.15\textheight]{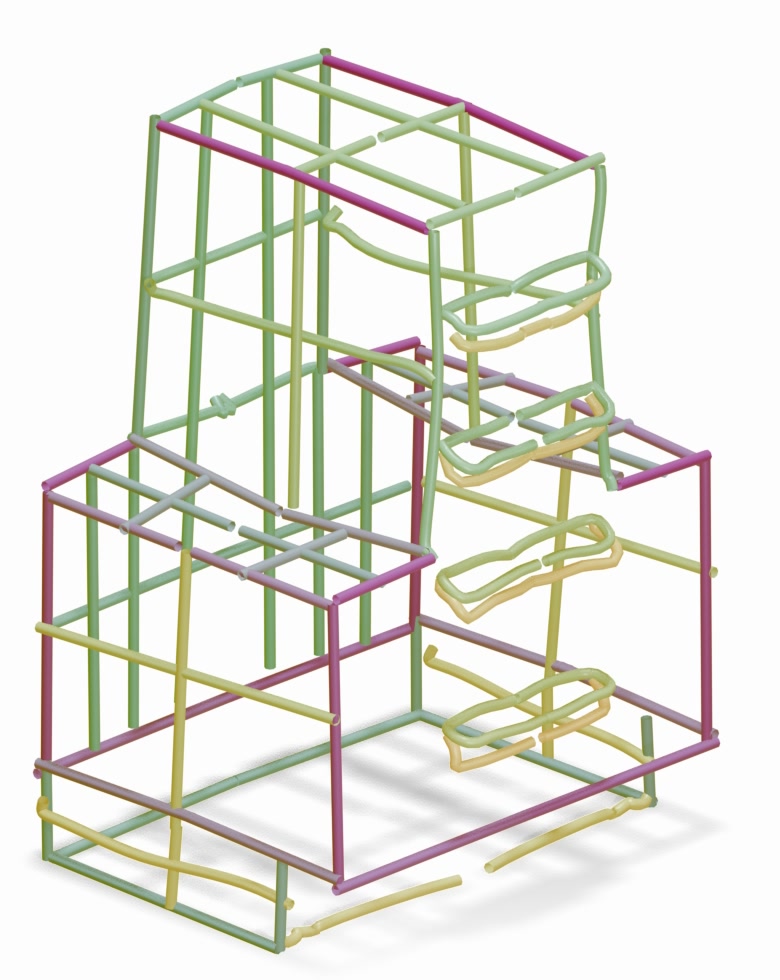}
\\[-4mm]
\rotatebox{90}{\small \bf \qquad Our prediction} &
\includegraphics[height=0.15\textheight]{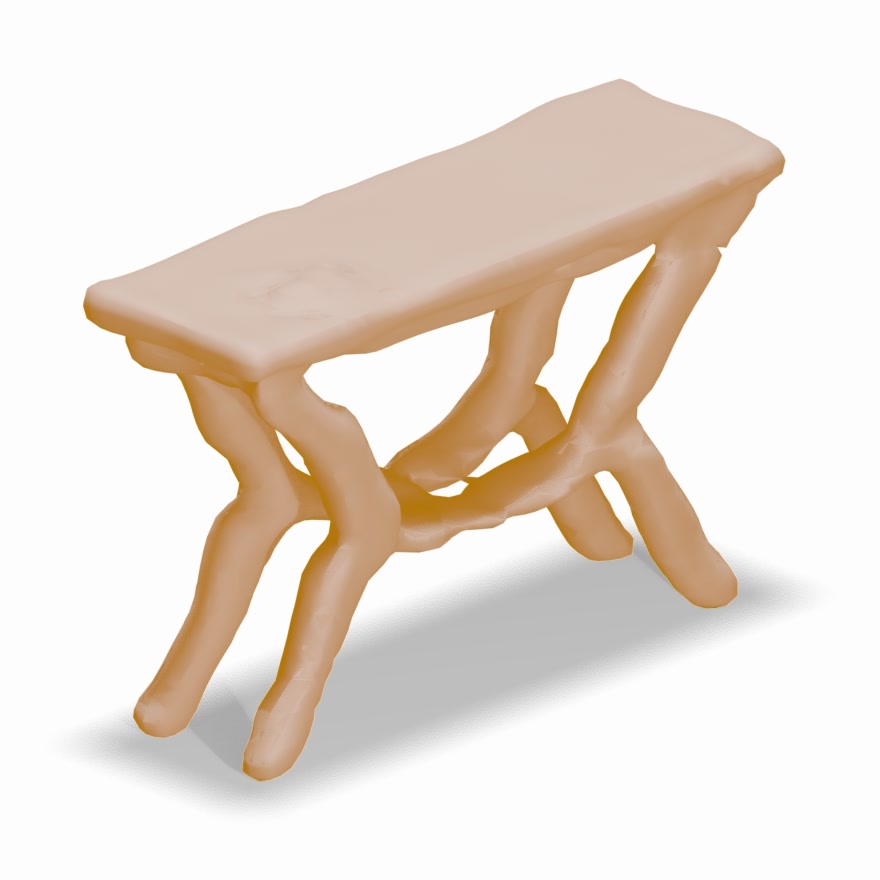} &
\includegraphics[height=0.15\textheight]{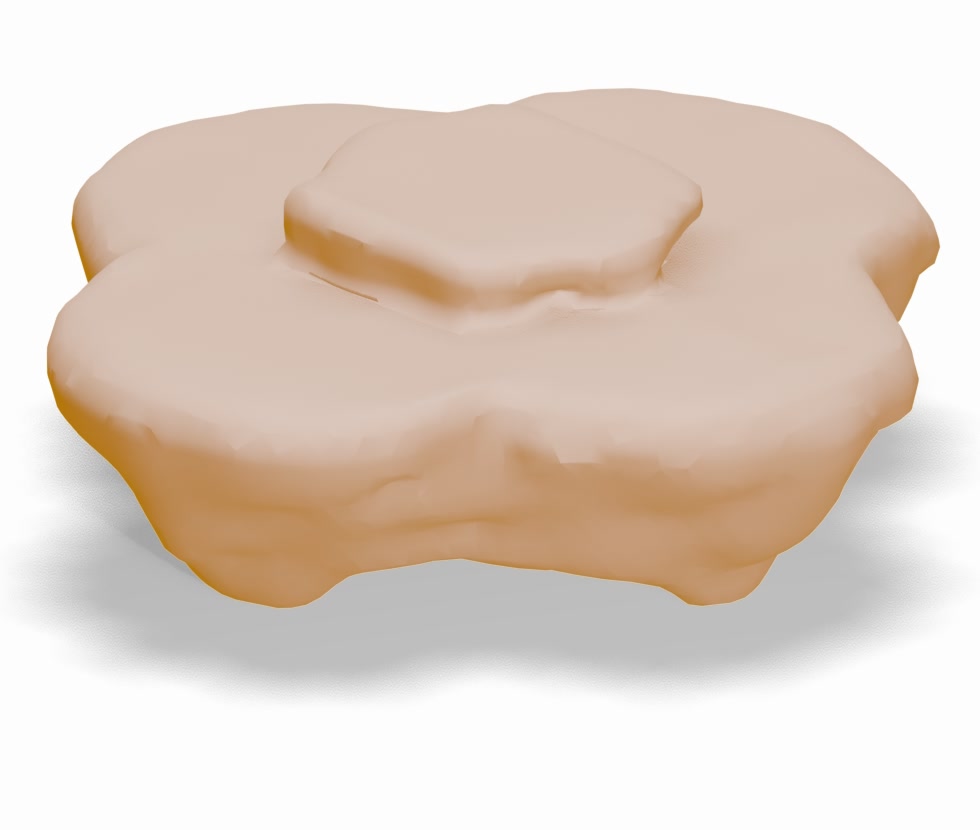} &
\includegraphics[height=0.15\textheight]{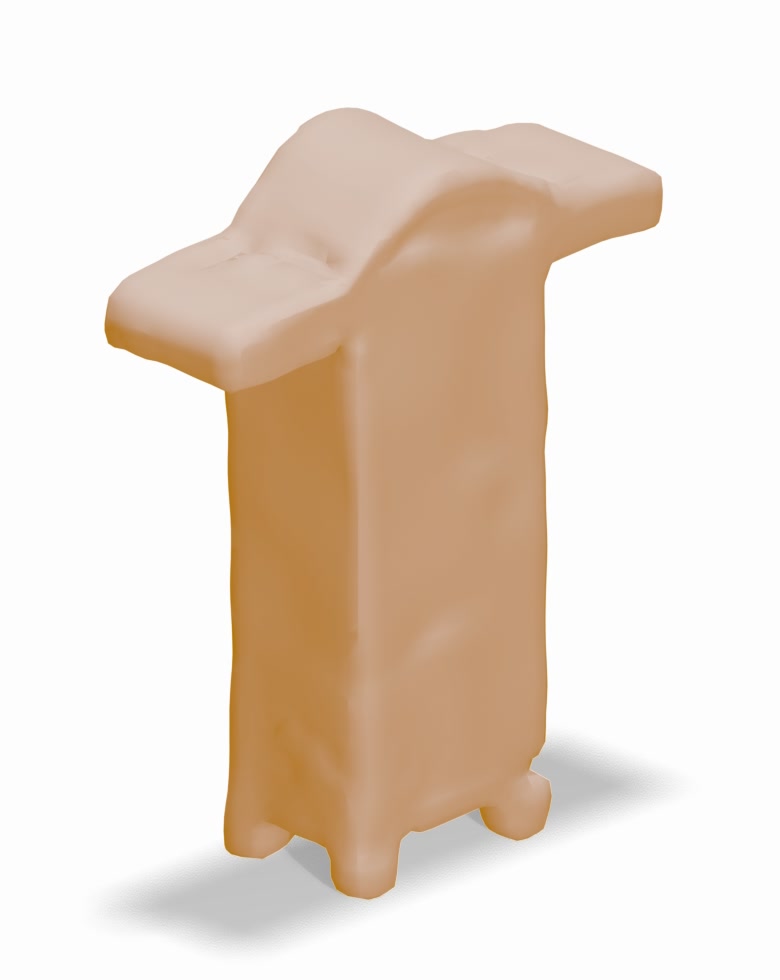} 
&
\includegraphics[height=0.15\textheight]{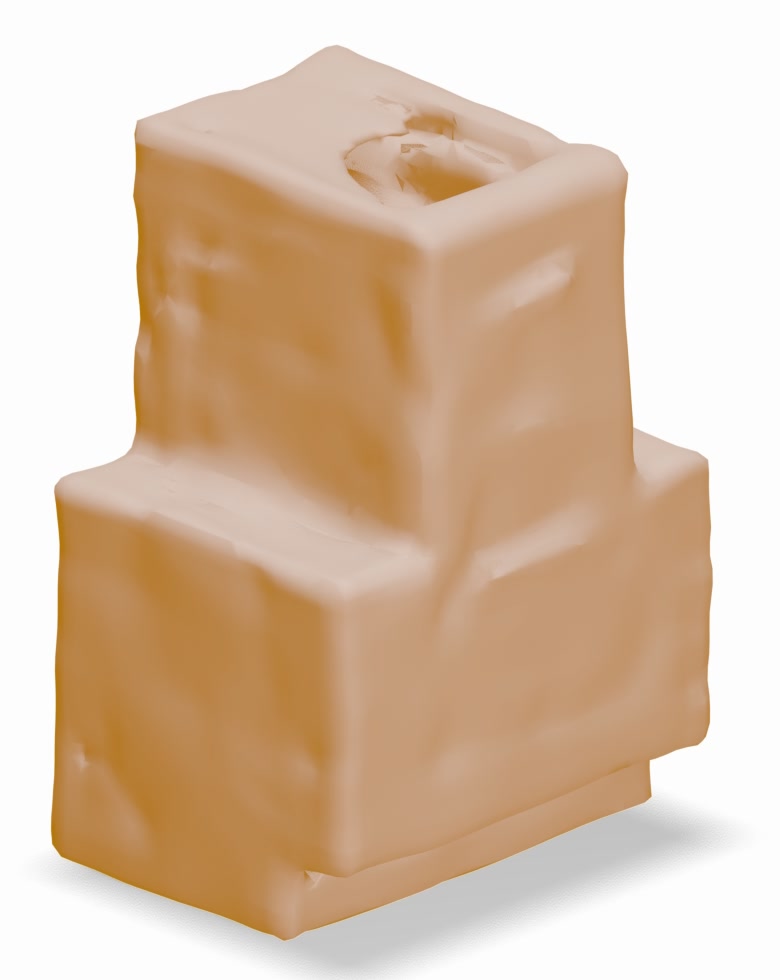}
\end{tabular}
}

%

%
    \caption{{\bf Shape Generation from Free-Hand Sketches.}     Our model generalizes well to free-hand sketches drawn without any reference shape for airplanes, chairs/sofas, tables, and cabinets, producing detailed and plausible reconstructions that reflect the user’s intent.
    }
    \label{fig:suppl_freehand}
\end{figure}

\subsection{Ablation Study}
\label{sec:ablation}
We perform an ablation study to quantify the contribution of our main design choices, summarized in \cref{tab:ablation}.
All experiments are conducted in a zero-shot synthetic-to-real setting on the \textit{chair} subset of our dataset.

\paragraph{Impact of Design Choices.} We first assess the importance of key components of our model:
\begin{compactitem}
    \item \textbf{w/o ordering.} 
    We remove stroke and point indices from the sketch encoder, keeping only $E_\text{spa}$ while discarding $E_\text{stroke}$ and $E_\text{point}$.
    Performance drops sharply, confirming that  \emph{order does matter}.
    \item \textbf{w/o augmentations.} 
    Disabling augmentations noticeably degrades performance, showing that these simple augmentations effectively improve robustness and prevent overfitting.
    \item \textbf{w/o pre-training.} 
    We skip pre-training on synthetic data and train only on the 200 real sketch–shape pairs from the fine-tuning chair set.
    The model collapses to trivial solutions, emphasizing the necessity of large-scale synthetic pre-training and the effectiveness of our fully automated sketch synthesis pipeline.
    Reaching a comparable data volume through manual sketching would require prohibitive human effort.
    \item \textbf{SketchBERT encoder.} 
    Replacing our encoder with a direct 3D extension of SketchBERT~\cite{lin2020sketchbert} leads to a substantial drop in accuracy, highlighting the importance of our design adaptation for 3D sequential data.
\end{compactitem}

\paragraph{Impact of Sketch Format.} We then compare representing sketches as ordered point–stroke sequences versus other common encodings:
\begin{compactitem}
   \item \textbf{Sketches as point clouds.} 
    We uniformly sample 1,024 points along all strokes and encode them using PointNet++~\cite{qi2017pointnet++}, following Luo~\etal~\cite{luo20233d}.
The clear performance degradation indicates that our sequential formulation, and not just the diffusion generator, accounts for much of the observed improvement.
    
    \item \textbf{Sketches as images.} 
    We convert each 3D sketch into a mesh and render five 2D views using \texttt{pyrender}~\cite{pyrender}.
    Following LAS-Diffusion~\cite{zheng2023locally}, the rendered images are encoded with a pretrained VGG network. The embeddings and corresponding camera poses condition a latent diffusion model trained to predict $64^3$ occupancy grids.
    For fairness, we retrain this model on our full synthetic dataset and apply the pretrained super-resolution module from~\cite{zheng2023locally} to upsample predictions to $128^3$ signed distance fields.
This variant yields a marked performance drop, as occlusions in the rendered views cause missing or distorted geometry.
\end{compactitem}

\paragraph{Generation Speed.} On a single consumer NVIDIA RTX~4090 GPU, generating a 3D shape from a sketch takes on average 6.61\,$\pm$\small{1.18}\normalsize\,s, including 26\,ms for sketch encoding, 6.55\,s for latent denoising, and 31\,ms for SDF decoding with Marching Cubes \cite{lorensen1998marching}. Training on our $20$k{+} synthetic sketches completes in roughly 50 hours, and fine-tuning on the real sketches adds an additional 10 hours.

\paragraph{Limitations.} Since our model is only supervised by its ability to denoise latent embeddings and uses a frozen 3D VQ-VAE, reconstruction quality and inference speed is ultimately bounded by the capacity of this encoder–decoder. In particular, the $64^3$ SDF resolution limits fine-grained geometric detail. Future work could lift this constraint by training the shape generator end-to-end at higher spatial resolutions.



\begin{table}[]
    \centering
    \caption{{\bf Ablation study.} 
We evaluate variants of our models trained on synthetic sketches and evaluated on real \textit{chair} sketches to show the contribution of each component. 
Discarding stroke order, augmentations, or pretraining significantly degrades performance, while alternative sketch format (point clouds or multiview) fail to capture 3D sequential structure effectively.}

\small 
    \begin{tabular}{lcc}
    \toprule
     Method   & F-score $\uparrow$ &  CD$\tiny{\times 1000}$ $\downarrow$ \\ \midrule
     Full model    &  \bf 56.8  & \bf 5.1 \\\greyrule
     w/o stroke ordering & 48.9 & 7.1  \\
     w/o augmentation & 49.0 & 6.6 \\
     w/o pretraining & 12.0 & \negphantom{9}99.9 \\
     SketchBert encoding & 50.4 & 6.3  \\\greyrule
     Sketches as point clouds \subref{fig:ablation:points} & 30.8 & \negphantom{2} 25.8 \\
     Sketches as Images \subref{fig:ablation:images}  & 23.8 & \negphantom{2}62.6 \\
     \bottomrule
    \end{tabular}
    \\[3mm]
\begin{tabular}{c@{}c}
\raisebox{-1cm}
    {\includegraphics[width=.15\linewidth]{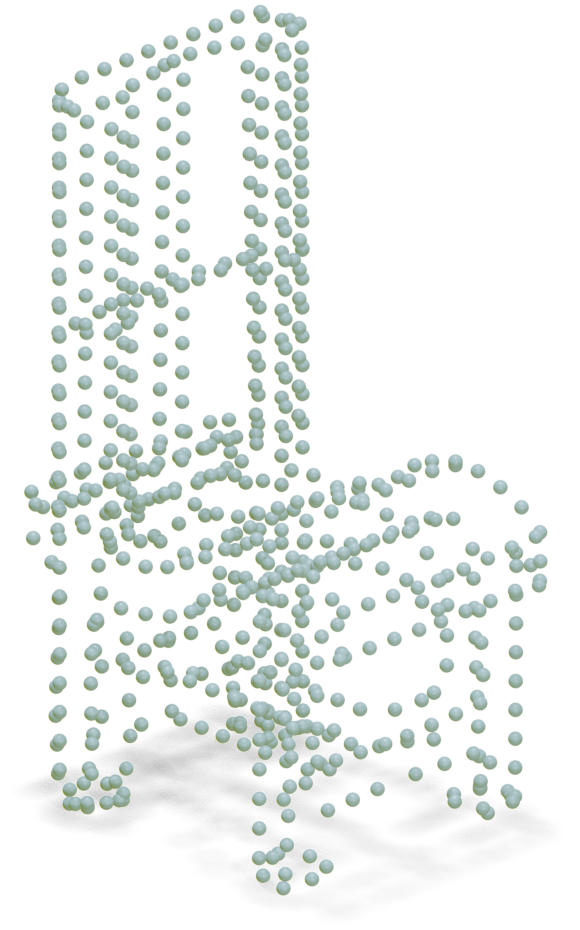}}
&
    \begin{tabular}{c@{\;}c@{\;}c@{\;}c@{\;}c}
        {\adjustbox{trim={0.35\width} {0.20\height} {0.30\width} {0.15\height}, clip}{
    \includegraphics[width=.3\linewidth]{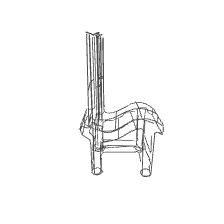}
        }}
        &
        {\adjustbox{trim={0.35\width} {0.20\height} {0.30\width} {0.15\height}, clip}{
    \includegraphics[width=.3\linewidth]{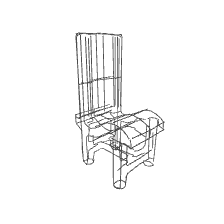}
        }}
        &
       {\adjustbox{trim={0.35\width} {0.20\height} {0.30\width} {0.15\height}, clip}{
    \includegraphics[width=.3\linewidth]{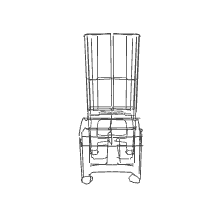}
        }}
    &
       {\adjustbox{trim={0.35\width} {0.20\height} {0.30\width} {0.15\height}, clip}{
    \includegraphics[width=.3\linewidth]{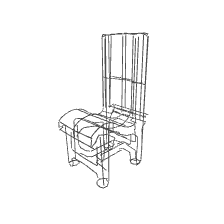}
        }}
            &
       {\adjustbox{trim={0.25\width} {0.20\height} {0.30\width} {0.15\height}, clip}{
    \includegraphics[width=.3\linewidth]{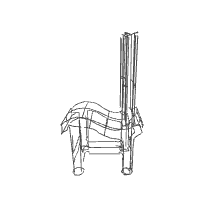}
        }}
        \end{tabular}
\\[-3mm]
\begin{subfigure}{0.2\linewidth}
    \caption{\makebox[0.20\linewidth][l]{Sketches as point clouds}}
    \label{fig:ablation:points}
\end{subfigure}
&
\begin{subfigure}{0.8\linewidth}
\captionsetup{justification=raggedleft,singlelinecheck=false}
    \caption{Sketches as images \qquad\qquad~}
    \label{fig:ablation:images}
\end{subfigure}
\end{tabular}
    \label{tab:ablation}
\end{table}


\section{Conclusion}
We introduced \textsc{VRSketch2Shape}, the first open-source, multi-category dataset and model for 3D shape generation conditioned on sequential VR sketches. 
Our contributions include an automatic pipeline for scalable synthetic sketch generation, a curated collection of real hand-drawn sketches with preserved drawing order, and a stroke-aware sketch encoder coupled with a diffusion-based shape generator. 
Extensive experiments show that explicitly modeling stroke order improves structural fidelity and generalization, enabling effective training even on synthetic data alone. 
\section*{Acknowledgment}
We thank Gege Gao for insightful discussions and Prof. Lorenz Hurni for their support and access to GPUs.
This work is supported by Hi! PARIS and ANR/France 2030 program (ANR-23-IACL-0005). The data collection process was funded by the Swiss National Science Foundation (SNSF) under the grant \emph{3D Sketch Maps} (Grant No. 202284).
{
    \small
    \bibliographystyle{ieeenat_fullname}
    \balance{
    \bibliography{main}
    }
}

 \clearpage
 \maketitlesupplementary
 \setcounter{page}{1}
\setcounter{page}{1}
\setcounter{section}{0}
\setcounter{table}{0}

\setcounter{figure}{0}

\renewcommand{\thefigure}{A-\arabic{figure}}
\renewcommand{\thetable}{A-\arabic{table}}
\renewcommand{\thesection}{A-\arabic{section}}
\vspace{1em}

\noindent

In this appendix, we further justify the design choice and evaluate the generalization capabilities of our model.
First, we provide additional ablation studies on Fourier features and sketch orders (\cref{sec:suppl_ablation}), and an experiment on shape completion (\cref{sec:completion}).
Next, we test sketches drawn without our surface-snapping tool (\cref{sec:suppl_unsnap}), followed by free-hand sketches created without any reference shape (\cref{sec:suppl_freehand}).
We then analyze the impact of the sketcher expertise (\cref{sec:suppl_expertise}) and the performance–speed tradeoff of our architecture (\cref{sec:perf}).
Finally, we provide additional qualitative illustrations of shape generation and sketch completion for both our method and competing baselines. 
\section{Additional Ablation}
\label{sec:suppl_ablation}
We further test the effectiveness of proposed 3D fourier features.
As shown in \cref{tab:ablation_rebuttal}[\textsc{a}], replacing 3D spatial Fourier features with raw coordinates leads to a clear performance drop, confirming their importance for encoding fine geometric detail. Replacing 1D Fourier encodings with fixed positional embeddings also degrades performance and restricts variable-length sequences.

We also test different ordering strategy. Needless to say,
our synthetic sketches are not intended to model human drawing behavior, but to provide effective supervision for learning a sketch-to-shape mapping.
Stroke order is therefore treated as an inductive bias, not as a claim about how humans draw.
For instance, DFS ordering empirically outperforms BFS (\cref{tab:ablation_rebuttal}[\textsc{b}]), but this does not necessarily reflect human sketching strategies.

We further evaluate the impact of order perturbations (\cref{tab:ablation_rebuttal}[\textsc{c}]):
reversing stroke order has little effect, scrambling strokes causes a moderate drop, while scrambling points leads to a clear degradation in F-score.
These results indicate that \emph{consistent sequential structure}, rather than a specific human-like ordering, is what benefits learning.

Importantly, order modeling is critical for partial sketches: when reconstructing shapes from only the first half of a human-drawn sketch, our order-aware model outperforms an order-agnostic variant by +6.6 F1-score.
We will therefore reframe our claim of “order matters” to these empirically supported points.

\section{Cross-Modal Shape Completion}
\label{sec:completion}
We evaluate our model’s ability to infer complete 3D shapes from partial sketches.
To simulate incomplete inputs during inference, we keep only the first fraction of points in the sketch sequence, preserving its natural drawing order, and pad the remainder with learned $\texttt{MASK}$ tokens.
\SEP~tokens are randomly inserted to mimic the natural stroke-length distribution of real sketches.
The model then predicts the full 3D shape directly from these partially masked sequences. 

\begin{figure*}[h]
    \centering
    \begin{tabular}{l@{\,}c@{\,}c@{\,}|c@{\,}c@{\,}|c@{\,}c}
\rotatebox{90}{\small \bf \qquad Sequential Sketch}
&
\includegraphics[width=0.15 \linewidth]{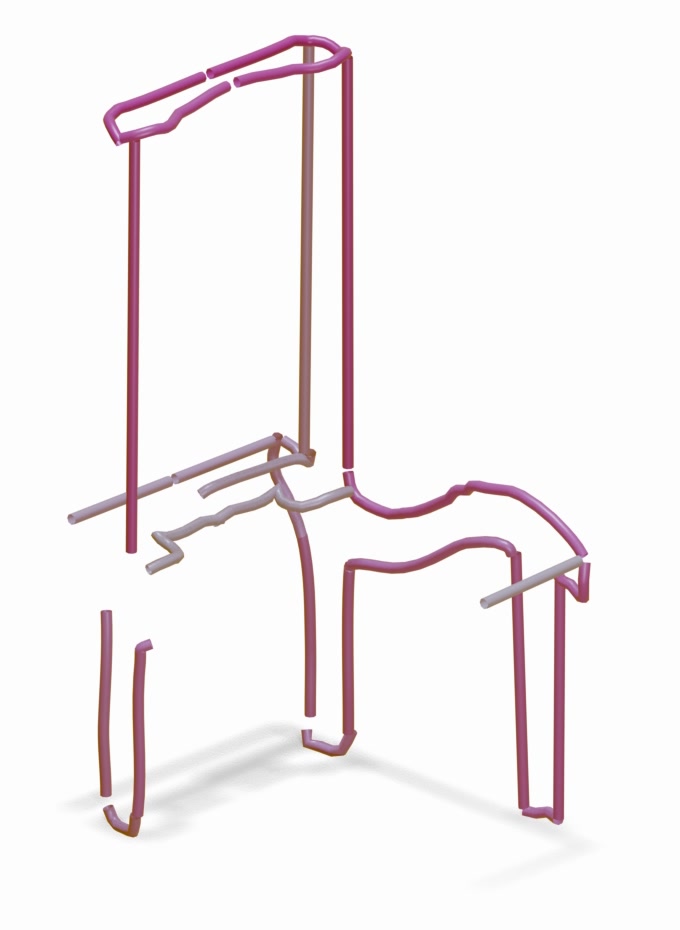}
&
\includegraphics[width=0.15\linewidth]{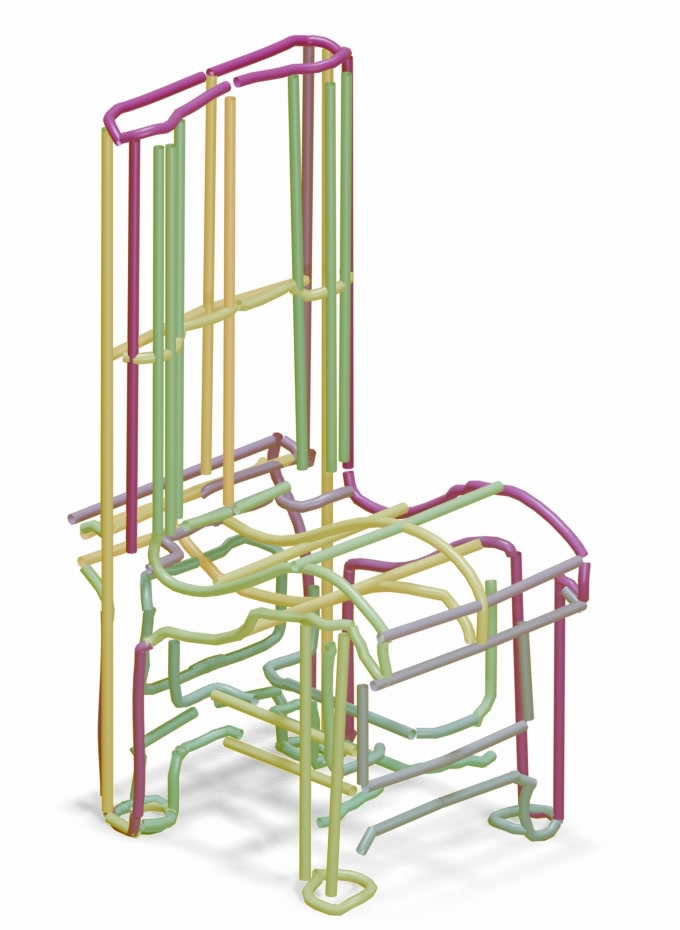}
&
\includegraphics[width=.15\linewidth]{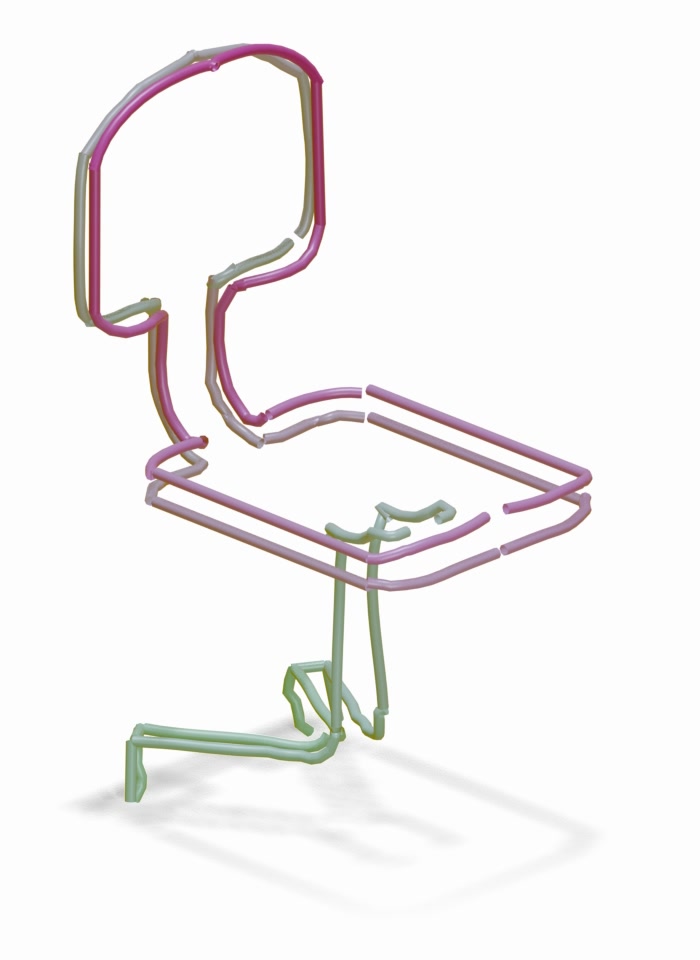}
&
\includegraphics[width=.15\linewidth]{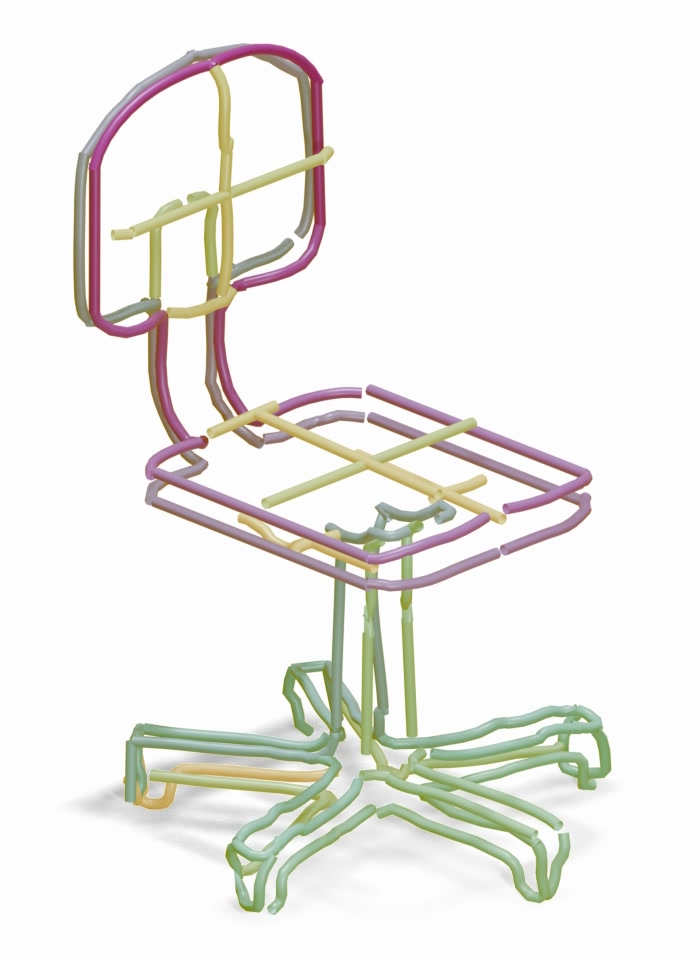}
&
\includegraphics[width=.15\linewidth]{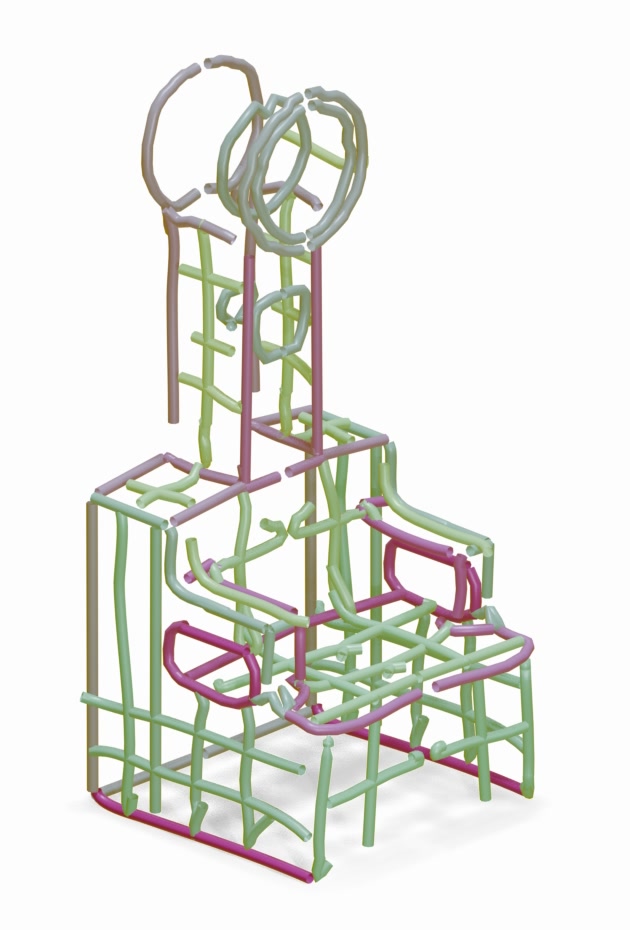}
&
\includegraphics[width=.15\linewidth]{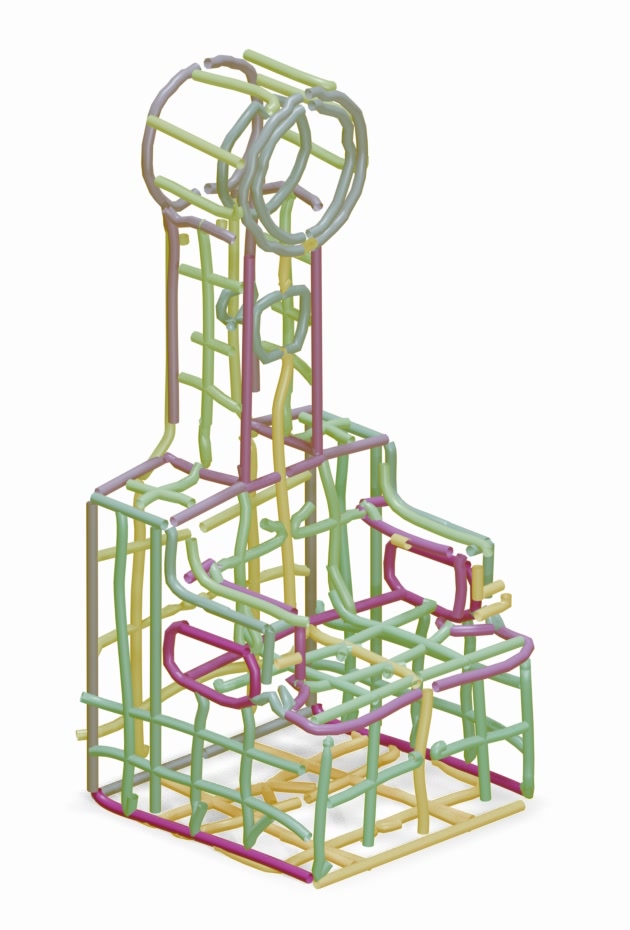}
\\[-1mm]
\rotatebox{90}{\small \bf \qquad Our prediction}
&
\includegraphics[width=.15\linewidth]{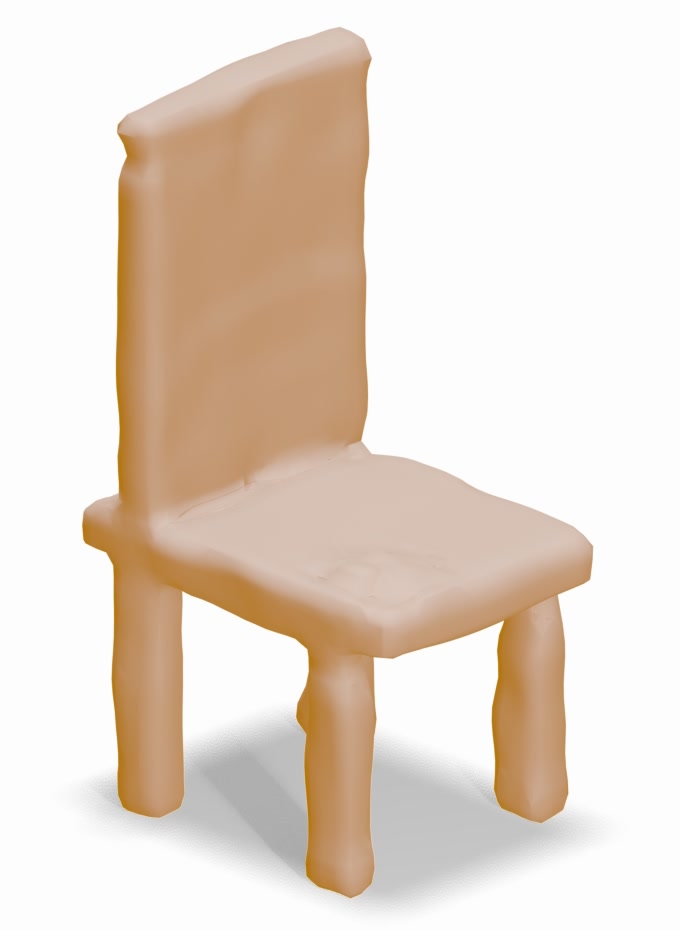}
&
\includegraphics[width=.15\linewidth]{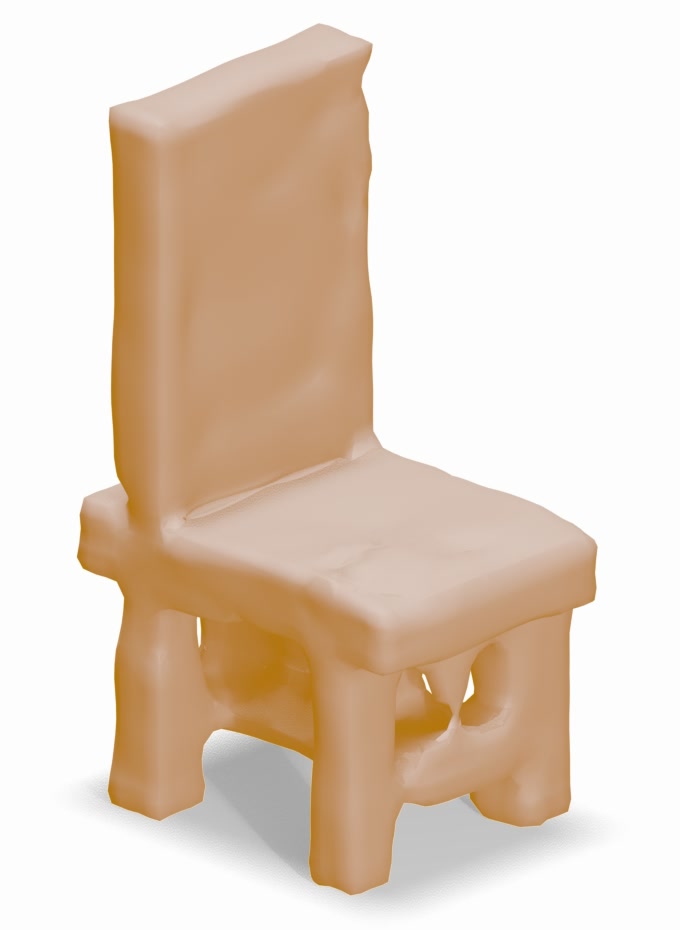}
&
\includegraphics[width=.15\linewidth]{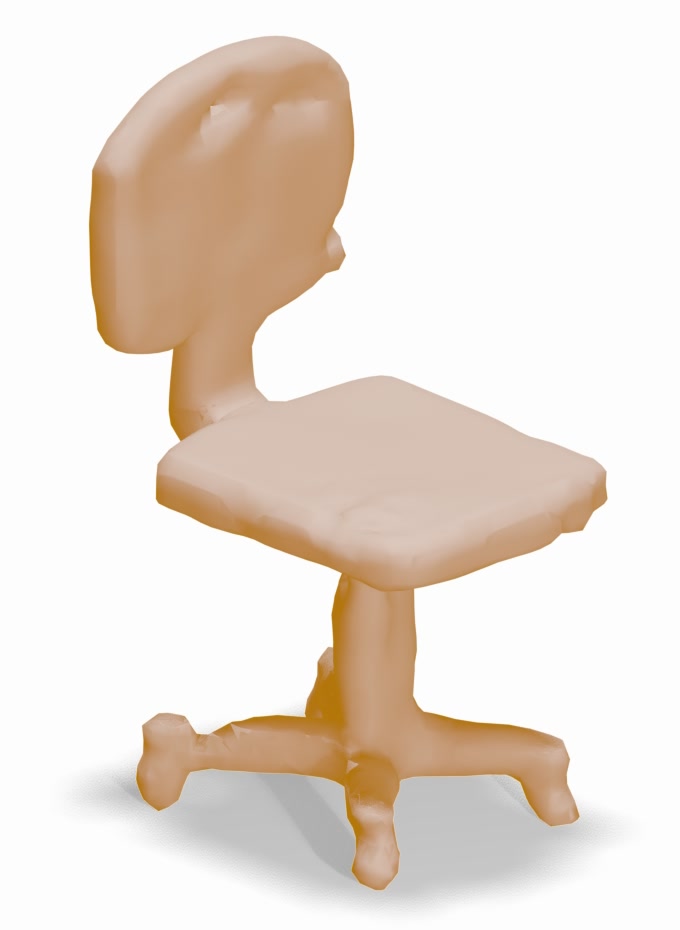}
&
\includegraphics[width=.15\linewidth]{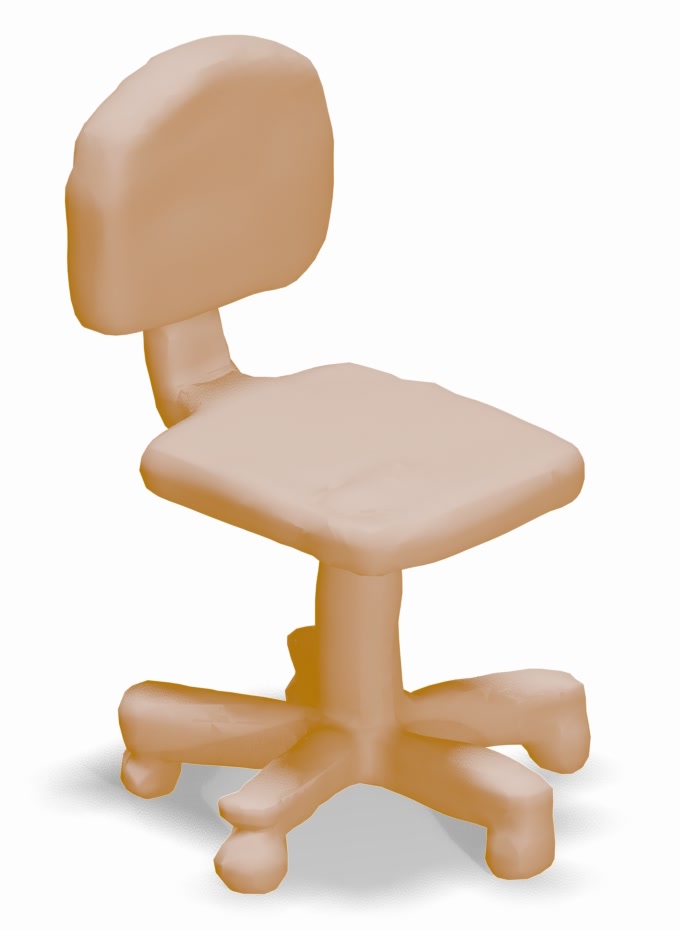}
&
\includegraphics[width=.15\linewidth]{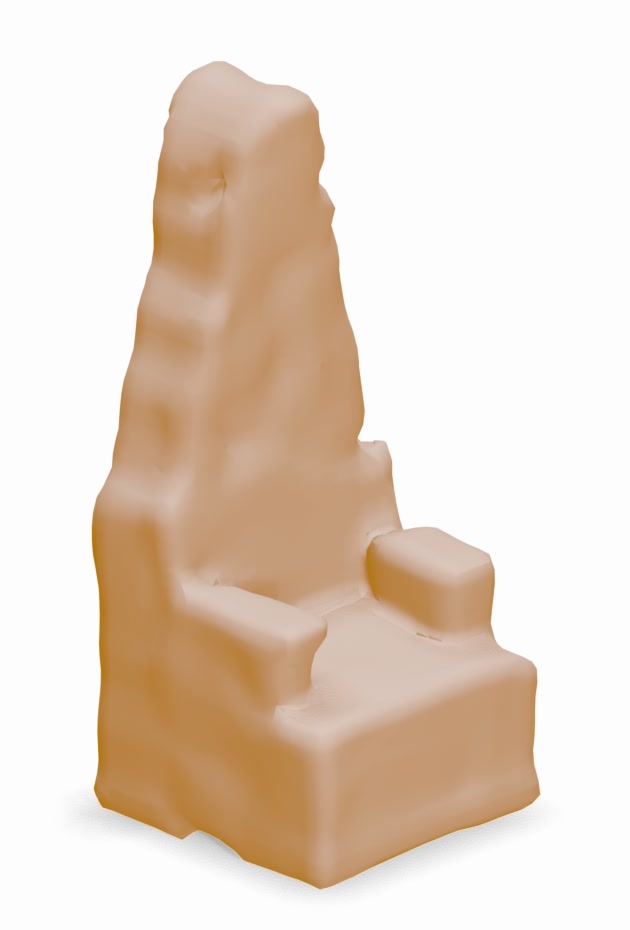}
&
\includegraphics[width=.15\linewidth]{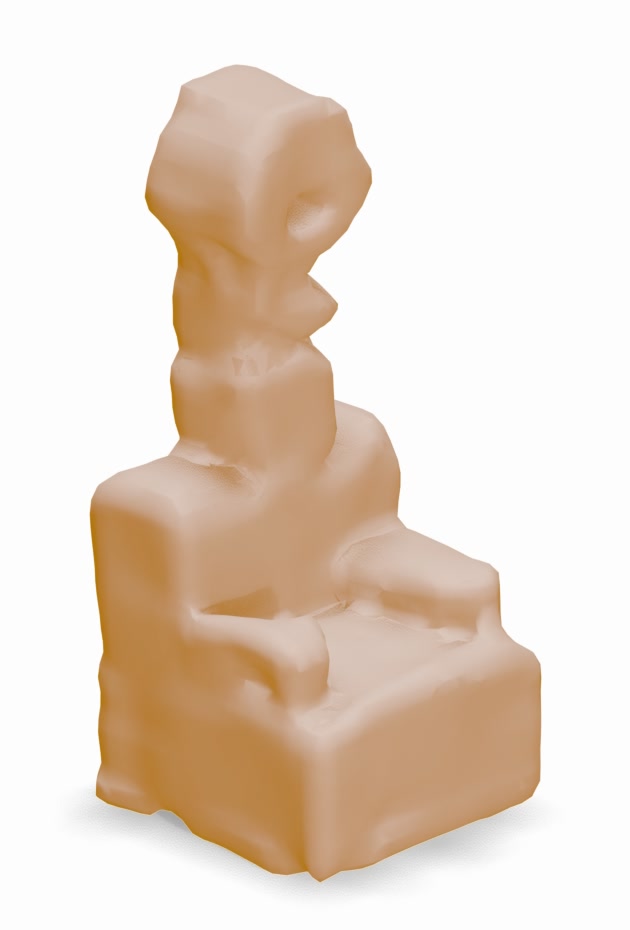}
\\[-1mm]
\rotatebox{90}{\small \bf \qquad Luo's \cite{luo20233d} prediction}
&
\begin{subfigure}{0.15\linewidth}
    \includegraphics[width=\linewidth]{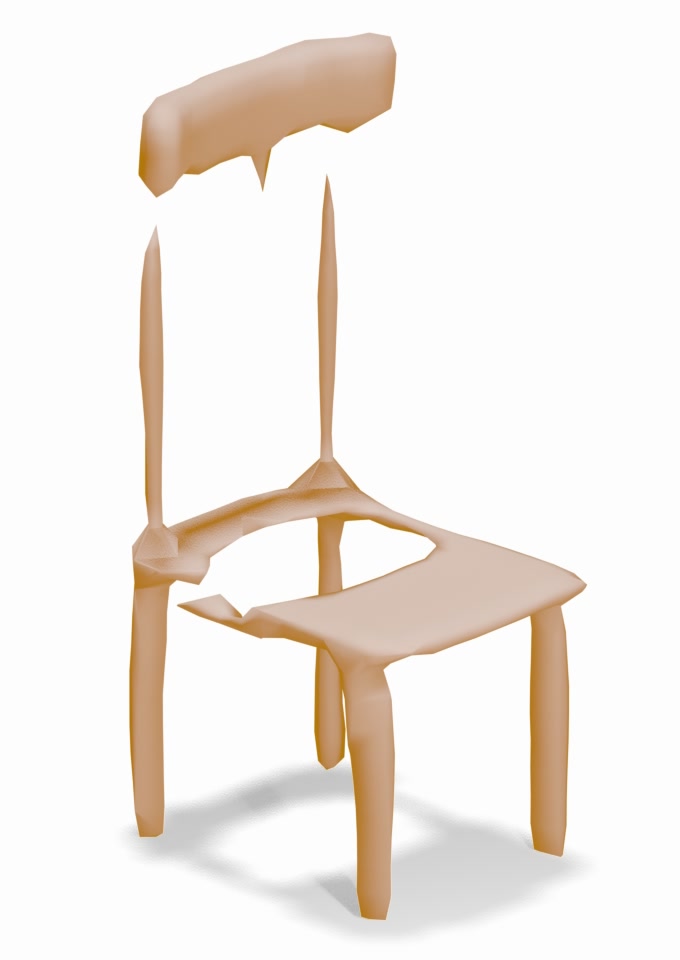}
    \caption{25\% sketch.}
    \label{fig:completion:b}
\end{subfigure}
&
\begin{subfigure}{0.15\linewidth}
    \includegraphics[width=\linewidth]{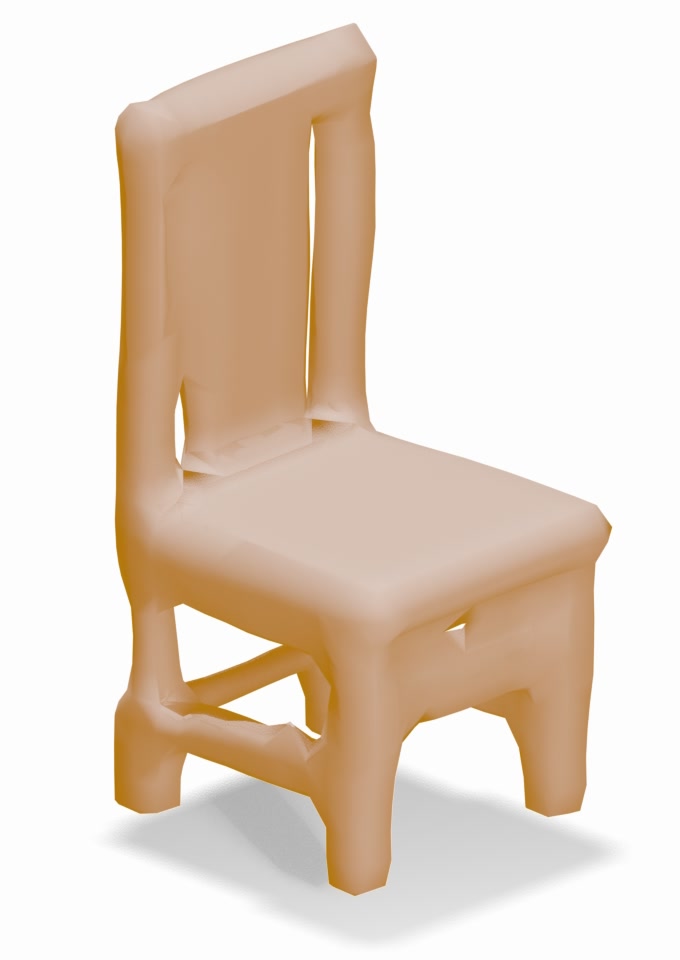}
    \caption{100\% sketch.}
    \label{fig:completion:b}
\end{subfigure}
&
\begin{subfigure}{0.15\linewidth}
    \includegraphics[width=\linewidth]{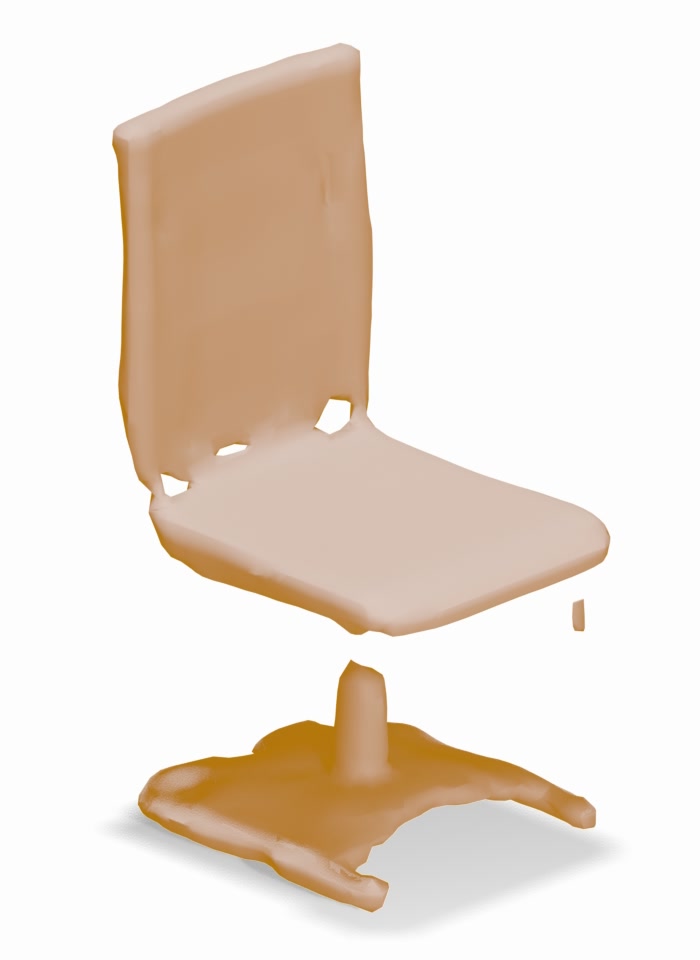}
    \caption{50\% sketch.}
    \label{fig:completion:b}
\end{subfigure}
&
\begin{subfigure}{0.15\linewidth}
    \includegraphics[width=\linewidth]{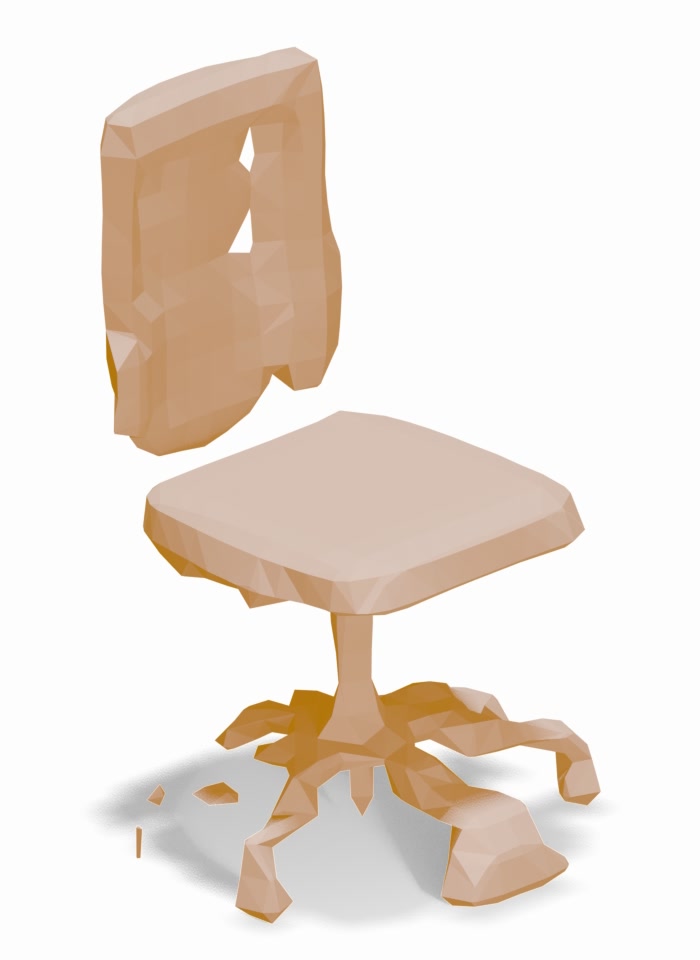}
    \caption{100\% sketch.}
    \label{fig:completion:b}
\end{subfigure}
&
\begin{subfigure}{0.15\linewidth}
    \includegraphics[width=\linewidth]{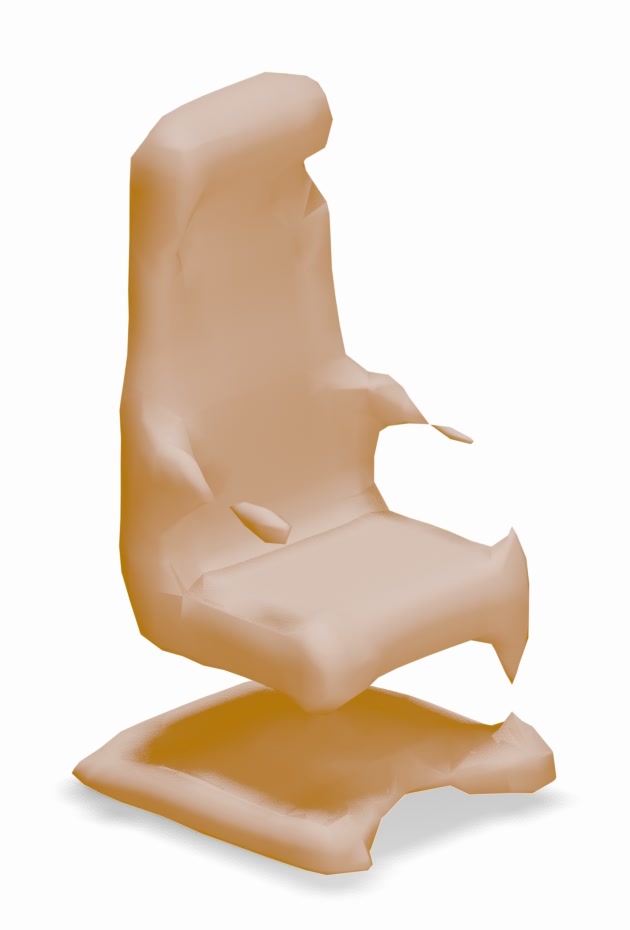}
    \caption{75\% sketch.}
    \label{fig:completion:b}
\end{subfigure}
&
\begin{subfigure}{0.15\linewidth}
    \includegraphics[width=\linewidth]{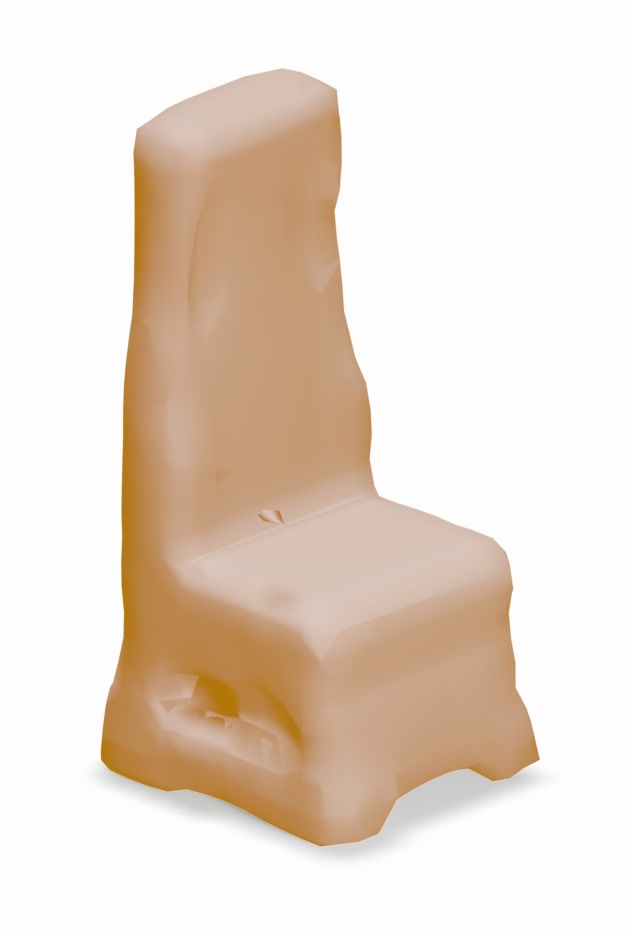}
    \caption{100\% sketch.}
    \label{fig:completion:b}
\end{subfigure}
\end{tabular}
    \vspace{-3mm}
    \caption{{\bf Sketch Completion Results.} 
Our model infers coherent 3D shapes even from highly partial sketches. 
As more strokes are provided, reconstructions become increasingly detailed and faithful to the target geometry.
    }
    \label{fig:suppl_completion}
\end{figure*}

We report completion results in \cref{fig:suppl_completion}.
Even when given only 25–50\% of the original stroke sequence, the model infers coherent geometry and progressively refines the structure as more strokes are revealed, confirming its strong internal shape priors and robustness to missing sketch information. In particular, the model is able to exploit the geometry of shapes to complete un-sketched parts, such as missing chair legs.

\begin{figure}[t]
    \centering
         \resizebox{\linewidth}{!}{
\begin{tikzpicture}
  \begin{axis}[
      width=1\linewidth,
      height=4.5cm,
      xlabel={Portion of points kept (\%)},
      ylabel={\textcolor{colCD}{$\leftarrow$ CD $\times1000$}},
      ymin=0, ymax=80,
      xtick=data,
      symbolic x coords={25,50,75,100},
      ytick={0,20,40,60,80},
      axis y line*=left,
      axis x line=bottom,
      yticklabel style={colCD},
      xlabel style={yshift=0ex},
      ylabel style={xshift=0em},
      enlargelimits=0.00,
      legend style={fill=none, at={(0.035,1)}, anchor=north west, legend columns=-1, inner sep=0pt},
      legend cell align=left,
  ]
    \addplot[colCD, dashed, thick, mark=*, mark size=2.2pt, mark options={solid}, forget plot] coordinates {
      (25,69.9) (50,25.9) (75,14.8) (100,14.2)
    };
    
    \addplot[colCD, thick, mark=diamond*, mark size=2.5pt, mark options={solid}, forget plot] coordinates {
      (25,52.5) (50,14.4) (75,4.6) (100,4.0)
    };

     \addlegendimage{black, thick, mark=diamond*}
    \addlegendentry{Ours}

    \addlegendimage{black, dashed, thick, mark=*, mark options={solid}}
    \addlegendentry{Luo \cite{luo20233d}}

  \end{axis}

  \begin{axis}[
      width=1\linewidth,
      height=4.5cm,
      axis y line*=right,
      axis x line=none,
      ylabel={\textcolor{colFONE}{F-score $\rightarrow$}},
      ymin=20, ymax=70,
      symbolic x coords={25,50,75,100},
      ytick={20,40,60},
      yticklabel style={colFONE},
      ylabel near ticks,
      enlargelimits=0.00,
      legend cell align=left,
  ]
    \addplot[colFONE, dashed, thick, mark=*, mark size=2.2pt, mark options={solid}] coordinates {
      (25,23.6) (50,33.7) (75,38.2) (100,37.1)
    };

    \addplot[colFONE, thick, mark=diamond*, mark size=2.5pt, mark options={solid}] coordinates {
      (25,37.2) (50,57.9) (75,64.0) (100,64.3)
    };
  \end{axis}
\end{tikzpicture}
}
         \vspace{-7mm}
        \caption{{\bf Sketch Completion Performance.} Performance remains high even when only partial sketches are provided.
        }
        \label{fig:completion_graph}
\end{figure}
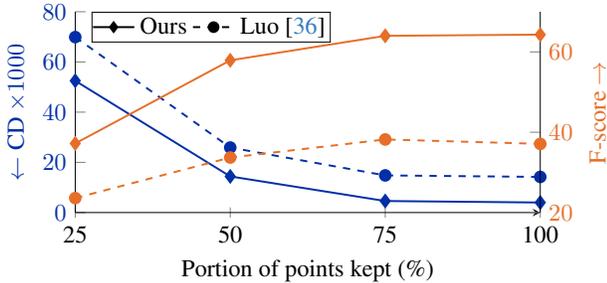

As reported in \cref{fig:completion_graph}, our model is able to reconstruct faithful shapes even from partial inputs, clearly outperforming point cloud–based baselines.
Interestingly, our model is able to reach near-maximum performance with only the first half of the drawn points. 
This trend reflects how annotators typically first draw global outlines of the shape before adding finer details.

\begin{table}[t]
    \caption{{\bf Additional Ablation.}} \vspace{-2mm}
    \label{tab:ablation_rebuttal}
    \centering
    \small{
    \begin{tabular}{lll cc} \toprule
    & Experiment & F-score $\uparrow$ &  CD$\tiny{\times 1000}$ $\downarrow$\\\midrule
     & Best & \bf 56.8 & \bf 5.1\\\greyrule
      \textsc{a}  & No 3D Fourrier & 52.1 & 5.6\\
      \textsc{a}  & No 1D Fourrier & 48.2 & 6.3 \\\greyrule
      \textsc{b} & BFS traversal & 47.8 & 6.4 \\
      \textsc{c} &  Reverse Order & 56.2 & \bf 5.0\\
      \textsc{c}  & Scrambled Strokes & 54.6 & 5.9 \\
      \textsc{c}  & Scrambled Points & 52.2 & 5.2
      \\\bottomrule
    \end{tabular}}
\end{table}

\section{Impact of Sketch Snapping}
\label{sec:suppl_unsnap}

Our surface-snapping tool helps users draw geometrically accurate sketches directly on reference shapes.
Because snapped sketches adhere to the underlying surface and remove user-induced alignment noise, they provide cleaner supervision during training and yield more reliable evaluations.
A natural concern, however, is whether models trained on such clean, snapped sketches might overfit to this idealized scenario and fail to generalize to sketches produced in the wild, without snapping assistance.

\begin{figure*}[h]
    \centering
    \begin{tabular}{c@{\,}c@{\,}c@{\,}c@{\,}c}
\includegraphics[width=0.16 \linewidth]{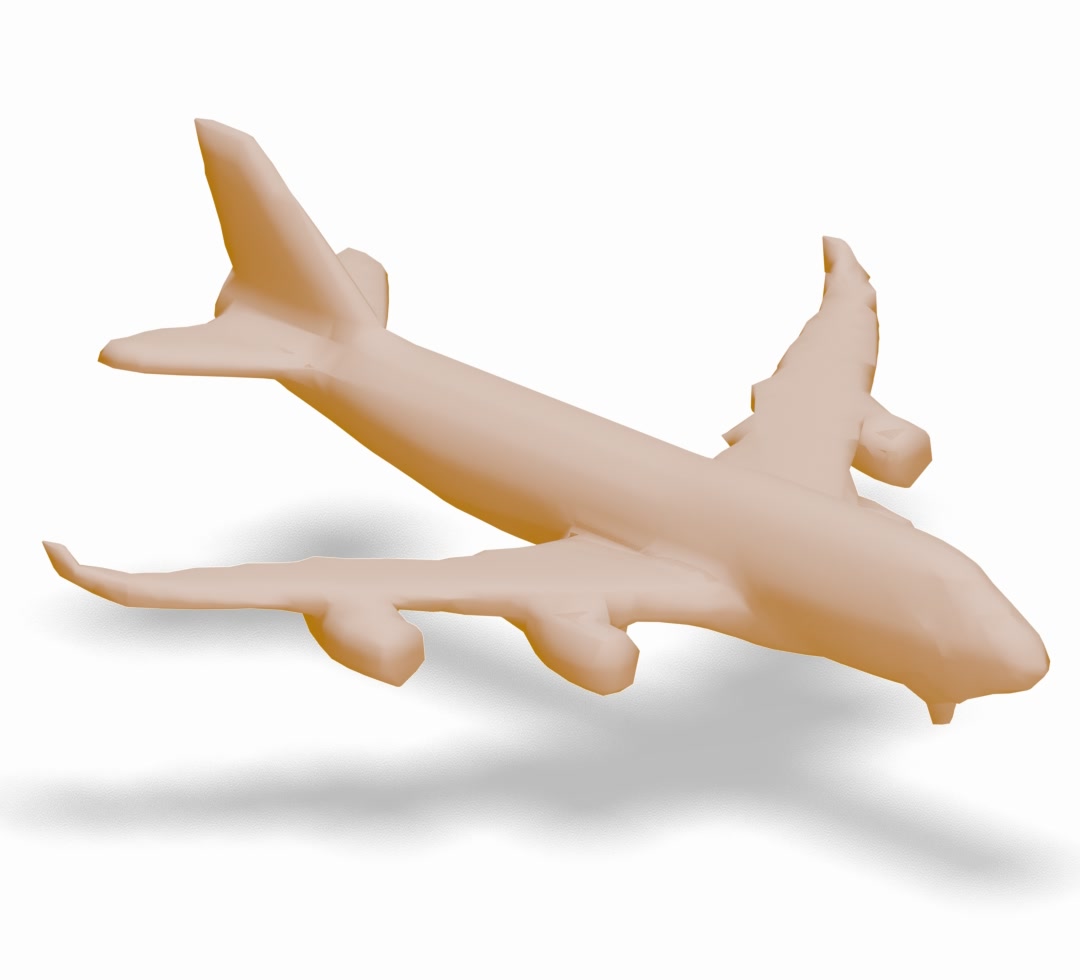}
&
\includegraphics[width=0.16\linewidth]{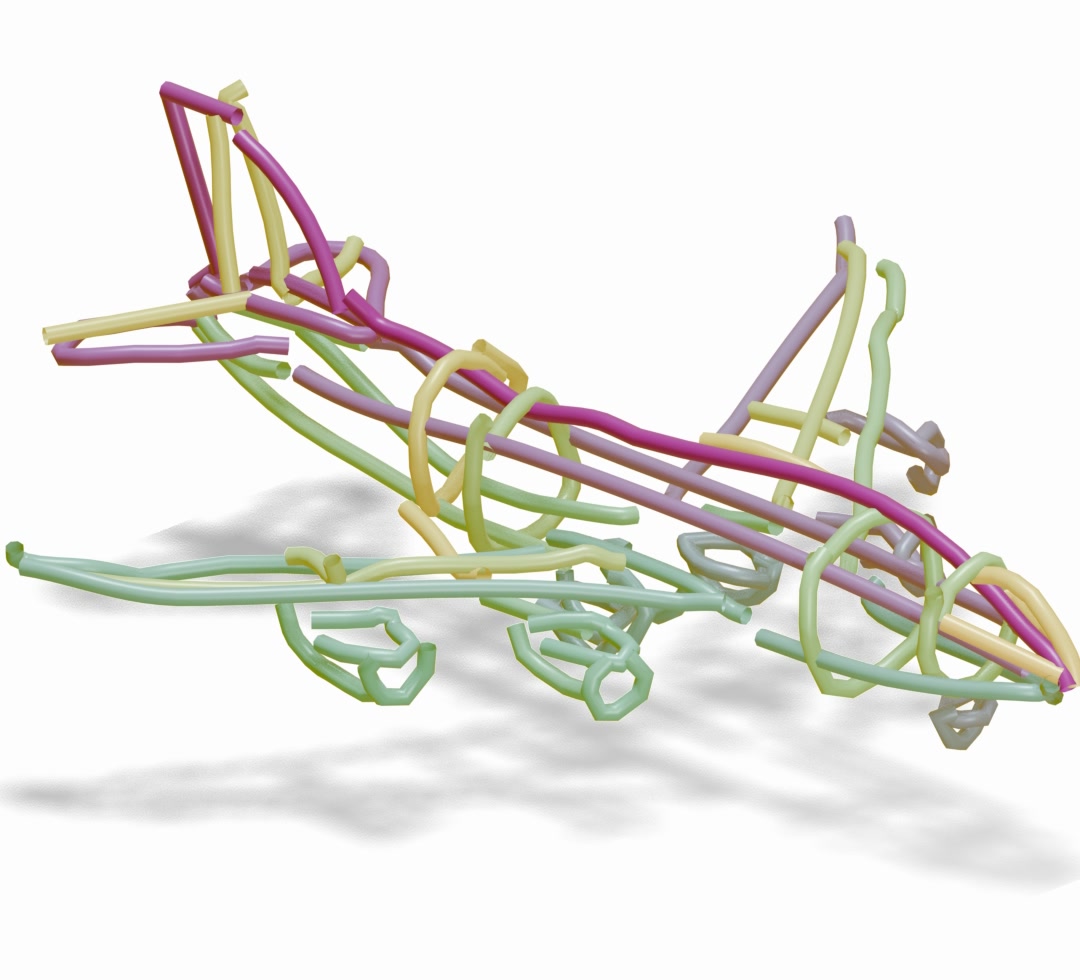}
&
\includegraphics[width=0.16\linewidth]{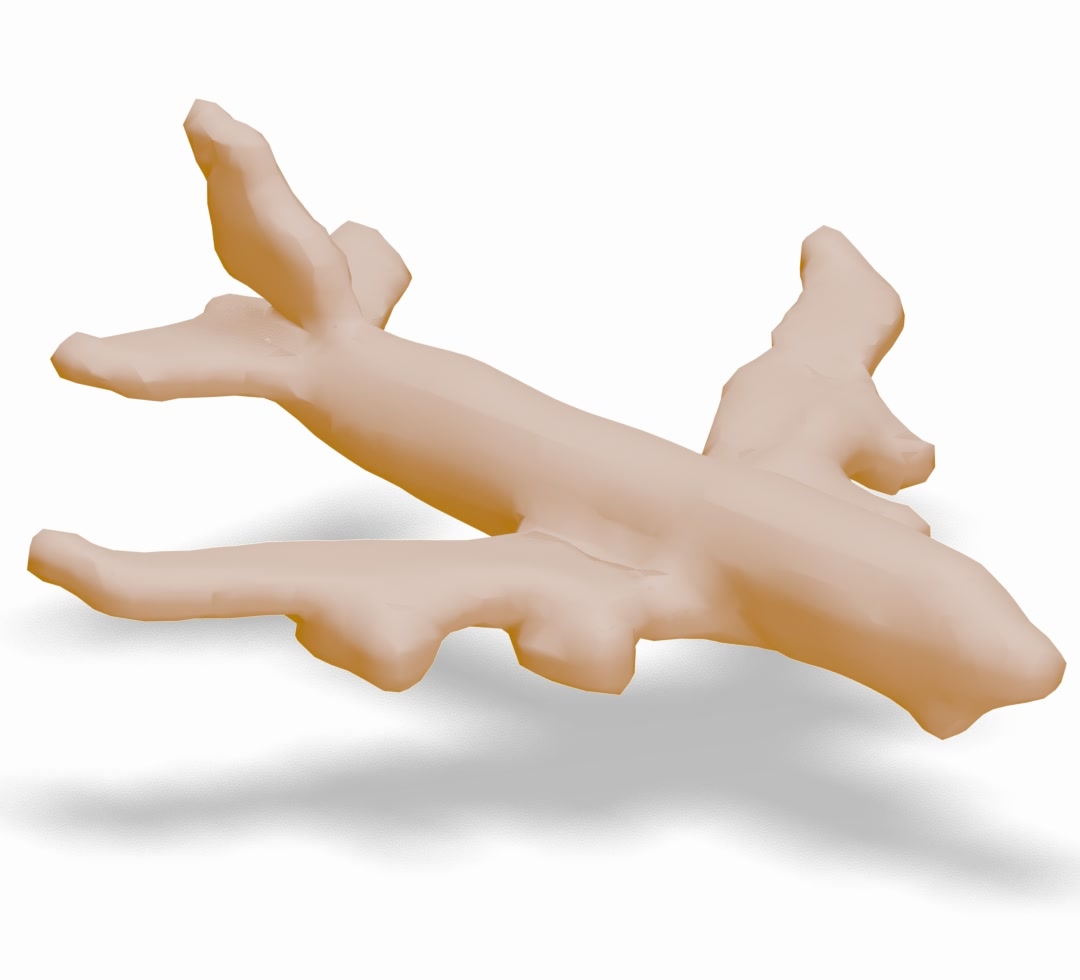}
&
\includegraphics[width=.16\linewidth]{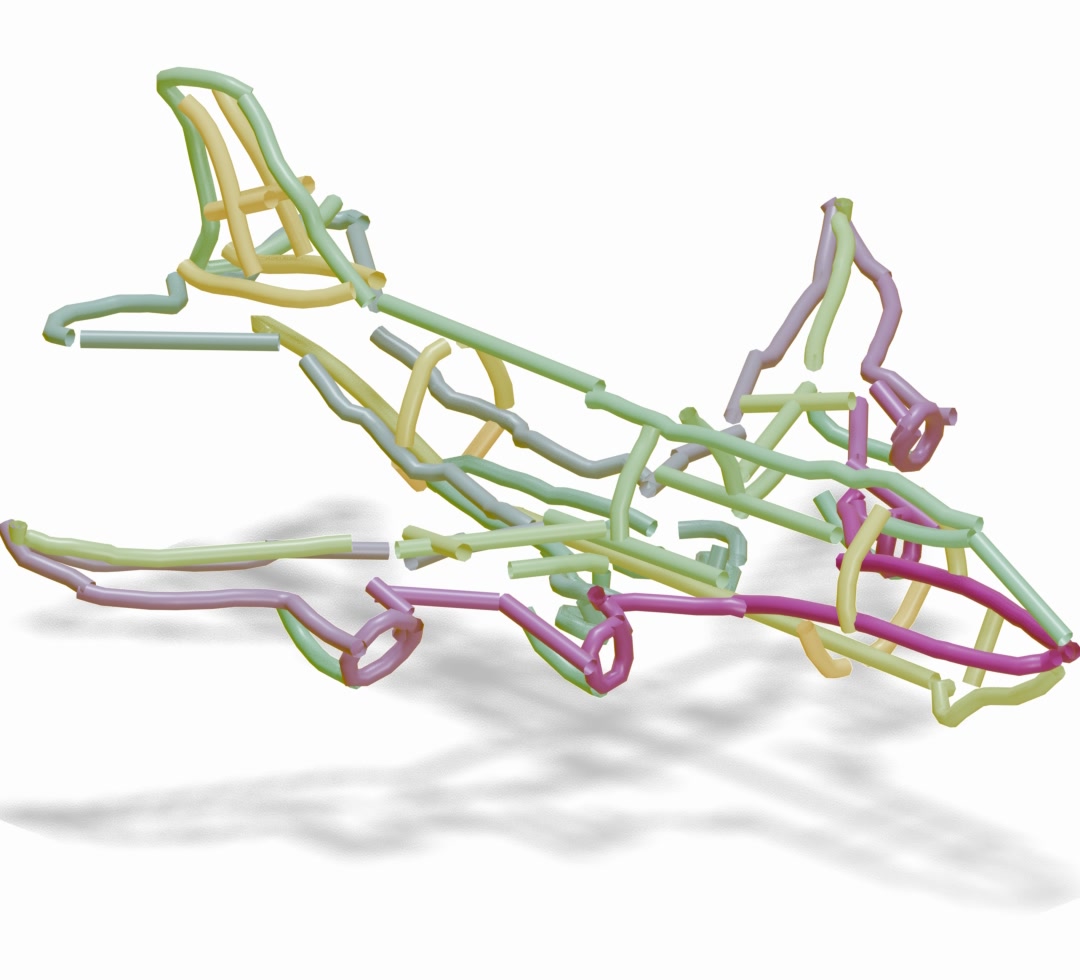}
&
\includegraphics[width=.16\linewidth]{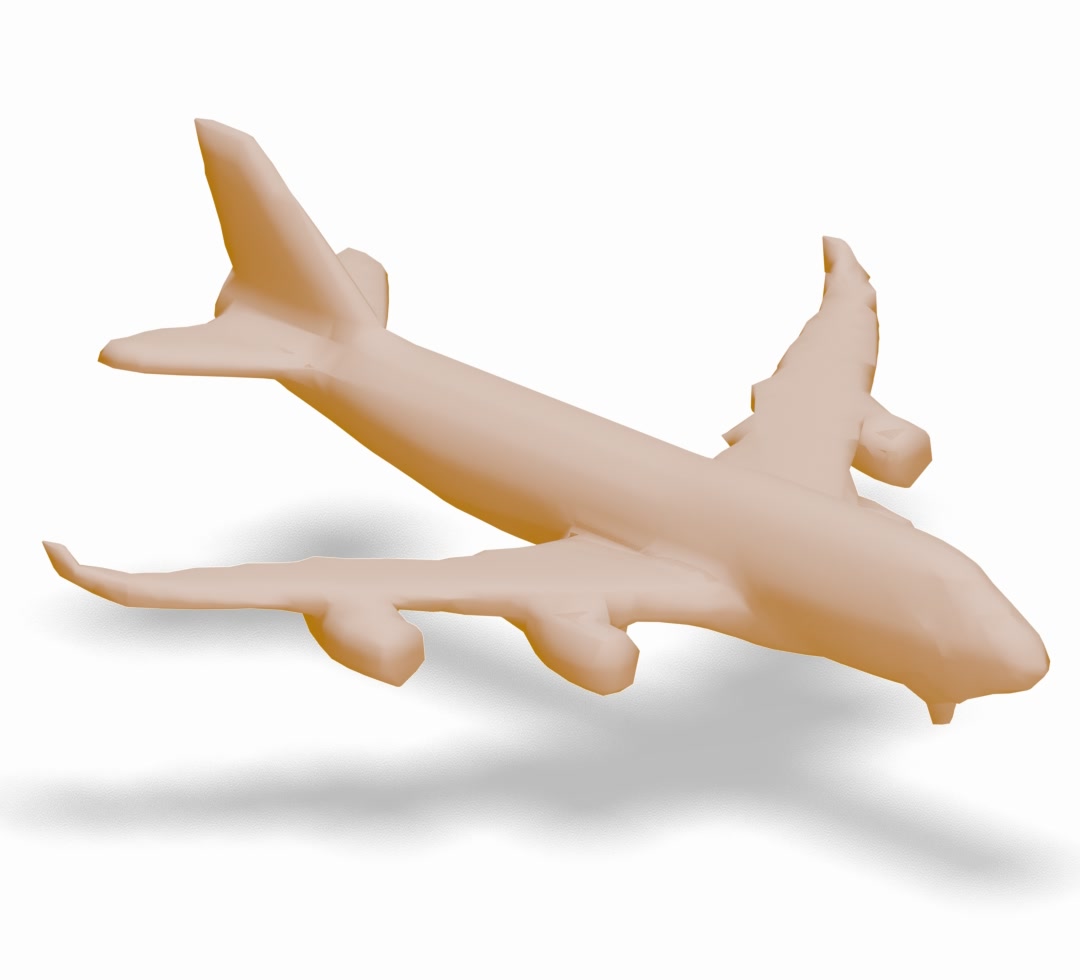}
\\[-1mm]
\includegraphics[width=.16\linewidth]{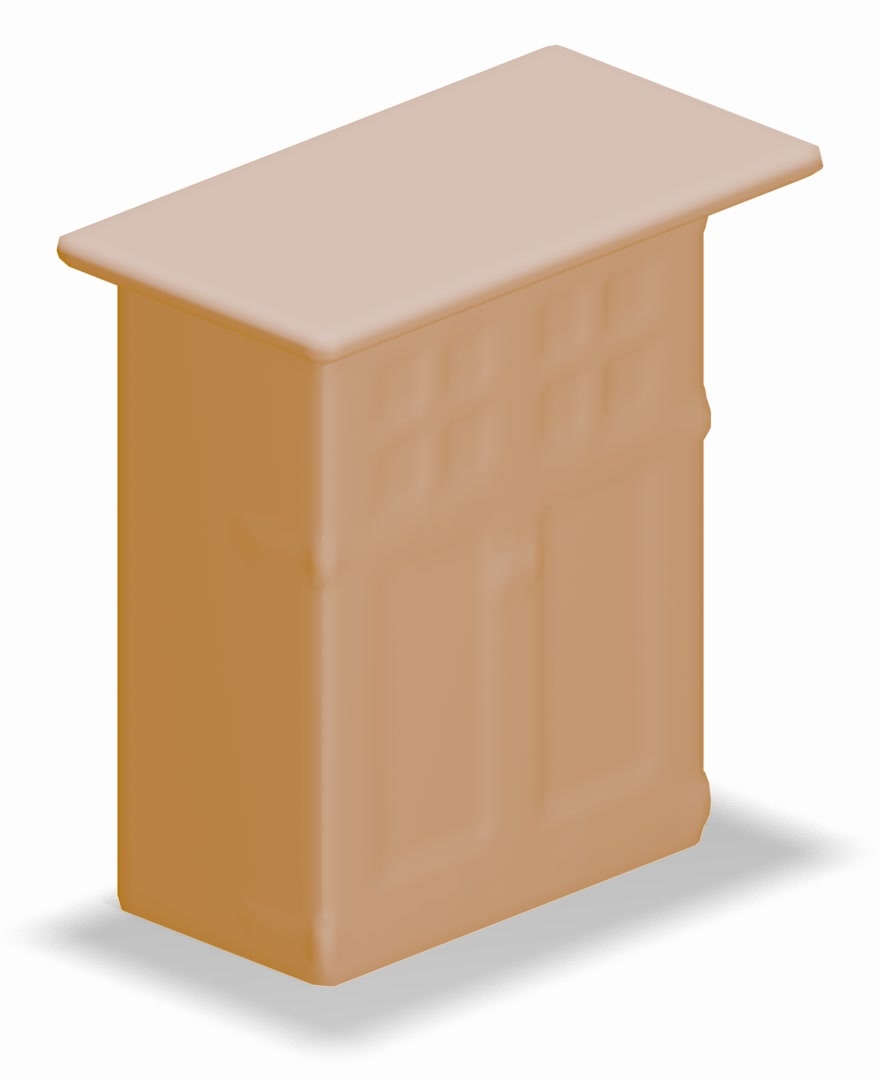}
&
\includegraphics[width=.16\linewidth]{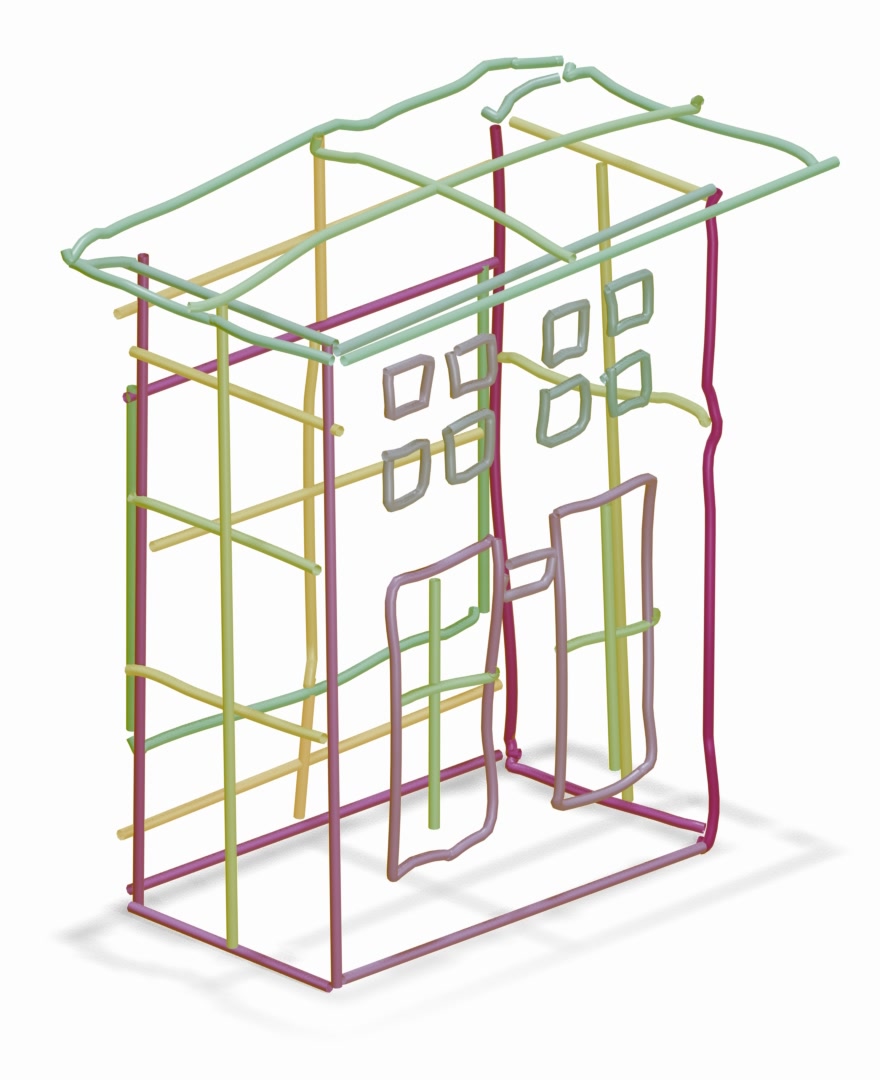}
&
\includegraphics[width=.16\linewidth]{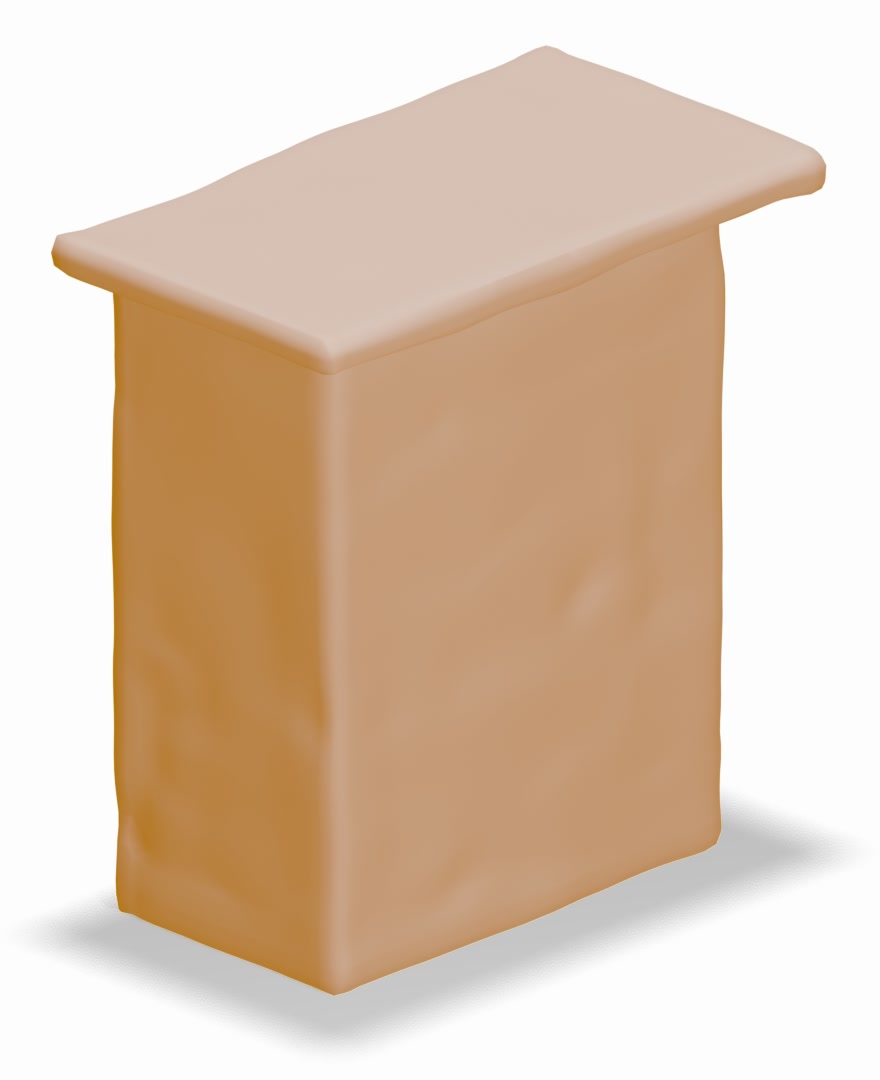}
&
\includegraphics[width=.16\linewidth]{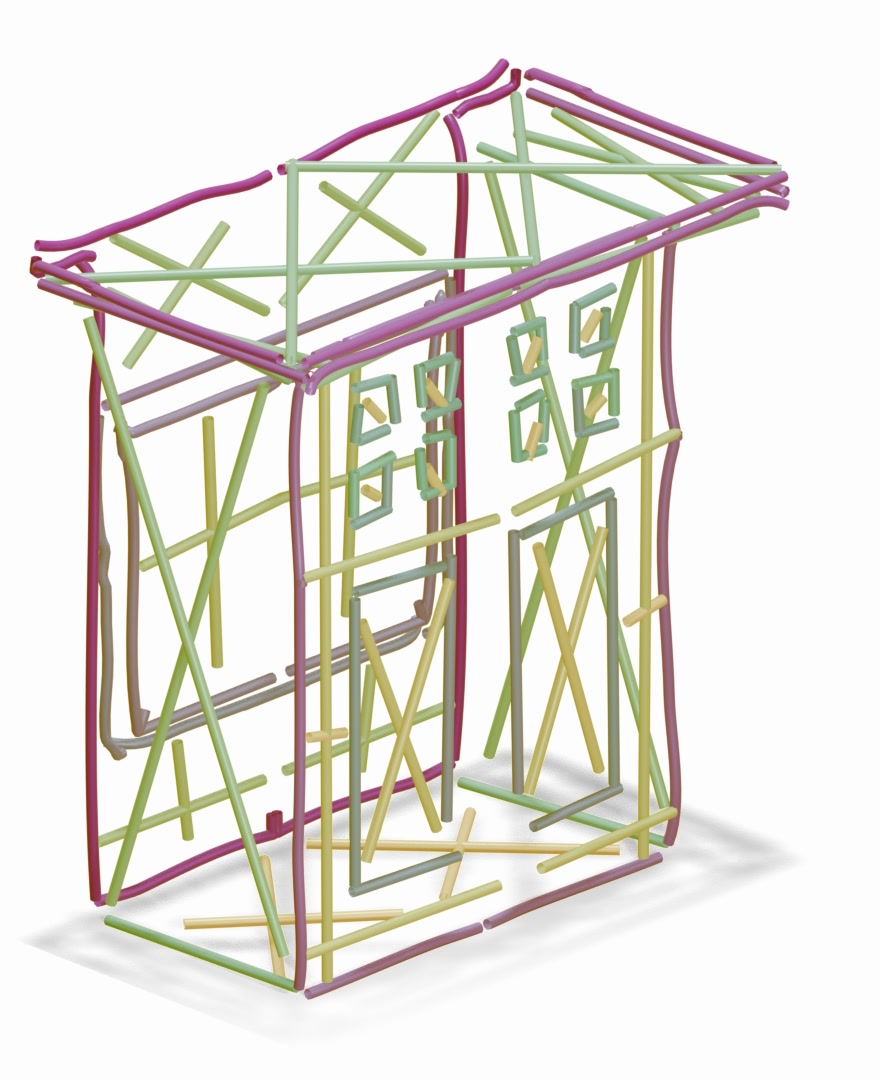}
&
\includegraphics[width=.16\linewidth]{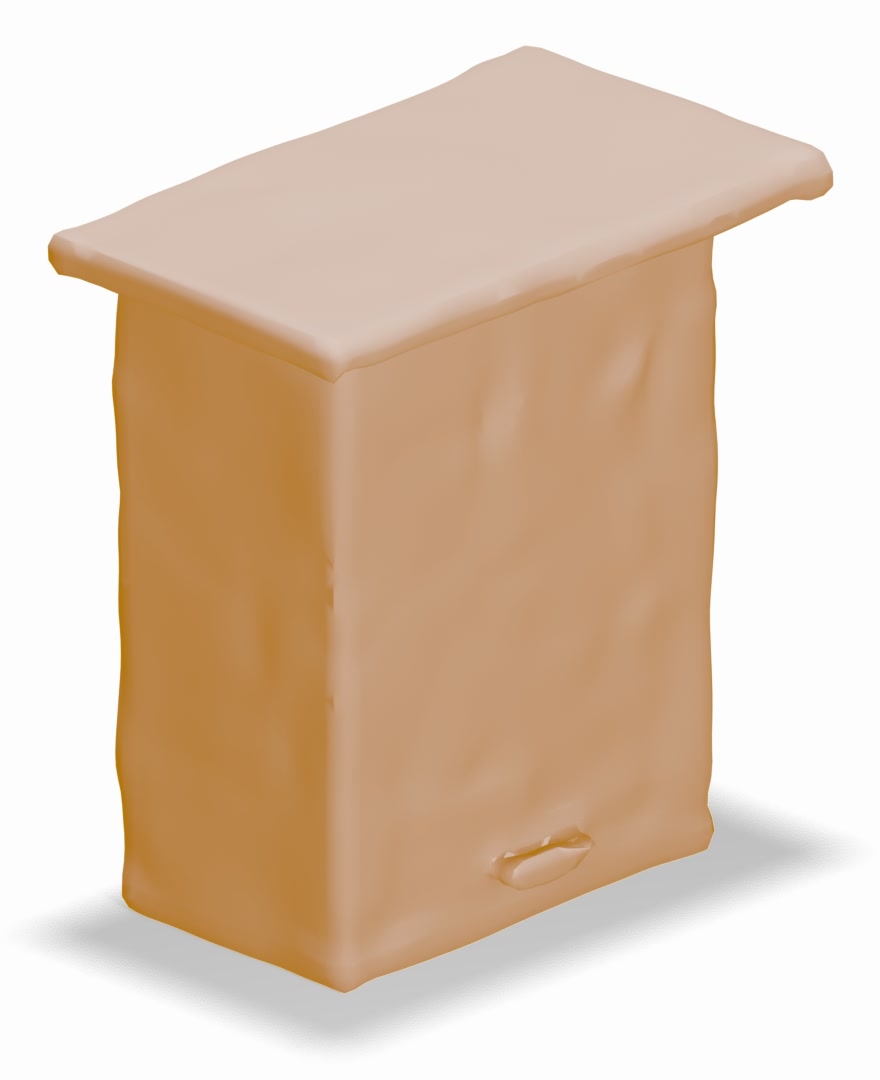}
\\[-1mm]
\includegraphics[width=.16\linewidth]{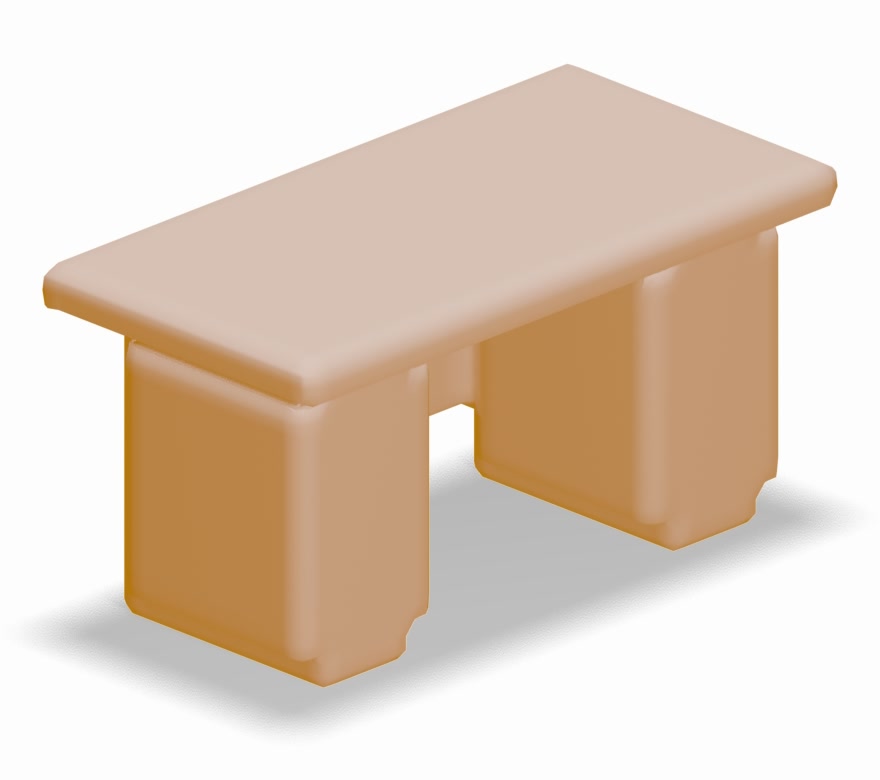}
&
\includegraphics[width=.16\linewidth]{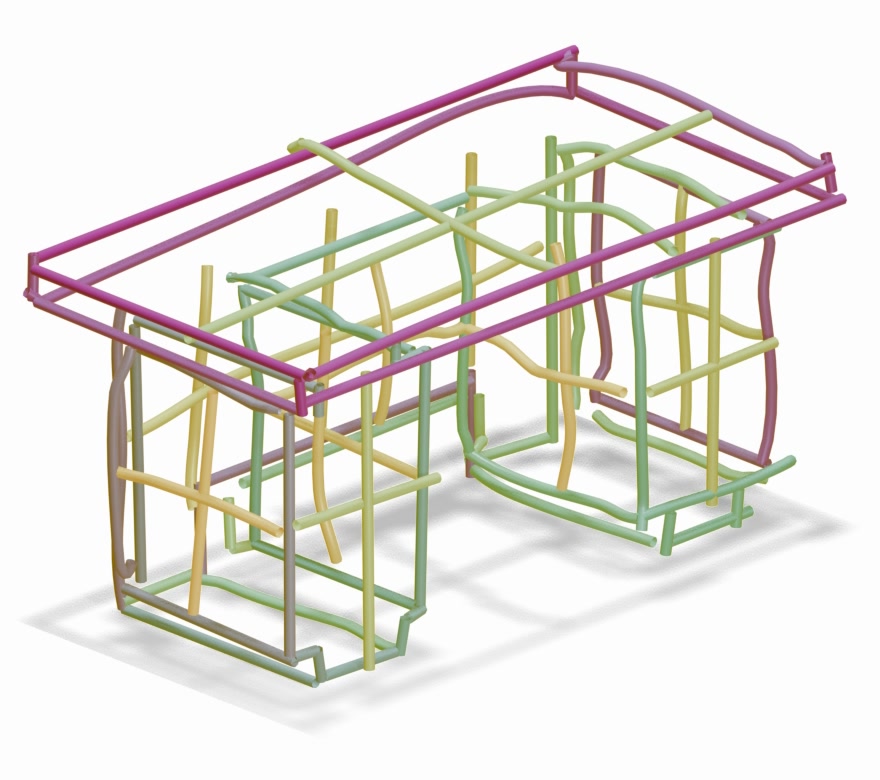}
&
\includegraphics[width=.16\linewidth]{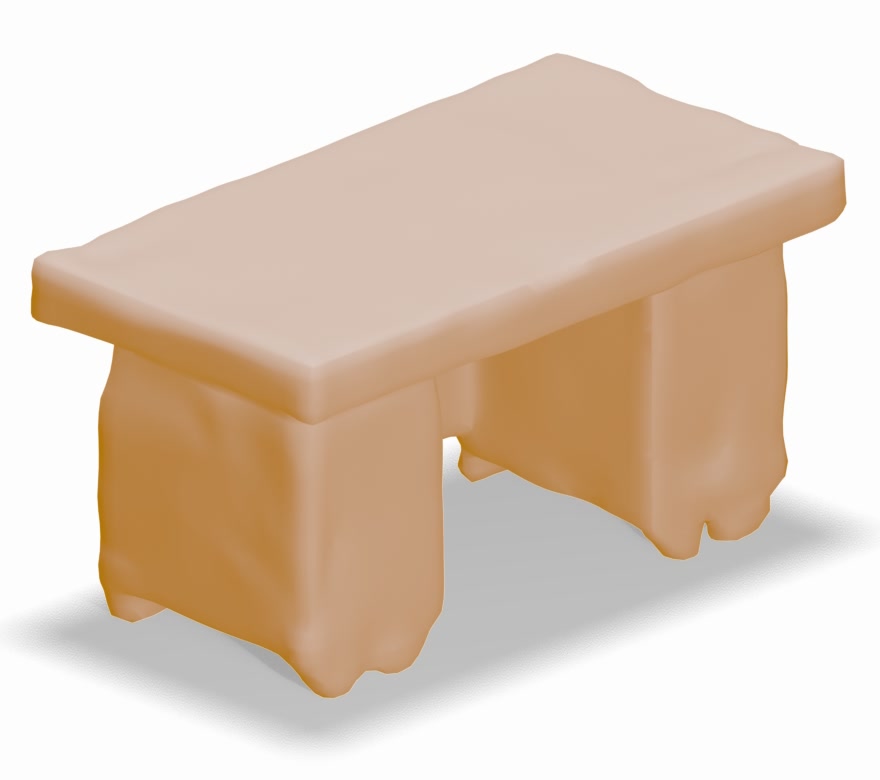}
&
\includegraphics[width=.16\linewidth]{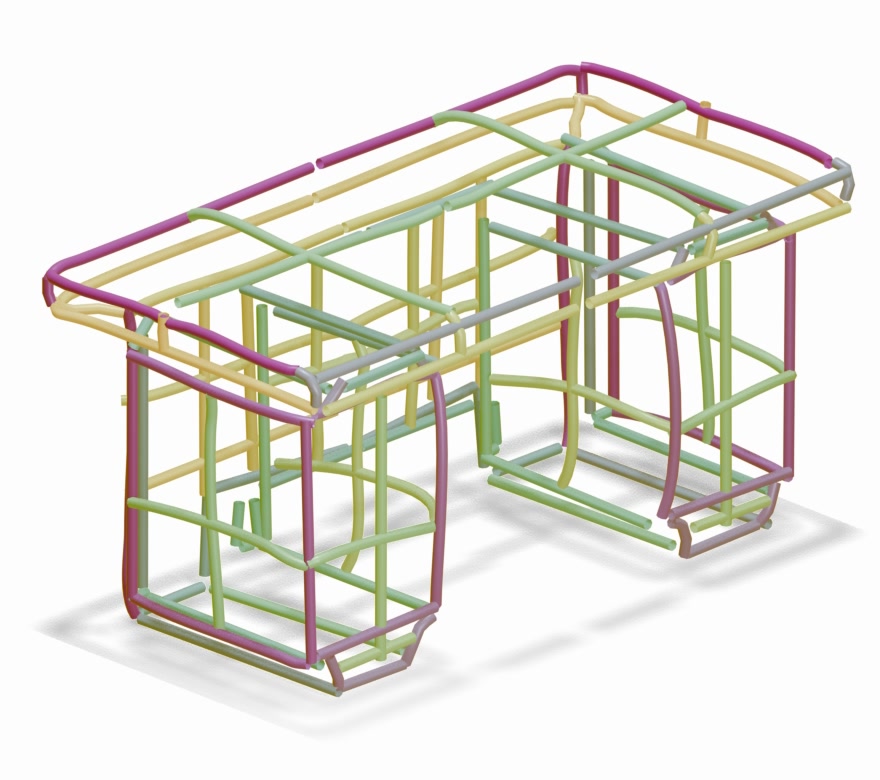}
&
\includegraphics[width=.16\linewidth]{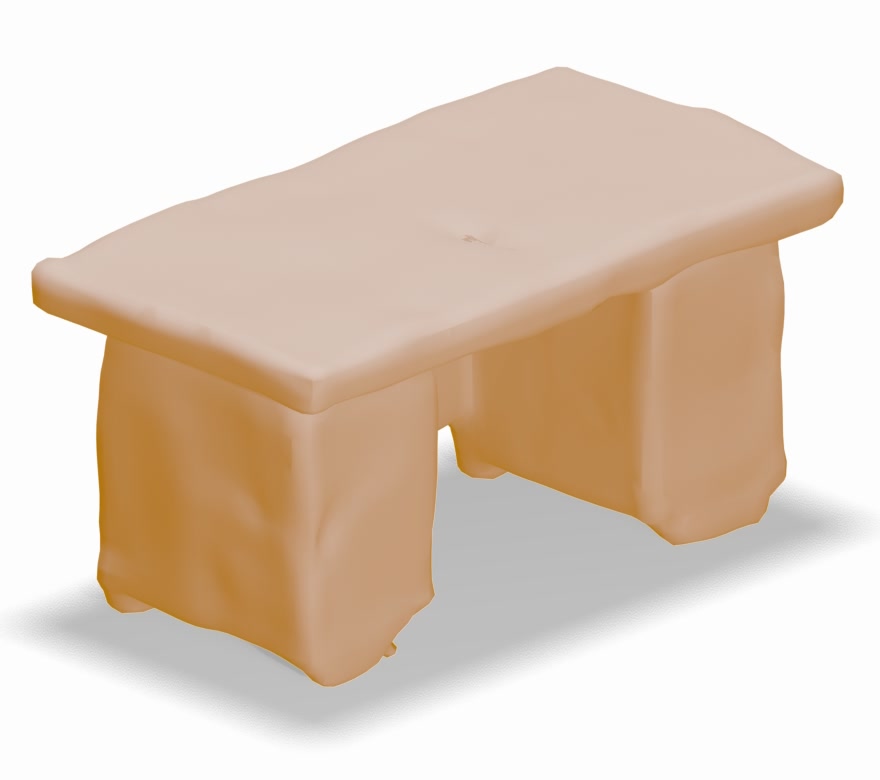}
\\[-1mm]
\begin{subfigure}{0.16\linewidth}
    \includegraphics[width=\linewidth]{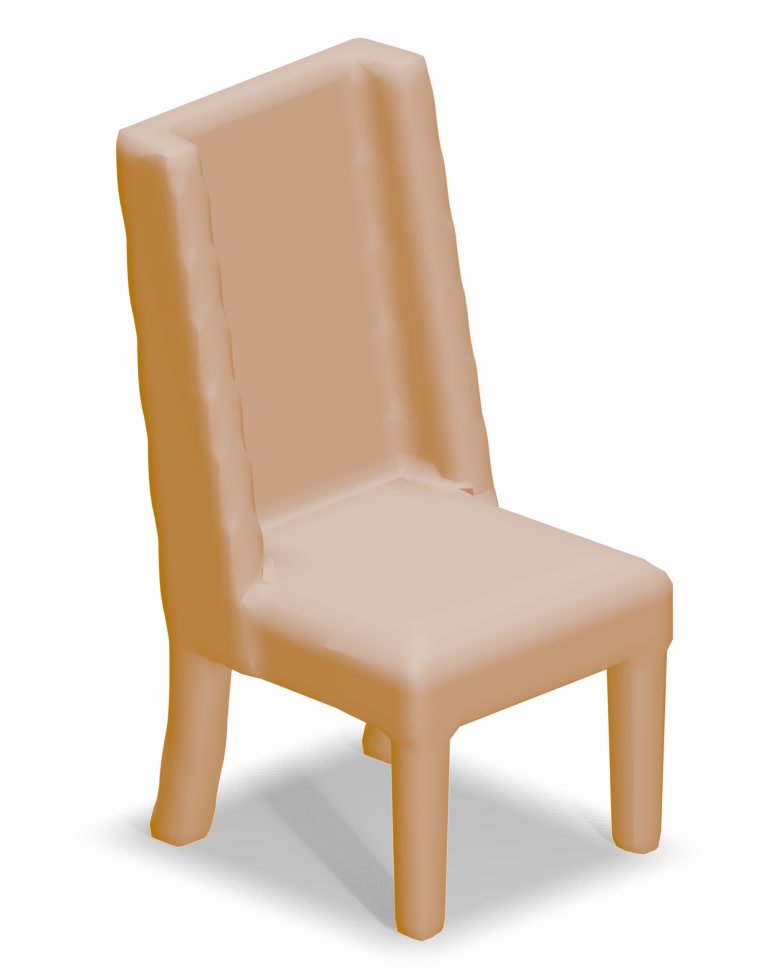}
    \caption{GT shape.}
    \label{fig:completion:b}
\end{subfigure}
&
\begin{subfigure}{0.16\linewidth}
    \includegraphics[width=\linewidth]{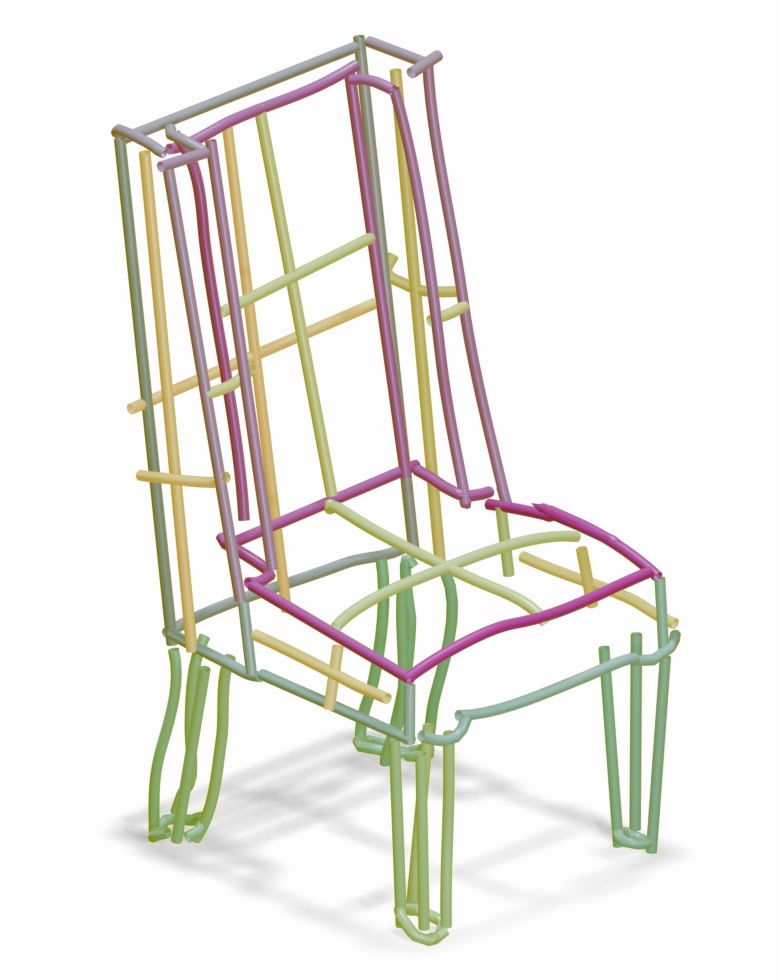}
    \caption{Unsnapped sketch.}
    \label{fig:completion:b}
\end{subfigure}
&
\begin{subfigure}{0.16\linewidth}
    \includegraphics[width=\linewidth]{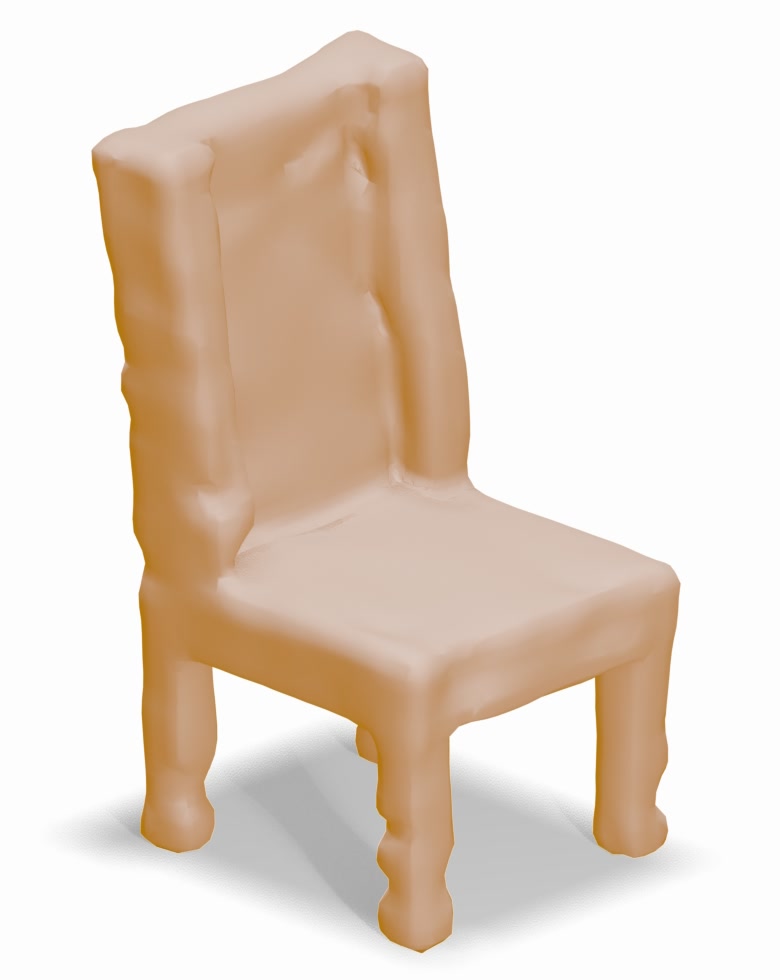}
    \caption{Our prediction.}
    \label{fig:completion:b}
\end{subfigure}
&
\begin{subfigure}{0.16\linewidth}
    \includegraphics[width=\linewidth]{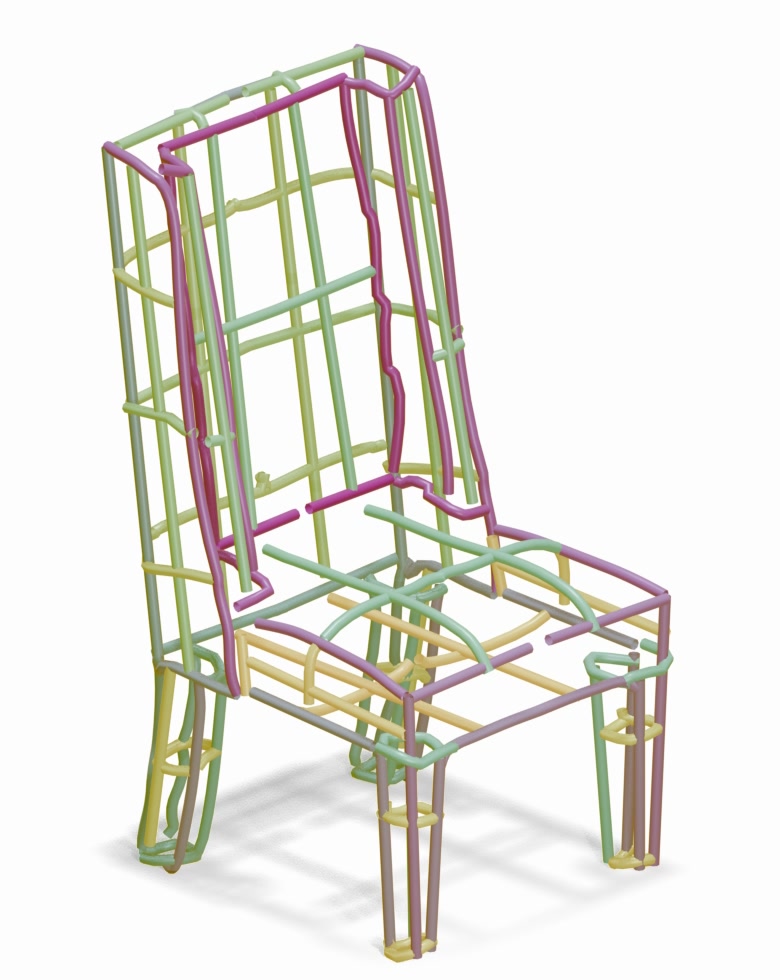}
    \caption{Snapped sketch.}
    \label{fig:completion:b}
\end{subfigure}
&
\begin{subfigure}{0.16\linewidth}
    \includegraphics[width=\linewidth]{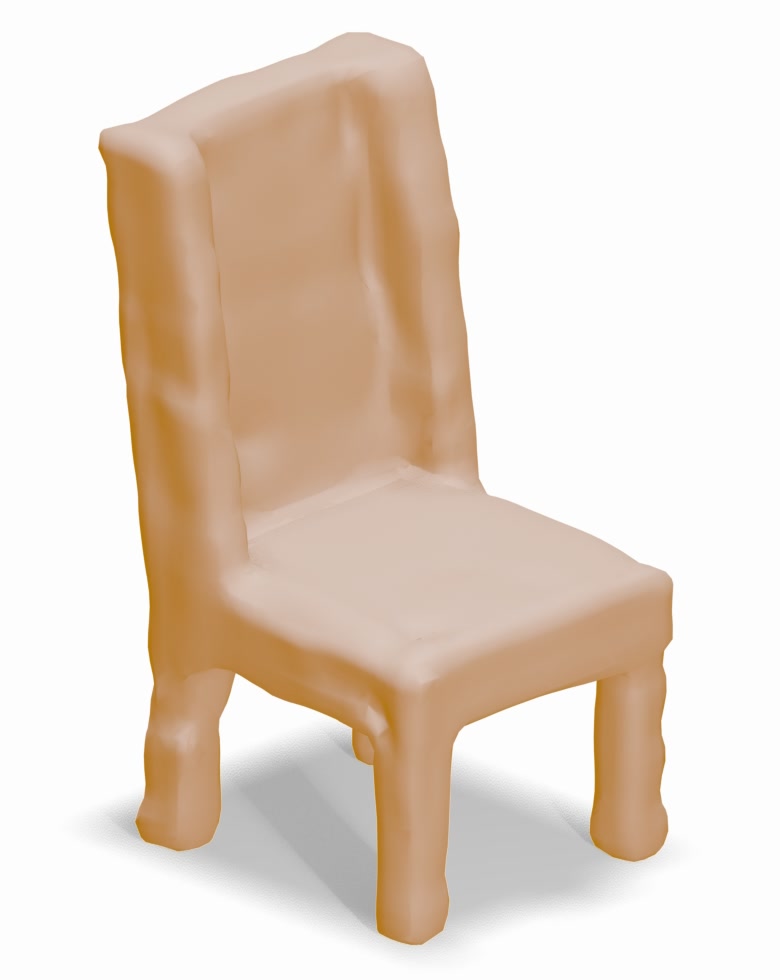}
    \caption{Our prediction.}
    \label{fig:completion:b}
\end{subfigure}
\end{tabular}
    \vspace{-3mm}
    \caption{{\bf Shape Generation from Sketches Without Snapping.} Sketches drawn without our snapping tool are noticeably noisier and less geometrically accurate, but our model can still generate coherent and plausible 3D shapes from them.
    }
    \label{fig:suppl_unsnap}
\end{figure*}

To assess this, we evaluate our model on sketches drawn without snapping and present qualitative results in \cref{fig:suppl_unsnap}.
Unsnapped sketches are visibly less precise and often contain wobbling or local distortions.
Despite this domain shift, our model still produces convincing and geometrically faithful shapes, demonstrating strong robustness to deviations from perfectly aligned input sketches.


\section{Evaluation on Free-Hand Sketches}
\label{sec:suppl_freehand}
We provide additional illustrations for reconstruction of sketches drawn without references in \cref{fig:suppl_freehand}. We observe  that despite clear stylistic and geometric differences from the training sketches, our model produces coherent, detailed, and semantically meaningful shapes that align well with the intent expressed in the free-hand inputs. By contrast, Luo~\etal~\cite{luo20233d} generalizes poorly and tends to overfit to certain shape priors.  
These findings indicate that the model has learned a robust sketch-to-shape mapping rather than overfitting to the constraints of reference-guided drawing.


\begin{figure*}[h]
    \centering
    \resizebox{\linewidth}{!}{
    \begin{tabular}{l@{\,}c@{\,}c@{\,}c@{\,}c@{\,}c@{\,}c@{\,}c}
\rotatebox{90}{\scriptsize \bf \qquad  Free-hand Sketch}&
\includegraphics[height=0.12\textheight]{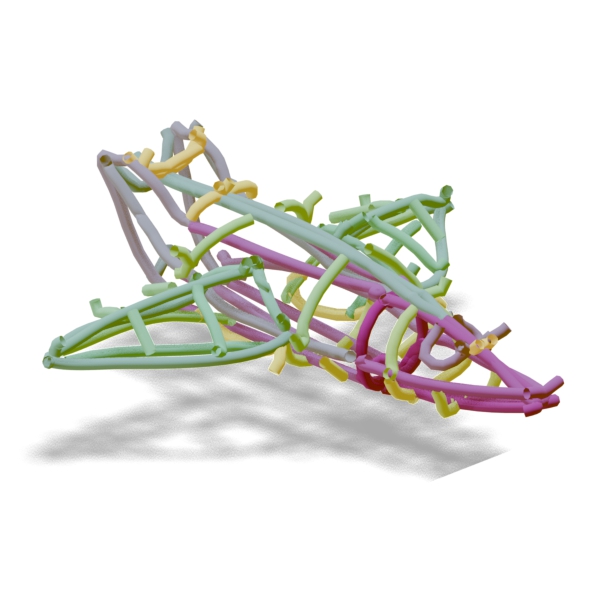} &
\includegraphics[height=0.12\textheight]{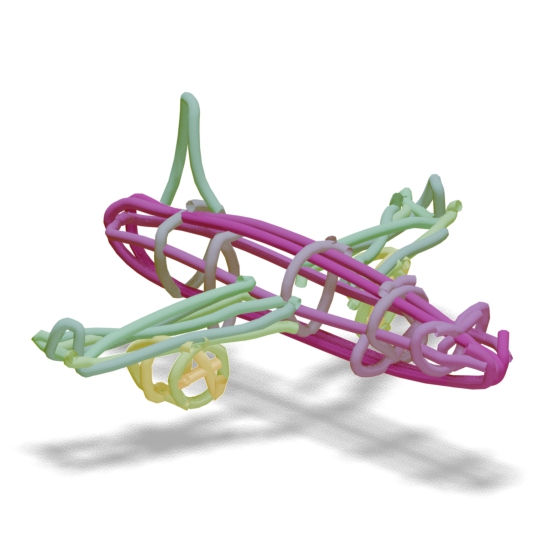} &
\includegraphics[height=0.12\textheight]{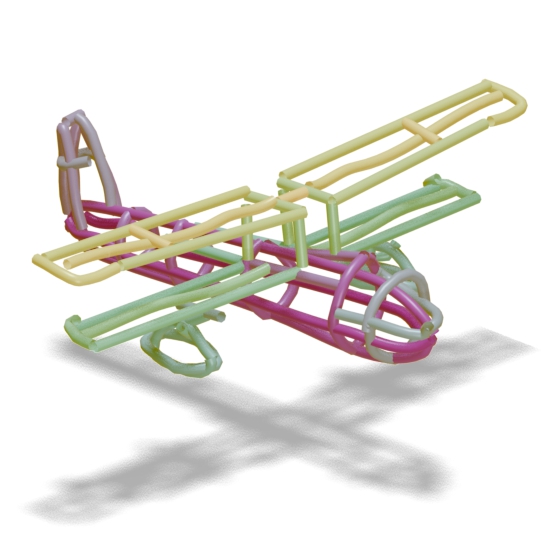} 
&
\includegraphics[height=0.12\textheight]{images/suppl/freehand2_airplane_sketch_clipped.jpg} 
&
\includegraphics[height=0.12\textheight]{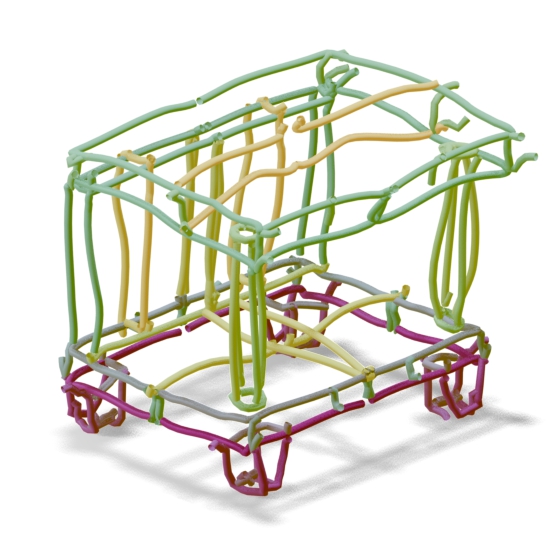} 
&
\includegraphics[height=0.12\textheight]{images/suppl/freehand7_cabinet_sketch_clipped.jpg} 
&
\includegraphics[height=0.12\textheight]{images/suppl/freehand8_cabinet_sketch_clipped.jpg} 
\\[-2mm]
\rotatebox{90}{\scriptsize \bf \qquad Our Prediction} &
\includegraphics[height=0.12\textheight]{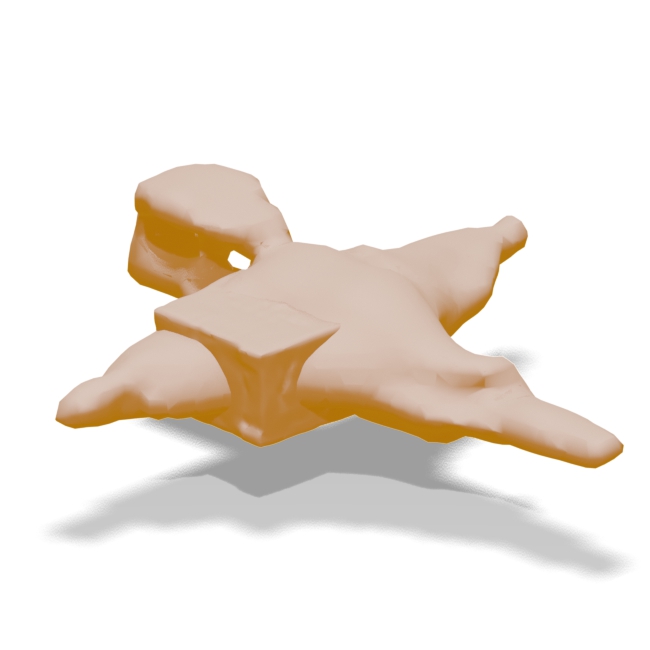} &
\includegraphics[height=0.12\textheight]{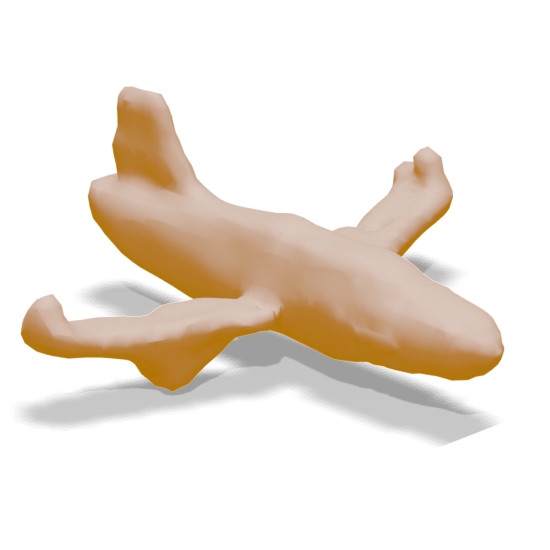} &
\includegraphics[height=0.12\textheight]{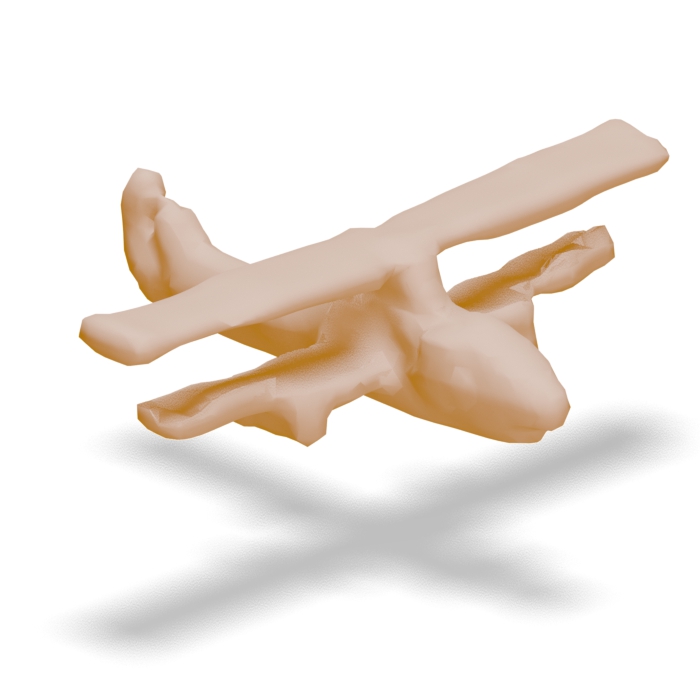} 
&
\includegraphics[height=0.12\textheight]{images/suppl/freehand2_airplane_ours_clipped.jpg} 
&
\includegraphics[height=0.12\textheight]{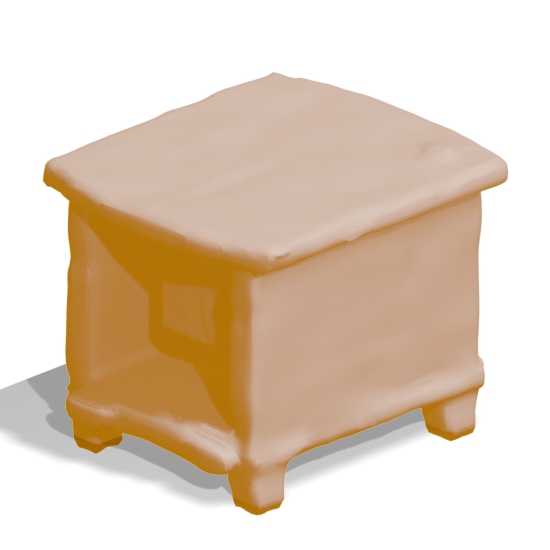} 
&
\includegraphics[height=0.12\textheight]{images/suppl/freehand7_cabinet_ours_clipped.jpg} 
&
\includegraphics[height=0.12\textheight]{images/suppl/freehand8_cabinet_ours_clipped.jpg} 
\\[-2mm]
\rotatebox{90}{\scriptsize \bf \qquad Luo's prediction} &
\includegraphics[height=0.12\textheight]{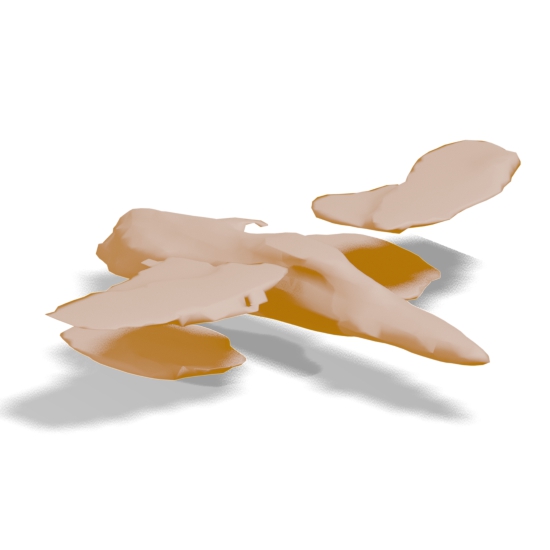} &
\includegraphics[height=0.12\textheight]{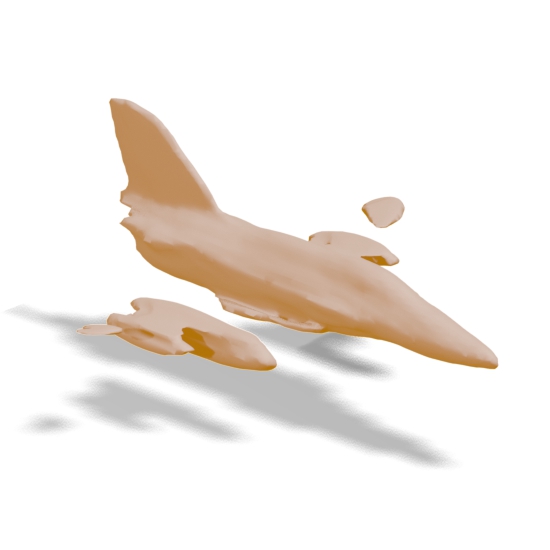} &
\includegraphics[height=0.12\textheight]{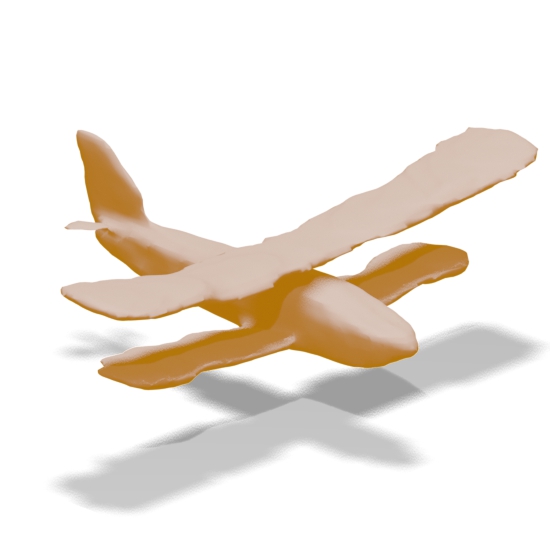} 
&
\includegraphics[height=0.12\textheight]{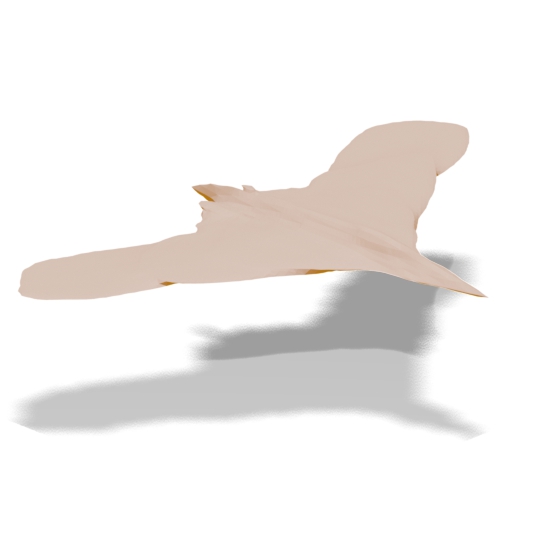} 
&
\includegraphics[height=0.12\textheight]{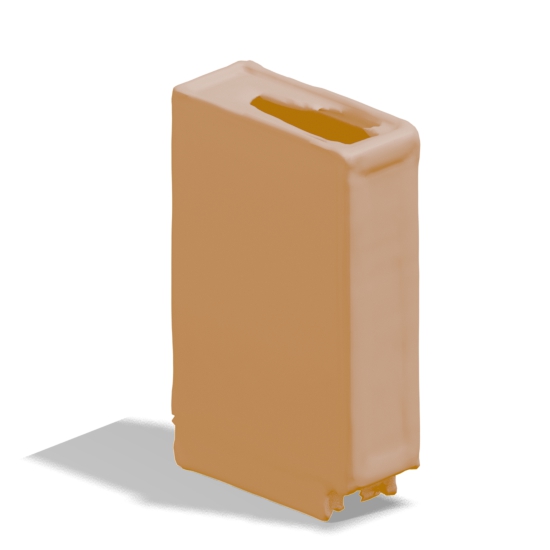} 
&
\includegraphics[height=0.12\textheight]{images/suppl/freehand_cabinet1_luo_clipped.jpeg} 
&
\includegraphics[height=0.12\textheight]{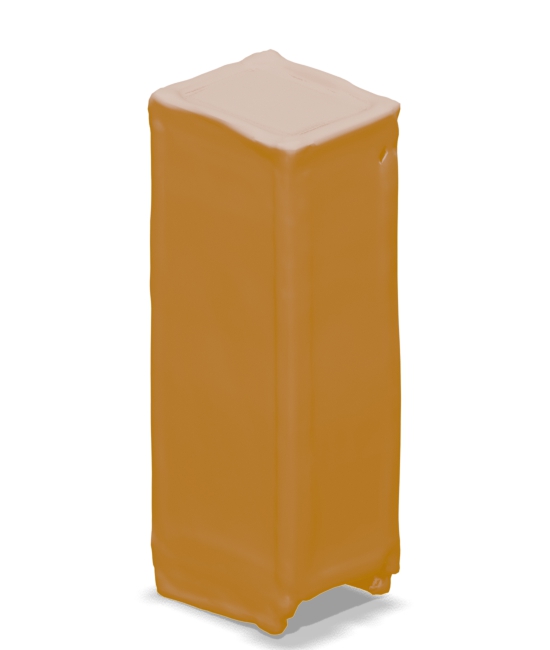} 
\\\greyrule
\rotatebox{90}{\scriptsize \bf \qquad  Free-hand Sketch}&
\includegraphics[height=0.12\textheight]{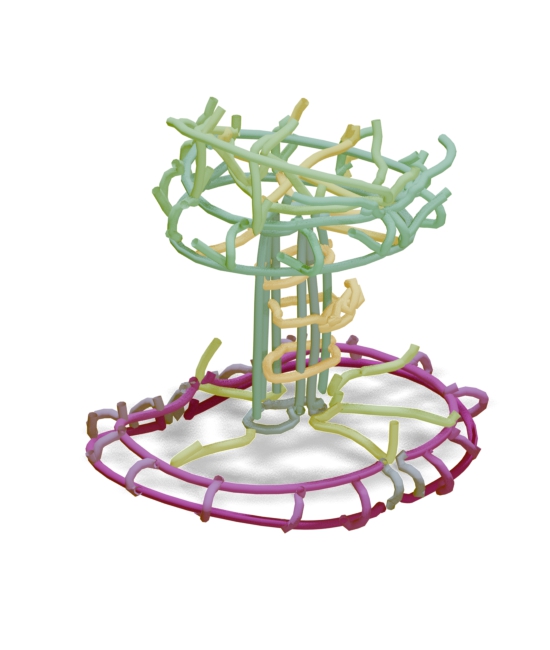} &
\includegraphics[height=0.12\textheight]{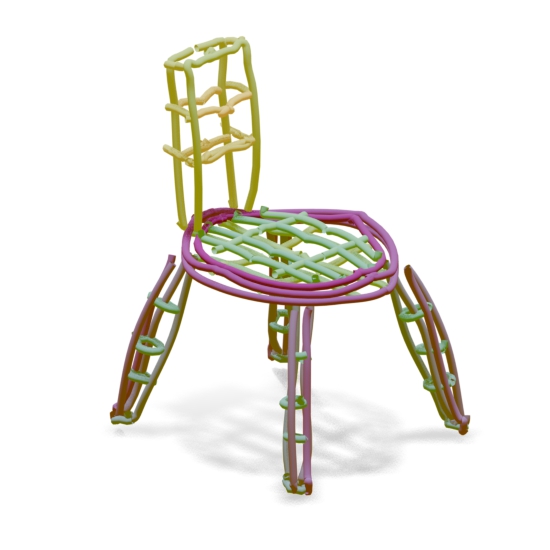} &
\includegraphics[height=0.12\textheight]{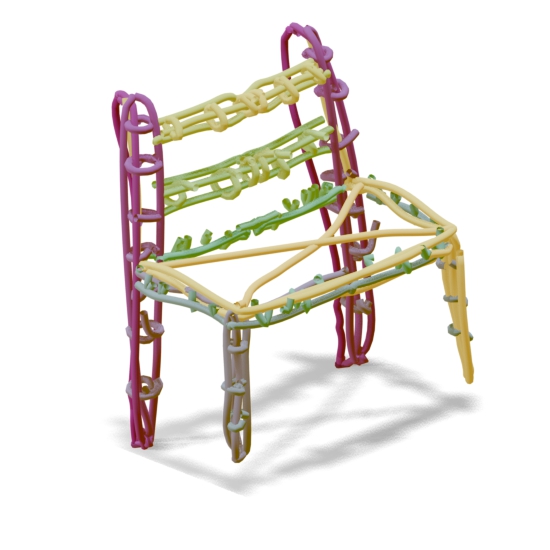} 
&
\includegraphics[height=0.12\textheight]{images/suppl/freehand3_chair_sketch_clipped.jpg} 
&
\includegraphics[height=0.12\textheight]{images/suppl/freehand4_chair_sketch_clipped.jpg} 
&
\includegraphics[height=0.12\textheight]{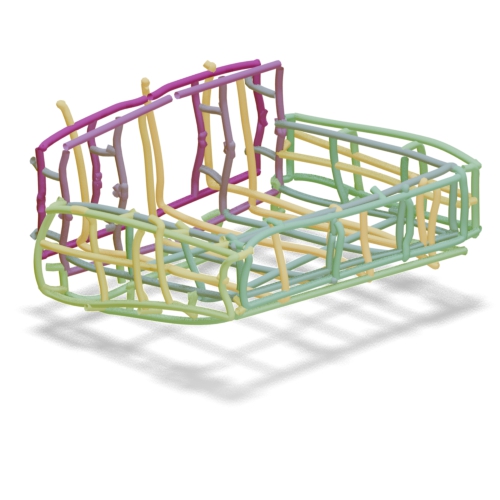} 
&
\includegraphics[height=0.12\textheight]{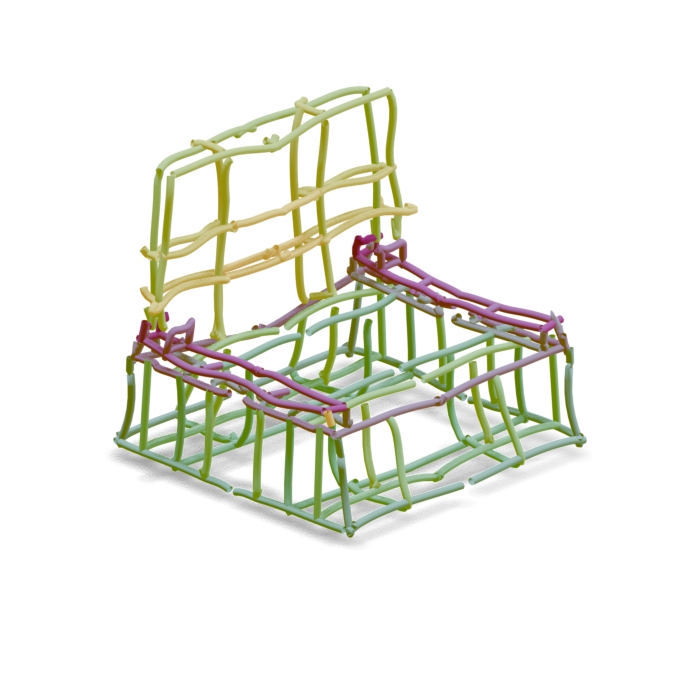} 
\\[-2mm]
\rotatebox{90}{\scriptsize \bf \qquad  Our Prediction}&
\includegraphics[height=0.12\textheight]{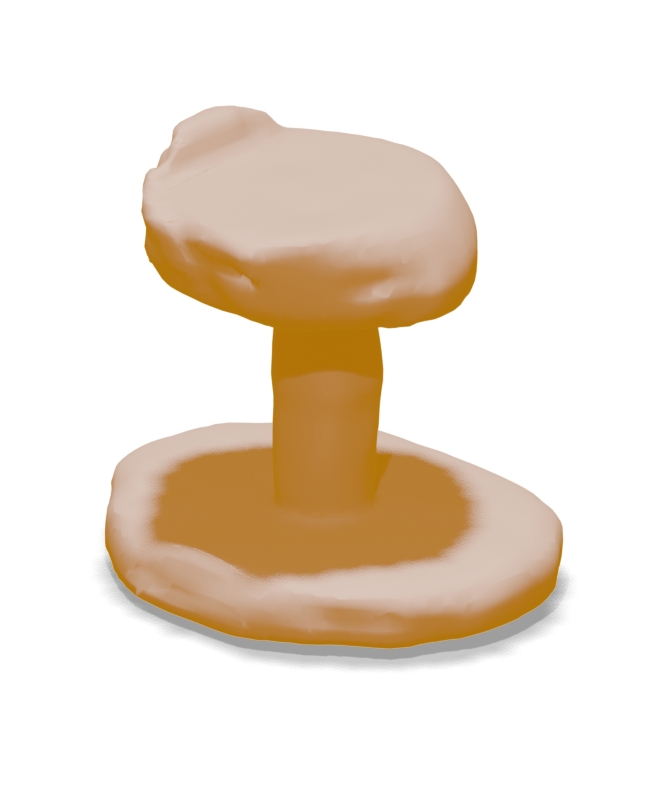} &
\includegraphics[height=0.12\textheight]{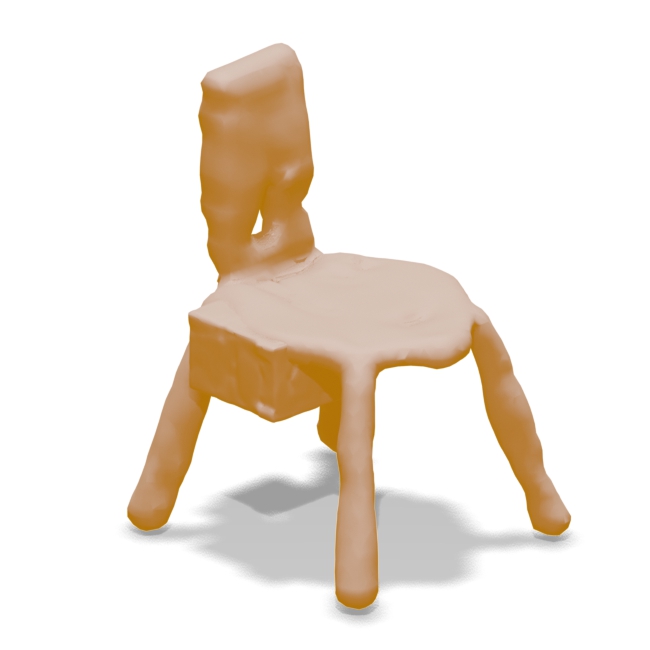} &
\includegraphics[height=0.12\textheight]{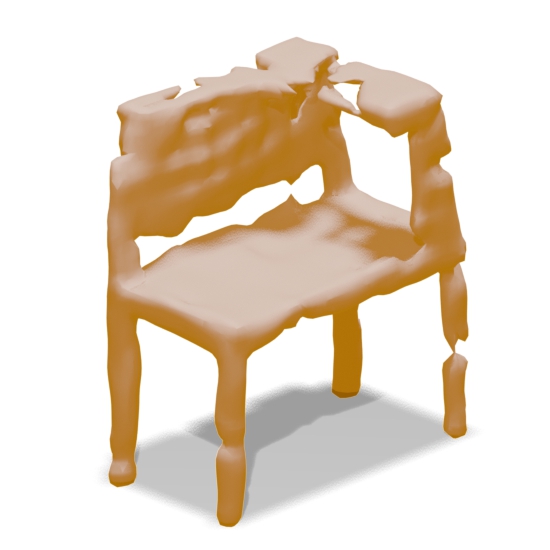} 
&
\includegraphics[height=0.12\textheight]{images/suppl/freehand3_chair_ours_clipped.jpg} 
&
\includegraphics[height=0.12\textheight]{images/suppl/freehand4_chair_ours_clipped.jpg} 
&
\includegraphics[height=0.12\textheight]{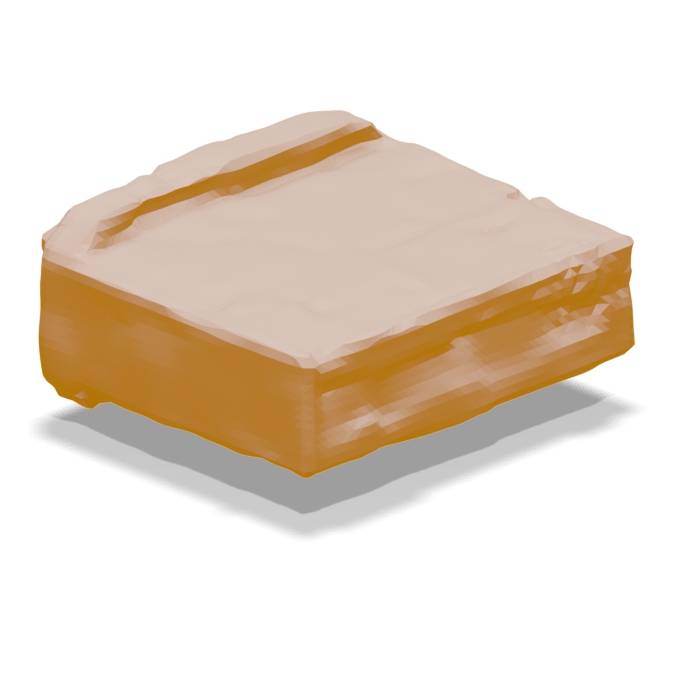} 
&
\includegraphics[height=0.12\textheight]{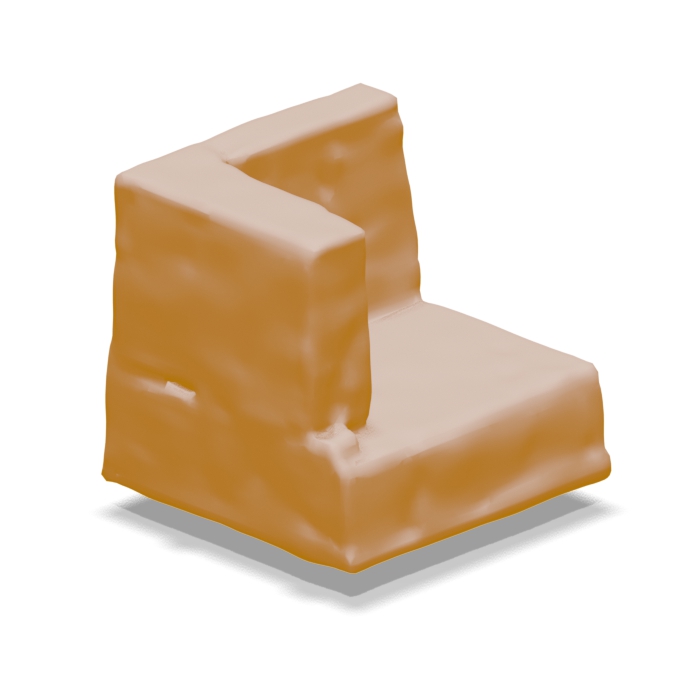} 
\\[-2mm]
\rotatebox{90}{\scriptsize \bf \qquad  Luo's Prediction}&
\includegraphics[height=0.12\textheight]{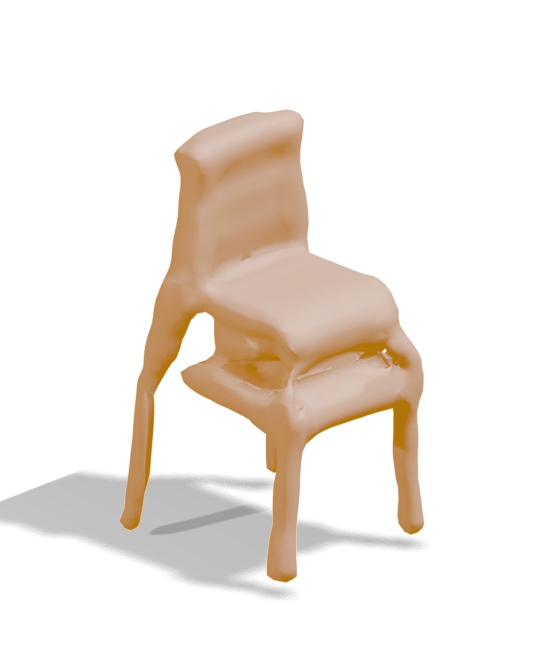} &
\includegraphics[height=0.12\textheight]{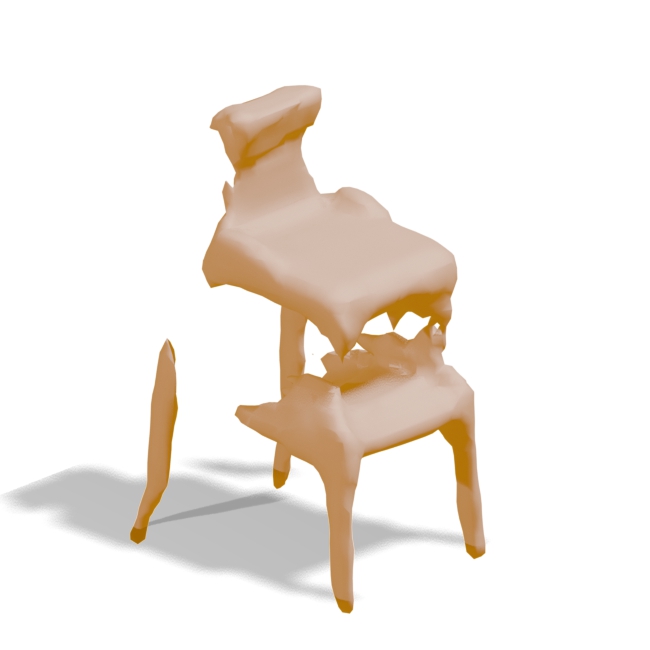} &
\includegraphics[height=0.12\textheight]{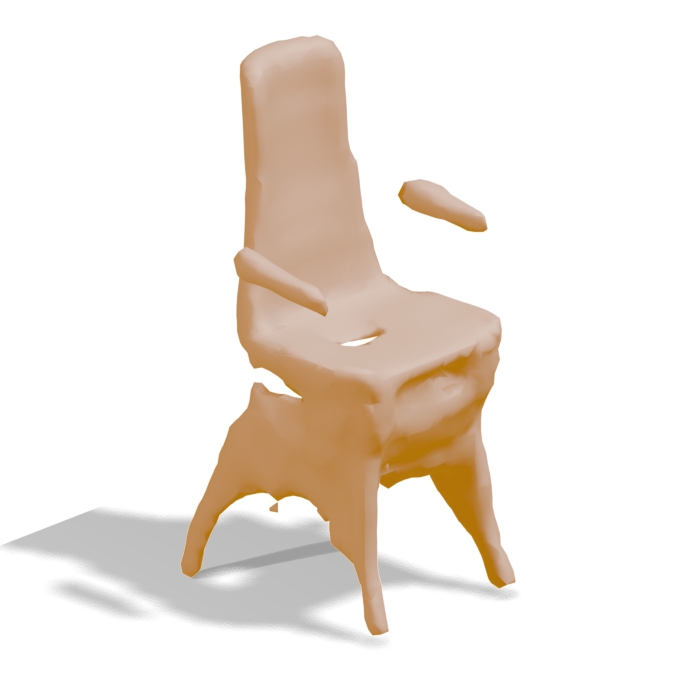} 
&
\includegraphics[height=0.12\textheight]{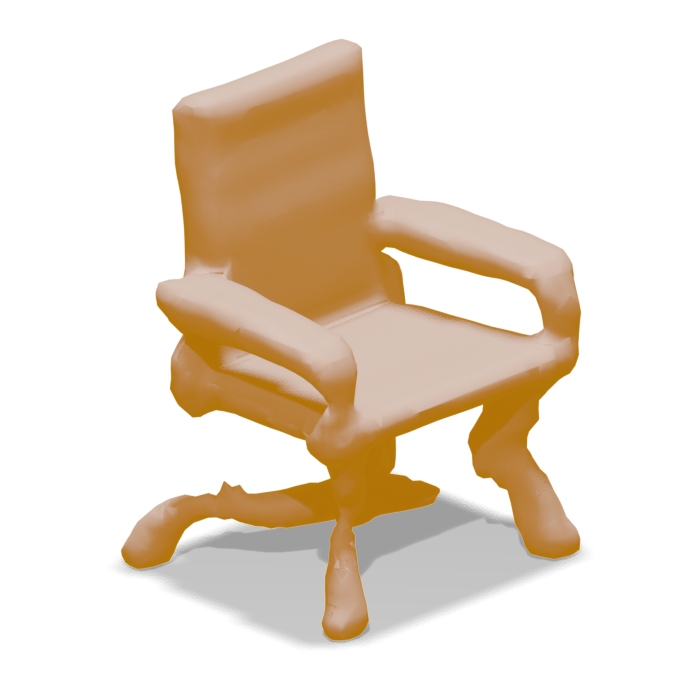} 
&
\includegraphics[height=0.12\textheight]{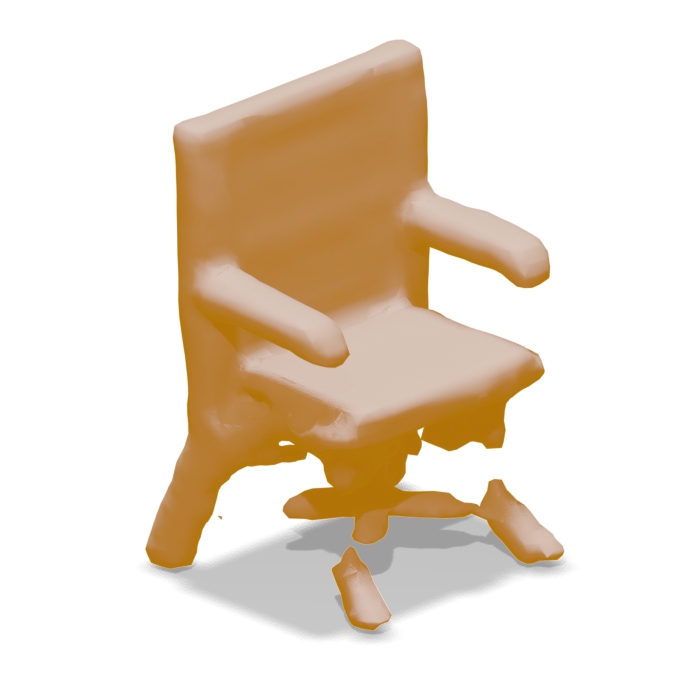} 
&
\includegraphics[height=0.12\textheight]{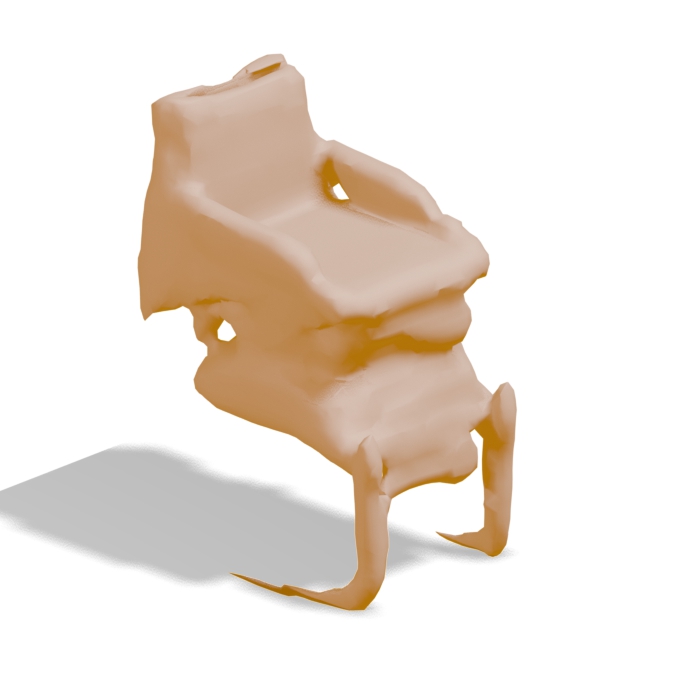} 
&
\includegraphics[height=0.12\textheight]{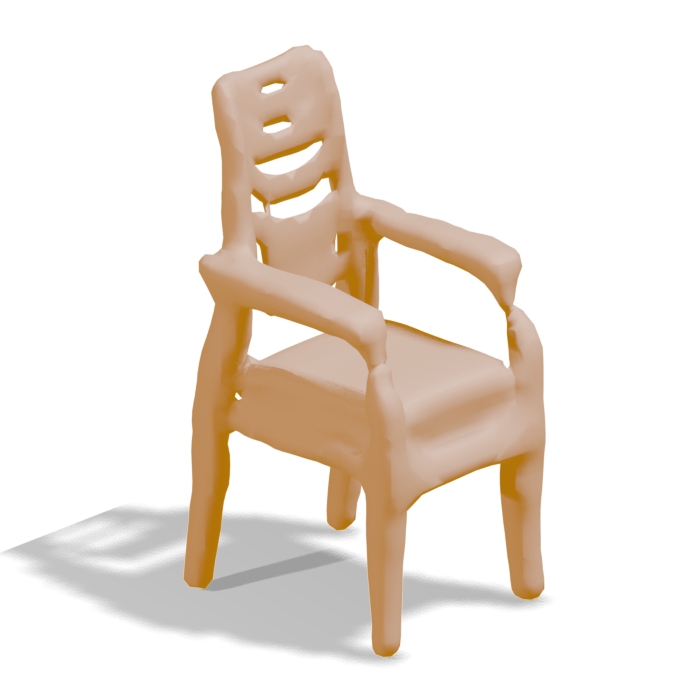} 
\\\greyrule
\rotatebox{90}{\scriptsize \bf \qquad  Free-hand Sketch}&
\includegraphics[height=0.12\textheight]{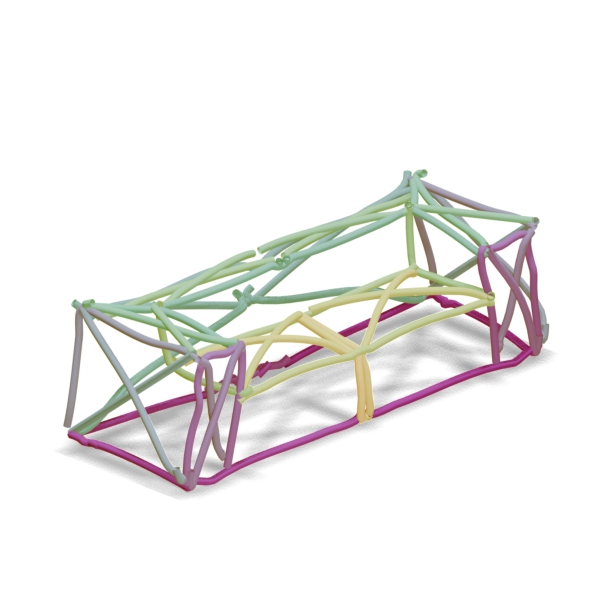} &
\includegraphics[height=0.12\textheight]{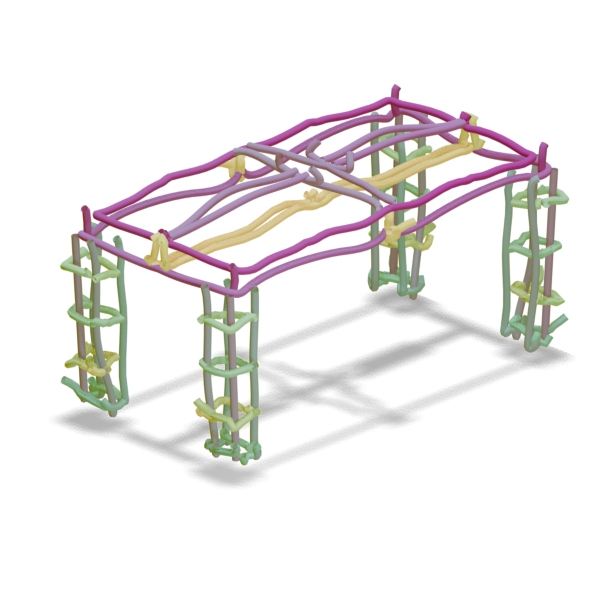} &
\includegraphics[height=0.12\textheight]{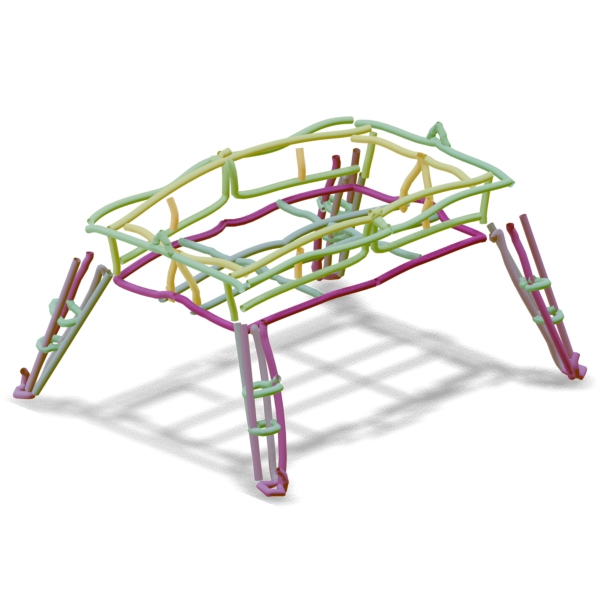} 
&
\includegraphics[height=0.12\textheight]{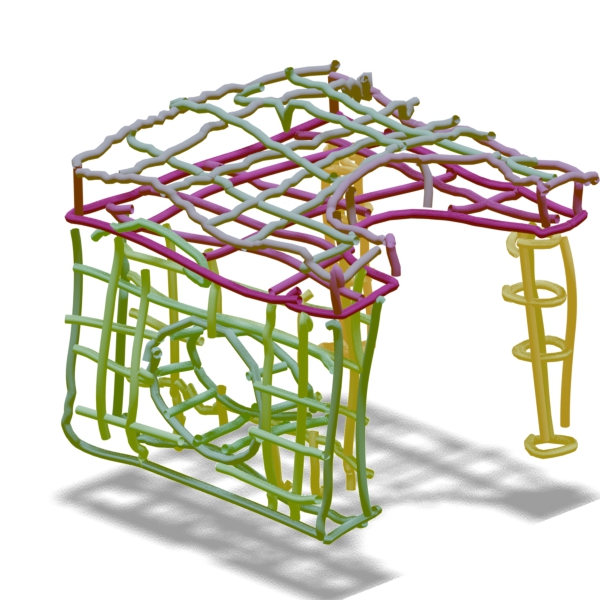} 
&
\includegraphics[height=0.12\textheight]{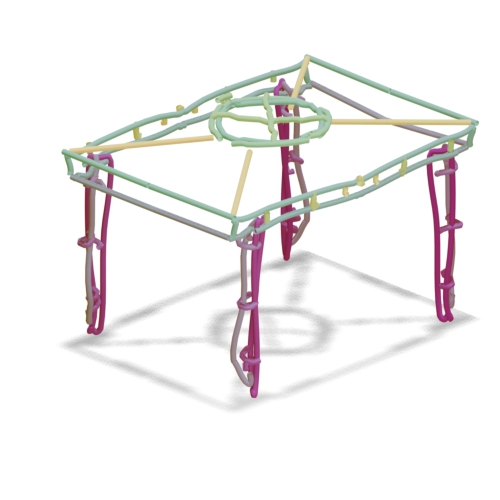} 
&
\includegraphics[height=0.12\textheight]{images/suppl/freehand5_table_sketch_clipped.jpg} 
&
\includegraphics[height=0.12\textheight]{images/suppl/freehand6_table_sketch_clipped.jpg} 
\\[-2mm]
\rotatebox{90}{\scriptsize \bf \qquad  Our Prediction}&
\includegraphics[height=0.12\textheight]{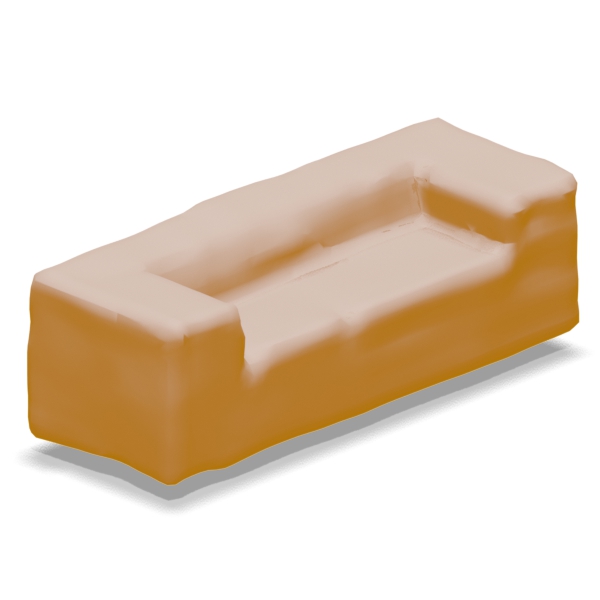} &
\includegraphics[height=0.12\textheight]{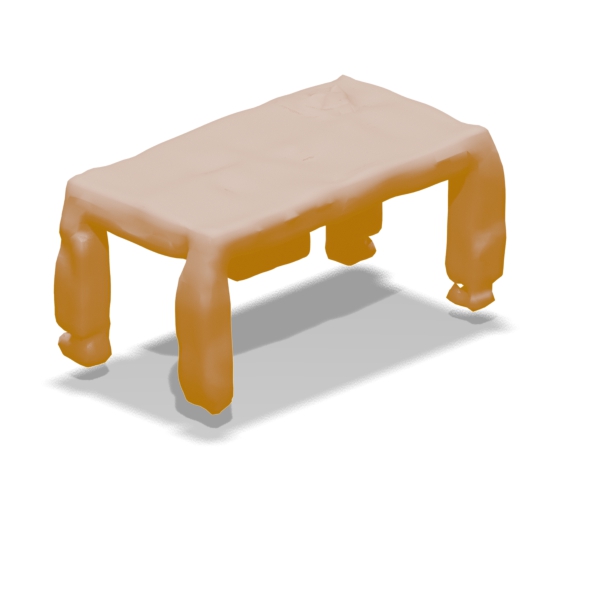} &
\includegraphics[height=0.12\textheight]{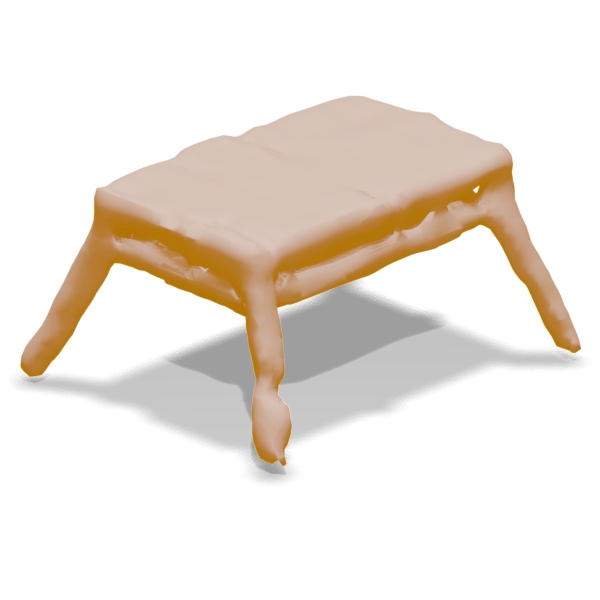} 
&
\includegraphics[height=0.12\textheight]{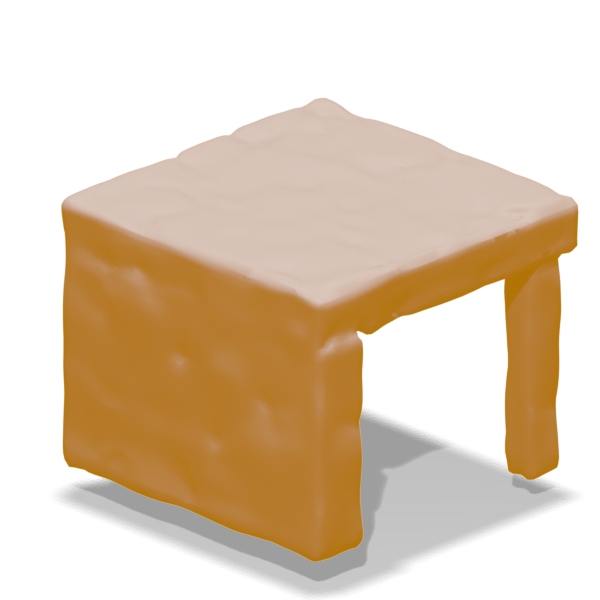} 
&
\includegraphics[height=0.12\textheight]{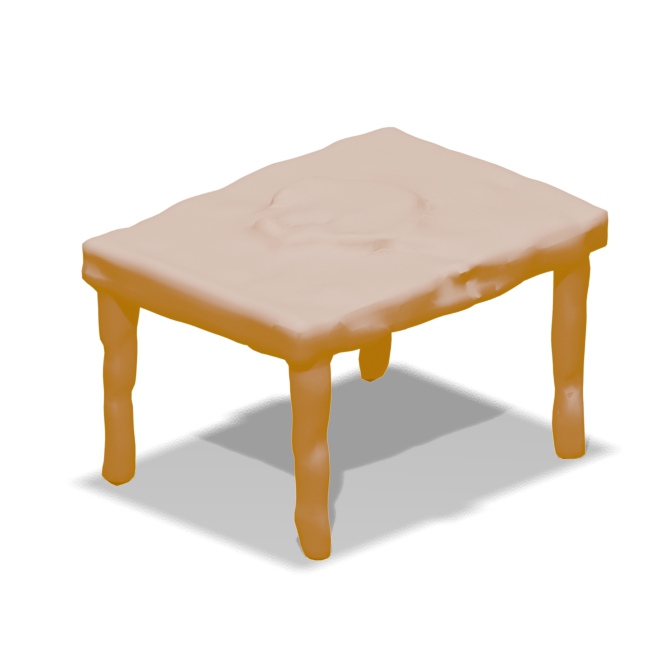} 
&
\includegraphics[height=0.12\textheight]{images/suppl/freehand5_table_ours_clipped.jpg} 
&
\includegraphics[height=0.12\textheight]{images/suppl/freehand6_table_ours_clipped.jpg} 
\\[-2mm]
\rotatebox{90}{\scriptsize \bf \qquad  Luo's Prediction}&
\includegraphics[height=0.12\textheight]{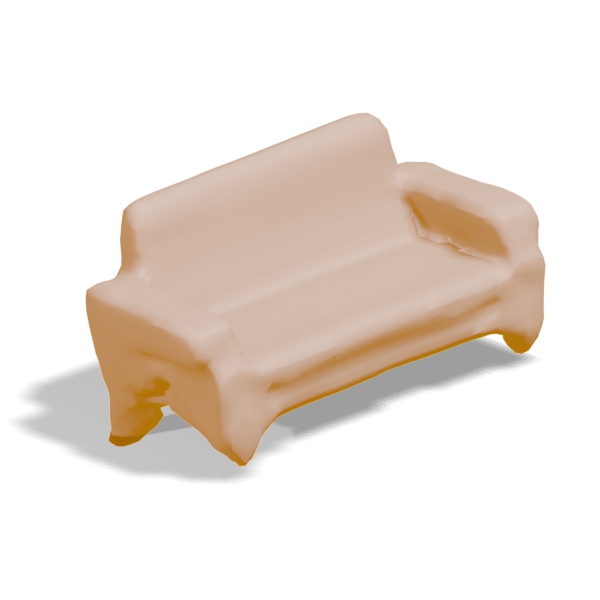} &
\includegraphics[height=0.12\textheight]{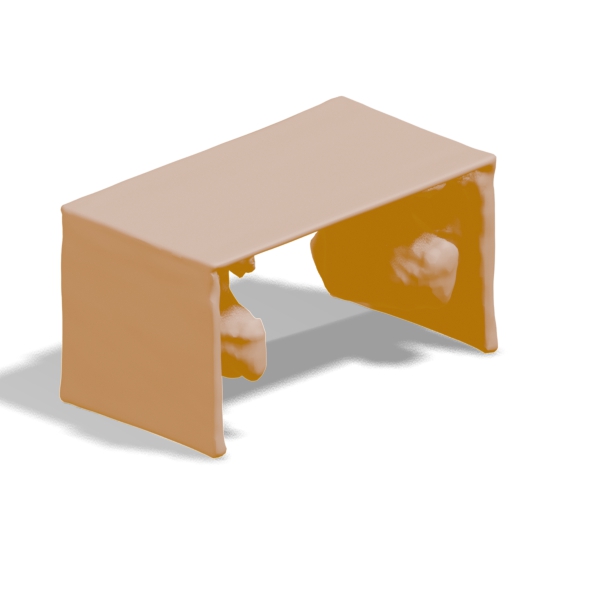} &
\includegraphics[height=0.12\textheight]{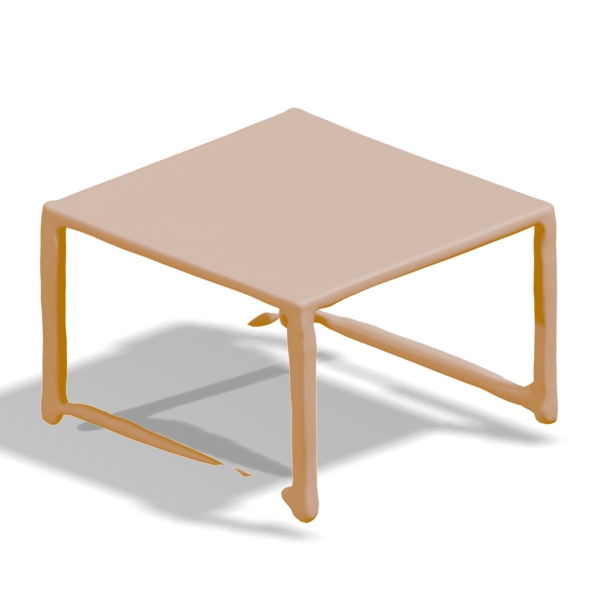} 
&
\includegraphics[height=0.12\textheight]{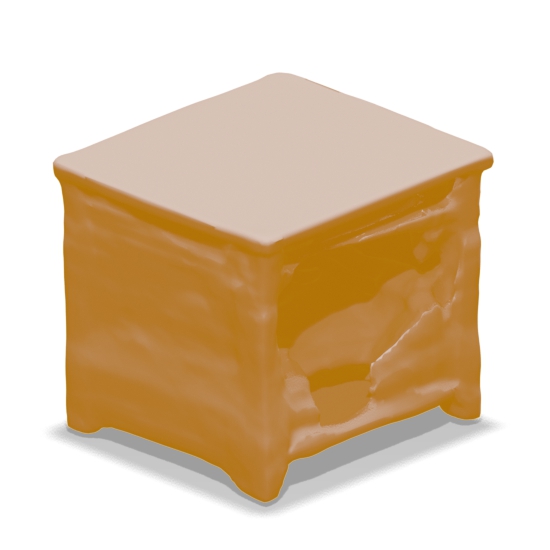} 
&
\includegraphics[height=0.12\textheight]{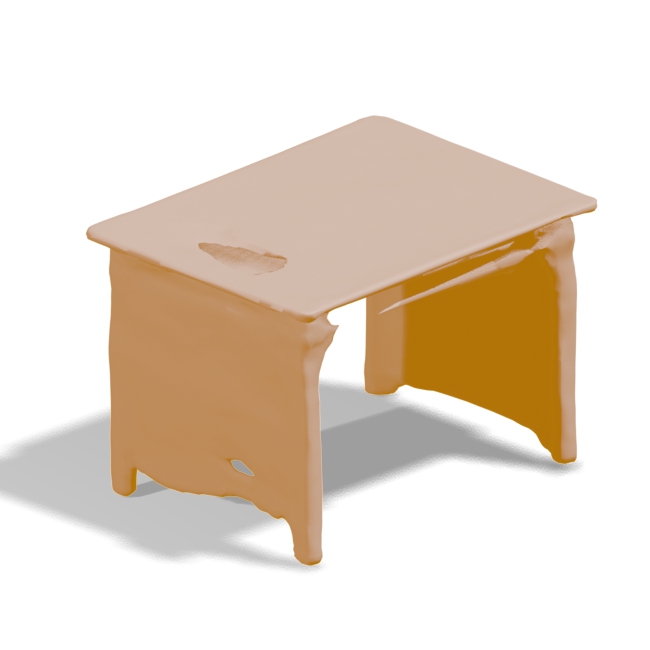} 
&
\includegraphics[height=0.12\textheight]{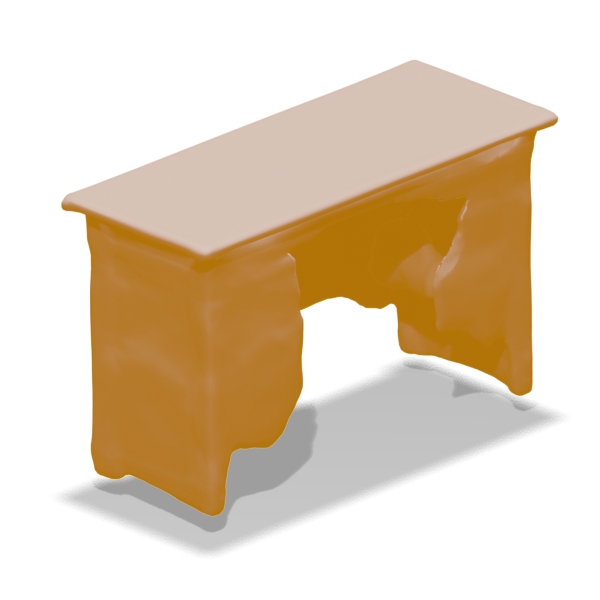} 
&
\includegraphics[height=0.12\textheight]{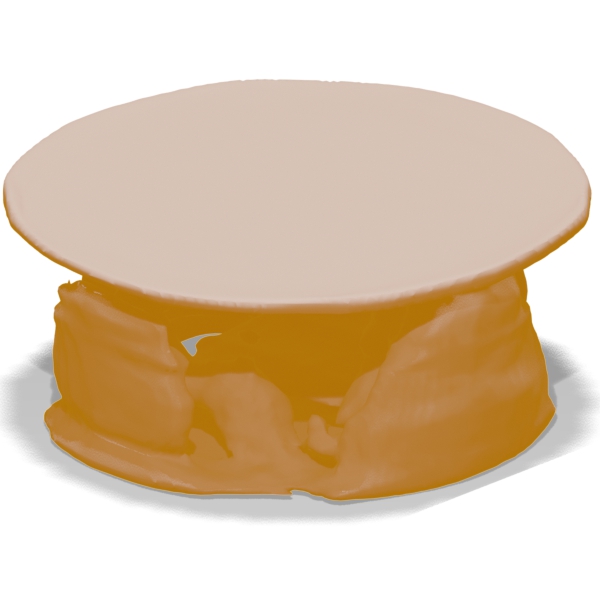} 
\end{tabular}

    }
    \vspace{-3mm}
    \caption{{\bf Shape Generation from Free-Hand Sketches.}     Compared with Luo~\etal~\cite{luo20233d}, Our model generalizes well to free-hand sketches drawn without any reference shape for airplanes, chairs/sofas, tables, and cabinets, producing detailed and plausible reconstructions that reflect the user’s intent.
    }
    \label{fig:suppl_freehand}
\end{figure*}

\section{Evaluation on Unseen Classes}
\label{sec:suppl_unseen}
A related concern is whether a model trained exclusively on ShapeNet~\cite{chang2015shapenet} categories truly learns a sketch-to-geometry mapping, or whether it merely exploits memorized class-specific priors.
If the latter were the case, it should struggle when faced with sketches depicting objects outside the training categories.

\begin{figure*}[h]
    \centering
    \resizebox{\linewidth}{!}{
    
\begin{tabular}{l@{\,}c@{\,}c@{\,}c@{\,}c@{\,}c@{\,}c@{\,}c}
\rotatebox{90}{\small \bf \qquad  Sketch}&
\includegraphics[height=0.12\textheight]{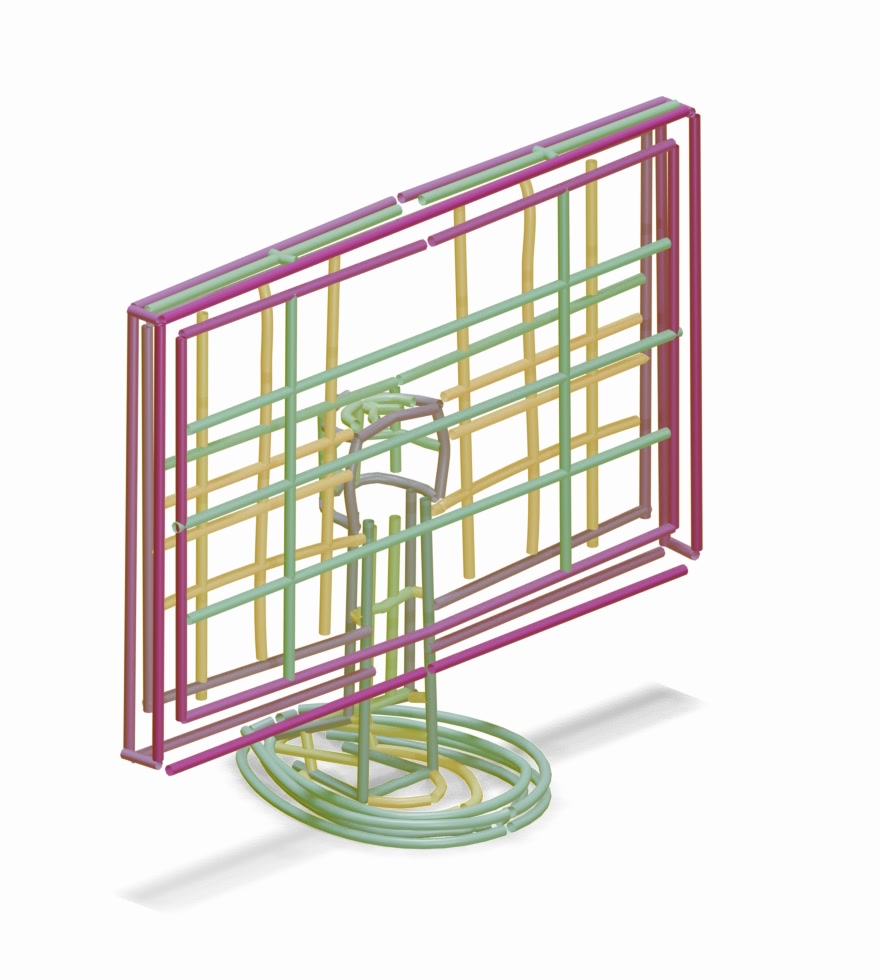} &
\includegraphics[height=0.12\textheight]{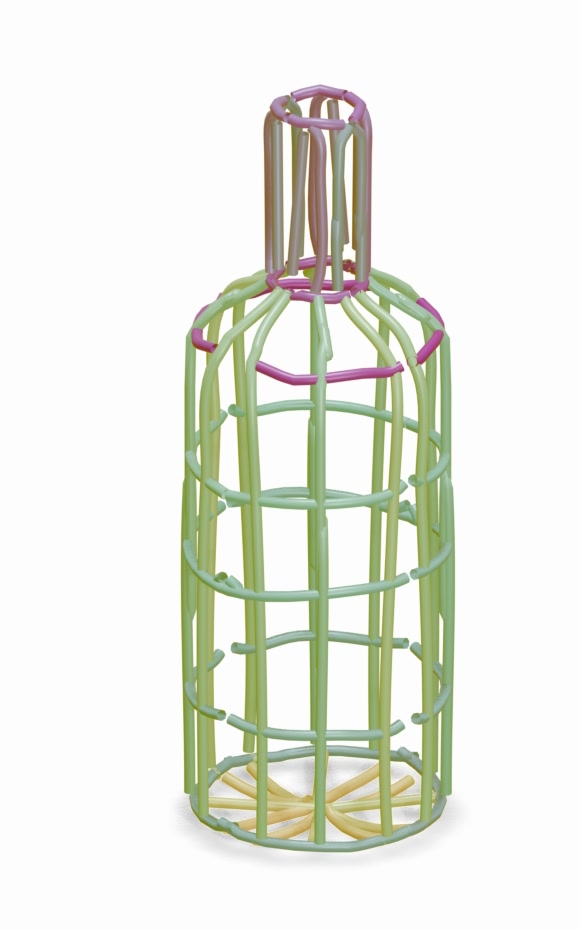} &
\includegraphics[height=0.12\textheight]{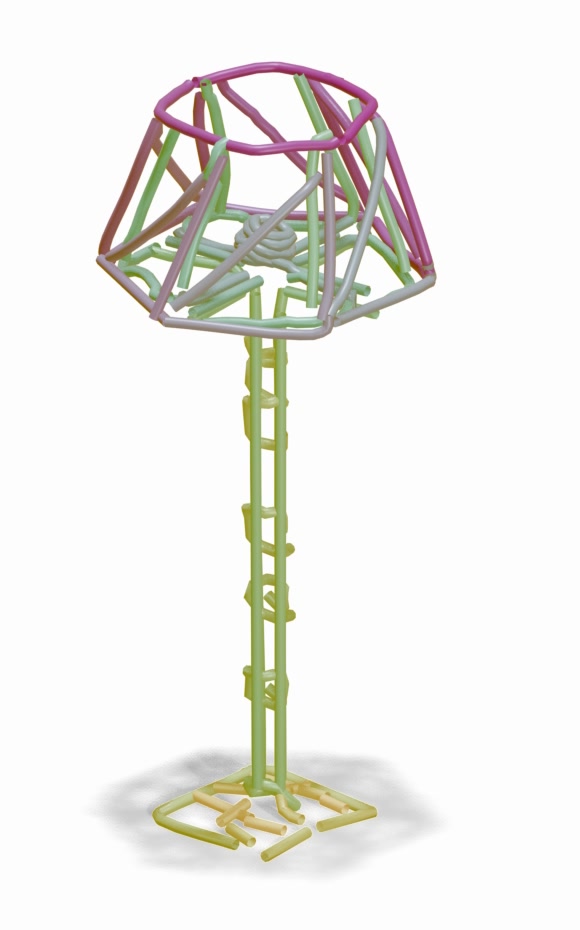} 
&
\includegraphics[height=0.12\textheight]{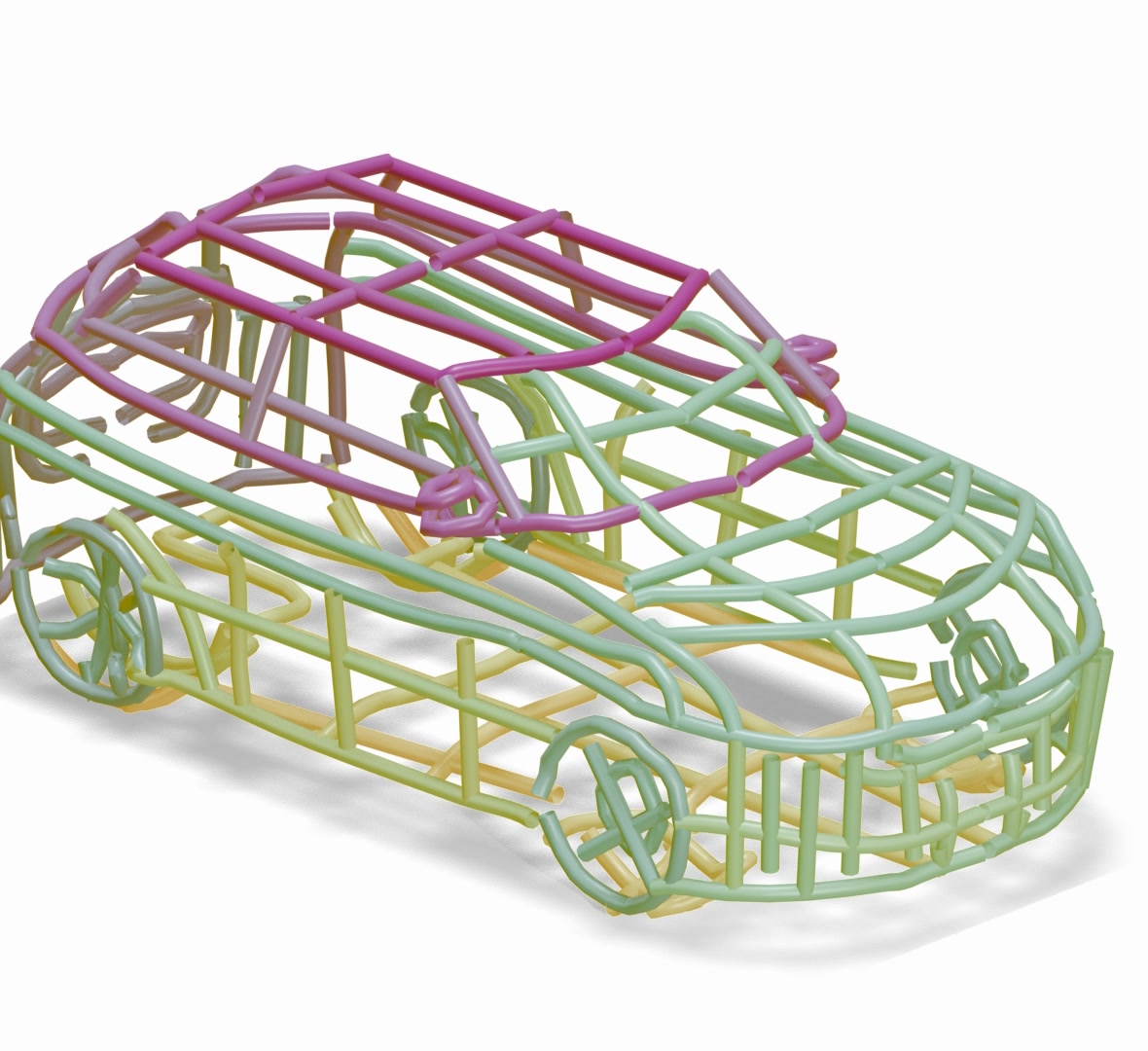} 
&
\includegraphics[height=0.12\textheight]{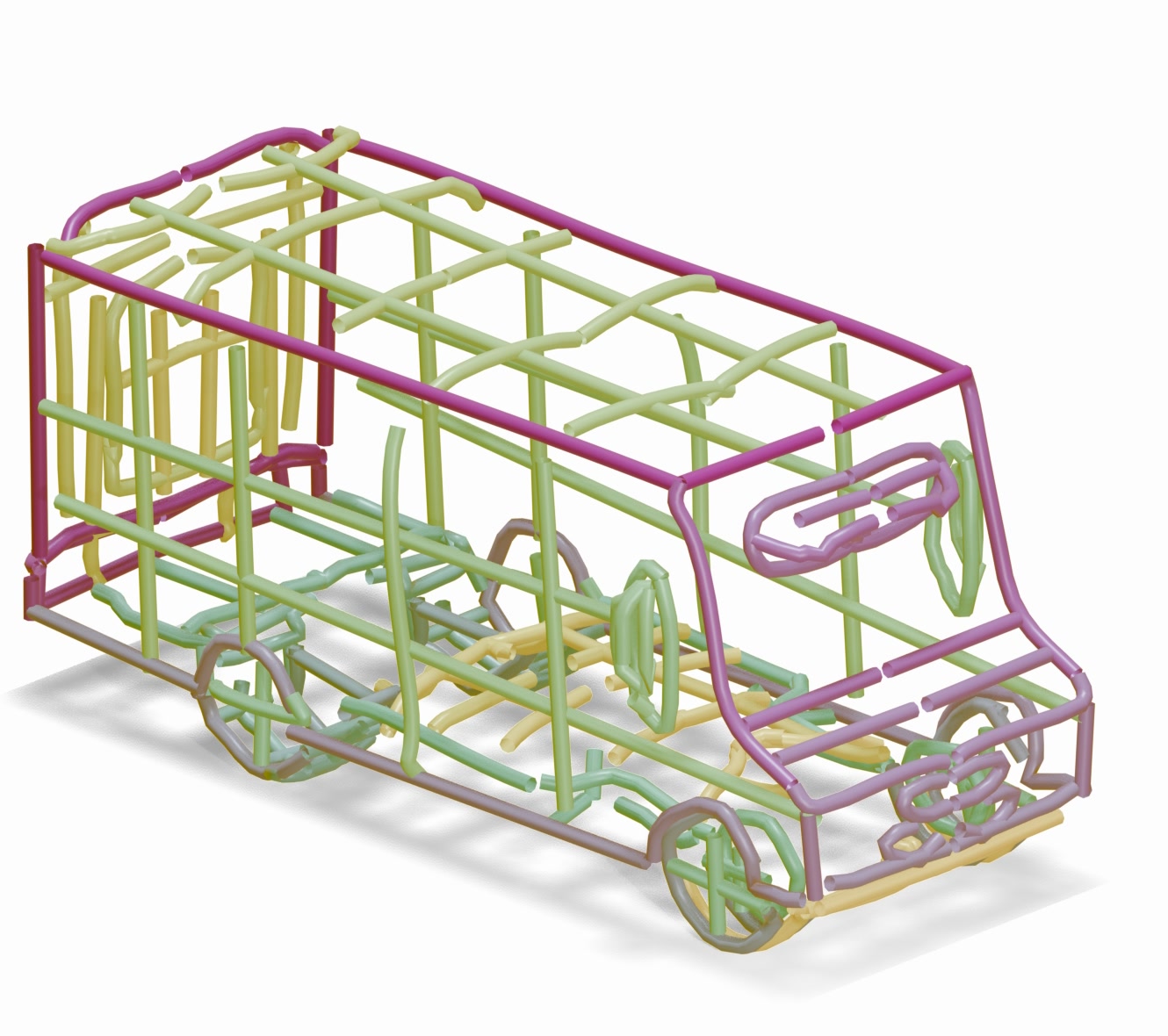} 
&
\includegraphics[height=0.12\textheight]{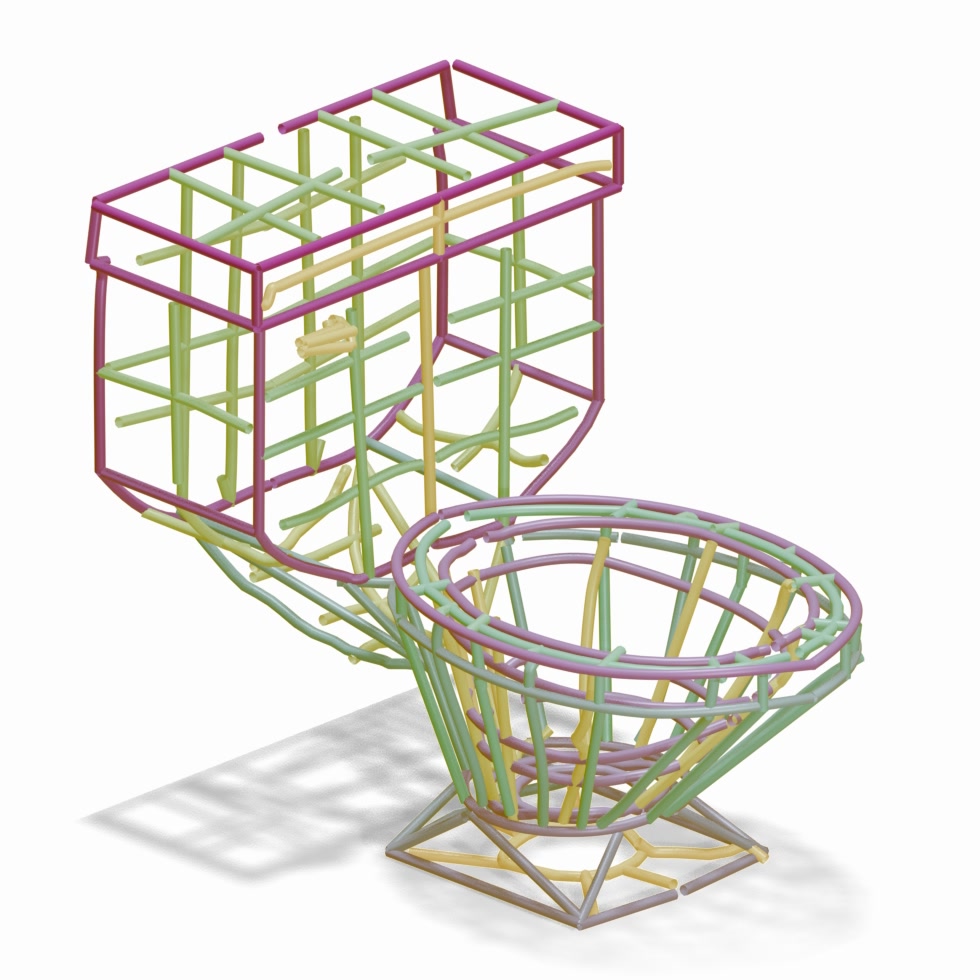} 
&
\includegraphics[height=0.12\textheight]{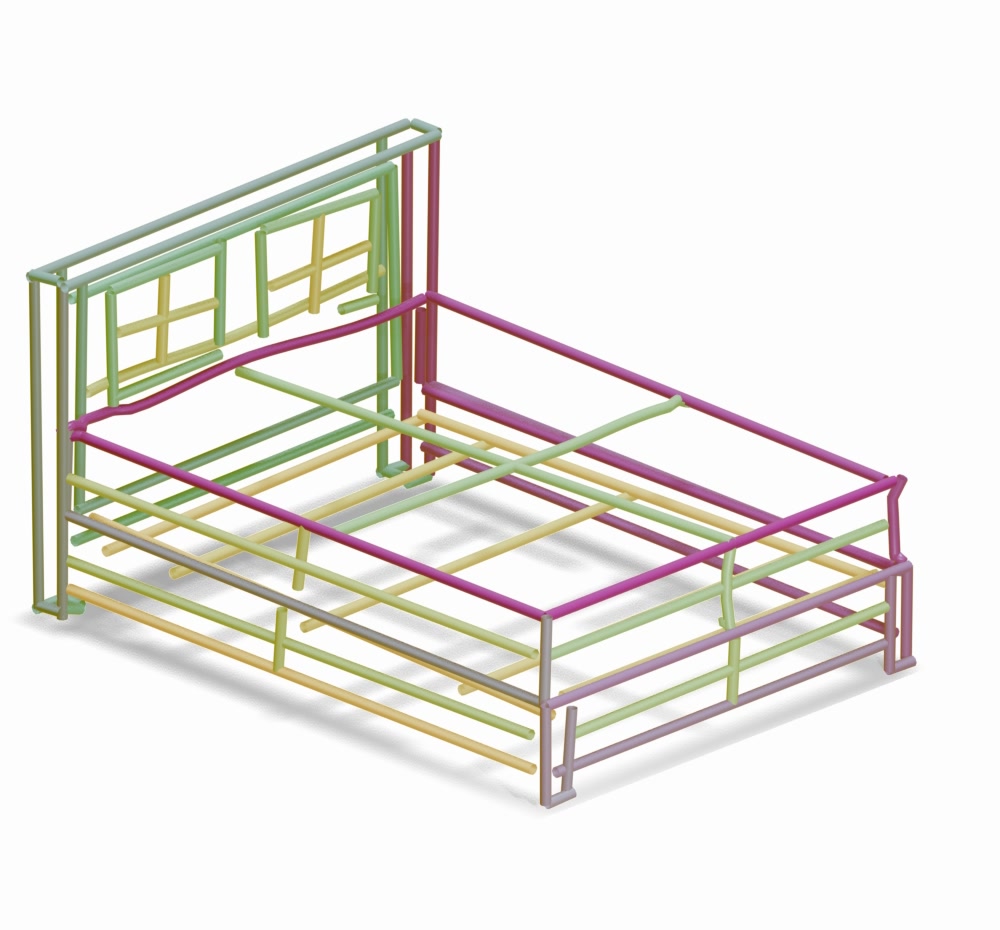} 
\\[-4mm]
\rotatebox{90}{\small \bf \qquad GT shape} &
\includegraphics[height=0.12\textheight]{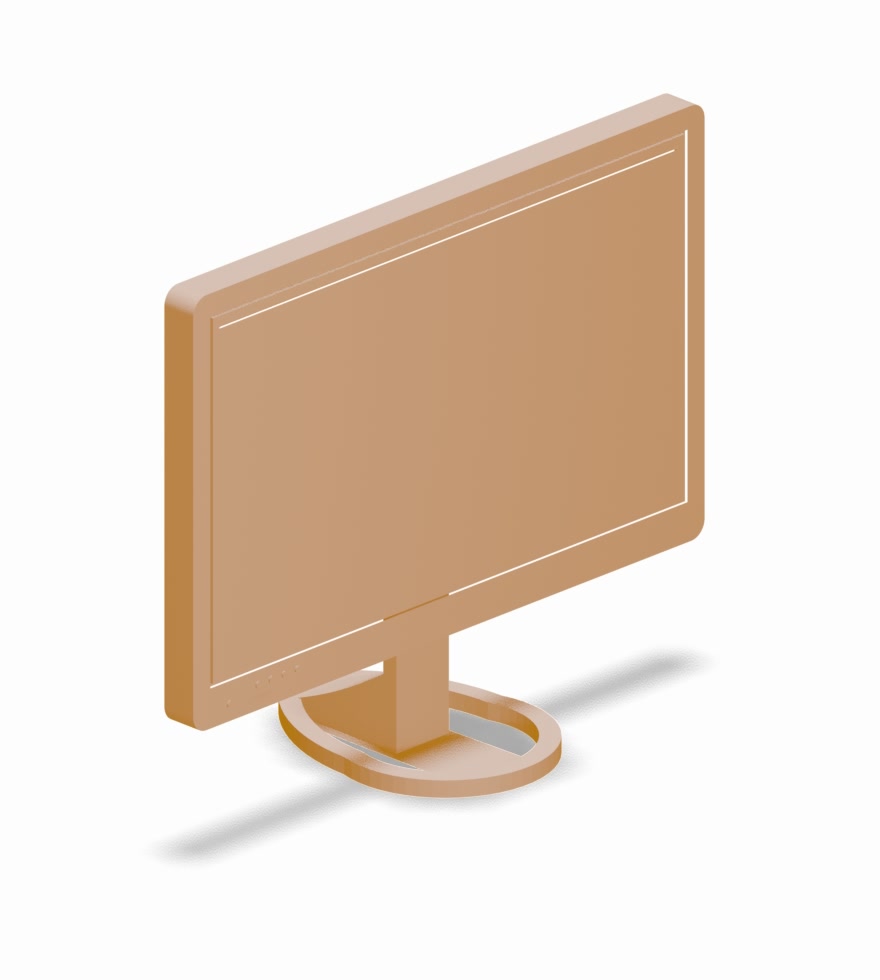} &
\includegraphics[height=0.12\textheight]{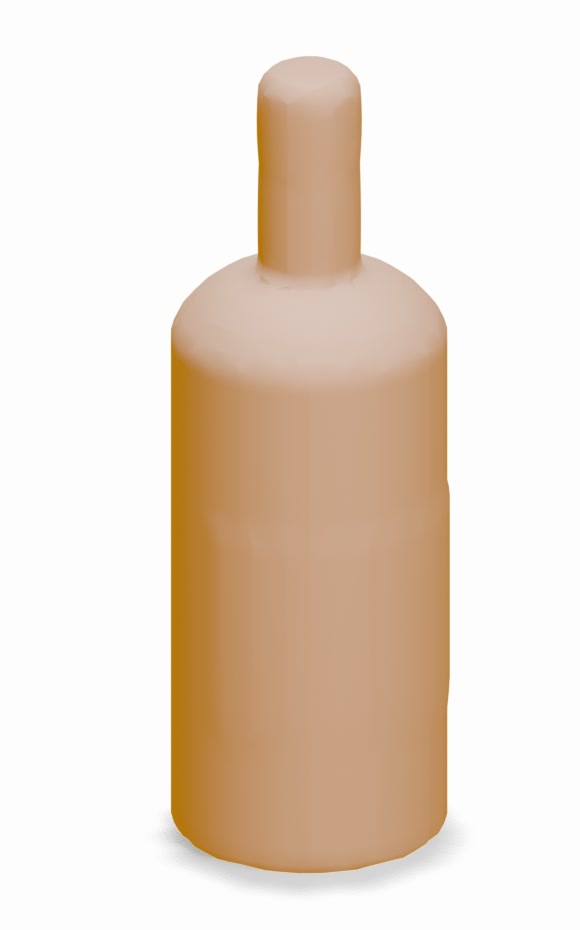} &
\includegraphics[height=0.12\textheight]{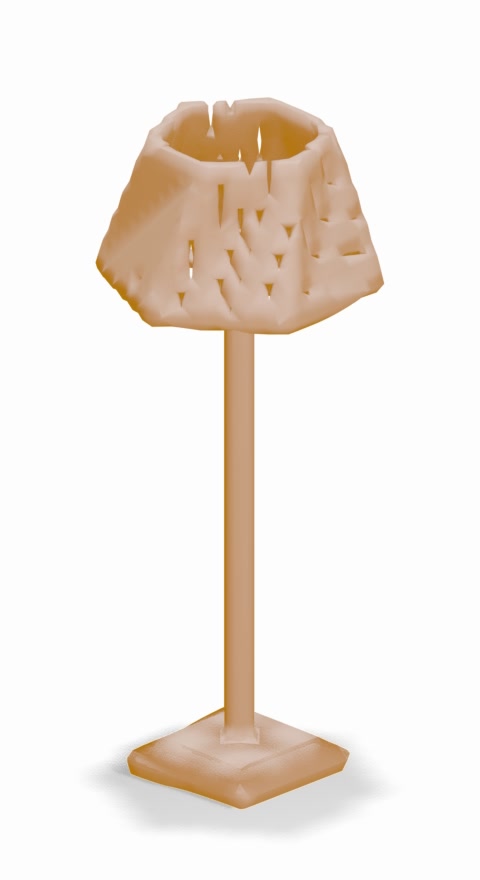} 
&
\includegraphics[height=0.12\textheight]{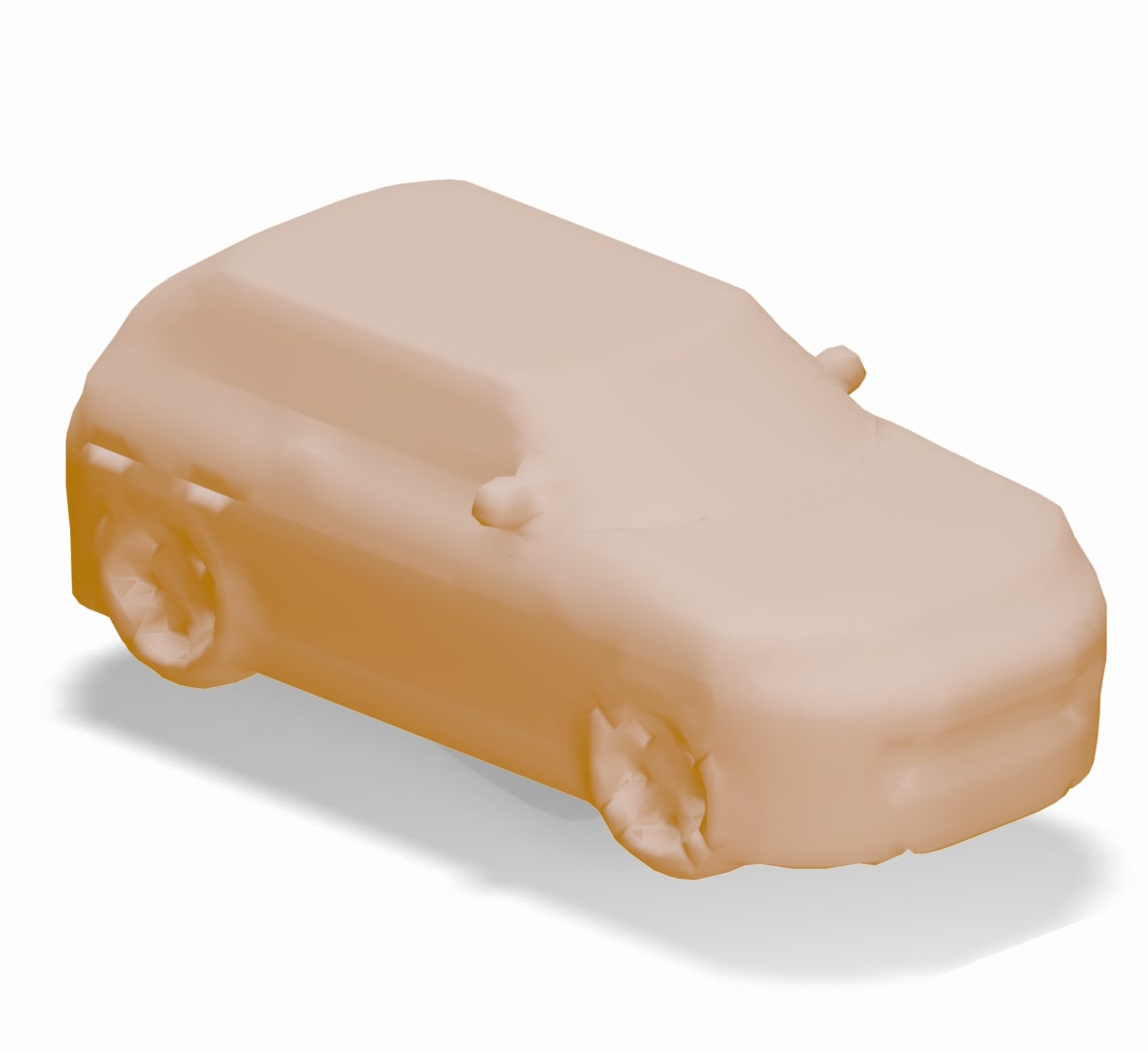} 
&
\includegraphics[height=0.12\textheight]{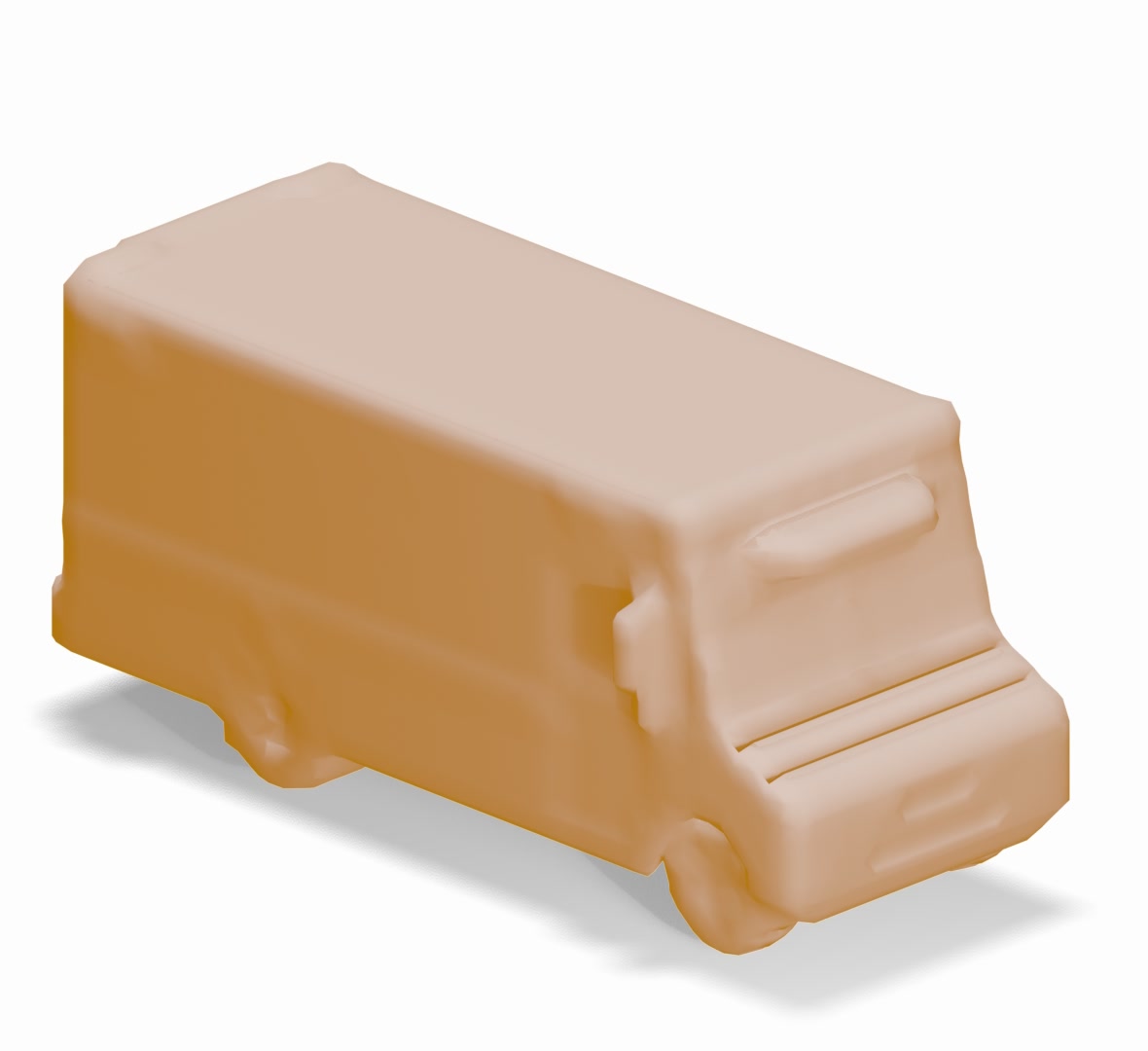} 
&
\includegraphics[height=0.12\textheight]{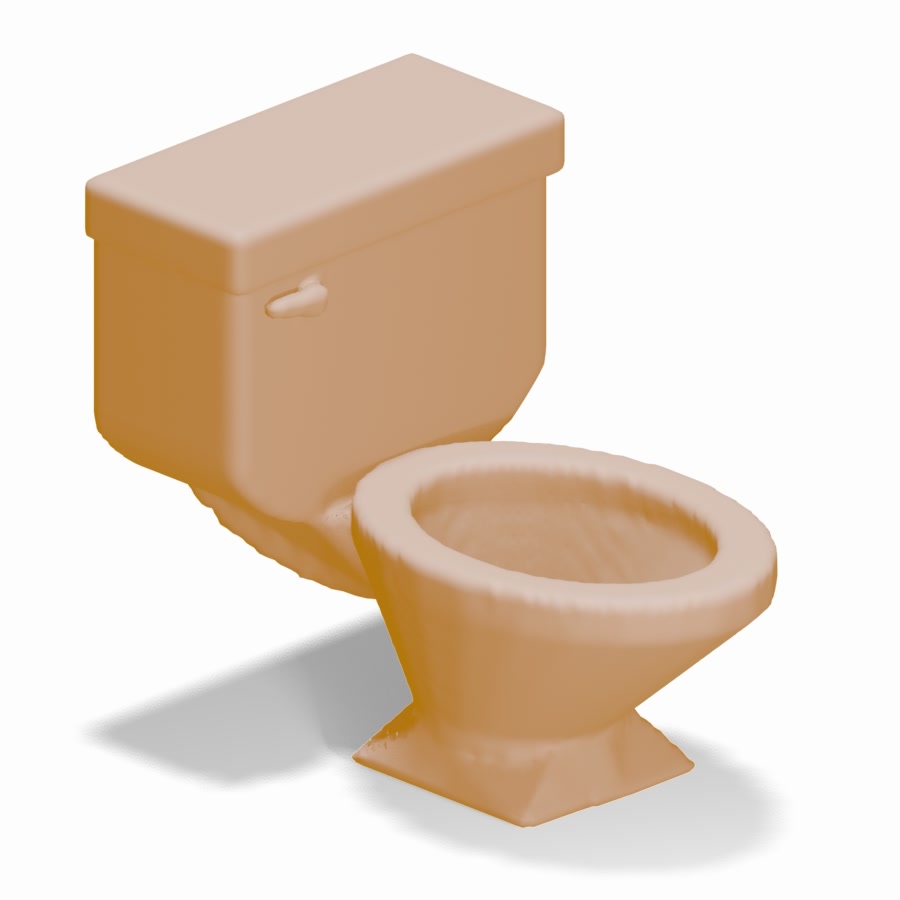} 
&
\includegraphics[height=0.12\textheight]{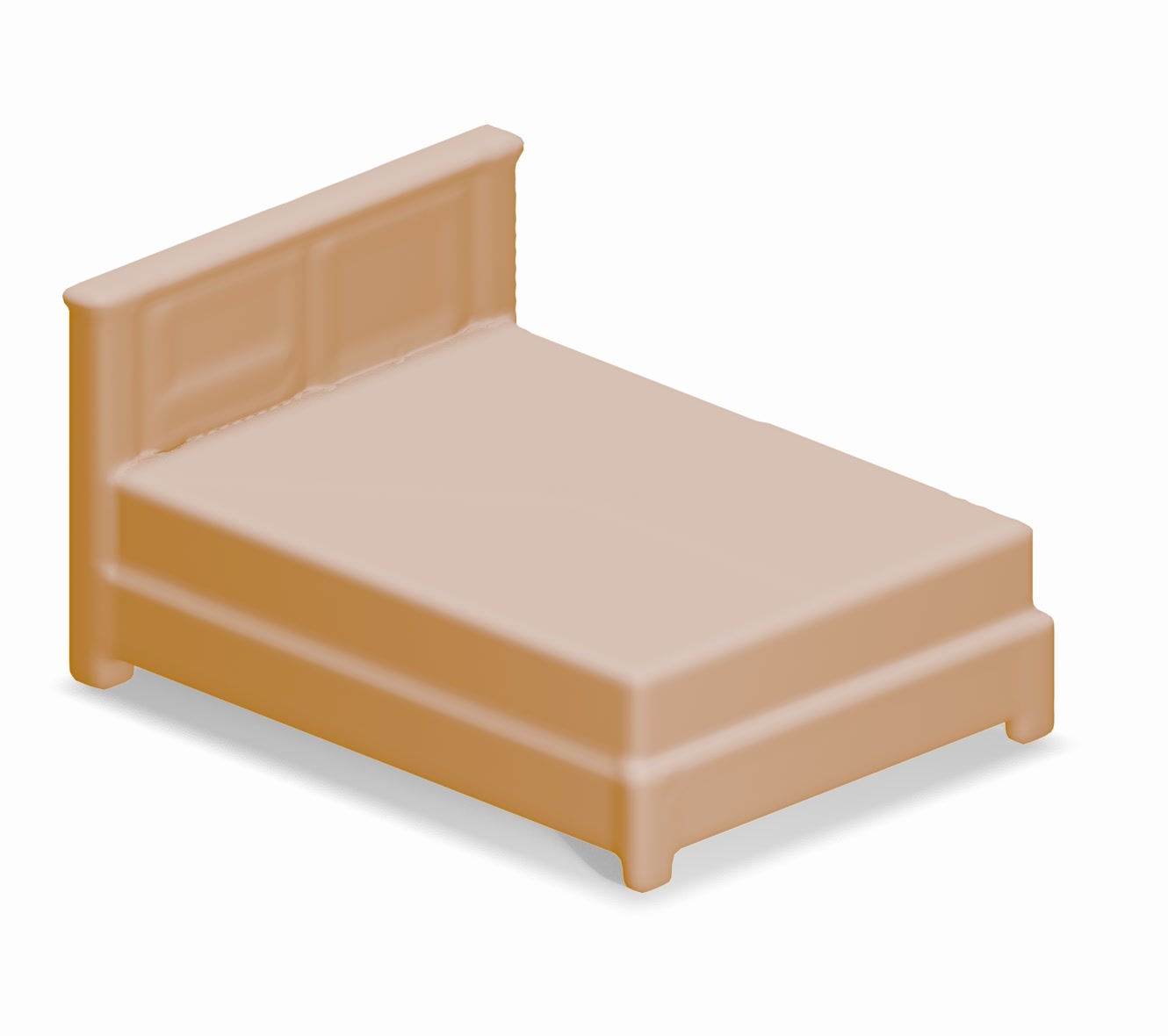} 
\\[-4mm]
\rotatebox{90}{\small \bf \qquad Our prediction} &
\includegraphics[height=0.12\textheight]{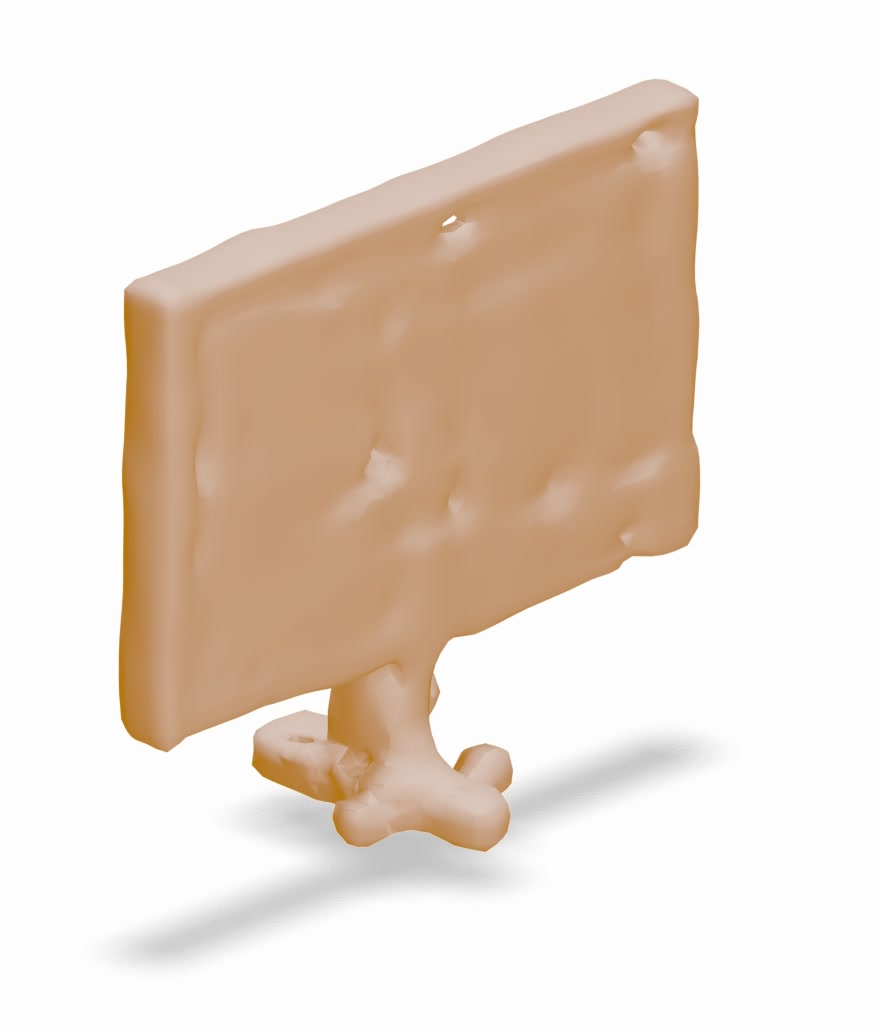} &
\includegraphics[height=0.12\textheight]{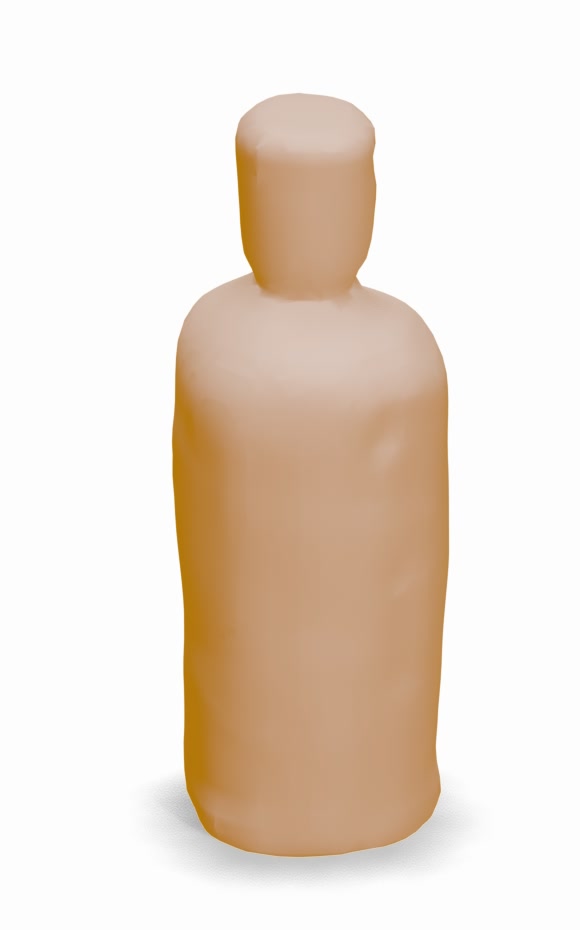} &
\includegraphics[height=0.12\textheight]{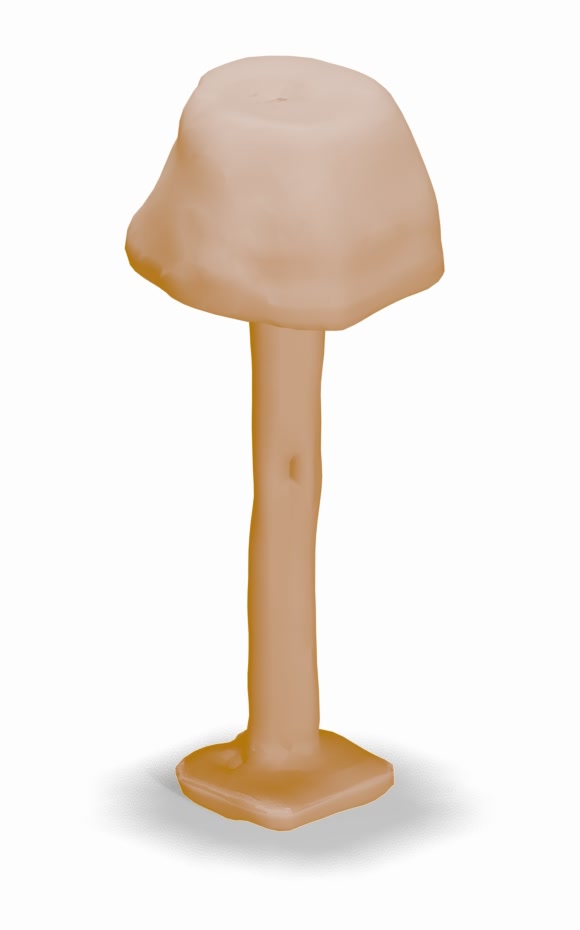} 
&
\includegraphics[height=0.12\textheight]{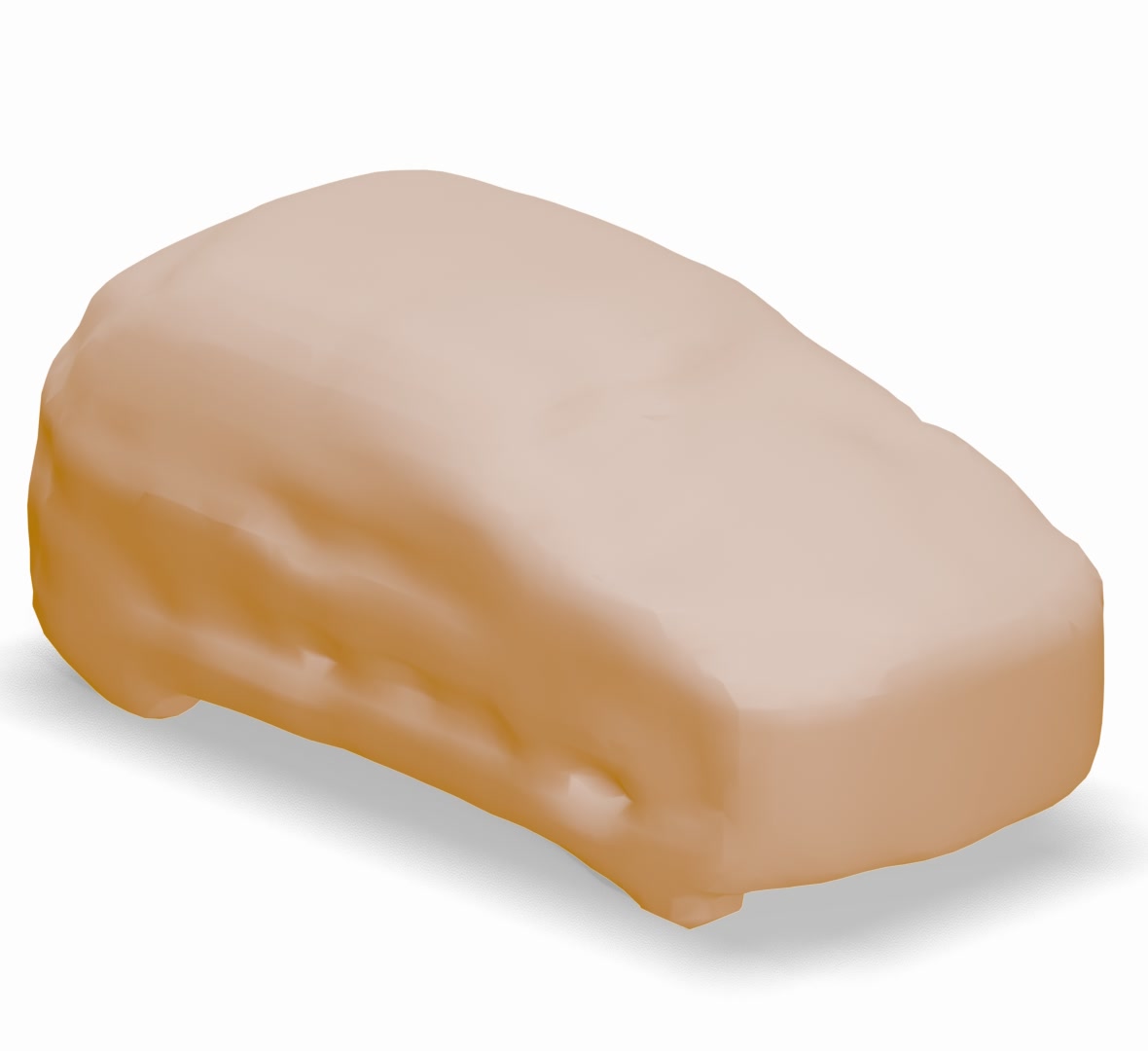} 
&
\includegraphics[height=0.12\textheight]{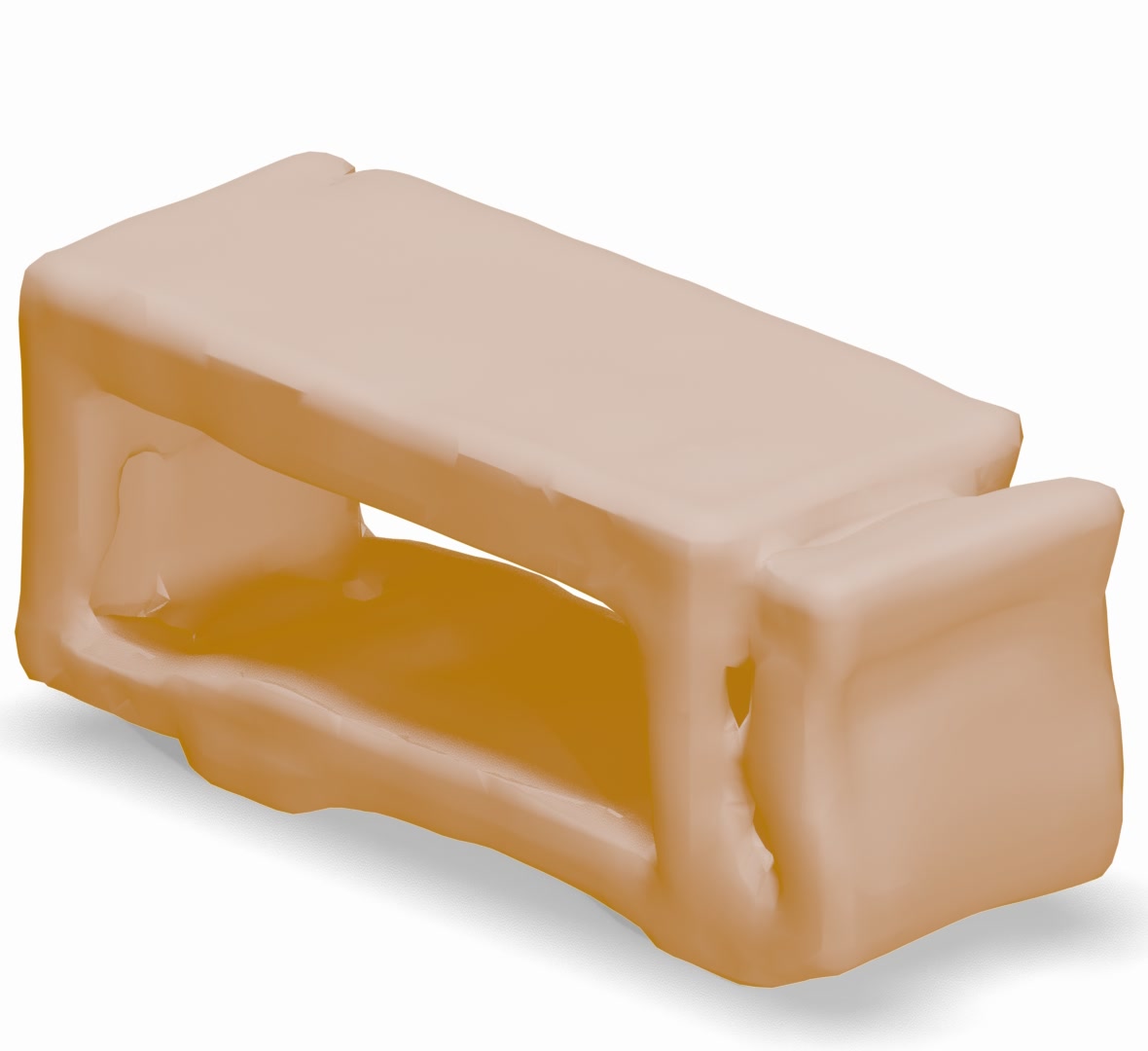} 
&
\includegraphics[height=0.12\textheight]{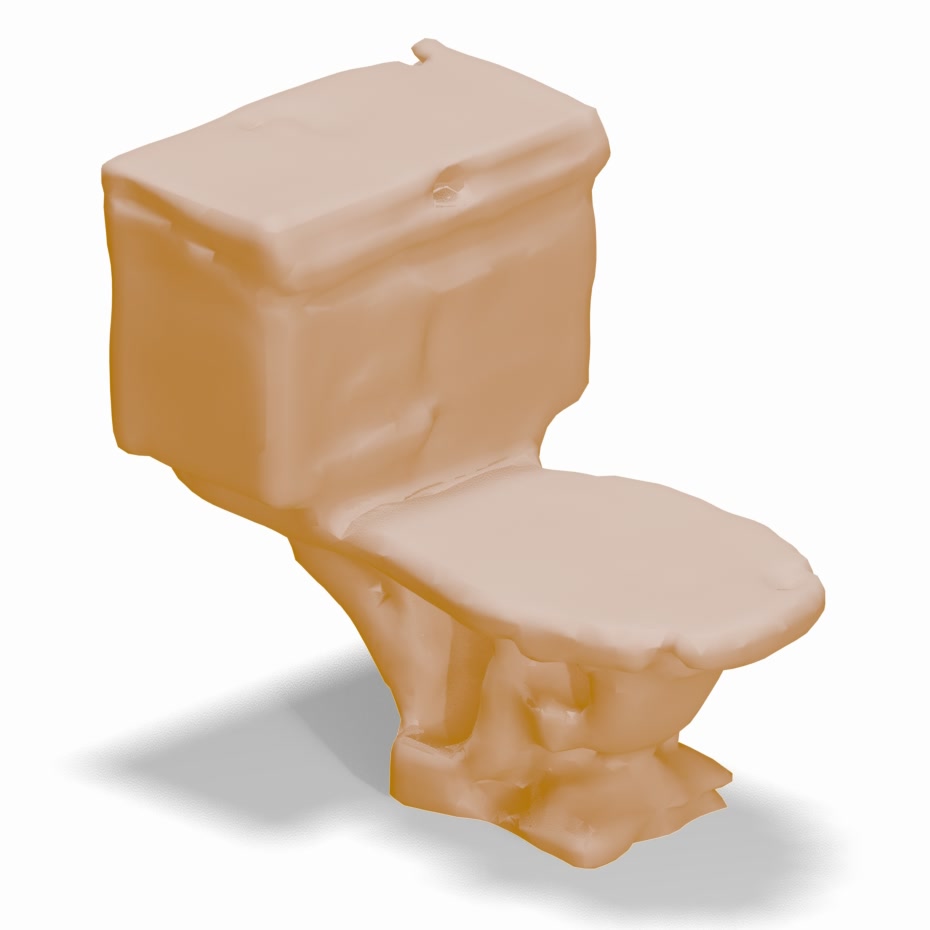} 
&
\includegraphics[height=0.12\textheight]{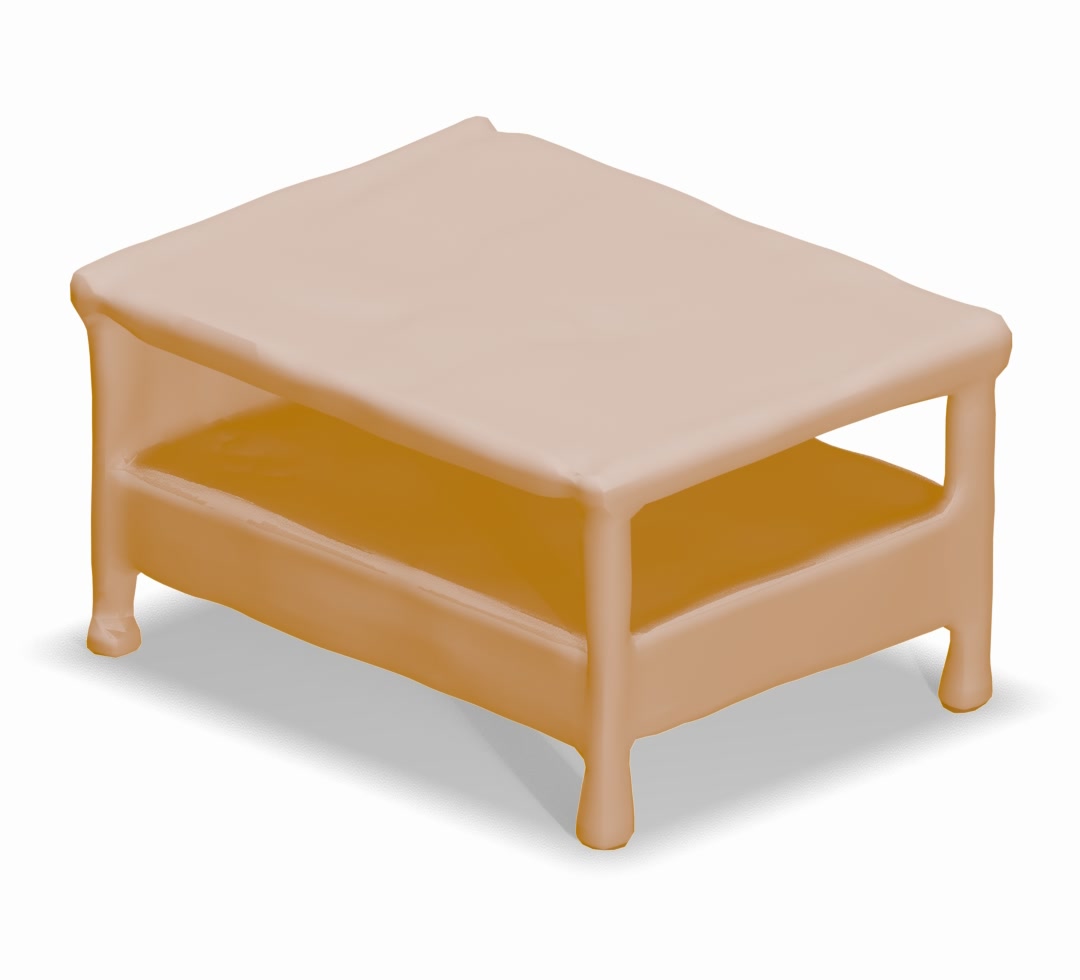} 
\end{tabular}

%

%
    }
    \vspace{-3mm}
\caption{{\bf Shape Generation from Unseen Classes.}
Results on sketches depicting object categories not present in the training data, including bottles, lamps, and cars from ShapeNet~\cite{chang2015shapenet}, and monitors, toilets, and beds from ModelNet~\cite{wu20153d}.
Despite the domain shift, our model generally produces plausible shapes aligned with the sketch intent.
However, some predictions reveal an overreliance on learned priors: trucks or beds may be completed into table-like structures.}
    \label{fig:suppl_unseen}
\end{figure*}

To investigate this, we evaluate our model on sketches of \emph{unseen categories} and \emph{unseen shape collections}.
Annotators produced VR sketches from ShapeNet classes excluded from training, as well as from the ModelNet dataset~\cite{wu20153d}.
Representative results are shown in \cref{fig:suppl_unseen}.

Overall, the model generalizes surprisingly well: for many unseen categories, the generated shapes are coherent, structurally consistent, and aligned with the intent of the sketch—despite never encountering such objects during training.
However, the influence of learned priors remains visible in edge cases; for example, a sketched truck or bed may be reconstructed as an empty table-like structure, or a toilet may be reconstructed with a closed lid as a chair-like structure, reflecting the dominance of furniture categories in the training set.

These results indicate that the model has indeed learned a meaningful sketch-to-shape mapping that transfers across datasets and categories, while also revealing the limits of its current shape diversity and the role of priors when sketch evidence is sparse or ambiguous.

\section{Impact of Sketcher Expertise}
\label{sec:suppl_expertise}
To evaluate the impact of the expertise of the sketcher, we evaluated sketches of 100 shapes drawn by both expert annotators and beginners. Non-expert sketches yield higher reconstruction accuracy ($72.2\scriptsize{\pm 2.6}$ vs.\ $66.3\scriptsize{\pm6.4}$ F-score),. We hypothesis that this is because experts introduce greater abstraction while beginners trace geometry more faithfully.



\section{Precision/Performance Tradeoff}
\label{sec:perf}
In the main paper, we reported results using 100 DDIM steps, which yield the best reconstruction quality but account for over 99\% of the inference time.  
This setting results in a latency of roughly 6 seconds per sketch, which may be impractical in interactive design scenarios.  

To assess whether fewer sampling steps provide a better speed–accuracy compromise, we evaluate our model with 10, 25, 50, and 100 DDIM steps.  
As shown in \cref{tab:perf}, performance remains remarkably stable even with as few as 10 steps, while inference becomes over 3$\times$ faster.  
This indicates that interactive or real-time use cases can run with drastically reduced sampling budgets at minimal loss in quality.

\begin{table}[t]
    \centering
    \begin{tabular}{cccc}
    \toprule
    DDIM step &  F-score $\uparrow$& CD $\times 1000$ $\downarrow$        & time (samples/s)\\ 
    \midrule
    10        &  69.24  & 5.04     &  2.26\\
    25        &  69.70  & 4.82     &  3.06\\
    50        &  69.74  & 4.89     &  4.47\\
    100       &  69.80  & 4.78     &  6.33\\
    \bottomrule
    \end{tabular}
    \caption{{\bf Performance/Speed Tradeoff.}   Reconstruction accuracy remains stable even with a small number of DDIM steps, while inference becomes substantially faster (computed with a batch size of 1).
}
    \label{tab:perf}
\end{table}

\section{Additional Qualitative Results}
\label{sec:sup_quali}
\begin{figure*}[h]
    \centering
        \resizebox{\linewidth}{!}{
\begin{tabular}{l@{\;}c@{\;}c@{\;}c@{\;}c@{\;}c@{\;}c@{\;}c@{\!\!}r}
\rotatebox{90}{\small \bf  \quad Real sketch }&
\includegraphics[height=0.10\textheight]{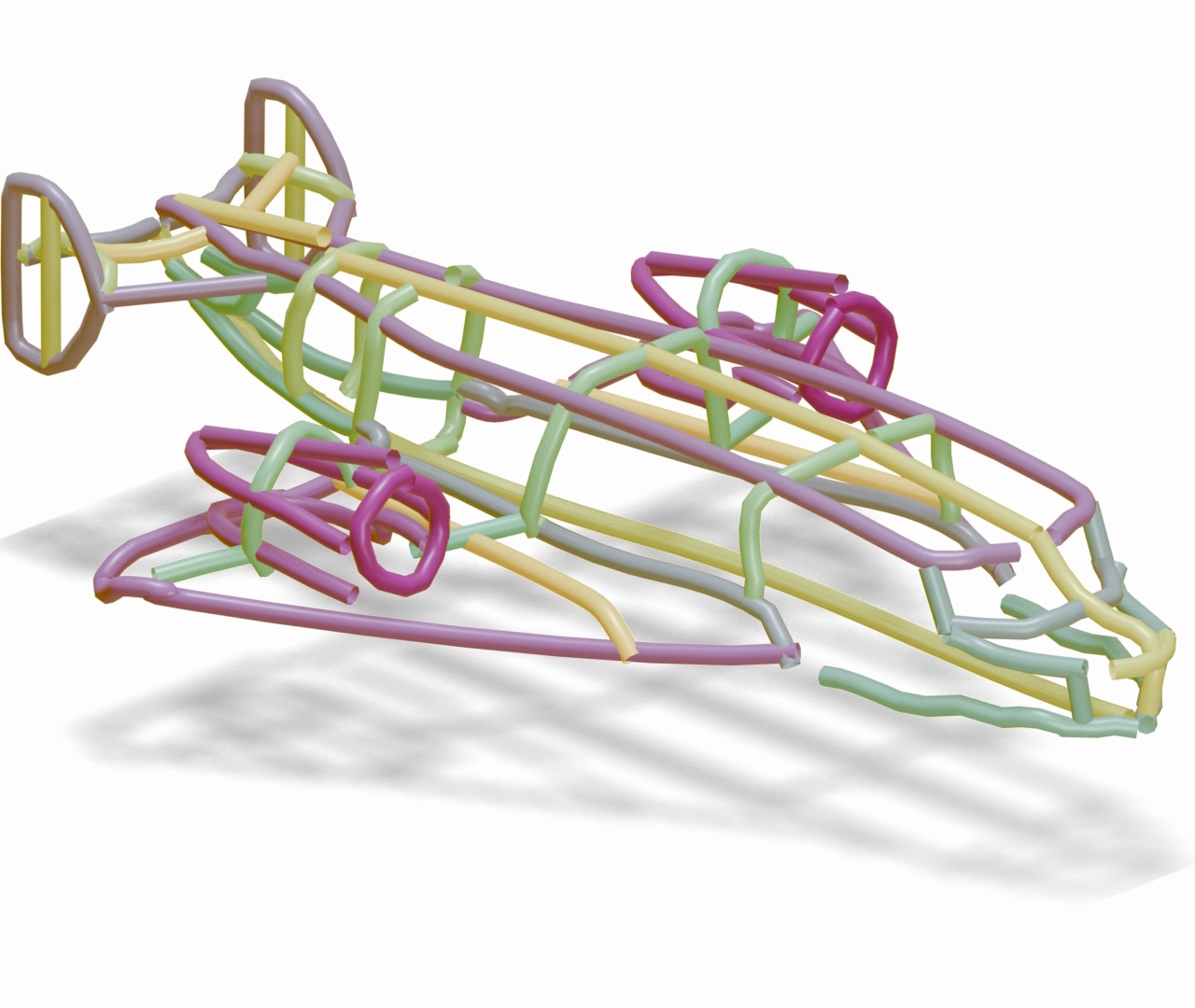} &
\includegraphics[height=0.10\textheight]{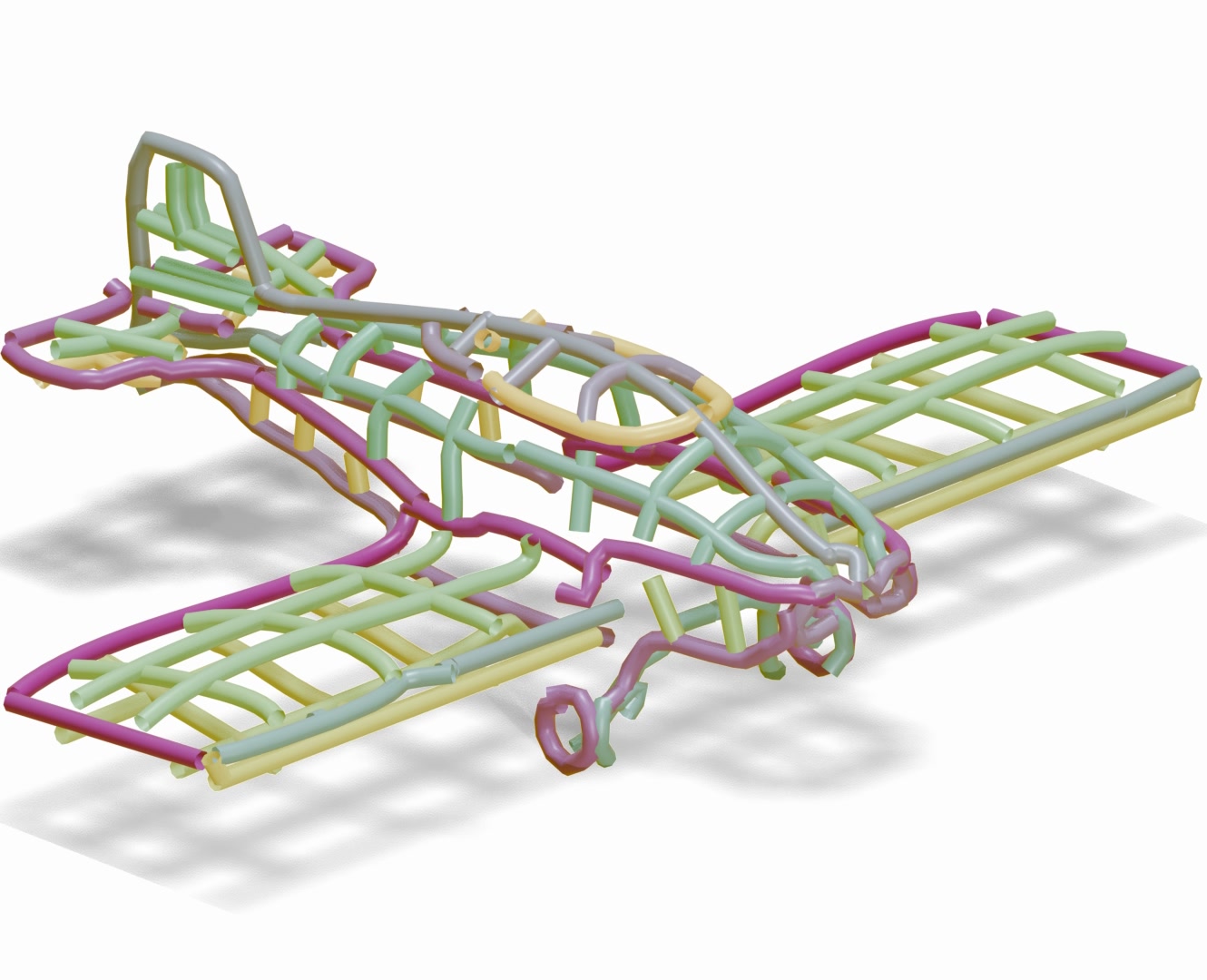} &
\includegraphics[height=0.115\textheight]{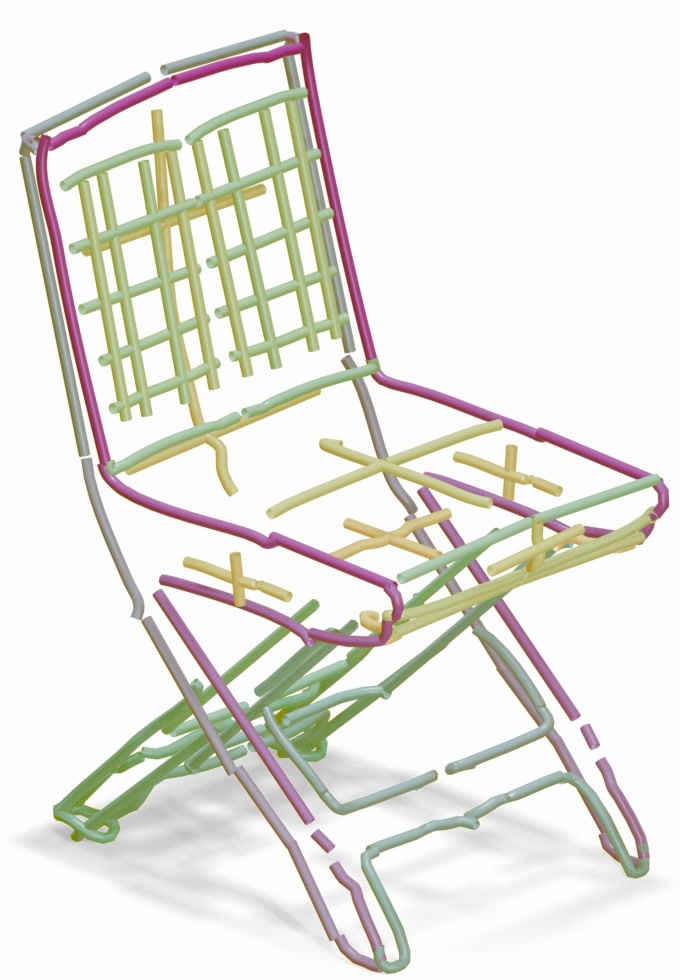} 
&
\includegraphics[height=0.115\textheight]{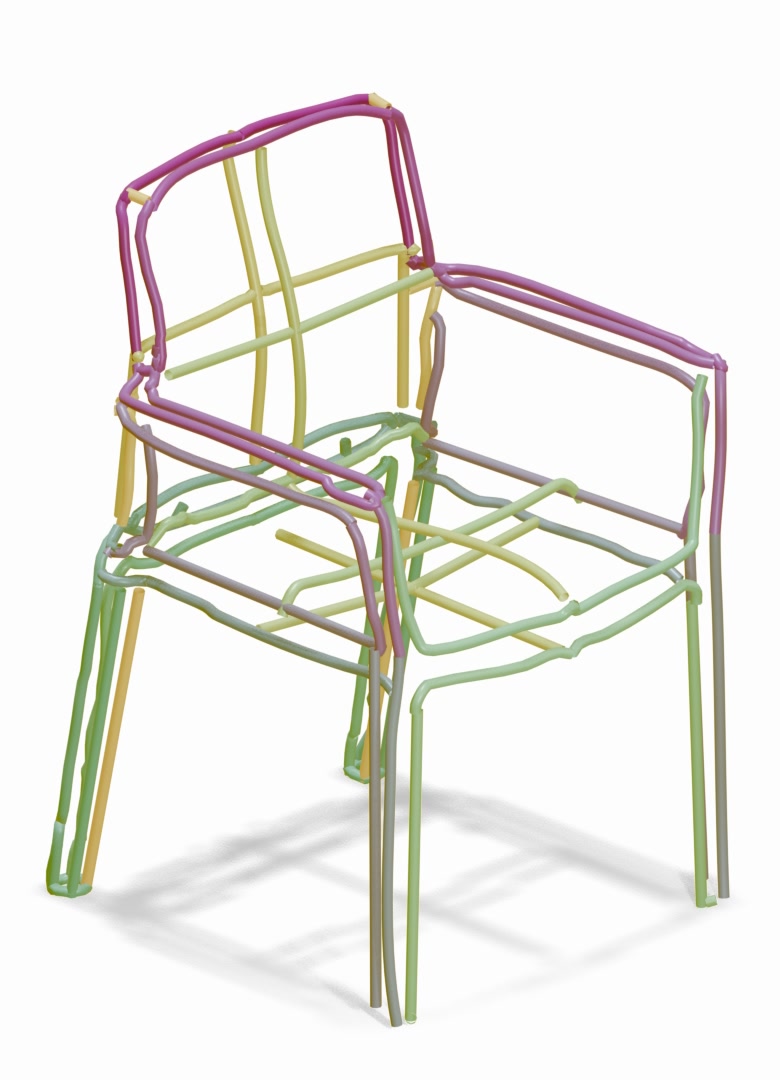} 
&
\includegraphics[height=0.115\textheight]{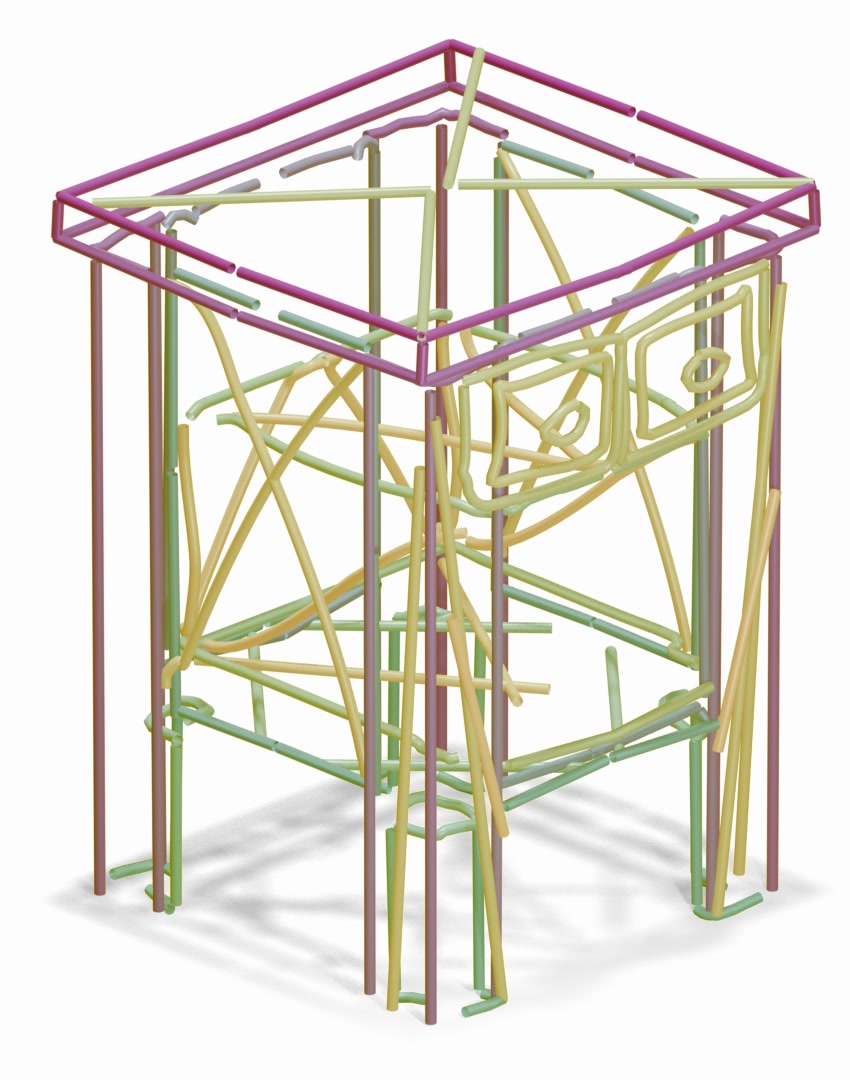}
&
\includegraphics[height=0.115\textheight]{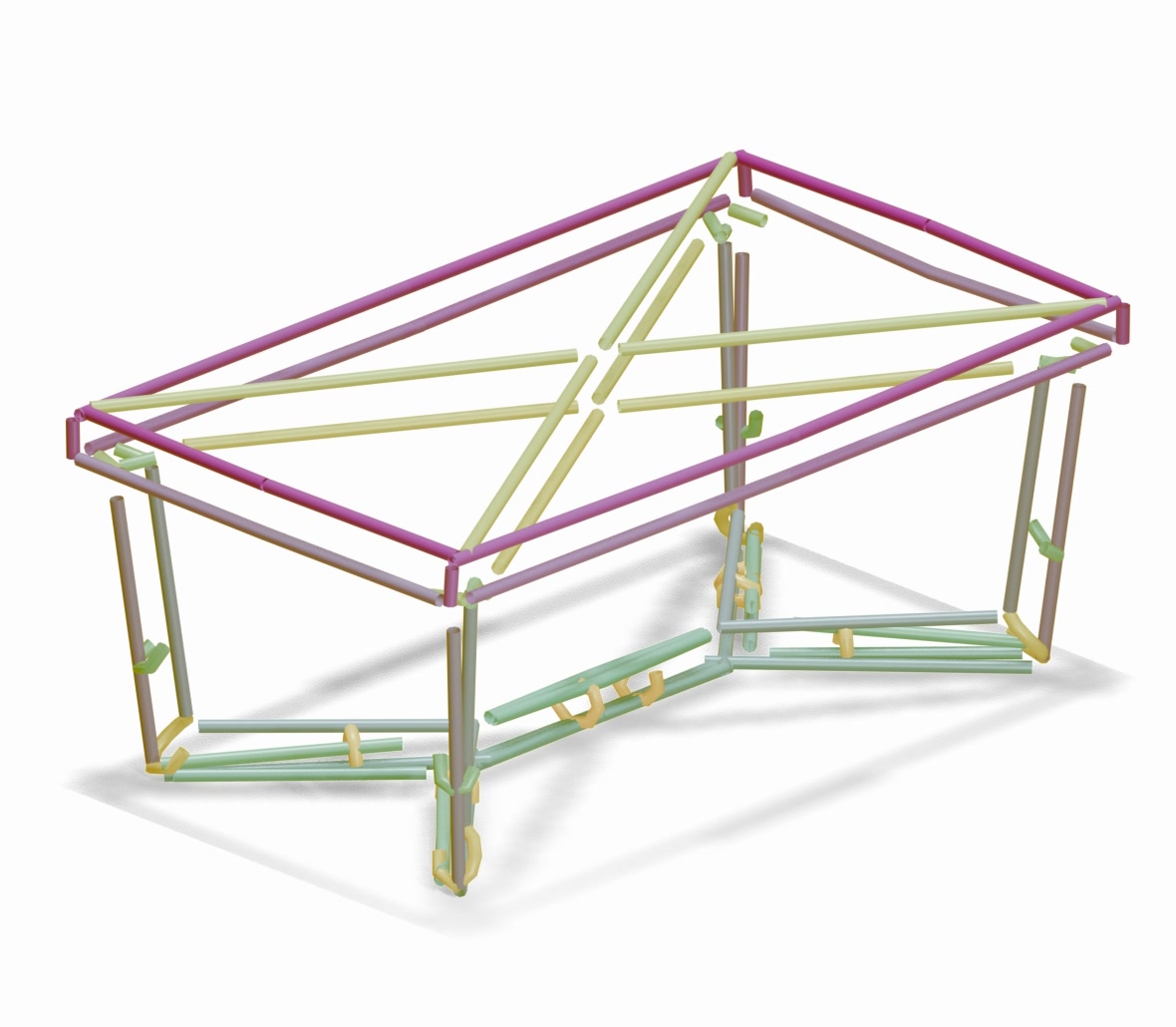}
&
\includegraphics[height=0.115\textheight]{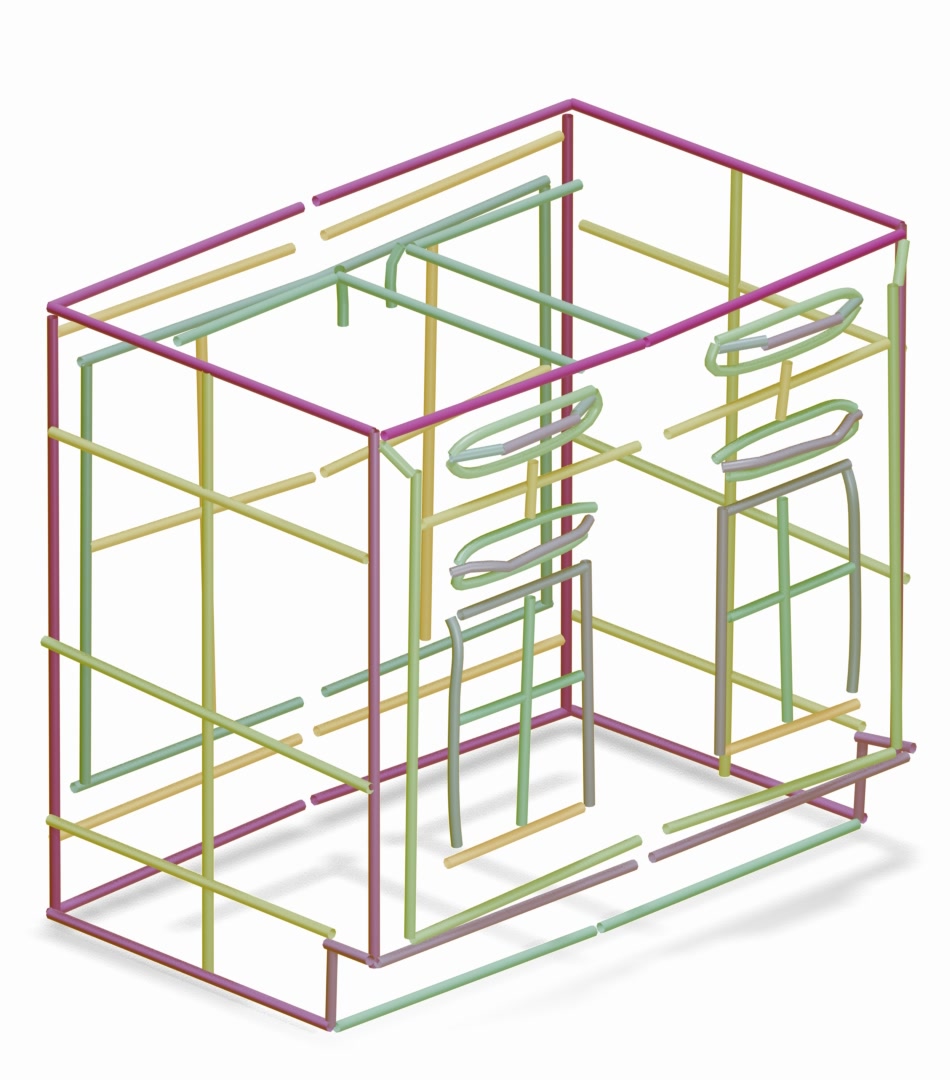}
&
\begin{tikzpicture}[baseline=-0.040\textheight]
      \node[draw=none, inner sep=0pt] (jet) at (0,0) {\rotatebox{0}{\def\colormapheight{50}\rotatebox{0}{\input{figures/jetcolorbar}}}};
    \node[above=-0mm of jet]{\scriptsize \!start};
    \node[below=-9.0mm of jet.south]{\scriptsize \!end};
\end{tikzpicture}
\\[-4mm]
\rotatebox{90}{\small \bf \qquad  GT shape} &
\includegraphics[height=0.10\textheight]{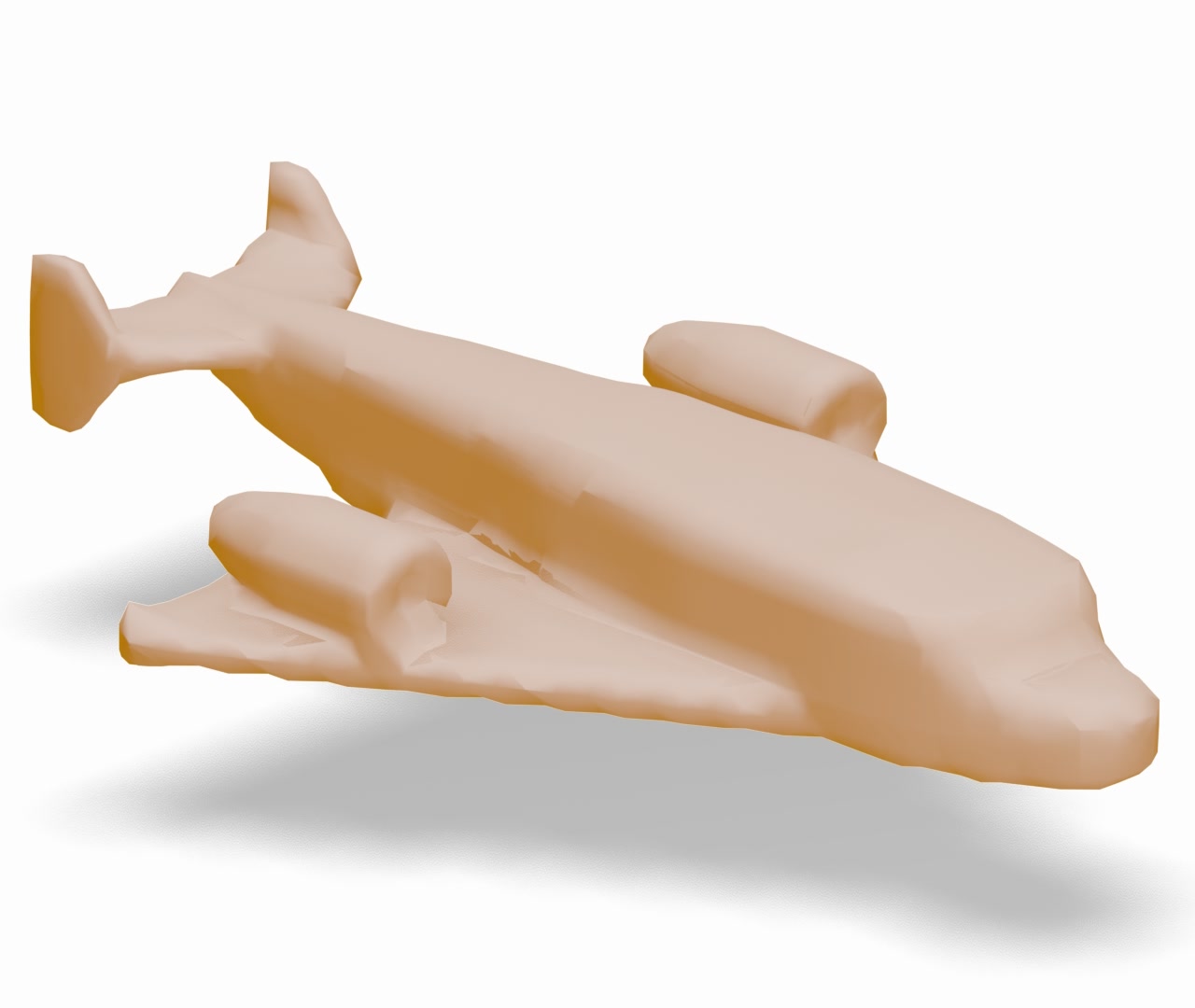} &
\includegraphics[height=0.10\textheight]{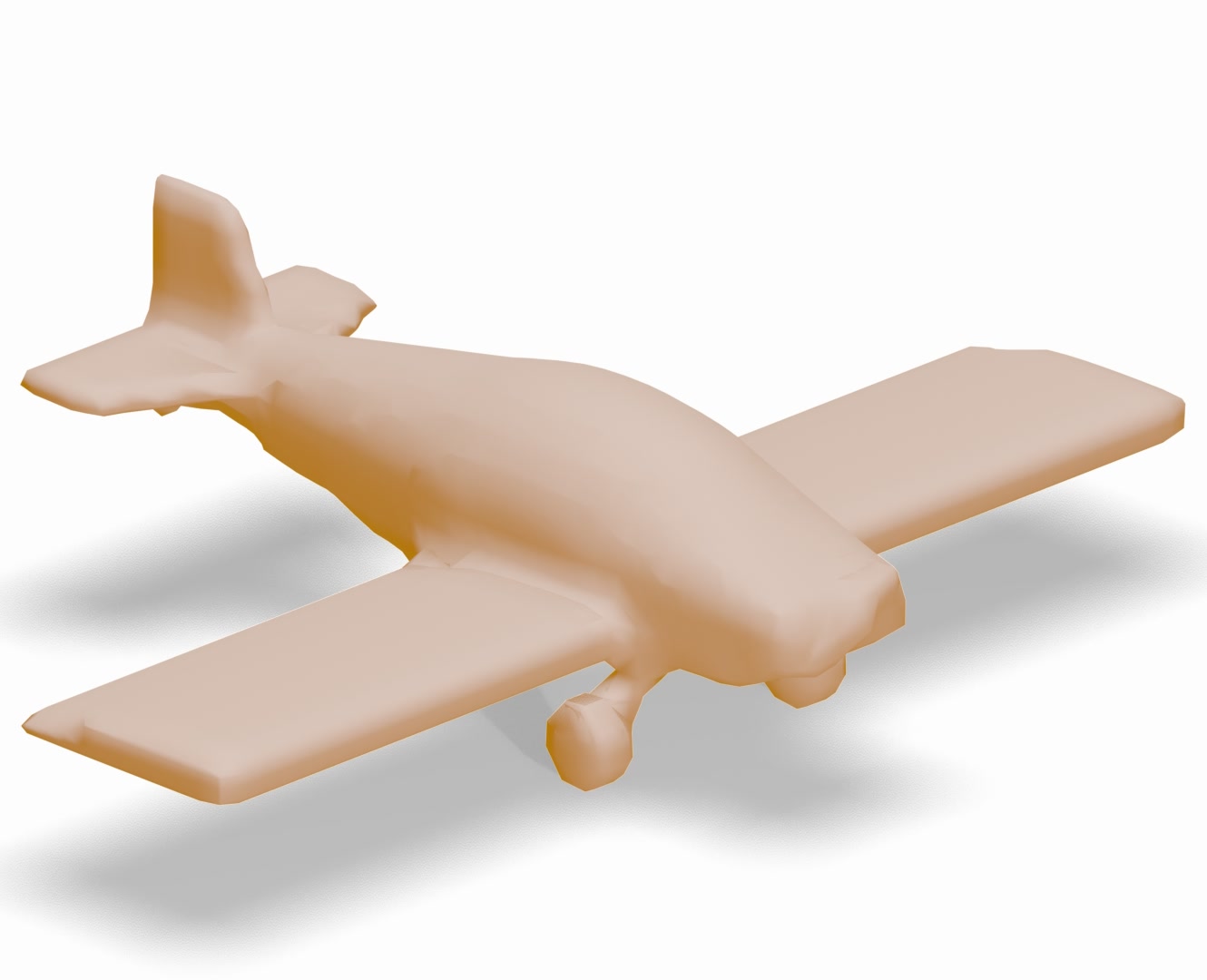} &
\includegraphics[height=0.115\textheight]{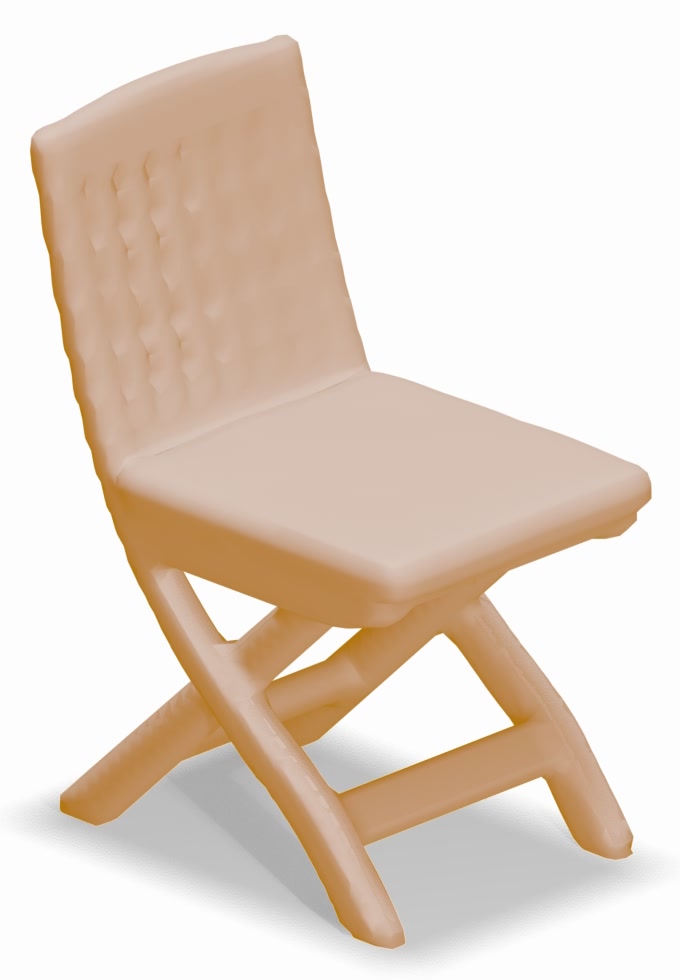} 
&
\includegraphics[height=0.115\textheight]{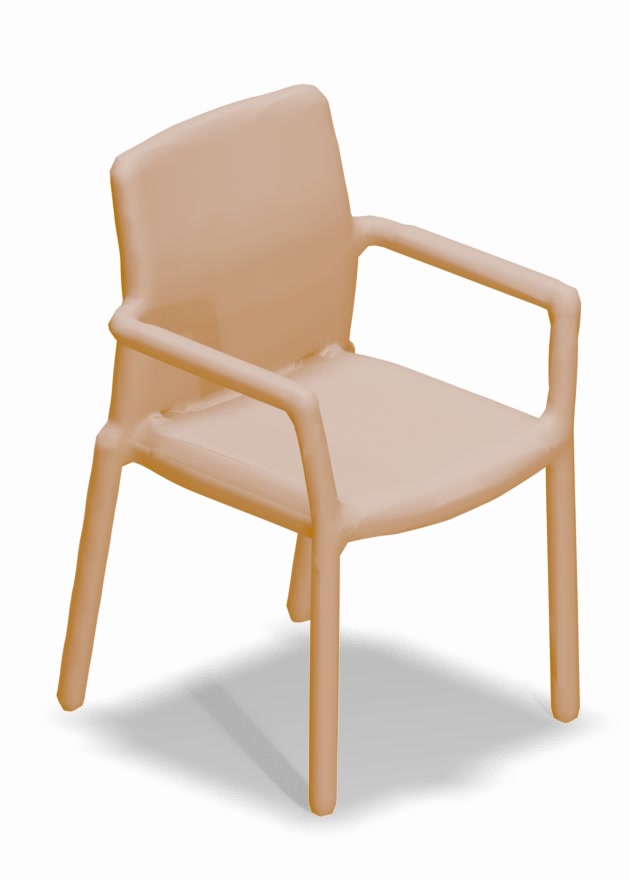}
&
\includegraphics[height=0.115\textheight]{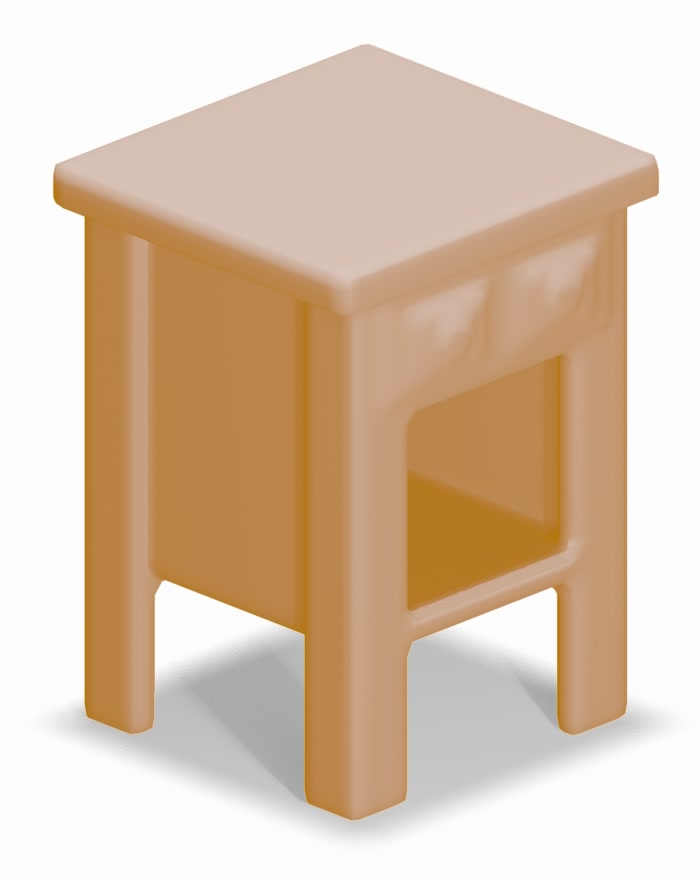}
&
\includegraphics[height=0.115\textheight]{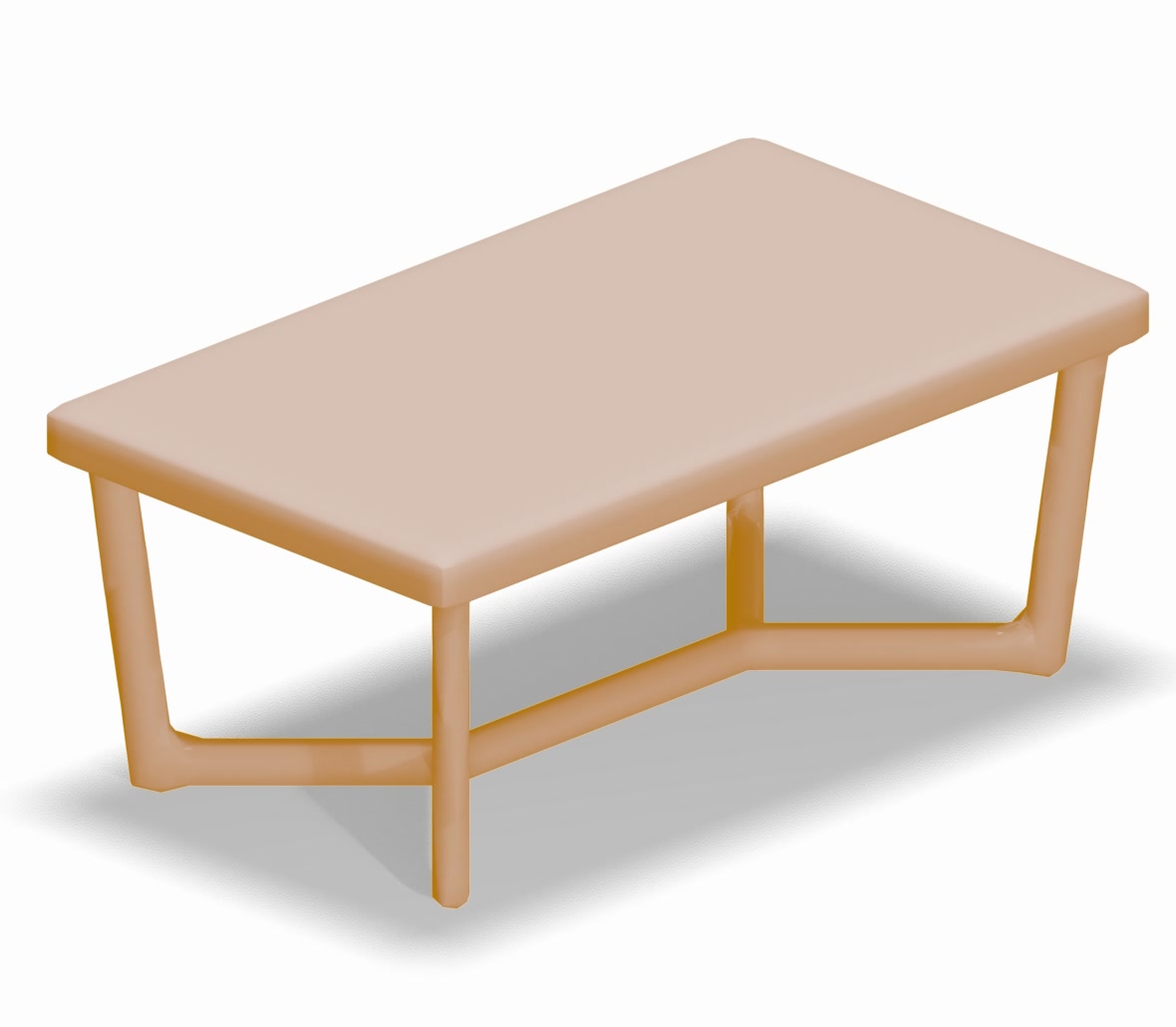}
&
\includegraphics[height=0.115\textheight]{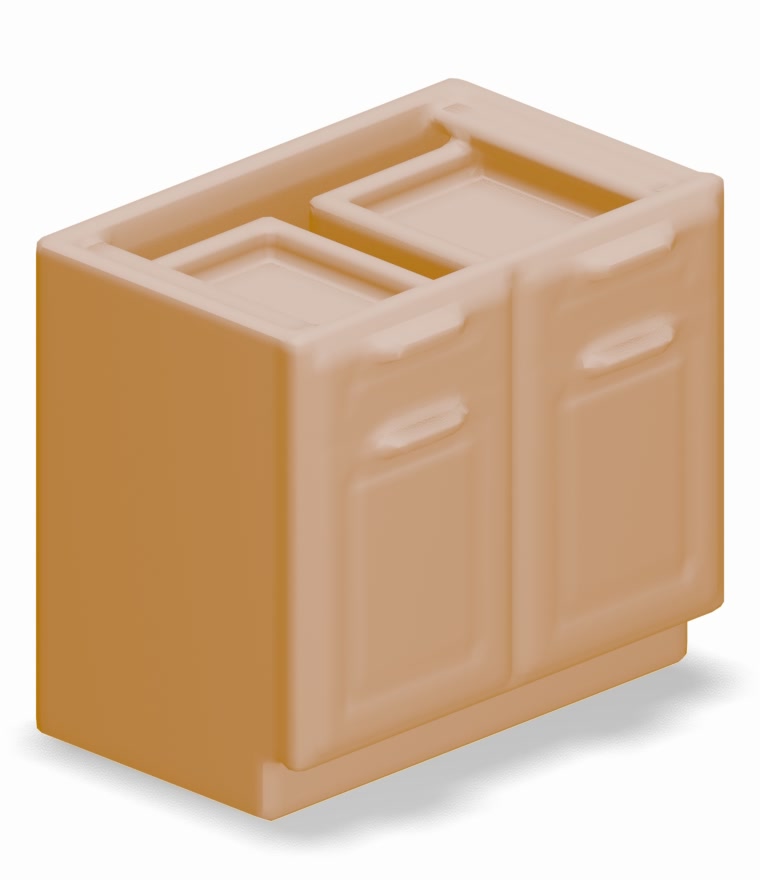}
\\
[-1mm]
\rotatebox{90}{\small \bf \;\; Our prediction} &
\includegraphics[height=0.10\textheight]{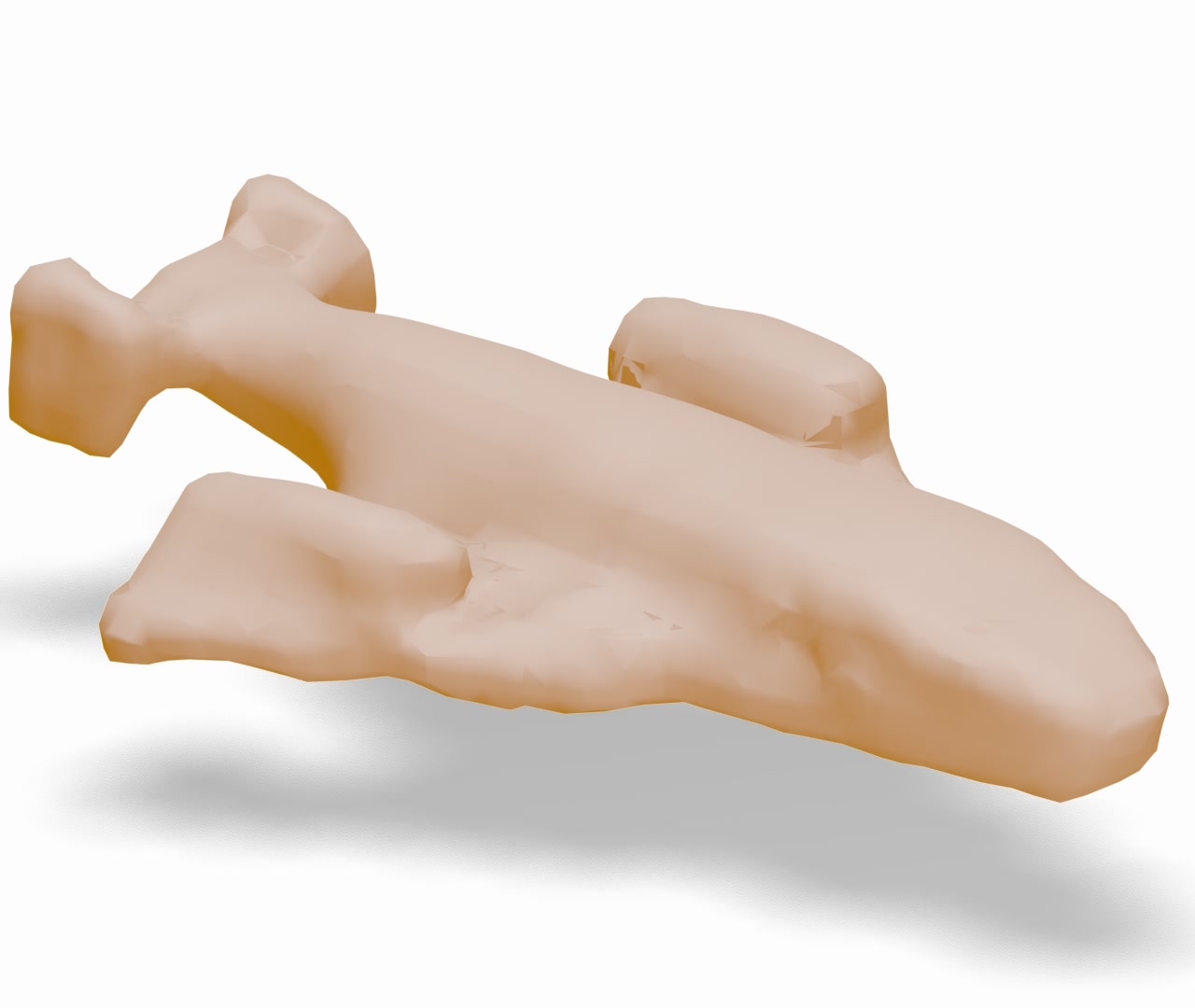} &
\includegraphics[height=0.10\textheight]{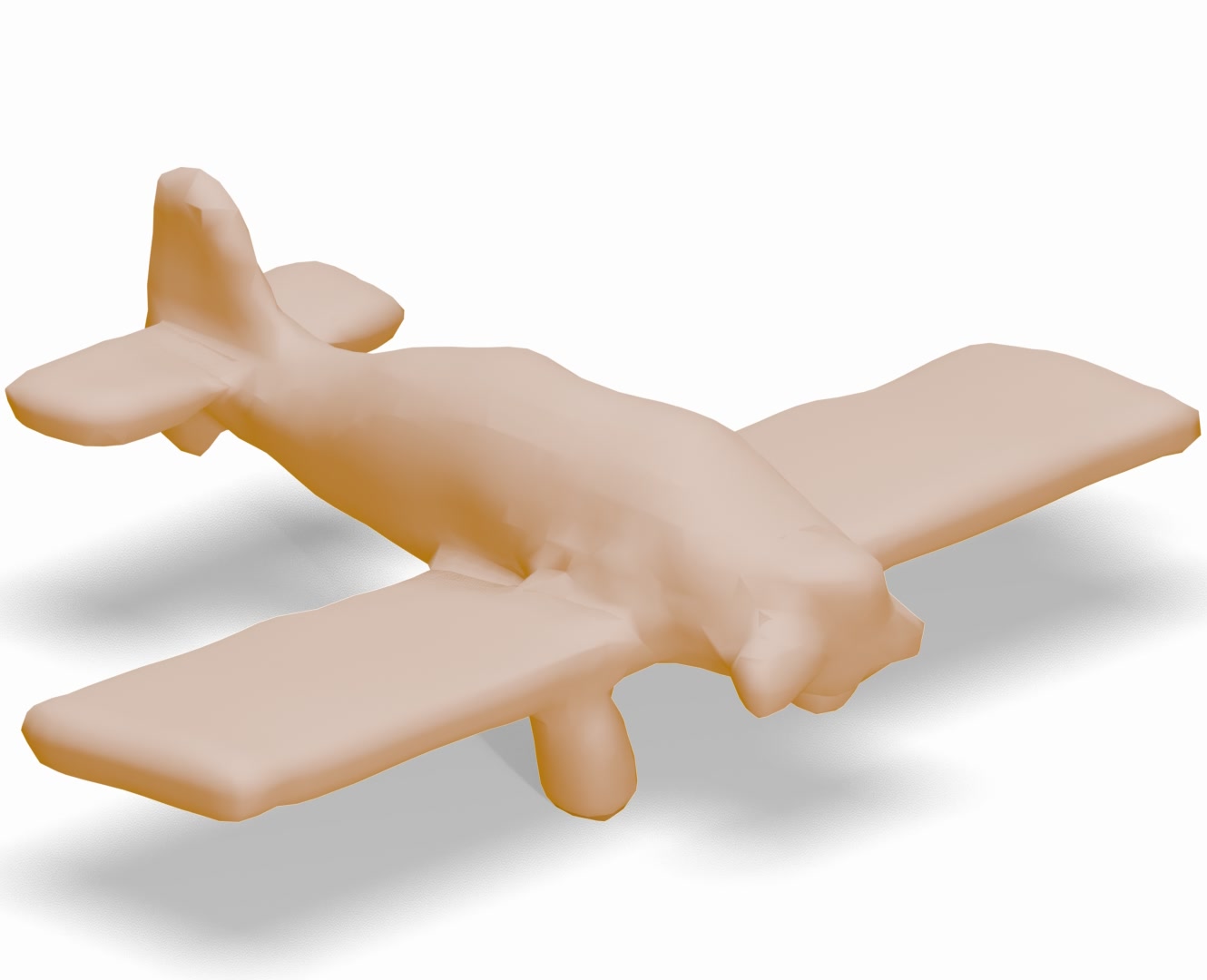} &
\includegraphics[height=0.115\textheight]{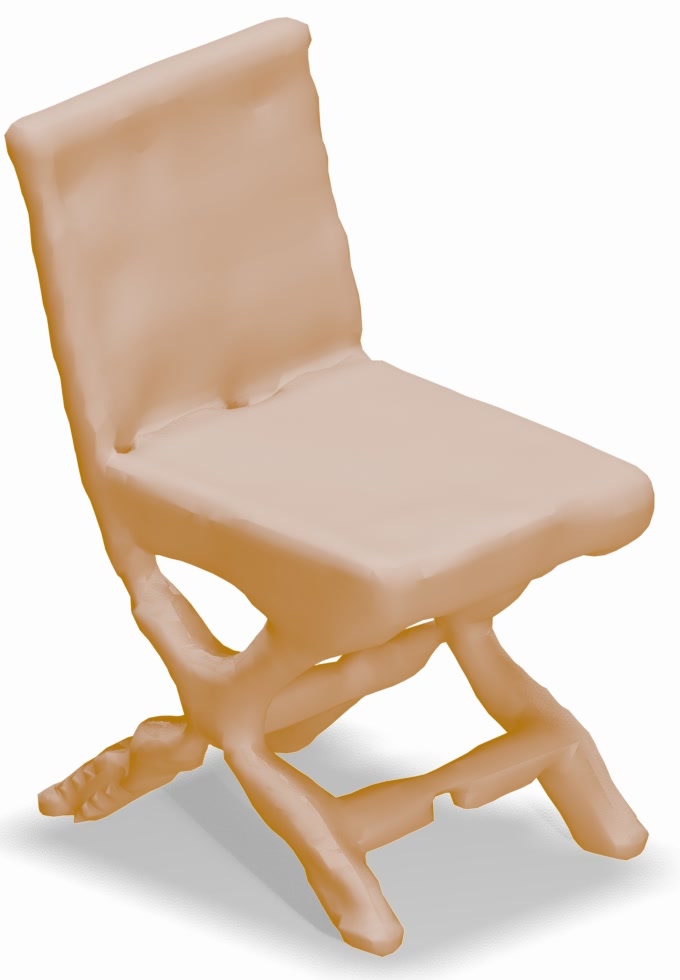}
&
\includegraphics[height=0.115\textheight]{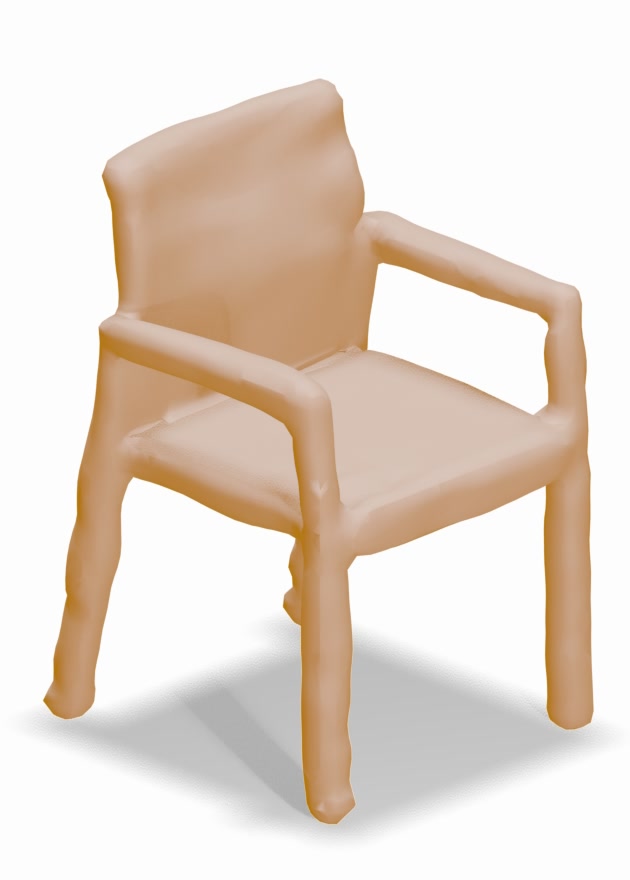} 
&
\includegraphics[height=0.115\textheight]{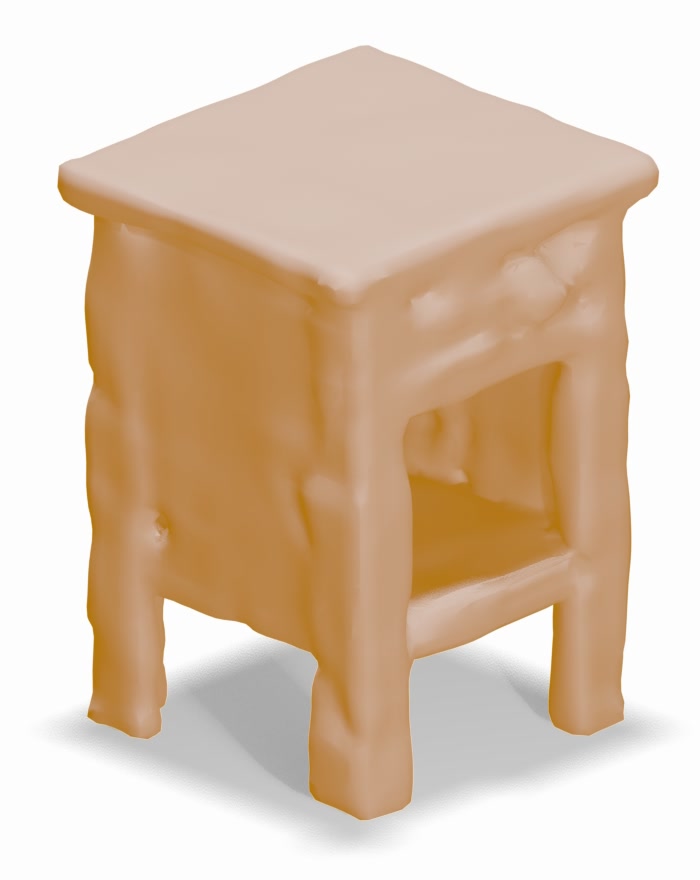}
&
\includegraphics[height=0.115\textheight]{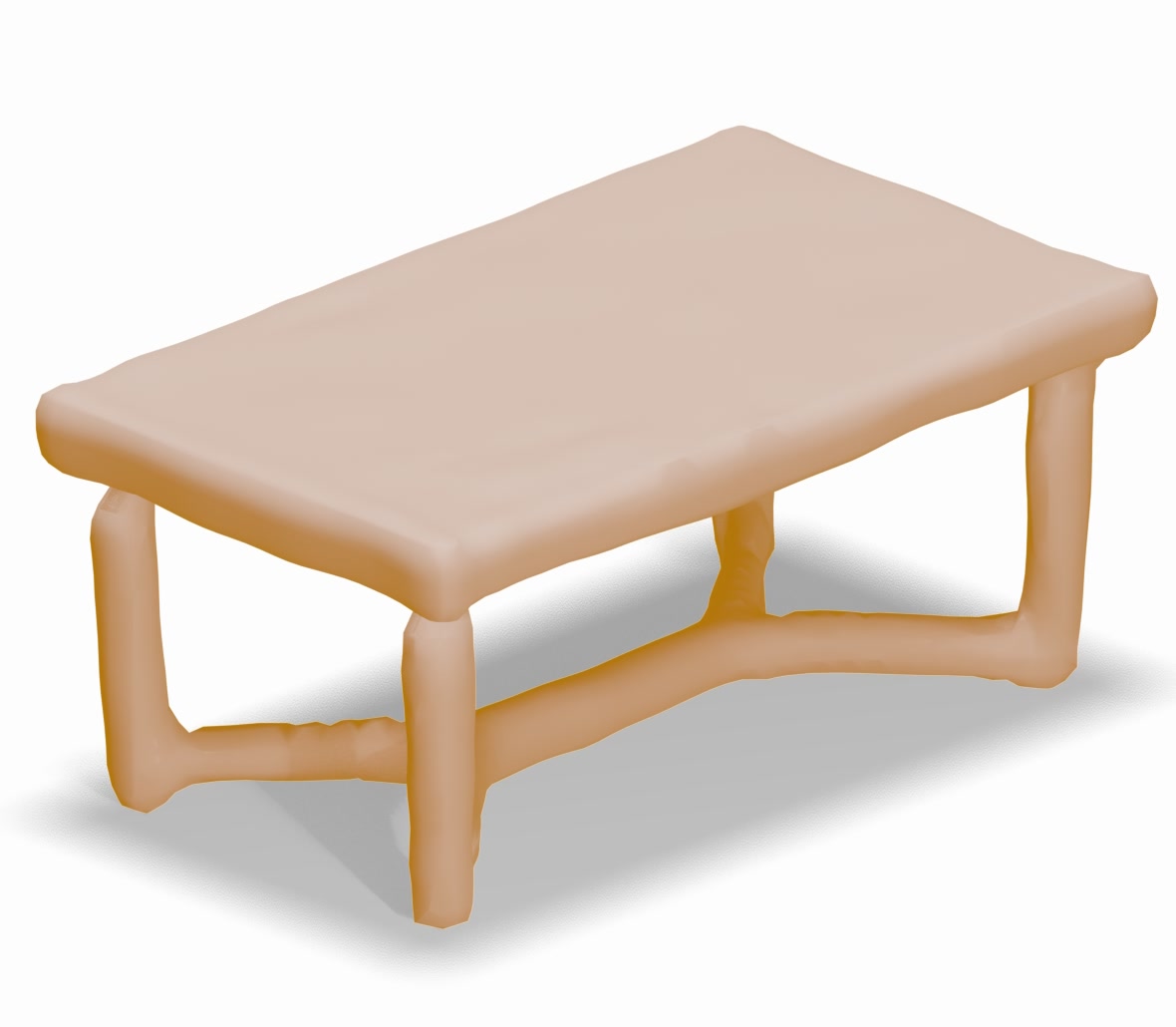}
&
\includegraphics[height=0.115\textheight]{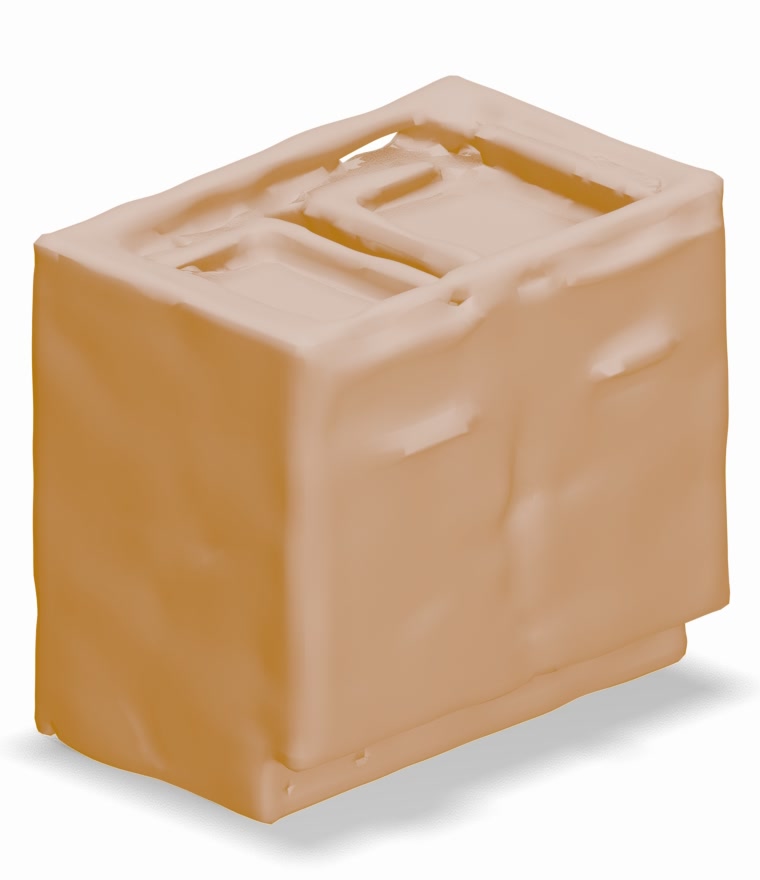}
\\
[-3mm]
\rotatebox{90}{\small \bf Luo's prediction} &
\includegraphics[height=0.10\textheight]{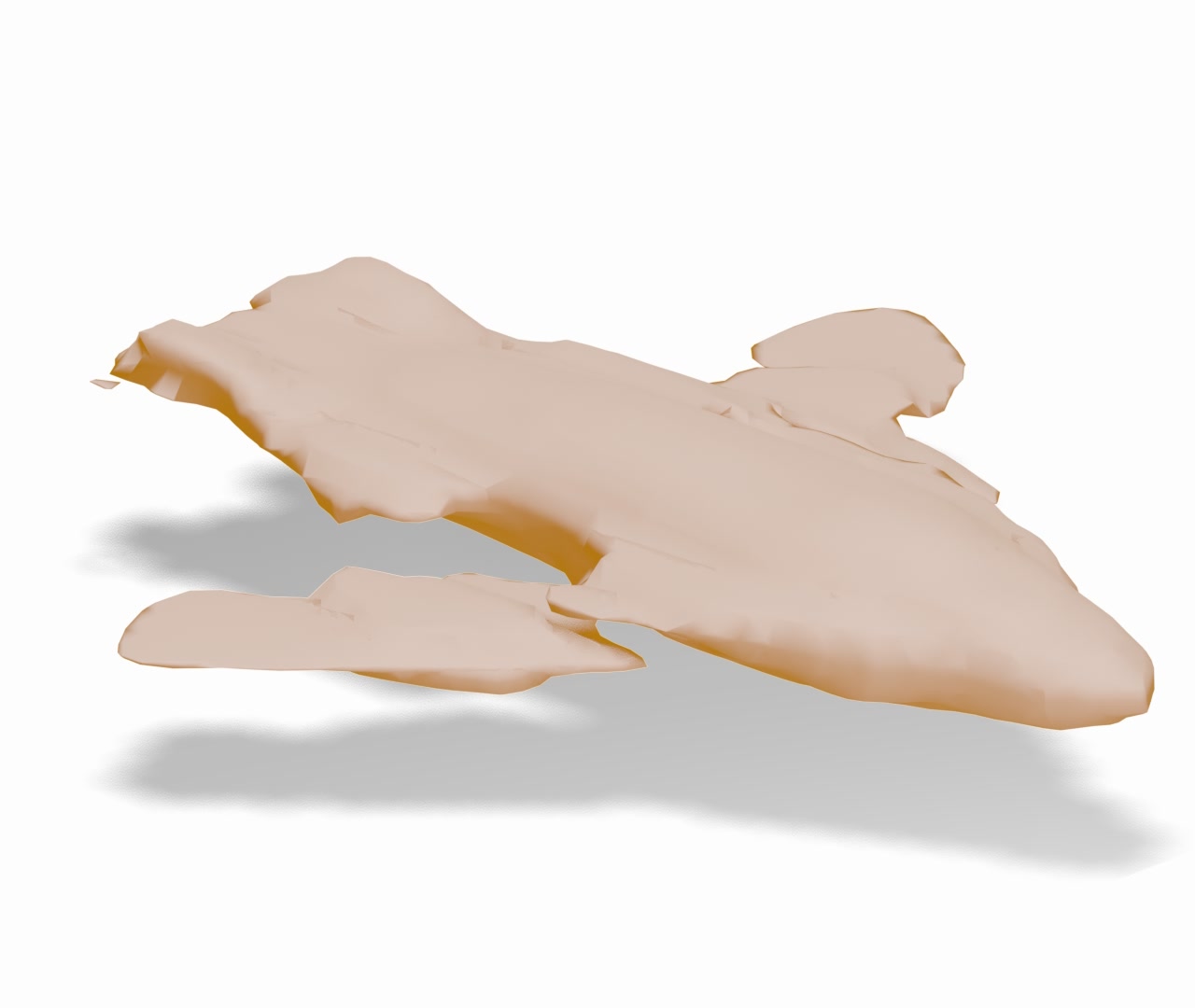} &
\includegraphics[height=0.10\textheight]{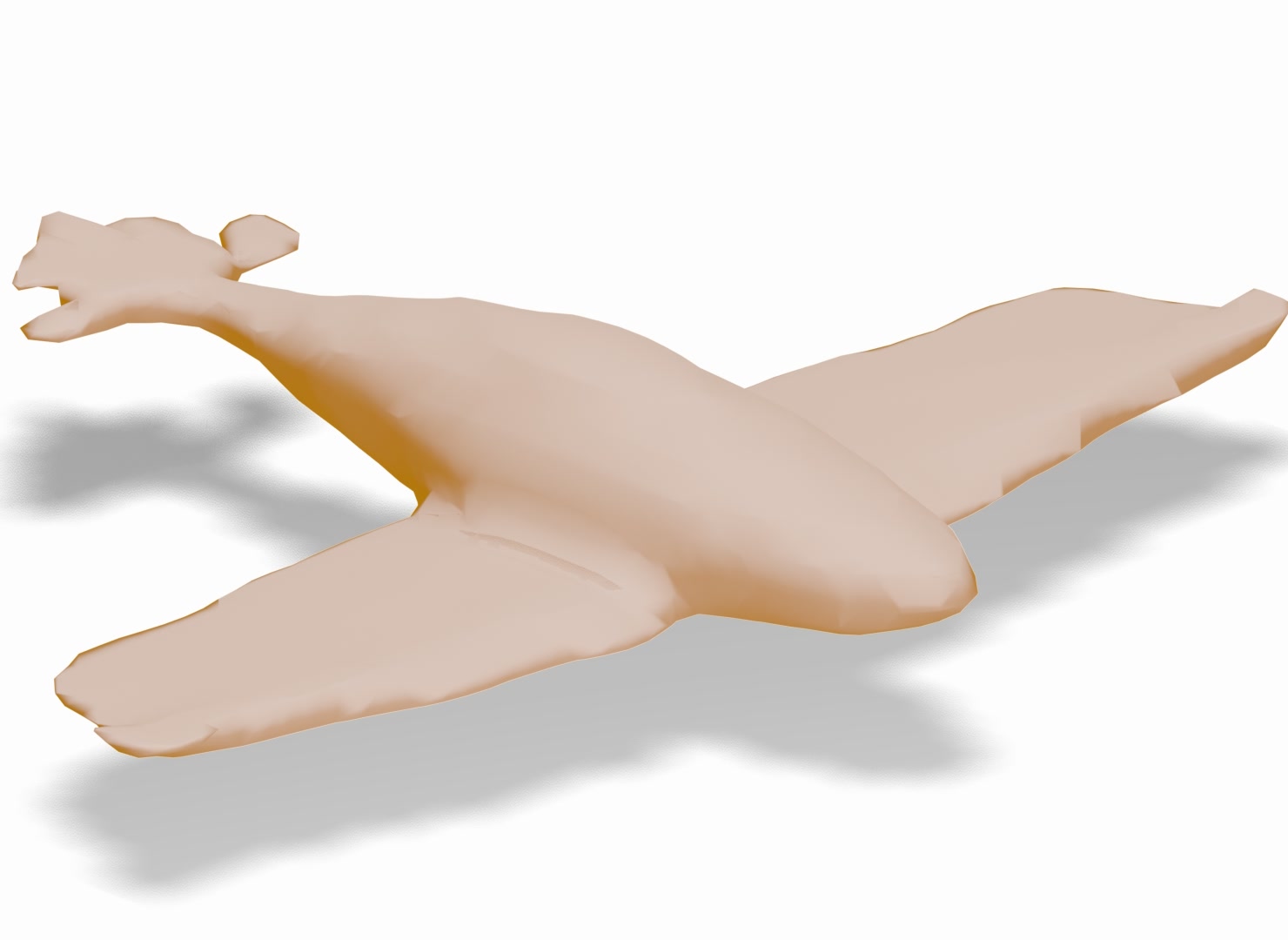} &
\includegraphics[height=0.115\textheight]{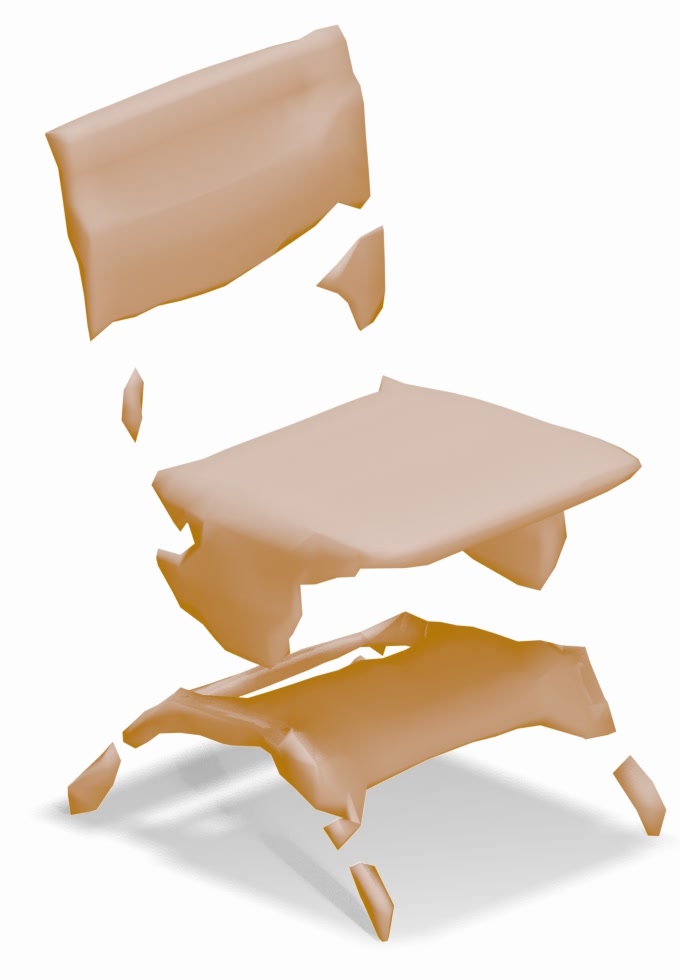}
&
\includegraphics[height=0.115\textheight]{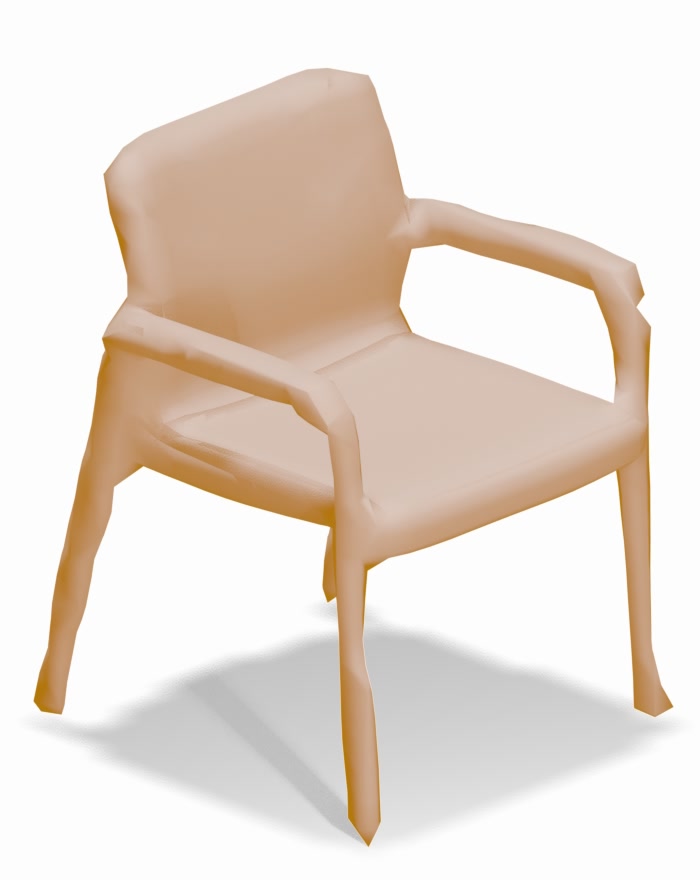} 
&
\includegraphics[height=0.115\textheight]{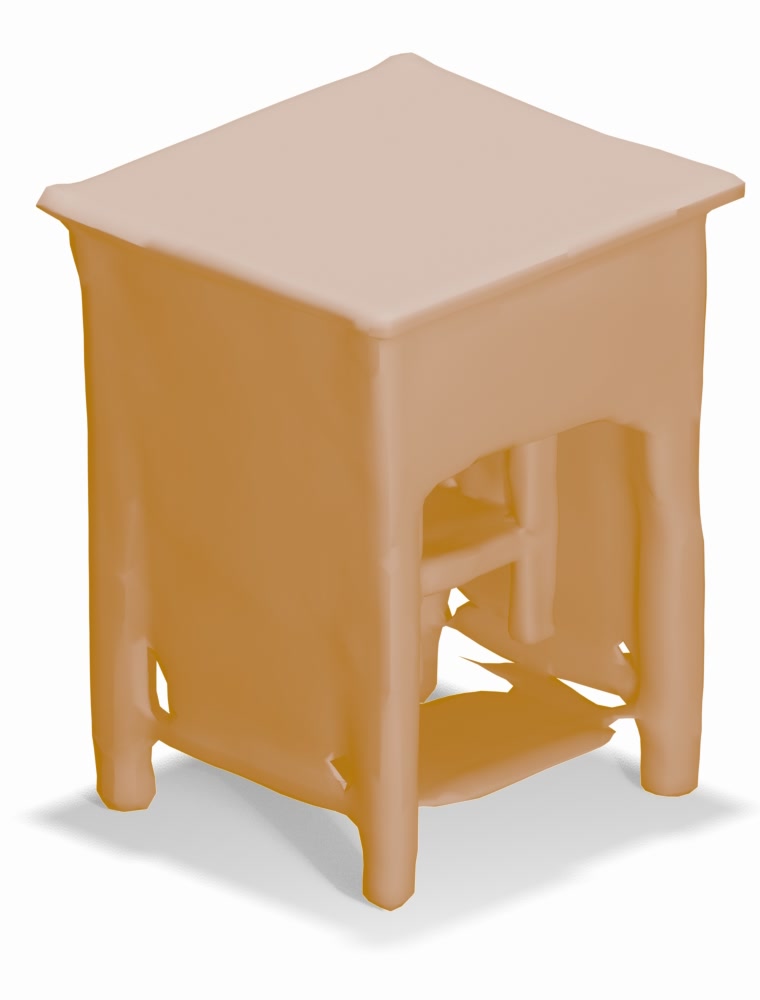}
&
\includegraphics[height=0.115\textheight]{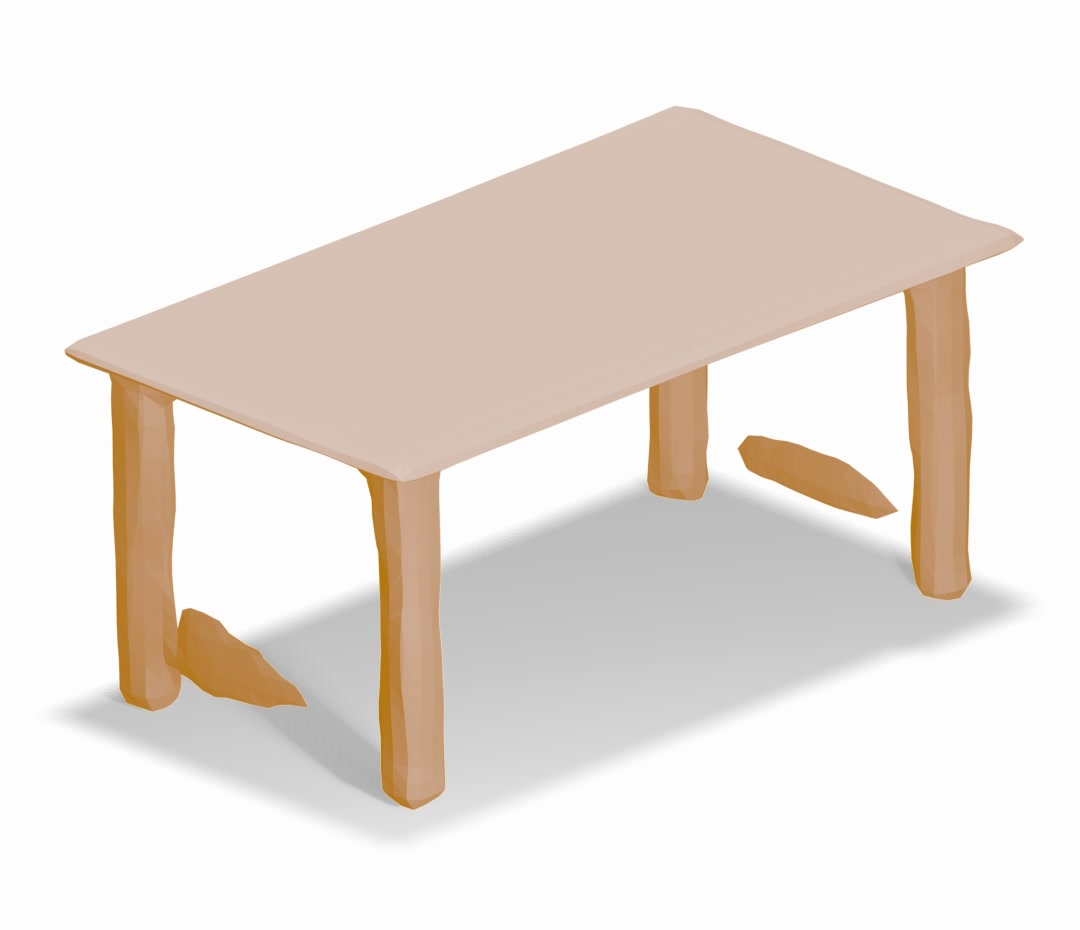}
&
\includegraphics[height=0.115\textheight]{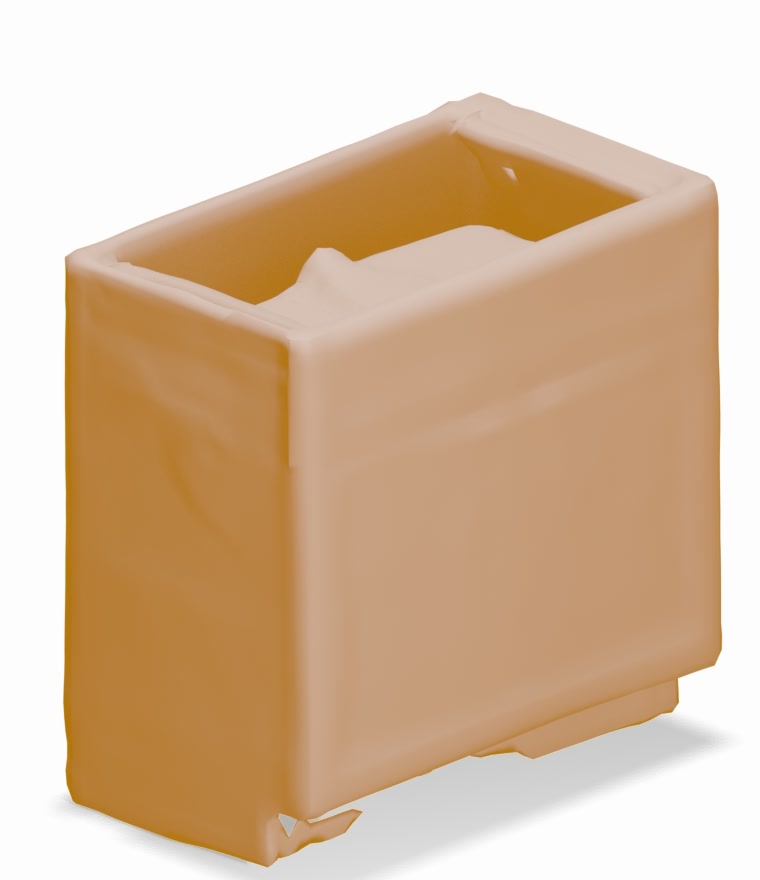}
\\
[-3mm]
\rotatebox{90}{\small \bf LAS-diffusion} &
\includegraphics[height=0.10\textheight]{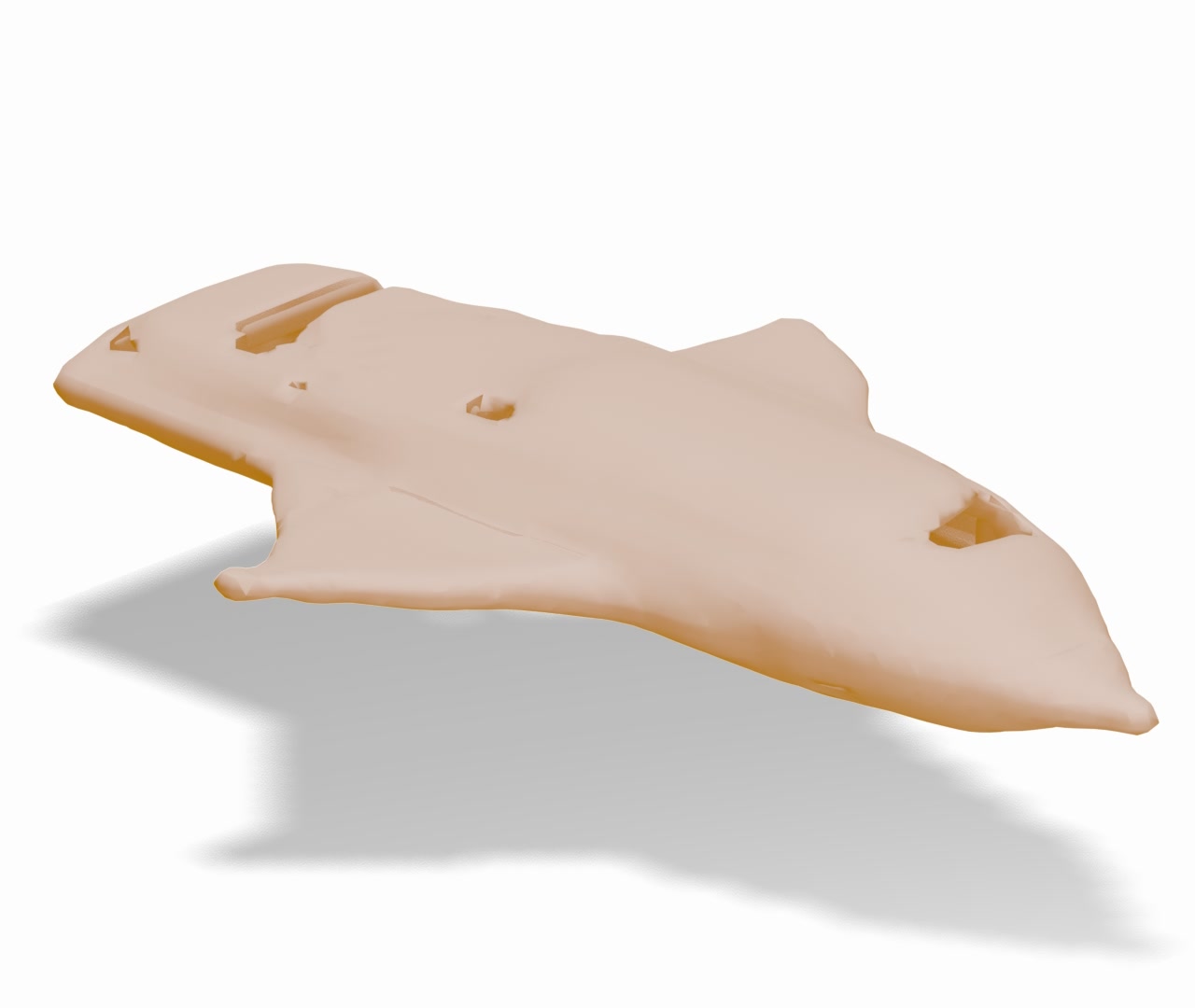} &
\includegraphics[height=0.10\textheight]{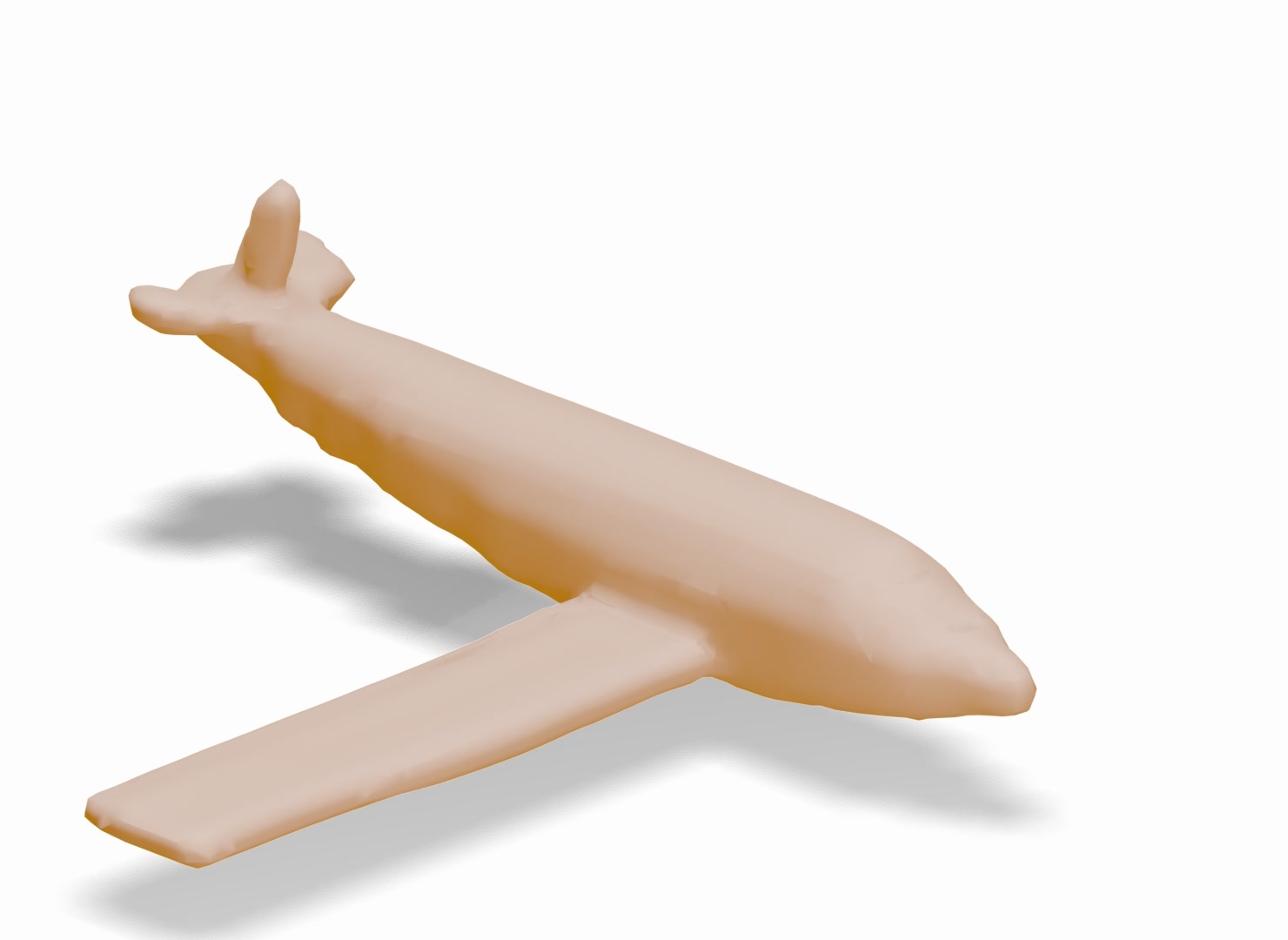} &
\includegraphics[height=0.115\textheight]{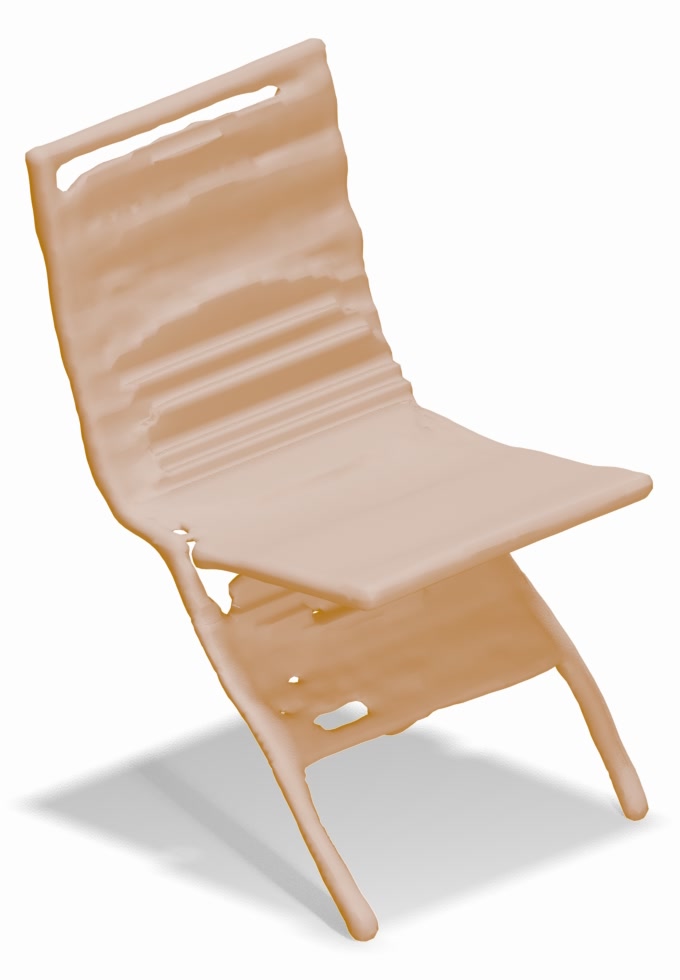}
&
\includegraphics[height=0.115\textheight]{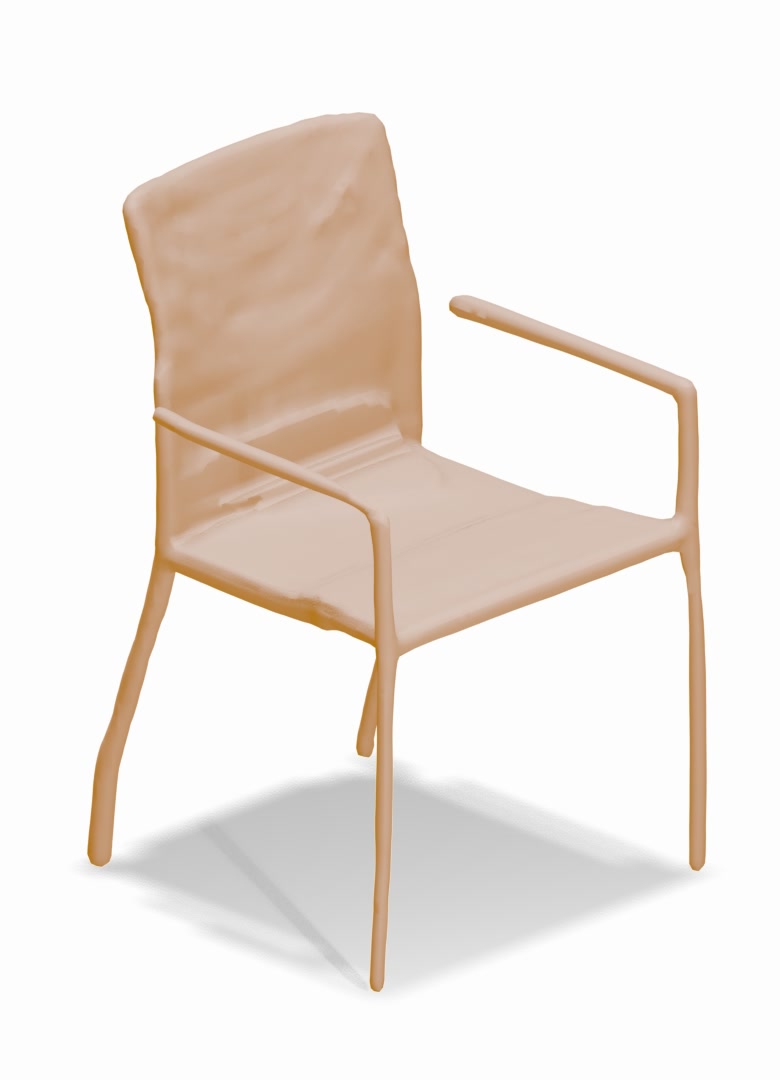} 
&
\includegraphics[height=0.115\textheight]{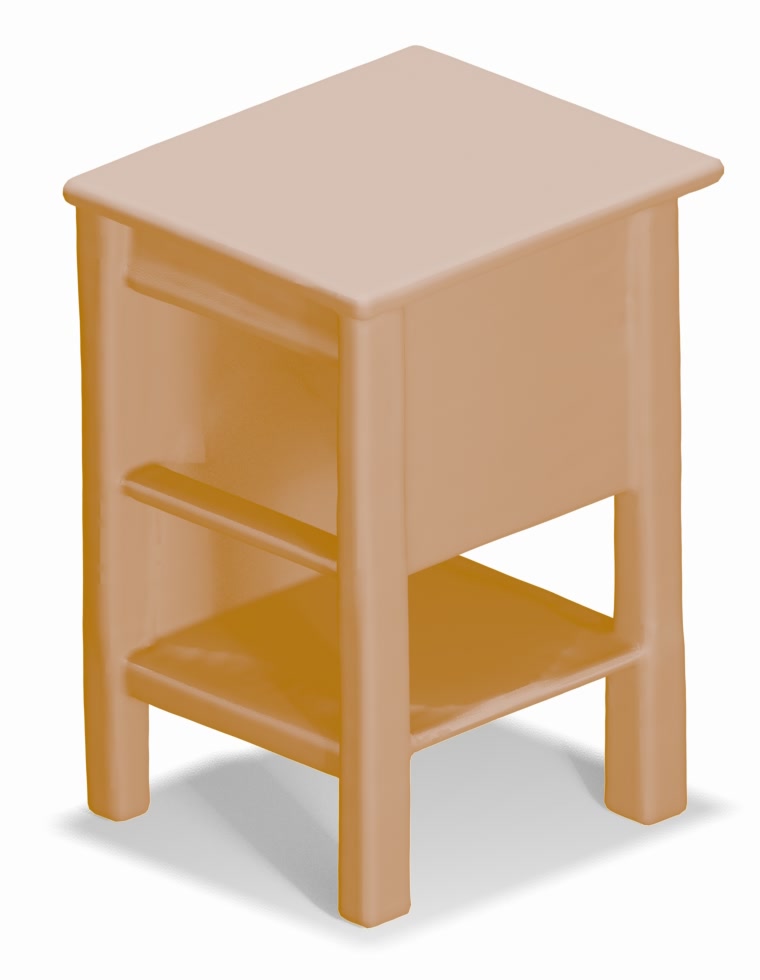}
&
\includegraphics[height=0.115\textheight]{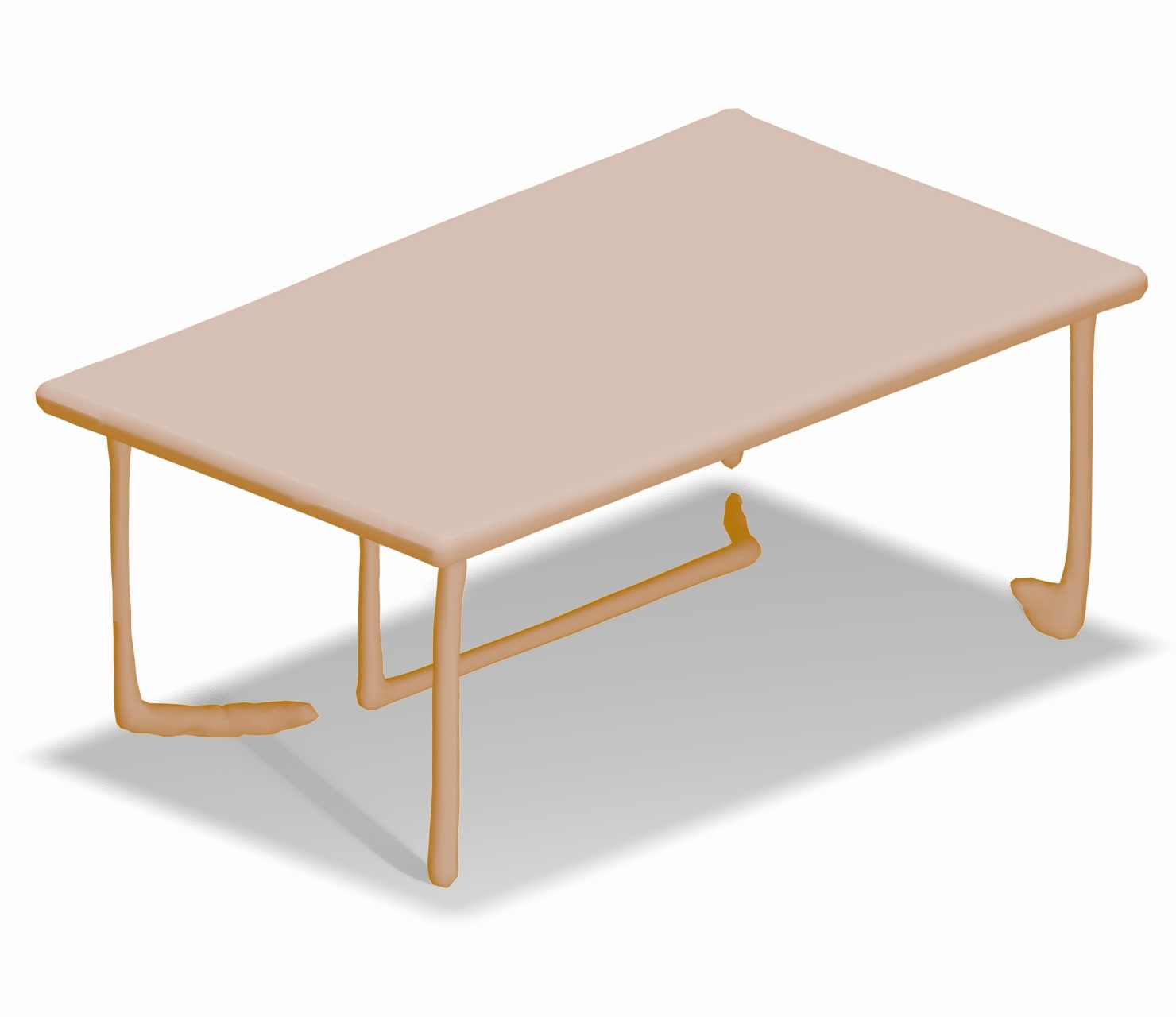}
&
\includegraphics[height=0.115\textheight]{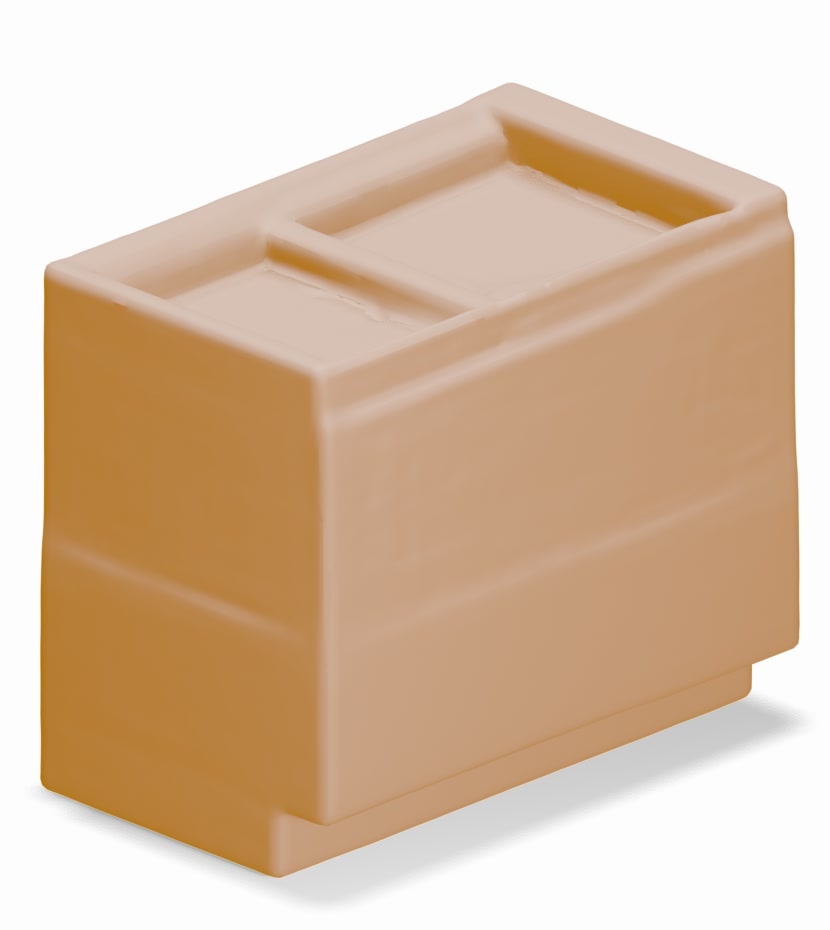}
\end{tabular}
}

%

%
        \vspace{-2mm}
\caption{{\bf Additional Qualitative Illustrations.}
Comparison between our method, Luo~\etal~\cite{luo20233d}, and LAS-Diffusion~\cite{zheng2023locally} on the real test set of \textsc{VRSketch2Shape}. 
}
\label{fig:suppl_quali}
\end{figure*}

We present additional comparisons with Luo~\etal~\cite{luo20233d} and LAS-Diffusion~\cite{zheng2023locally} in \cref{fig:suppl_quali}.
Although LAS-Diffusion yields smooth surfaces, it struggles with occlusions and fails to generate complete geometry, consistent with its higher Chamfer errors. Our method produces more detailed, structurally accurate shapes across diverse sketches.


\end{document}